\PassOptionsToPackage{french,english}{babel}
\documentclass[
12pt, 
oneside, 
english, 
singlespacing, 
liststotoc, 
parskip, 
headsepline, 
]{MastersDoctoralThesis} 

\usepackage[english,french]{babel}
\AtBeginDocument{}
\usepackage[utf8]{inputenc} 
\usepackage[T1]{fontenc} 
\usepackage{url}
\usepackage[skip=0pt]{caption}
\usepackage{palatino} 
\usepackage{natbib}
\usepackage{multirow}
\usepackage{qtree}
 \usepackage{longtable}
 \usepackage[final]{pdfpages}
 \usepackage{lscape}
 \usepackage{array,makecell}
 \usepackage[toc,page]{appendix}
 \usepackage{pdflscape}
\newcolumntype{P}[1]{>{\arraybackslash}p{#1}}
\newcolumntype{M}[1]{>{\arraybackslash}m{#1}}
\usepackage{floatrow}
\usepackage[framemethod=TikZ]{mdframed}
\mdfdefinestyle{MyFrame}{%
    linecolor=black,
    outerlinewidth=0.1pt,
    roundcorner=0pt,
    innertopmargin=5pt,
    innerbottommargin=5pt,
    innerrightmargin=5pt,
    innerleftmargin=5pt,
    font=\small,
    backgroundcolor=white}
\usepackage{listings}
\usepackage{algorithmicx}
\usepackage{algorithm}
\usepackage{algpseudocode}
\usepackage{algorithm,algorithmicx,algpseudocode}
\algdef{SE}[DOWHILE]{Do}{doWhile}{\algorithmicdo}[1]{\algorithmicwhile\ #1}%
\makeatletter
\algrenewcommand\ALG@beginalgorithmic{\small}
\makeatother

\usepackage{setspace}

\makeatletter
\renewcommand\footnotesize{%
   \@setfontsize\footnotesize\@ixpt{11}%
   \abovedisplayskip 8\p@ \@plus2\p@ \@minus4\p@
   \abovedisplayshortskip \z@ \@plus\p@
   \belowdisplayshortskip 4\p@ \@plus2\p@ \@minus2\p@
   \def\@listi{\leftmargin\leftmargini
               \topsep 4\p@ \@plus2\p@ \@minus2\p@
               \parsep 2\p@ \@plus\p@ \@minus\p@
               \itemsep \parsep}%
   \belowdisplayskip \abovedisplayskip
}
\makeatother
\usepackage[shortlabels]{enumitem}

\usepackage{bbding}
\usepackage{pifont}
\usepackage{wasysym}
\usepackage{amssymb}
\usepackage{hyperref}
\usepackage{minitoc}
\usepackage{titlesec}
\titlespacing*{\section}
{0pt}{2.5ex plus 1ex minus .2ex}{0.8ex plus .2ex}
\titlespacing*{\subsection}
{0pt}{2.5ex plus 1ex minus .2ex}{0.8ex plus .2ex}

\titlespacing*{\subsubsection}
{0pt}{2.5ex plus 1ex minus .2ex}{0.8ex plus .2ex}

\usepackage[autostyle=true]{csquotes} 
\usepackage{ragged2e}
\addto\captionsfrench{}

\thesistitle{Contributions to the Improvement of Question Answering Systems in the Biomedical Domain} 
\supervisor{Prof. Said \textsc{Ouatik El Alaoui}} 
\examiner{} 
\degree{Philosophy Doctor (PhD) } 
\author{Mourad \textsc{Sarrouti}} 
\addresses{} 

\subject{Computer Science} 
\university{\href{http://www.usmba.ac.ma/}{Sidi Mohamed Ben Abdelah University}} 
\department{\href{http://department.university.com}{Laboratory of Informatics and Modeling}} 
\group{\href{http://researchgroup.university.com}{Research Group Name}} 
\faculty{\href{http://faculty.university.com}{Faculty of Sciences Dhar El Mahraz}} 

\AtBeginDocument{
\hypersetup{pdftitle=\ttitle} 
\hypersetup{pdfauthor=\authorname} 
\hypersetup{pdfkeywords=\keywordnames} 
}

\begin{document}

\frontmatter 

\pagestyle{plain} 


\includepdf[offset=-0.0cm -0.0cm]{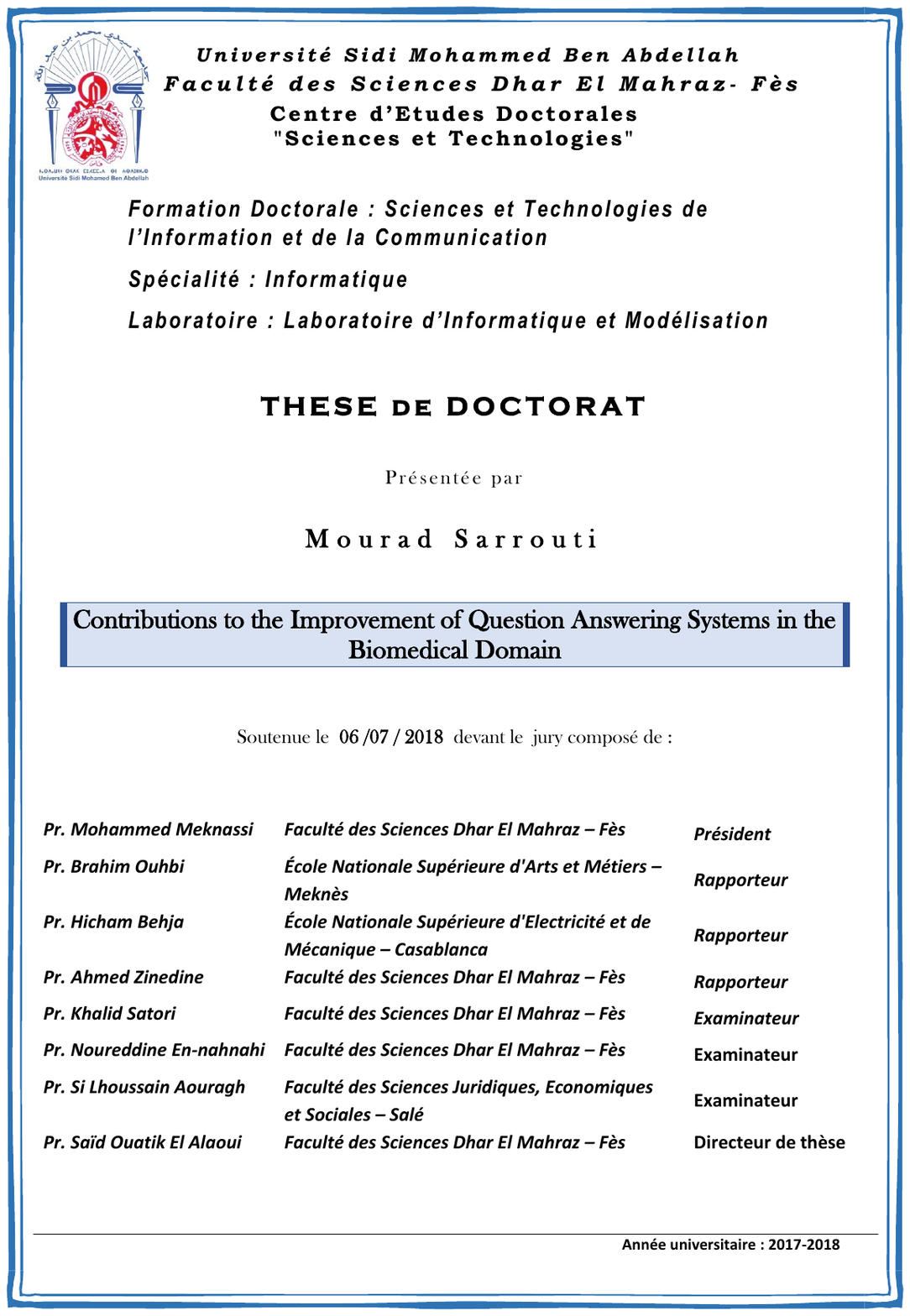}
\begin{titlepage}
\begin{center}

{\scshape\LARGE \univname\par}\vspace{1.5cm} 

\includegraphics[width=4cm, height=3cm]{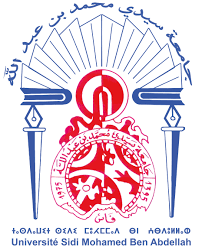}\vspace{1.5cm} 

\textsc{\Large PhD Thesis}\\[0.5cm] 

\HRule \\[0.4cm] 
{\Large \bfseries \ttitle\par}\vspace{0.4cm} 
\HRule \\[1.5cm] 

\begin{minipage}[t]{0.4\textwidth}
\begin{flushleft} \large
\emph{Author:}\\
\href{https://sites.google.com/site/mouradsarrouti/home}{\authorname} 
\end{flushleft}
\end{minipage}
\begin{minipage}[t]{0.4\textwidth}
\begin{flushright} \large
\emph{Supervisor:} \\
\href{https://scholar.google.fr/citations?user=ToOococAAAAJ&hl=fr&oi=ao}{\supname} 
\end{flushright}
\end{minipage}\\[3cm]

\vfill

\large \textit{A thesis submitted in fulfillment of the requirements\\ for the degree of \degreename}\\[0.3cm] 
\textit{in the}\\[0.4cm]

\textit{``Computer Science and Informatics'' specialty}\\ [2cm]


\vfill

{\large 06/07/2018}\\[4cm] 

\vfill
\end{center}
\end{titlepage}


\cleardoublepage





\dedicatory{
\begin{flushright}
To my parents who taught me to move on \ldots

\end{flushright}
}

\begin{acknowledgements}
\addchaptertocentry{\acknowledgementname} 
\begin{flushright}
``\emph{We all change. When you think about it, we’re all different people all
through our lives, and that’s okay, that’s good, you gotta keep moving, so long
as you remember all the people that you used to be. I will not forget one line of
this. Not one day. I swear. I will always remember when the Doctor was me.}'' The Doctor (Doctor Who, The Time of the Doctor)
\end{flushright}

Many people contributed in the elaboration of this PhD thesis in one way or
another, without whom this work probably could not have been completed, for
this reason I will take this opportunity to express my immense gratitude to each
and every one of them.

First and foremost, I would like to express my deepest gratitude to my
supervisor, Pr. Said Ouatik El Alaoui, for giving me the opportunity to join the
LIM laboratory, for choosing the subject and the scientific orientation of this
thesis work, for providing me the freedom to pursue my research, for carrying
out stimulating discussions despite his busy schedule. I would also like to
thank him for his personal and human qualities which have also contributed a lot
to the accomplishment of this work. Doing research is one of the most enriching
activities I have done in my life, but it is also a hard and demanding endeavor
that, without proper guidance, can lead to frustration and discouragement. I
would like to thank him for always having his door open for lively discussion
and for his all valuable comments and suggestions during the accomplishment of
this thesis work. His enthusiasm motivated me a lot.

I particularly thank Pr. Mohammed Meknassi, the director of the LIM
laboratory, for his presence and dedication to the service of his laboratory
despite all his responsibilities. I cannot but record my deep gratitude and
appreciation to Pr. Mohammed El Hassouni, the vice dean of research,
cooperation and partnerships of the Faculty of Sciences Dhar ElMahraz.

I owe a huge thank you to my examining committee members: Pr. Brahim
Ouhbi, Pr. Hicham Behja and Pr. Ahmed Zinedine for their time, interest, and
helpful comments, as well as the three members of my oral defense committee,
Pr. Khalid Satori, Pr. Noureddine En-Nahnahi and Pr. Si Lhoussain Aouragh, for
their time, invaluable comments and advices.

Special thanks are given to all my friends who were willing to put into time and
effort to proof read various drafts of the technical writings related to this thesis.

I would extend my warm thanks to my family, friends and colleagues who gave
me continuous encouragement and uninterrupted stimulus throughout all my
years at university. I especially appreciate the unconditional support of my
parents without whom I would not have made it through my PhD studies. No
words would be enough to reward your sacrifices and patience.

Thank you all.

\begin{flushright}
\emph{Fez, June 22, 2018.}
\end{flushright}

\end{acknowledgements}

\selectlanguage{english}
\begin{abstract}

\addchaptertocentry{\abstractname} 

This thesis work falls within the framework of question answering (QA) in the biomedical domain where several specific challenges are addressed, such as specialized lexicons and terminologies, the types of treated questions, and the characteristics of targeted documents. We are particularly interested in studying and improving methods that aim at finding accurate and short answers to biomedical natural language questions from a large scale of biomedical textual documents in English. QA aims at providing inquirers with direct, short and precise answers to their natural language questions. A typical QA system can be viewed as a pipeline composed of four main components including (1) question analysis and classification, (2) document retrieval, (3) passage retrieval, and (4) answer extraction. We consider that the improvement of such fundamental dimensions of the usefulness of QA has to take into account and solve the problems lying in each of these components. In this Ph.D. thesis, we propose four contributions to improve the performance of QA in the biomedical domain. In our first contribution, we propose a machine learning-based method for question type classification to determine the types of given questions which enable to a biomedical QA system to use the appropriate answer extraction method. We also propose an another machine learning-based method to assign one or more topics (e.g., pharmacological, test, treatment, etc.) to given questions in order to determine the semantic types of the expected answers which are very useful in generating specific answer retrieval strategies. In the second contribution, we first propose a document retrieval method to retrieve a set of relevant documents that are likely to contain the answers to biomedical questions from the MEDLINE database. We then present a passage retrieval method to retrieve a set of relevant passages to questions. In the third contribution, we propose specific answer extraction methods to generate both \emph{exact} and \emph{ideal} answers. Finally, in the fourth contribution, we develop a fully automated semantic biomedical QA system called SemBioNLQA which is able to deal with a variety of natural language questions and to generate appropriate answers by providing both \emph{exact} and \emph{ideal} answers. SemBioNLQA is derived from our established methods.  Our proposals are evaluated on a common experimental design that considers large standard and well-known datasets for biomedical questions and answers provided by the BioASQ challenge. The experimental results support the validity of our contributions. In addition, a subsystem of SemBioNLQA was presented at the 2017 BioASQ challenge and was one of the challenge winners.

\keywordnames{\textbf{Keywords: }}{Question answering, Biomedical text mining, Information retrieval, Natural language processing, Machine learning, Semantic approach.}

\end{abstract}

\selectlanguage{french}
\begin{abstract}
\addchaptertocentry{\abstractnom}

Ce travail de thèse s’inscrit dans le cadre des systèmes de questions-réponses (SQR) dans le domaine biomédical où plusieurs défis spécifiques à ce domaine sont relevés tels que le lexique et la terminologie, le type des questions posées, et la particularité des documents traités. Nous nous intéressons particulièrement à l’étude et l’amélioration des méthodes permettant de déterminer les réponses précises à des questions biomédicales exprimées en langage naturel (en anglais). Les SQR visent à fournir, à partir d’une collection de documents, des réponses succinctes et précises à des questions en langage naturel. Généralement, ils sont constitués de quatre composantes principales : (1) l'analyse et la classification des questions, (2) la sélection des documents pertinents, (3) la recherche des passages pertinents, et (4) l'extraction des réponses. Dans ce travail de thèse, nous apportons quatre contributions. Dans la première, nous avons proposé une méthode de classification permettant de déterminer le type de la question. Etant basée à la fois sur les patrons lexico-syntaxiques et l’apprentissage automatique, celle-ci est exploitée par le SQR dans la phase d’extraction de la réponse appropriée à une question formulée en langage naturel. Dans le but de déterminer le type sémantique de la réponse attendue (i.e., un ou plusieurs sujets), nous avons proposé une variante de cette méthode qui s’appuie sur d’autres caractéristiques de type lexical, morpho-syntaxique et sémantique. Le type sémantique d’une question biomédicale peut être pharmacologie, analyse, traitement, etc. Cette information permet de réduire le nombre de documents parcourus lors de la recherche des réponses. La deuxième contribution consiste à suggérer une méthode de recherche des documents pertinents à la question à partir de la base de données MEDLINE. Nous avons également proposé une alternative permettant la recherche des passages (i.e., extraits des documents) pertinents susceptibles de contenir les réponses candidates aux questions biomédicales. La troisième contribution propose des méthodes d’extraction des réponses appropriées permettant de générer à la fois les réponses \emph{exactes} et \emph{idéales}. En fin, dans la quatrième contribution, l’ensemble des méthodes proposées sont développées et intégrées au sein d’un système global de questions-réponses, appelé SemBioNLQA. Celui-ci accepte en entrée une variété de questions et retourne des réponses \emph{exactes} et \emph{idéales}. L'ensemble des contributions sont évaluées en utilisant la collection standard de questions fournies par la compagne d'évaluation BioASQ. Les résultats obtenus montrent l'intérêt de notre propos. De plus, un sous-système de SemBioNLQA a été présenté au challenge BioASQ 2017 et classé parmi les premiers vainqueurs.

\keywordnames{\textbf{Mots clés: }}{Systèmes de questions-réponses, Fouille de texte biomédical, Recherche d’information, Traitement automatique de la langue, Apprentissage automatique, Approche Sémantique.}
\end{abstract}
\selectlanguage{english}

\dominitoc
\tableofcontents 
\singlespacing

\listoffigures 

\listoftables 


\begin{abbreviations}{ll} 

\textbf{AI} & \textbf{A}rtificial \textbf{I}ntelligence\\
\textbf{BioNER}& \textbf{Bio}medical \textbf{N}amed \textbf{E}ntity \textbf{R}ecognition\\
\textbf{BioNLP}& \textbf{Bio}medical \textbf{N}atural \textbf{L}anguage \textbf{P}rocessing\\
\textbf{BOB}& \textbf{B}ag-\textbf{o}f-\textbf{B}igrams\\
\textbf{BOCST}& \textbf{B}ag-\textbf{o}f-\textbf{U}MLS \textbf{C}oncept and \textbf{S}emantic \textbf{T}ypes\\
\textbf{BOBNE}& \textbf{B}ag-\textbf{o}f-\textbf{B}iomedical \textbf{N}amed \textbf{E}ntities\\
\textbf{BOSDR}& \textbf{B}ag-\textbf{o}f-\textbf{S}yntactic \textbf{D}ependency \textbf{R}elations\\
\textbf{BOS} &\textbf{B}ag-\textbf{o}f-\textbf{S}tems\\
\textbf{BOW}& \textbf{B}ag-\textbf{o}f-\textbf{W}ords\\

\textbf{CLEF} & \textbf{C}ross \textbf{L}anguage \textbf{E}valuation \textbf{F}orum\\
\textbf{EBI} & \textbf{E}uropean \textbf{B}ioinformatics \textbf{I}nstitute \\
\textbf{IE} & \textbf{I}nformation \textbf{E}xtraction\\
\textbf{IR} & \textbf{I}formation \textbf{R}etrieval\\
\textbf{MAP} & \textbf{M}ean \textbf{A}verage \textbf{P}recision\\
\textbf{MeSH} & \textbf{Me}dical \textbf{S}ubject \textbf{H}eadings \\
\textbf{MRR}& \textbf{M}ean \textbf{R}eciprocal \textbf{R}ank\\
\textbf{NaCTeM} &\textbf{Na}tional \textbf{C}entre for \textbf{Te}xt \textbf{M}ining \\
\textbf{NCBI} & \textbf{N}ational \textbf{C}enter for \textbf{B}iotechnology \textbf{I}nformation\\
\textbf{NCBO} & \textbf{N}ational \textbf{C}enter for \textbf{B}iomedical \textbf{O}ntology\\
\textbf{NCI} &\textbf{N}ational \textbf{C}ancer \textbf{I}nstitute\\
\textbf{NER}& \textbf{N}amed \textbf{E}ntity \textbf{R}ecognition\\
\textbf{NIH}&  \textbf{N}ational \textbf{I}nstitutes of \textbf{H}ealth\\
\textbf{NLM}& \textbf{N}ational \textbf{L}ibrary of \textbf{M}edecine\\
\textbf{NLP} & \textbf{N}atural \textbf{L}anguage \textbf{P}rocessing\\
\textbf{NTCIR} & \textbf{N}II-NACSIS \textbf{T}est \textbf{C}ollection for \textbf{IR} systems\\
\textbf{OBO} &\textbf{O}pen \textbf{B}iomedical \textbf{O}ntology \\
\textbf{POS}& \textbf{P}art-\textbf{O}f-\textbf{S}peech\\
\textbf{PMID}&\textbf{P}ub\textbf{M}ed \textbf{ID}entifier\\
\textbf{QA} & \textbf{Q}uestion \textbf{A}nswering\\
\textbf{SemBioNLQA}& \textbf{Sem}antic \textbf{Bio}medical \textbf{N}atural \textbf{L}anguage \textbf{Q}uestion \textbf{A}nswering\\
\textbf{SNOMED CT}& \textbf{S}ystematized \textbf{N}omenclature of \textbf{M}edicine \textbf{C}linical \textbf{T}erms \\
\textbf{SVM}& \textbf{S}upport \textbf{V}ector \textbf{M}achine\\
\textbf{TFIDF}& \textbf{T}erm \textbf{F}requency \textbf{I}nverse \textbf{D}ocument \textbf{F}requency\\
\textbf{TREC} & \textbf{T}ext \textbf{RE}trieval \textbf{C}onference \\
\textbf{UMLS} & \textbf{U}nified \textbf{M}edical \textbf{L}anguage \textbf{S}ystem \\
\textbf{UTS}&\textbf{U}MLS \textbf{T}erminology \textbf{S}ervices\\

\end{abbreviations}





\mainmatter 

\pagestyle{thesis} 



\chapter*{General Introduction} 
\addcontentsline{toc}{chapter}{General Introduction}
\label{Chapter1} 
\markboth{General Introduction}{General Introduction}

\newcommand{\keyword}[1]{\textbf{#1}}
\newcommand{\tabhead}[1]{\textbf{#1}}
\newcommand{\code}[1]{\texttt{#1}}
\newcommand{\file}[1]{\texttt{\bfseries#1}}
\newcommand{\option}[1]{\texttt{\itshape#1}}


\section*{Motivation}

\addcontentsline{toc}{section}{Motivation}

\begin{figure}[h!]
\captionsetup{justification=justified}
\graphicspath{{Figures/}}
\centering
\includegraphics[width=14.5cm, height=8cm]{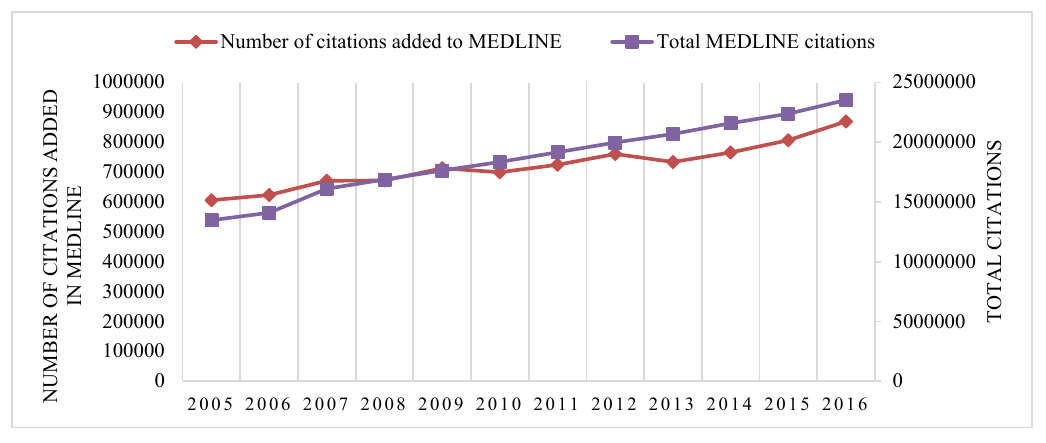}
\caption[Total number of citations added to MEDLINE by fiscal year]{Total number of citations added to MEDLINE by fiscal year: 2005-2016. These statistics come from the official site of the U.S. NLM: \url{https://www.nlm.nih.gov/bsd/index_stats_comp.html\#footnote1)}}
\label{fig:MEDLINEstatistics}
\end{figure}

Over the course of the last decades, research in the biomedical field has received a fast growing interest from the research community, reflected by an exponential growth in the peer-reviewed scientific literature. MEDLINE\footnote{MEDLINE: \url{https://www.nlm.nih.gov/pubs/factsheets/medline.html}}, the U.S. National Library of Medicine (NLM) bibliographic database, contains more than 24 million references to journal articles in life sciences with a concentration on biomedicine. As shown in Figure~\ref{fig:MEDLINEstatistics} which shows the total number of citations added to MEDLINE by fiscal year, hundreds of thousands of citations are added to the database each year. For instance, more than 806,000 and 869,000 citations were added in 2015 and 2016 respectively. These statistics come from the official site of the U.S. NLM \footnote{U.S. National Library of Medicine: \url{https://www.nlm.nih.gov/bsd/index_stats_comp.html\#footnote1)}}.

Unsurprisingly, this explosive rise in the amount of written scientific knowledge has made it difficult to absorb all relevant information even for experts in their field of interest. In this sense, recent reports such as the one presented by
\cite{hristovski2015biomedical}, have highlighted that the most widely used are specialized and domain-specific classical information retrieval (IR) systems, also known as search engines, such as PubMed\footnote{PubMed: \url{https://www.ncbi.nlm.nih.gov/pubmed/}} which gives access to the MEDLINE database. To given queries, usually expressed in terms of some keywords or concepts, traditional IR systems return a large number of citations that are potentially relevant. Indeed, the set of the retrieved documents represents an answer size that is still too large to identify the precise information readily. Moreover, in traditional IR, the users have often to deal with the burden of studying and filtering the returned citations of their queries so as to find the precise information they were looking for.
In this context, an evaluation study presented by \citet{ely2000taxonomy} showed that physicians spent on average less than two minutes looking for information to answer clinical questions, although many of their questions remained unanswered. In another study by \citet{Hersh283}, at least 30 minutes are needed on average for medical and nurse practitioner students to answer clinical questions using MEDLINE.

Retrieving short, accurate answers to a given natural language question from the ever-increasing volume of biomedical literature is the key to creating high-quality systematic reviews that support the practice of evidence-based medicine \citep{Sarker_2016,Ji_2017} and improve the quality of patient care \citep{ely2002obstacles,Kopanitsa_2017,Kropf_2017}. However, \citet{elyjhon} have shown that physicians do not seek answers to many of their natural language questions, often suspecting a lack of usable information. The most commonly reported obstacles to the search of an answer were the doctors's doubt that an answer existed and also the failure of the selected database to provide an answer.

To reduce searching, filtering and browsing time and effort while maximizing usefulness of that scientific knowledge, more accurate systems are needed such as question answering systems \citep{Wren_2011}. Question answering (QA), unlike classical IR, aims at providing information seekers with precise and specific answers, by automatically analyzing thousands of articles based on information extraction (IE) and natural language processing (NLP) techniques, instead of providing a large number of citations that are potentially relevant for the natural language questions posed by the inquirers \citep{lee2006beyond,athenikos2010biomedical,Bauer_2012,neves2015question}. Furthermore, QA systems, which can help users locate useful information quickly, take questions expressed in natural language (e.g., why, how, when, etc.) and extract precise answers by linguistically and semantically processing both questions and data sources under consideration. For example, for the biomedical question ``What is the name for anorexia in gymnasts?'', a QA system would provide a particular name as an answer , i.e., Anorexia Athletica, for anorexia in gymnasts.

Due to the importance of QA systems in the biomedical domain, a great number of research groups in the areas of IR, NLP and artificial intelligence have dedicated great efforts in the development of new methodologies to automatically answer biomedical questions from the scientific
literature. This is the scope of this thesis work reported here.

\section*{Background}
\addcontentsline{toc}{section}{Background}

Question answering in the open domain, a longstanding challenge widely studied, has received much attention by the IR community, initiated by the QA Track in Text REtrieval Conference (TREC\footnote{TREC: \url{http://trec.nist.gov/}}) evaluations, which takes place regularly every year since 1999 \citep{voorhees1999trec}. Since then, various open-domain QA systems have been proposed and developed \citep{voorhees1999trec,roberts2002information,Monz_2003,gaizauskas2004information,Collins_Thompson_2004,voorhees2005trec,Teufel}. Nevertheless,
only few efforts have been made in the biomedical domain due to a variety of reasons and conditions. As has been extensively documented in recent survey on biomedical QA \citep{athenikos2010biomedical}, open-domain QA is concerned with the questions that are not restricted to any domain, while in restricted-domain QA such as the biomedical one, the application domain provides a context for the QA process. \cite{athenikos2010biomedical} have also summarised the main characteristics of QA in the biomedical domain in three points: (1) large-sized textual corpora, (2) highly complex domain specific terminology, and (3) domain-specific format and typology of questions. The most basic obstacle is that there exists a vast amount of textual data, especially in the form of scientific articles that is constantly and rapidly increasing as new scientific articles are published every day in the biomedical domain. Furthermore, not only these data are generally expressed in natural language, which makes its automated processing more difficult and complex, but also terminological variations and synonymy make question answering generally more difficult for the biomedical domain.

Recently, especially since the introduction of biomedical QA Track at the BioASQ\footnote{BioASQ challenge: \url{http://www.bioasq.org/}} challenge \citep{tsatsaronis2012bioasq}, biomedical QA systems have attracted an increasing level of interest by researchers \citep{balikas2014results,balikas2015results,krithara2016results,Nentidis_2017}. Despite this considerable progress, there is a general awareness that there are still many open challenges and controversial issues that affect the current state of biomedical QA systems. The current state of the existing biomedical QA systems does not deal with all types of questions and answers. Furthermore, until recently only few integral QA systems such as the ones presented in \citep{lee2006beyond,cruchet2009trust,gobeill2009question,Cao_2011,abacha2015means,Kraus_2017} can automatically retrieve answers to biomedical questions written in natural language. While such systems have proven to be quite successful at answering biomedical questions, they accept a limited amount of question types and, furthermore, they provide a limited amount of answer types. For instance, some of them \citep{lee2006beyond,cruchet2009trust,Cao_2011}  only handle definition questions or return solely short summaries as answers for all types of questions, and the most of the other ones do not deal with yes/no questions which are one of the most complicated question types to answer as they are seeking for a clear ``yes'' or ``no'' answer. Moreover, the biomedical QA systems still require further efforts in order to improve their performance in terms of precision to currently supported question and answer types.

\section*{Research Goals}
\addcontentsline{toc}{section}{Research Goals}

The aim of this thesis work is to propose new methods in biomedical QA so as to enable a user to find an accurate answer to his human natural language question. A typical QA system can be viewed as a pipeline composed of many components, including question classification, document retrieval, passage retrieval, and answer extraction each of which has to deal with specific challenges and issues \citep{HIRSCHMAN_2001,athenikos2010biomedical,neves2015question,abacha2015means}. We consider that the improvement of such fundamental dimensions of the usefulness of question answering has to take into account the problems lying in each of the aforementioned components. It is a matter of proposing, designing, and evaluating a system which is able to automatically identify the types of questions, relevant documents, relevant passages, and precise answers. To achieve this purpose, we have pursued the following research goals.

\begin{itemize}

 \item \textbf{Research Goal 1} is the proposal of sophisticated methods for improving QA performance in the biomedical domain through the combination of NLP/IR techniques, machine-learning methods, and domain-specific knowledge resources.

 \item \textbf{Research Goal 2} is to deal with all types of natural language questions and answers. As stated in the motivation section, the current state of the existing biomedical QA systems does not deal with all types of questions and answers.

 \item \textbf{Research Goal 3} is to develop a fully automated semantic biomedical QA system which is able to accept a variety of natural language questions and to generate appropriate answers by providing both exact and ideal answers.
\end{itemize}

\section*{Contributions}
\addcontentsline{toc}{section}{Contributions}

\begin{figure}[h!]
\captionsetup{justification=justified}
\graphicspath{{Figures/}}
\centering
\includegraphics[width=16.5cm, height=20cm]{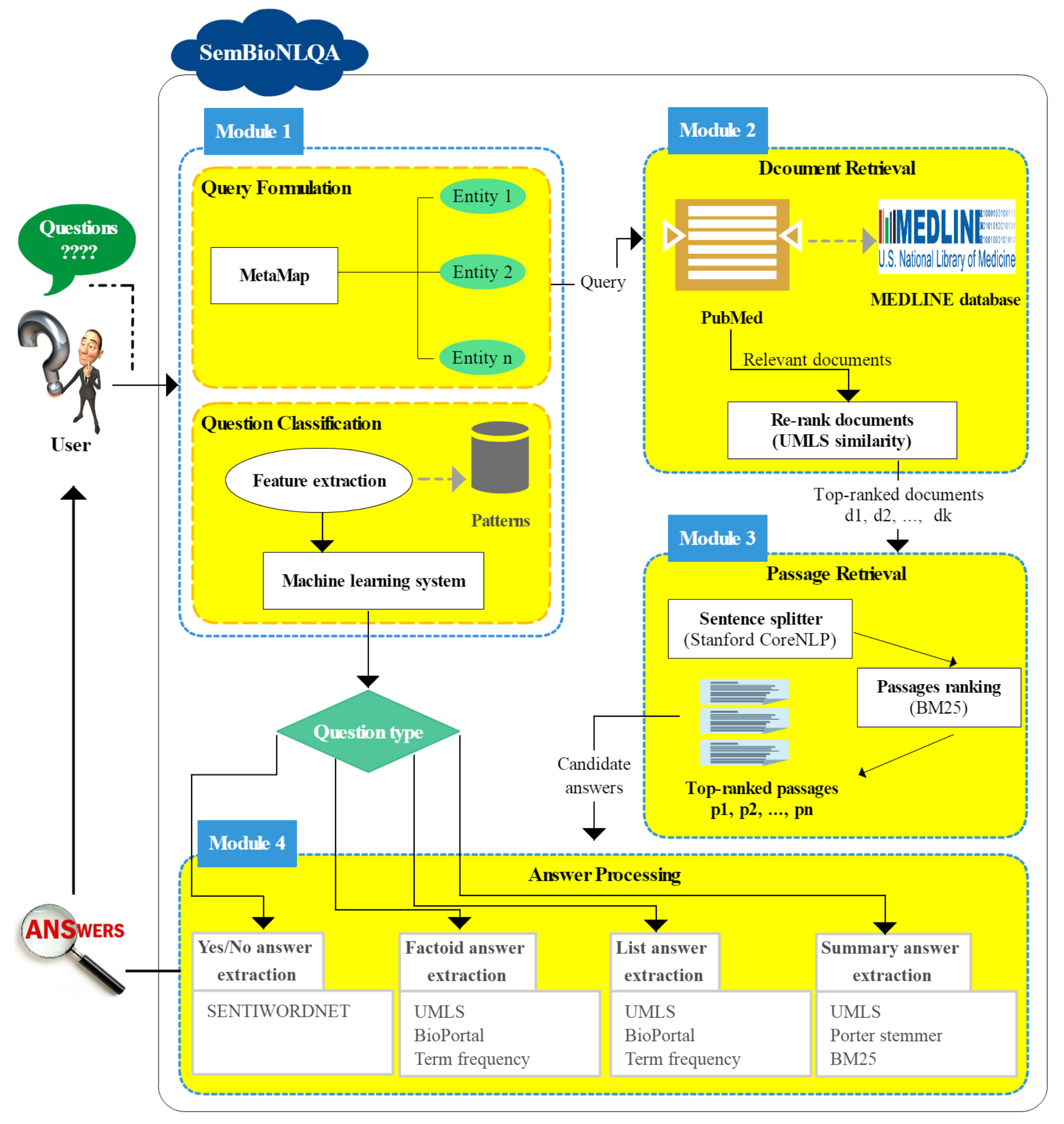}
\caption[Generic architecture of the proposed question answering system for the biomedical domain]{Generic architecture of the proposed question answering system in the biomedical domain.}
\label{fig:QAA1}
\end{figure}

By achieving the goals noted above, this thesis work makes various contributions to the field of question answering for the biomedical domain. This thesis work proposes new methods to improve QA performance in the biomedical domain through the combination of NLP techniques, machine-learning methods, and domain-specific knowledge resources. The proposed system, as shown in Figure~\ref{fig:QAA1}, is composed of many components which are: (1) question classification and query reformulation, (2) document retrieval, (3) passage retrieval, and (4) answer extraction.

The first component receives the input entered by the user, i.e., a natural language question (e.g., what, why, where, etc.), and includes preprocessing of the question, identification of the question type and the expected answer format to be required, as well as building a query which is an input to document retrieval, the second component. A document retrieval system is used to retrieve documents satisfying the query. After that, top-ranked passages are extracted from top-ranked documents by the passage retrieval, the third component. The output of this component is a set of top-ranked passages which is used as a set of candidate answers and as input to the last component, answer extraction. In this component, the candidate answers are matched against the expected answer type generated by the first component and ranked by how well they satisfy the user question using an appropriate answer extraction method. Finally, the top-ranked candidate answers and the raw texts from which the answers were extracted are shown to the user.

This thesis work mainly revolves around the following contributions:

The \textbf{first contribution} of this thesis work is the proposal of two machine learning based methods for question classification in biomedical QA. The first method \citep{kdir15,Sarrouti_MIM_2017} consists at identifying the type of a given biomedical question in order to determine the expected answer format. It is based on our predefined set of handcrafted lexico-syntactic patterns and machine learning algorithms. The second method \citep{Sarrouti_IBRA_2017}, which is based on lexical, syntactic and semantic features for machine learning algorithms, allows classifying questions into topics in order to filter out irrelevant answer candidates.

The \textbf{second contribution} of this thesis work is, at first, the proposal of a document retrieval method \citep{Sarrouti_2016} which retrieves relevant citations to a given biomedical question from the MEDLINE database. The proposed method first builds the query by extracting biomedical concepts, then uses a specialized IR system that gives access to the MEDLINE database to retrieve relevant documents, and finally ranks them based on a semantic similarity. We also propose an alternative \citep{Sarrouti_2017} based on a probabilistic IR model and biomedical concepts to retrieve and extract a set of relevant passages (i.e., snippets) from the retrieved documents to given biomedical questions.

The \textbf{third contribution} of this thesis work is the proposal of effective methods for answer extraction from passages that potentially contain answers through the use of semantic knowledge, NLP techniques and statistical techniques. The first answer extraction method \citep{Sarrouti_yes_2017}, based on a sentiment lexicon, aims at generating the exact answers to yes/no questions. The second method uses a biomedical metathesaurus to provide the exact answers suited for factoid and list questions which require with respectively a biomedical entity and a list of them as answers. The third method, aiming at retrieving the ideal answers (i.e., short summaries of relevant information) to biomedical questions, is based on a probabilistic IR model and biomedical concepts \citep{Sarrouti_bioasq_2017}.

The \textbf{fourth contribution} of this thesis work is the development of a fully automated semantic biomedical QA system named SemBioNLQA which is aimed to be able to accept a variety of natural language questions and to generate appropriate answers by providing both exact and ideal answers.  SemBioNLQA, which is a fully automatic system, includes innovative methods previously proposed in question classification, document retrieval, passage retrieval and answer extraction components. It is derived from our previously established contributions in each of the aforementioned components.

\section*{Structure}
\addcontentsline{toc}{section}{Structure}

The overview of this thesis is structured as follows:

\begin{itemize}

\item \textbf{Chapter~\ref{Chapter2}} presents general concepts of biomedical natural language processing and text mining. First, we describe some natural language processing tasks we use as constructing blocks for the new methods, such as sentence splitting, word tokenization, n-grams, part-of-speech tagging, stemming and lemmatization, and parsing. Then, we introduce biomedical text mining tasks, such as information retrieval, document classification, named entity recognition, relation or information extraction, document summarization and question answering. We also describe a broader classification framework involving QA systems based on different criteria, and important factors that distinguish restricted-domain QA from open-domain.

\item \textbf{Chapter~\ref{Chapter3}} describes the state-of-the-art on the topics of interest for this thesis work. First, we provide a brief introduction to question answering and its generic architecture. Next, we describe the main characteristics of QA in the biomedical domain, and the resources available for biomedical QA. Then, we present related work and discussion about the main QA approaches with a particular focus on the biomedical domain. Finally, we describe the setup of the experiments of our contributions. In particular, we describe in details the datasets and evaluations metrics used for the assessment of our contributions.

\item \textbf{Chapter~\ref{Chapter4}} describes in details the methods we propose for question classification in biomedical QA. We first describe our proposed method for biomedical question type classification. This method aims at classifying questions into one of the four categories: yes/no questions, factoid questions, list questions and summary questions. We then present our method for question topic classification. This method aims at automatically assigning one or more general topics (e.g., treatment, test, etc.) to clinical questions. Evaluation and results are also presented for each method.

\item \textbf{Chapter~\ref{Chapter5}} presents our contributions for the improvement of both document retrieval and passage retrieval in biomedical QA. We first describe our document retrieval system in which we propose a new document re-ranking method. We then describe our passage retrieval method which retrieves the relevant passages/snippets that are likely to contain the answer for a given biomedical question. Evaluation and results are also presented for each contribution.

\item \textbf{Chapter~\ref{Chapter6}} describes in details the methods we propose for the exact and ideal answer extraction. It also presents our semantic QA system, called SemBioNLQA. Evaluation and results are presented for each contribution as well.

\item \textbf{\hyperref[Chapter7]{\textcolor{black}{Conclusion and Future Work}}} offers the conclusion of this thesis work and indicates directions for future works which are either in progress or those we plan to carry out for the continuity of the systems that have been developed during this thesis work.


\item \textbf{Appendix~\ref{AppendixA}} contains a list of BioASQ questions used for the manual evaluation.

\item \textbf{Appendix~\ref{Chapter8}} contains a detailed summary of this thesis in French.

\item \textbf{\hyperref[AppendixB]{\textcolor{black}{Publications of the Author}}} contains publications of the author that are related to this thesis work.

\end{itemize} 
\setcounter{mtc}{7}

\chapter{Biomedical Natural Language Processing and Text Mining} 

\label{Chapter2} 

\minitoc

This chapter presents an introduction to natural language processing (NLP) and text mining, particularly in the biomedical domain. It also describes a broader classification framework involving QA systems based on different criteria, and important factors that distinguish restricted-domain QA from open-domain.



\section{Introduction}
\label{Chapter2_2.1}

The volume of published biomedical scientific articles, and therefore the underlying knowledge base, is expanding at an increasing rate. \citep{Cohen_2005}. For instance, the MEDLINE\footnote{MEDLINE: \url{https://www.nlm.nih.gov/pubs/factsheets/medline.html}} 2016 database contains over 24 million citations, and it is growing at the rate of 700,000 new scientific articles per year. With such explosive growth of biomedical literature, it is extremely challenging to absorb all relevant information even within one's own field of biomedical research. To handle such amount of information, the interest for the biomedical text mining from biomedical texts (e.g., scientific articles and electronic health records) has experienced a huge increment by the research community. Text mining and NLP refer to understanding and analyzing natural human language by using computer methods and tools. In particular, most text mining applications rely, to varying degrees, on NLP techniques and tools.

In this chapter we will review some concepts and techniques we use as constructing blocks for the new methodologies we present and which are referred throughout this thesis work. We start this chapter by describing in section~\ref{Chapter2_2.2} some related tasks to the natural language processing, such as sentence splitting, word tokenization, n-grams, part-of-speech tagging, stemming and lemmatization, and parsing. We also describe the most widely used NLP toolkits/platforms for the most common NLP tasks. We then proceed
with the presentation in section~\ref{Chapter2_2.3} of the biomedical text mining tasks, such as information retrieval, document classification, named entity recognition, relation or information extraction, document summarization, and question answering. We next describe in section~\ref{Chapter3_2} a broader classification framework involving QA systems based on different criteria. We finally present in section~\ref{Chapter3_3} important factors that distinguish restricted-domain QA from open-domain.

\section{Natural Language Processing}
\label{Chapter2_2.2}
Natural Language Processing (``NLP'' or ``BioNLP'' for the biomedical domain) is differentiated from text mining in that while NLP attempts to analyse and understand the meaning of text as a whole, text mining focus on solving a specific problem by discovering relevant information or explicit and implicit facts in text (possibly using some NLP methods in the process) \citep{Cohen_2005}. Indeed, NLP is generally a component of a text mining application that performs some pre-processing steps that essentially helps a machine ``read'' text \citep{Friedman_2013,Sun_2017,Kreimeyer_2017}.  In this section, we describe some of these pre-processing steps that are usually present on most of the text mining applications.

\subsection{Sentence tokenization}

The determination of the sentences that constitute the text using a special sentence splitter, also known as sentence boundary, is one of the first operations that are usually carried out for the text processing. This operation is necessary and important as most of the text mining systems require their input to be segmented into sentences for a number of reasons (e.g., the difficulties in dealing with anaphora). Although sentence segmentation seems to be an easy task when considering punctuation marks as separator such as the period, punctuation marks are often ambiguous. For example, a period may denote an abbreviation or decimal point not the end of a sentence. In this thesis work, Stanford CoreNLP, OpenNLP and Lingpipe sentence splitters have been used for separating the sentences in the abstracts of biomedical documents which where used for extracting the answers to biomedical questions.

\subsection{Word tokenization}

Tokenizing text into a sequence of tokens, which separates the sentence into an ordered list of words, is often the second operation in a NLP pipeline. This step is usually necessary of the text mining applications that require word-level granularity since the token is considered as the smallest unit in the text. Here again, word tokenization seems like a straightforward process, by considering white spaces as separator. However, it might not be so obvious since simple space delimitation does not deal with some of the subtleties of natural language text especially for the biomedical text where there are a lot of words (or at least phrasal lexical entries) that contain parentheses, hyphens, and so on. In this thesis work, Stanford CoreNLP and OpenNLP tokenizers a have been widely used for segmenting the tokens.

\subsection{N-grams}
N-grams, a sequence of tokens, are extensively used in many text mining tasks.  They are basically a set of co-occurring words, characters or subsets of characters within a given window. They also known as unigram, bigram or trigram in the case of sequences of size 1, 2 or 3, respectively. In this thesis work, bigrams have been used when referring to sequences of words in a sentence and they have been served as one of features of the machine learning algorithms developed for question classification. For instance, the question ``What is the molecular pathogenesis of Spinal Muscular Atrophy?'' might be represented by the bigrams \{What is, is the, the molecular, molecular pathogenesis, pathogenesis of, of Spinal, Spinal Muscular, Muscular Atrophy, Atrophy ?\}.

\subsection{Stemming and lemmatization}
For morphological reasons, a word might be expressed in different morphological forms which are in general the inflectional or derivational forms. The inflectional variations of the root word are its related forms with the same part-of-speech tag (i.e., word class) such as singular/plural variation, past/present tense variation and so on. For example, ``fish'' and ``fishes'' are inflected forms of the base or root ``fish''. On the other hand, derivational forms of a word are related to its variations with different POS tags. For example, ``fisher'' (noun) and ``fishing'' (verb) are derived forms for the root ``fish''. The importance to reduce inflectional and derivational forms of a word to the root form lies in that it would be useful for a search for the root word to retrieve and return textual documents that contain its different forms in the set. In a NLP pipeline, both stemming and lemmatization are aiming at transforming inflected and sometimes derived words to their common base form. Indeed, lemmatisation is closely related to stemming and both are special cases of normalization.  The main difference is that a stemmer reduces the word forms to the same stem even if the stem is not identical to the valid root of the word, whereas in lemmatization it will return the linguistically form of a word known also as the lemma, which must be a valid word. One of the most used stemmers is the Porter stemmer \citep{Porter_1980} which has been also used in many parts of this thesis work. Common stemming techniques and existing stemmers are described in \citep{jivani2011comparative}.

\subsection{Part-of-speech tagging}

Part-of-speech (POS) tagging, which is a process whereby tokens are sequentially associated with syntactic labels, also known as syntactic tags, word classes, or syntactic categories, is one of the most used syntactic analyzer tools. It indicates whether the word is a noun, a verb, an adjective or an adverb, for example. Knowing POS tags of words helps text mining applications to solve some problems lies in the syntactic disambiguation of each word in the text. Indeed, many English words can be used in more than one part-of-speech, which makes part-of-speech tagging an important block of syntactic parsing. For example, the word ``chat'' might refer to a noun (e.g., to have a chat) or a verb (to chat) depending on the context. The POS taggers provided by the Stanford toolkits have been widely used in many steps of this thesis work.

\subsection{Dependency parsing}
Dependency parsing or syntactic parsing, the task of analyzing the grammatical structure of a sentence by establishing syntactic dependency relations among the words, is a modern parsing technique. The main concept of dependency parsing is to construct a dependency parse tree of a sentence where the nodes represent linguistic unit (words) and the edges represent binary
asymmetric relations (called dependencies) between words. These structure trees are useful in various NLP applications such as question answering in which they play a vital role in the syntactic analysis of questions and answers. In this thesis work, the dependency parsing has been used in the question classification task as part of the question answering pipeline.

\subsection{Available NLP softwares}

Due to the importance and the necessity of the text processing for many text mining applications, various NLP toolkits/tools have been developed and made available for free to the text mining researchers. These tools support the most common NLP tasks. We list and describe here the most widely used toolkits:

\begin{itemize}
  \item The Standford CoreNLP\footnote{Standford CoreNLP: \url{https://stanfordnlp.github.io/CoreNLP/}.} \citep{Manning_2014}, a freely available natural language software, integrates many NLP tools including tokenizers, POS tagger, syntactic parsers, the named entity recognizer, the coreference resolution system, sentiment analysis, bootstrapped pattern learning, and the open information extraction tools. The Stanford CoreNLP is one of the most widely used tools for natural language processing. Accordingly, the Stanford CoreNLP has been widely used in this thesis work especially for tokenization, POS tagging and syntactic parsing tasks.
  \item LingPipe\footnote{LingPipe: \url{http://alias-i.com/lingpipe}.} is a freely available tool kit for the text processing using computational linguistics. It includes many NLP tasks such as sentence detection, POS tagging, named entity recognition, word sense disambiguation, language identification, sentiment analysis,  etc. LingPipe has been also used in this thesis work, especially for sentence detection.

  \item OpenNLP \footnote{OpenNLP: \url{https://opennlp.apache.org/}.} is also a freely available tool to support the most common NLP tasks such as, sentence tokenization, word tokenization, POS tagging, named entity extraction, chunking, parsing, language detection and coreference resolution. OpenNLP has been also used in this thesis work for sentence tokenization.

\end{itemize}

\section{Text Mining Tasks}
\label{Chapter2_2.3}
Text mining is the process of identifying and extracting high-quality information from natural language text \citep{Jackson_2007}. This information may be explicitly or implicitly stated in the text \citep{Zweigenbaum_2007}. On this definition, any system that extract and discover explicit or implicit knowledge hidden in text, or perform processing that are mandatory prerequisites for doing so, would be considered as a text mining application. This would include a range of application types such as named entity recognition, text classification, relation or information extraction, information retrieval, text classification, text summarization and question answering. In this section, we define these main tasks of text mining.


\subsection{Named entity recognition}
Named entity recognition (NER) is a task that recognizes sequences of words in a natural language text which are the names for a specific type of thing, such as person, locations and company names, or drug, gene and protein names. This task is an important area of research as it allows more complex text-mining tasks to be addressed \citep{de_Bruijn_2002,Cohen_2005}. For instance, identifying entities in text allows for further extraction of semantic relationships, retrieving documents related to a certain type of thing (e.g., gene), and answering many natural language questions. Due to a variety of reasons, NER is a challenging task, especially for biomedical text (also known as BioNER) where biological or medical terms should be identified \citep{Simpson_2012}. In the biomedical domain, not only a vast amount of biomedical entities exists but also new entities are rapidly being discovered \citep{Yeh_2005}. This ever-increasing list of semantically relevant terms is a real challenge for simple text matching algorithms that use only a dictionary based approach since these dictionaries can never be complete as long as new scientific discoveries, achievements, and inventions are made all the time. Another challenge of NER for biomedical text is that the same concept may be expressed using different biomedical entities (e.g., PTEN and MMAC1 refer the same gene). Because of the potential utility of biomedical entities in many biomedical text mining applications, various BioNER systems have been developed such as MetaMap \citep{aronson2001effective} that relies on a biomedical metathesaurus constructed for the unified medical language system (UMLS).

\subsection{Relation or information extraction}

Information extraction, or more recently relation extraction, is the task that aims at extracting the basic facts from the literature. These facts typically take the form of semantic relationships between a pair of entities of given types. In their simplest form, relationships among entities are binary, involving only the pair-wise associations between two entities. However, relations can involve more than two entities, and extraction of these complex relations are also known as event extraction \citep{Zweigenbaum_2007}. On this definition, biomedical relation extraction is a subfield of the information extraction field that handles only with entities related to the biomedical domain, such as genes, proteins or diseases. Indeed, biomedical relation extraction is a key and an important task of text mining since it takes part in many biomedical processes, and various investigations and efforts have been carried out to this matter. For instance, a typical biomedical relation or information extraction might extract affirmations about gene-gene interactions \citep{Cordell_2009,Jiang2013}, protein-protein interactions \citep{Krallinger_2008,jhon_2012} or drugs-drugs interactions \citep{Ferdousi_2017}, or might also extract assertions about relationships between diseases and drugs \citep{van_Mulligen_2012,Jang_2016}.

By far, in addition to binary relation extraction, event extraction is one of the most popular kind of information extraction.
In this context, many challenges (evaluation forums) have taken place in the last years for evaluating information extraction systems in the biomedical domain, such as the BioCreAtIve\footnote{BioCreAtIve: \url{http://www.biocreative.org/tasks/}} challenge \citep{Yeh_2005,Krallinger_2008,Krallinger_2011,Islamaj_Dogan_2017}, the i2b2\footnote{i2b2: \url{https://www.i2b2.org/index.html}} challenge \citep{Uzuner_2011}, BioNLP Shared Task\footnote{BioNLP Shared Task: \url{http://2016.bionlp-st.org/home}} \citep{kim2011overview,nedellec2013overview,Kim_2012,nedellec2013overview,nedellec2016proceedings}, BioASQ\footnote{BioASQ: \url{http://www.bioasq.org/}} task on funding information extraction from biomedical literature \citep{Nentidis_2017}, and Text Analysis Conference (TAC\footnote{TAC: \url{https://tac.nist.gov/2017/}} 2017) track on adverse drug reaction extraction from drug labels.

\subsection{Text classification}

Text classification also known as text categorization is the task of classifying a natural language text into one or more predefined categories (also known as classes or labels) according to their content \citep{Cohen_2005}. Narrowing this definition, we define biomedical text classification as the task of the assignment of one or more classes to a biomedical text. In a typical text classification workflow, the information of interest is not specified explicitly by the users and, instead, they give a stream of textual documents that have been found to contain the characteristics of interest (the positive training set), and another collection that does not (the negative training set). Text classification systems first use feature sets extracted from the training data, such as a set of words that may allow determine positives from negatives and then apply those features to new textual documents using some kind of decision-making algorithm. Automatic text classification has many practical applications, including indexing for biomedical document retrieval \citep{Dram__2016}, biomedical named entity recognition \citep{Gridach_2017}, relation extraction \citep{Zheng_2017}, word sense disambiguation \citep{Jimeno_Yepes_2017}, and question classification \citep{Sarrouti_MIM_2017} by detecting the topics a questions covers. The approaches for text classification may be classified in methods based on rules, machine or deep learning, and hybrid which combine both rules and machine or deep learning.

\subsection{Text clustering}

Unlike text categorization (or text classification) which is a kind of ``supervised'' learning where the classes or labels are known beforehand and predefined in advance for each training textual document, text clustering, the application of cluster analysis to textual documents, is ``unsupervised'' learning. In text clustering, there is no predefined label or ``category,'' but groups of textual documents that contain the same characteristics are sought. Indeed, document clustering algorithms (e.g., K-Means) use and require only some approaches of distances or similarity between pairs of documents such as Euclidean distance, Jaccard Coefficient and Cosine Similarity. Although text clustering have seen relatively little application in biomedical text mining applications, it is important and useful for a variety of information needs and applications such as organizing documents for better information retrieval, classification, summarization and analysis \citep{Boyack_2011}.

\subsection{Text summarization}

Text summarization is the task of identifying the salient aspects of individual documents or groups of documents and present these aspects succinctly and coherently. In the biomedical domain, due to the continuous growth of textual information in the form of scientific articles, clinicians notes, electronic health records and a variety of other input types, automatic text summarization has been widely applied for aiding researchers and health care professionals to determine the shorter text that conveys the most important information contained in an increasingly large volume of biomedical text \citep{Simpson_2012}. In this context, summarization has been widely applied to journal articles, clinical notes, and a variety of other input types. For instance, one text summarization system, MITRE'S MiTAP, does multi-document summarization of epidemiological reports, online news, newswire feeds, email, television news, and radio news to identify and detect disease outbreaks. Summarization has been also widely to answer biomedical questions which requires single paragraph-sized (the most relevant information from concepts, articles, snippets, and triples) text as answers.

\subsection{Information retrieval}

Information retrieval (IR) is defined as a domain concerned with ``the structure, analysis, organization, storage, searching, and retrieval of information'' \citep{salton1968automatic}. On this definition, biomedical IR can be defined as ``the structure, analysis, organization, storage, searching, and retrieval of biomedical information''. A typical IR pipeline consists of a user, a collection of documents and an IR system. Given a collection of textual documents (e.g., web pages, biomedical scientific articles and medical reports) and a user's information need expressed as some sort of query, IR also known as document retrieval is the task of searching and returning the most relevant documents.
The information contained in documents can also be images, audio, video, etc. However, IR research has concentrated on retrieval of text written in natural language, a reasonable emphasis given the importance and large volume of textual data \citep{salton1986introduction,greengrass2000information,chowdhury2010introduction}. Due to the immense and increasing volume of biomedical literature, not only biomedical IR systems have become important for end-users, such as physicians, biologists, biochemists, and biomedical researchers searching directly for relevant citations to their queries, but also plays an important role in many other text mining applications. The identification of relevant documents for a given subject of interest is a preceding step for other text mining tasks, such as text classification, text summarization, and question answering. Several factors make this a challenging task: (1) the information in the documents is generally expressed in an unstructured form; (2) documents are usually written in natural language; and (3) very often, the documents may cover a wide range of topics \citep{mitra2000information}. Under these circumstances, IR is a key issue in text mining, and many efforts have been dedicated to this matter. PubMed is an example of a biomedical IR system for scientific articles; Google is an IR system for web pages.

\subsection{Question answering}

Unlike traditional IR systems (also known as search engines), where a list of potentially relevant documents from large text collections to a given query, question answering (QA) refers to the task of searching and returning the short, specific answers to questions usually expressed in natural language.  Since QA needs the use of more complex natural language processing methodologies than employed by information retrieval systems for linguistically and semantically processing both the questions and data sources so as to generate precise answers, QA systems go beyond search engines and regarded as the next generation for high accuracy information retrieval \citep{Zweigenbaum_2007}. Like automatic text summarization, QA is a process directed towards helping researchers, biologists and physicians in quickly and reliably managing the enormous growth of textual information in the biomedical domain \citep{Simpson_2012}. Due to the importance of retrieving the precise answers to natural language questions in the biomedical domain, developing and improving QA systems are desirable. Accordingly, this thesis work investigates and presents sophisticated methods for performing question answering in the biomedical domain.

In the following section~\ref{Chapter3_2}, we present the main classification criteria of question answering systems. We also describe in section~\ref{Chapter3_3} important factors that distinguish restricted-domain QA from open-domain.

\section{Question Answering Systems Classification}
\label{Chapter3_2}

Several criteria for classifying available large number of QA systems have been proposed. For instance, the task of generating and extracting answers to natural language questions is related to the application domain \citep{molla2007question}. Questions asked by users may require general answers on open, general topic; others may need specific answers from a specific, restricted domain. The nature of a specific domain impacts the kinds of questions posed and answers that can be expected. Therefore, the application domain may be considered as a basis of classification of QA systems.

In addition to the application domain, QA is directly related to the types of addressed questions and answers. Based on the complexity of addressed questions and challenges relied in answer generation, \citet{moldovan2003performance} have classified QA systems into five classes:

\begin{itemize}
  \item Systems able to answer factoid questions.
  \item Systems employing simple reasoning techniques.
  \item Systems that extract and generate the answers from different resources.
  \item Systems able to answer questions based on previous interactions with the user.
  \item Systems capable to perform an analogical reasoning.
\end{itemize}

The first class of QA systems covers a large number of systems that deal with WH-questions (what, when, which, who, how, etc.). It groups approaches that extract answers as pieces of text from one or more documents. The second class of systems is a subclass of the first category where the extraction of answers requires logical inferences. In the third category of systems, the answer is scattered in several documents and therefore answer fusion is required. The fourth category of systems exploits previous interactions with the user in order to extract answers. Finally, the last type of systems that are considered more complex, has the ability to answer prediction questions. The answers of such questions are not explicitly stated in the documents. In such systems, the answers are generated using analogical reasoning techniques.

Although the type of input and output seems to be an important criterion for classifying QA systems, this approach does not take this criterion into consideration since it uses the same type of input (natural language questions) and the same type of output (text segments). In addition, not only the techniques and strategies used for question analysis and classification, documents, and answer extraction are not taken as criteria for classification, but also this classification does not explain the sources from which the answers are generated or extracted. In this context, some authors have proposed approaches that take advantage of the input types, output types and resources as criteria for classifying semantic QA systems that uses knowledge bases and ontologies to generate the answers. For example, \citet{lopez2011question} have surveyed a state of the art on semantic QA systems and classified them into categories based on four dimensions: (1) the input or type of questions they deal with, (2) the resources from which the answers are extracted (structured versus unstructured data), (3) the scope (restricted-domain versus open-domain), and (4) how they copes with the search environment problems. These categories include:

\begin{itemize}
  \item Natural language interfaces to structured data on databases.
  \item QA over semi-structured data (e.g., health records).
  \item QA over textual documents (unstructured data).
  \item QA over ontologies (structured semantic data).
\end{itemize}

\cite{athenikos2010biomedical} have proposed a classification approach for classifying semantic knowledge-based QA systems into three categories:

\begin{itemize}
  \item Semantics-based QA systems.
  \item Inference-based QA systems.
  \item Logic-based QA systems.
\end{itemize}

Semantics-based approaches take advantage of semantic metadata encoded in structured semantic resources such as ontologies and knowledge bases in order to produce answers to questions; inference-based approaches derive answers by employing extracted semantic relationships, and logic-based approaches exploit explicit logical forms and theorem proving techniques to generate and extract answers.
This classification remains ambiguous, however, as the three categories seem to be connected and may have many intersections with each other, which does not allow to effectively categorize and group the different QA systems. Based on this classification, the author further have reviewed and classified medical QA approaches and biological QA approaches.

\citet{mishra2016survey} have identified eight criteria in support of classifying QA approaches based on the literature surveyed. These criteria include:
\begin{itemize}
  \item Application domains for which QA systems are developed (open domain QA or restricted-domain QA).
  \item Types of questions asked by the users, e.g., factoid, list, confirmation questions, etc.
  \item Types of analyses performed on users' questions and source documents, e.g., morphological analysis, syntactical
analysis, semantic analysis, etc.
  \item Types of data consulted in data sources: structured data, semi-structured data, and structured data.
  \item Characteristics of data sources such as source size and language.
  \item Types of representations used for questions and their matching functions (e.g., algebraic model, probability models, etc.).
  \item Types of techniques used for retrieving answers, e.g., NLP techniques, information retrieval techniques, etc.
  \item Types of answers returned by QA systems (extracted text, snippets or other multimedia information, and generated answer).
\end{itemize}

In the following section we discuss the main factors that distinguish between open-domain and restricted-domain QA.

\section{Open-Domain vs. Restricted-Domain Question Answering}
\label{Chapter3_3}
Question answering may be open-domain and restricted-domain, also known as general-domain and closed-domain QA, respectively. Open-domain QA deals with questions from diverse arias, whereas restricted-domain QA answers questions under a specific field. QA has been initially presented in the open-domain, and more recently in restricted domains \citep{HIRSCHMAN_2001}. Open-domain QA research area has received much attention from the IR community, initiated by the QA Track in TREC evaluations, which takes place regularly every year since 1999 \citep{voorhees1999trec}. Since then, various systems have been developed and participated in TREC and other evaluation companies such as NTCIR \citep{kando2002overview} and CLEF \citep{Magnini_2005}.

As \citet{molla2007question} point out in their review on restricted-domains QA, there are several factors that distinguish QA in the restricted-domains from general, open-domain QA systems. These factors include: (1) the size of the data, (2) domain context, and (3) resources. The size of the text corpus available for open-domain QA tends to be quite large, which explains and justifies the use redundancy-based answer generation and extraction methods. In contrast to open-domain QA, in restricted-domain QA, the size of the data is relatively small
and varies from domain to domain, as well as redundancy-based techniques are likely to have a weaker effect and would not be useful for a specific domain with a small corpus size. Although terminological variations and synonyms create many challenges for text mining and natural language processing applications in restricted domain, QA is advantaged by the limited scope of questions that a domain-specific terminology and types of
questions provides. Indeed, the kinds of questions posed in a restricted domain especially by users who are experts in the domain are in general more complex and difficult than these asked by users in open domain. Experts of a specific domain usually use specific terminologies and pose domain-specific types of questions that require specific processing of questions and answers. Finally, another important distinction between open-domain and restricted-domain QA relies in the existence of specific resources such as tools, methods and domain specific information for restricted domains that can be exploited in the QA process.

\section{Summary of the Chapter}
\label{Chapter2_2.5}
In this chapter, we have defined a serie of tasks that are actively used and studied in both biomedical natural language processing and text mining. Even if these tasks seem to be unconnected, they are usually correctly combined together to establish and develop an analysis workflow to solve the issues that occur in the area of text mining.

We started this chapter by presenting an introduction to biomedical natural language processing and text mining. We next described in details NLP tasks which are usually used for shallow linguistics processing tasks, such as sentence tokenization, word tokenization, n-grams, part-of-speech tagging, stemming and lemmatization, as well as dependency parsing. We have also presented and described the most widely used NLP toolkits/platforms for the most common NLP tasks.
We then presented and defined a range of text mining application types, such as named entity recognition, text classification, information extraction, information retrieval, text classification, text clustering, text summarization, as well as question answering which is the scope of this thesis work. We finally described a broader classification framework involving QA systems based on different criteria, and important factors that distinguish restricted-domain QA from open-domain.

In the following chapter, we will present in details the state-of-the-art on question answering systems with a particular focus on the biomedical domain.


\chapter{State-of-the-Art in Biomedical Question Answering} 

\label{Chapter3} 
\setcounter{secnumdepth}{4}
\setcounter{minitocdepth}{2}
\minitoc

This chapter presents background information to contextualise the topics of interest of this thesis work. In section~\ref{Chapter3_1}, a brief introduction is provided to question answering and its typical pipeline. In section~\ref{Chapter3_4}, the main characteristics of question answering in the biomedical domain are shown. In section~\ref{Chapter3_5}, the main resources that can be exploited for QA in the biomedical domain are presented. In section~\ref{Chapter3_6}, we review the related work to question answering with a particular focus on the biomedical domain. In section~\ref{Chapter3_8}, a synthesis is presented. Finally, the evaluation metrics we use for the assessment of the proposed methods for question answering in the biomedical domain are described in section~\ref{Chapter3_7}.



\section{Introduction}
\label{Chapter3_1}

The search for short specific answers to questions written in natural language is one of the major challenges in the field of information retrieval. The question answering issue has been addressed by the artificial intelligence community in the literature since the 1960s. Early experiments in this direction implemented question answering systems based on knowledge bases written manually by experts in their own field of interest, such as BASEBALL \citep{Green_1961}, LUNAR \citep{Woods_1973} that operate in open domain. The BASEBALL system answers questions posed about the dates, locations or results of Baseball games played in the American league over one season. LUNAR which is one of the first sciences question answering system, was designed to assist the geologic analysis of stones returned by the Apollo mission.

Question answering (QA), unlike traditional IR aims at directly providing the short, precise answers to questions asked by inquirers in natural language, by employing complex IE and NLP techniques, instead of returning a large number of documents that are potentially relevant \citep{voorhees1999trec}. QA systems aim at directly producing and providing precise answers rather than entire documents to users questions, by automatically analyzing thousands of articles, ideally in less than a few seconds. Such systems, which can help users locate useful information quickly, need linguistically and semantically processing of both users questions and data sources in order to extract the relevant information. In particular, question answering differs from traditional IR in three main aspects: (1) information needs (i.e., queries) are expressed  using natural language instead of a set of keywords; (2) results depend directly to what has been specifically requested, be it a precise answer (e.g., a named entity) or a short summary (3) answers are generated based on the integration of unstructured and structured data, e.g., textual documents and knowledge bases, respectively. While the intended answer is usually a piece of text, in few cases, the format of the answer may also be a multimedia information.

While the early research works on QA in the domain of artificial intelligence (AI) dates far back to the 1960s, several investigations involving QA within IR and IE communities have been made by the introduction of the QA Track in Text REtrieval Conference (TREC\footnote{TREC: \url{http://trec.nist.gov/}}) evaluations in 1999 \citep{voorhees1999trec}. Since then, methods and tools have been developed for generating and extracting answers for the three types of natural language questions supported by TREC fora, including, factoid questions, list questions, and definitional questions. Most research works in the field of QA, as fostered by TREC and other similar evaluation fora such as NII-NACSIS Test Collection for IR systems (NTCIR\footnote{NTCIR: \url{http://research.nii.ac.jp/ntcir/index-en.html}}) \citep{kando2002overview} and Cross Language Evaluation Forum (CLEF\footnote{CLEF: \url{http://clef.isti.cnr.it/}}) \citep{Magnini_2005}, have so far been focused on open-domain QA.

Due to the continuous, enormous growth of textual data especially in the form of scientific articles in the biomedical domain, recently, the field of QA systems in the biomedical domain has witnessed a growing interest among researchers. Such systems can quickly and reliably assimilate relevant information from a multitude of biomedical textual resources. Furthermore, question answering is a text mining task directed towards aiding researchers and health care professionals in managing the exponential growth of textual information in the biomedical domain.

\begin{figure}[th]
\captionsetup{justification=justified}
\graphicspath{{Figures/}}
\centering
\includegraphics[width=16cm, height=11.5cm]{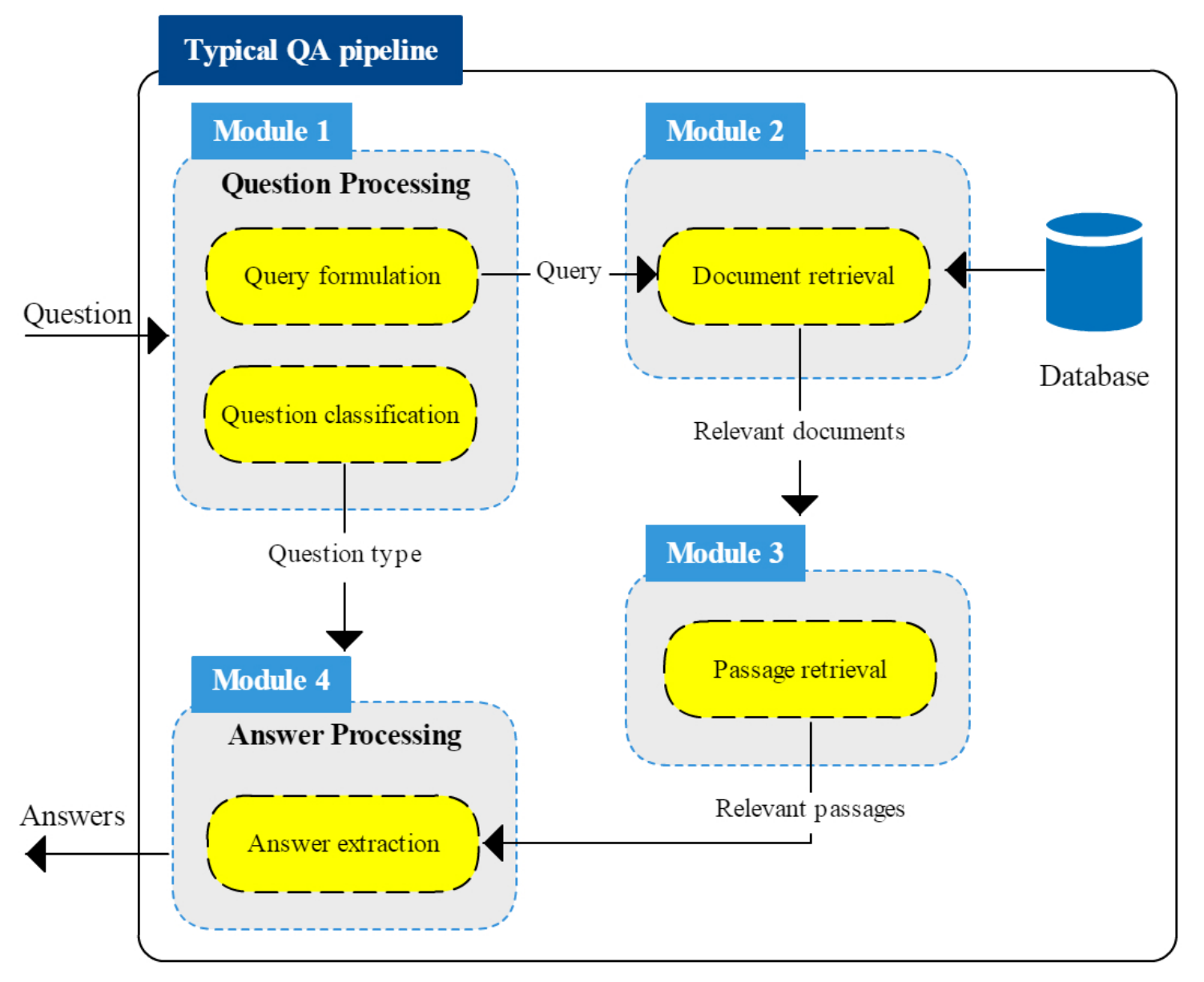}
\caption[Typical architecture of a question answering system]{Typical architecture of a question answering system.}
\label{fig:QAP}
\end{figure}

Natural language QA systems are typically composed of four main components including (1) question analysis and classification, (2) information or document retrieval, (3) passage retrieval and (4) answer extraction, as shown in Figure~\ref{fig:QAP} \citep{HIRSCHMAN_2001}.

The input to a QA system is a natural language question. Based on the linguistic and semantic  processing of the question, question analysis and classification identifies the type of question being posed to the system, and the type of the expected answer it should generate. Often, there may be more processes involved at this phase, such as named entity recognition and relation extraction. Output from this stage is one or more features of the question for use in subsequent stages. It then constructs a query from the input question to be fed into the third component, information retrieval. In the information retrieval component, the system inputs the query into a document retrieval engine so as to find a set of relevant documents satisfying the query. Out of the retrieved and selected documents, the passage retrieval component consists of retrieving relevant passages (text segments). The retrieved passages set may be narrowed down to a smaller set of most relevant passages/snippets of text, which constitute the input for the last component, answer extraction. Finally, in the answer extraction component, based on the type of the question and the expected answer format generated by the question analysis and classification component, the final answer(s) are extracted and generated from the candidate answers passages selected by the passage retrieval component. The output of the question answering system consists of the top ranked answer(s), associated with the raw text from which the answers were extracted.

The biomedical-domain QA is one of the more popular restricted-domains QA. The following section~\ref{Chapter3_4} provides the main characteristics that distinguish QA in the biomedical domain from open-domain QA.

\section{Characteristics of Question Answering in the Biomedical Domain}
\label{Chapter3_4}

Several characteristics distinguish the biomedical domain as an application domain for QA from general, open-domain and other restricted-domains QA. Based on the three factors discussed in the previous section, namely, the size of the data, domain-dependent context, and domain-specific resources, \citet{athenikos2010biomedical} point out in their review of biomedical QA the characteristic features of QA in the biomedical domain. These characteristics include:

\begin{enumerate}
  \item Large-sized corpora.
  \item Highly complex domain-specific terminology.
  \item Domain-specific lexical, terminological, and ontological resources.
  \item Tools and methods for exploiting the semantic information embedded in the above resources.
  \item Domain-specific format and typology of questions.
\end{enumerate}

First, biomedical QA is characterized and challenged by the large-sized of unstructured information especially in the form of scientific articles. Retrieving the accurate answers to potential questions from a multitude of biomedical sources tends to be quit complex. Second, the prominent use of highly complex domain-specific terminology is both an advantage and a challenge for biomedical QA. In general, the amount of terminological variations and synonyms make biomedical text mining relatively difficult. However, biomedical QA may take advantage of the particularity and limited scope of natural language questions that a domain-specific terminology gives. Third, domains that are specific, limited, and complex are most likely to have available resources such as methasaurus and ontologies that can be used in the QA process. These resources are generally developed to help biomedical domain users to retrieve information using text mining applications as well as categorize and group the biomedical knowledge. In addition to available resources, the tools and techniques required for employing the semantic information they contain enable for deep question and answer processing. Lastly, biomedical QA may benefit from the domain-specific formatting and typology of questions so as to use and develop answer processing techniques strategies for each specific question type.

Due to the importance and unique characteristics of QA in the biomedical domain, recently, several research works have been presented in the different stages of QA. These works which include the important methods and techniques are reviewed and discussed in the section~\ref{Chapter3_6}.

\section{Resources for Biomedical Question Answering}
\label{Chapter3_5}

While the biomedical domain poses a notable challenge for answering natural language questions, with highly complex domain-specific terminology and the enormous growth of textual information especially in the form of scientific articles and electronic health records, it also provides various resources that can be very beneficial for QA, as Zweigenbaum \citep{zweigenbaum2003question} also notes in his overview on QA in biomedicine. Here we review some of the well-known and relevant resources for QA in the biomedical domain.

\subsection{Corpora}

The primary resource for text-based QA in the biomedical domain is obviously text. In this context, MEDLINE was the first and still remains the primary resource for biomedical QA and in general for biomedical text mining. The MEDLINE\footnote{MEDLINE: \url{https://www.nlm.nih.gov/pubs/factsheets/medline.html}} database  maintained by the U.S. National Library of Medicine (NLM), and contains citations to journal articles in the life sciences with a particular focus on biomedicine. The 2016 MEDLINE contains over 24 million bibliographic references published from 1946 to the present in over 5,623 worldwide scientific journals. Indeed, hundreds of thousands of references are added to the database each year. For example, more than 806 thousand were added in 2016.

Perhaps more importantly, abstracts and sometimes full text of MEDLINE citations can be downloaded using the Entrez Programming Utilities\footnote{Entrez Programming Utilities (E-Utilities), Encyclopedia of Genetics, Genomics, Proteomics and Informatics: \url{https://doi.org/10.1007/978-1-4020-6754-9_5383}} for text mining purposes. For instance, PubMed\footnote{PubMed: \url{https://www.ncbi.nlm.nih.gov/pubmed/}}, a free service provided by the NLM under the U.S. National Institutes of Health (NIH) and accessible through the National Center for Biotechnology Information (NCBI), gives access to MEDLINE for searching abstracts of biomedical literature. The result of a PubMed search is a list of references (including authors, title, source, and often an abstract) to journal articles and an indication of free full-text availability. PubMed Central\footnote{PubMed Central: \url{https://www.ncbi.nlm.nih.gov/pmc/}} is a full-text archive of biomedical and life science articles, maintained by the NIH. Alternatively, subsets of MEDLINE documents can be obtained from the archives of individual research teams that share their annotated collections, and community-wide large-scale evaluations that use MEDLINE citations. For example, the OHSUMED\footnote{OHSUMED Test Collection. Available at: \url{http://trec.nist.gov/data/t9_filtering/}} collection contains all MEDLINE citations in 270 biomedical journals published over a five-year period (1987–1991) as well as a more recent collection provided in TREC Genomics\footnote{TREC genomics track data. Available at: \url{http://ir.ohsu.edu/genomics/data.html}} Track that contains ten years of MEDLINE documents (1994–2003).

Also available in the biomedical domain is topically-annotated collections of MEDLINE abstracts such as the GENIA corpus \citep{Kim_2003}, the earlier BioCreAtIve collections \citep{Hirschman_2005,Krallinger_2008,Islamaj_Dogan_2017}, and more recent set of the BioASQ challenge \citep{tsatsaronis2012bioasq}. The GENIA corpus which contains 1999 MEDLINE abstracts, is syntactically and semantically annotated for part-of-speech, coreference, biomedical concepts and events, cellular localization, and protein reactions. The BioCreAtIve data sets derived from the BioCreAtIve challenge which concerned with the extraction of biologically significant entities names and useful information (e.g., protein - functional term associations) from the literature. The BioASQ collections provided by the BioASQ challenges contains standard data sets for evaluating semantic indexing, question answering and information extraction systems in the biomedical domain. The BioASQ data sets are currently the most thoroughly annotated collection of MEDLINE abstracts and biomedical questions. The BioASQ challenges include biomedical text mining tasks relevant to text classification, information retrieval, question answering from texts and structured data, multi-document summarization and many other areas.

\subsection{Lexical, terminological, and ontological resources}

Due to the need of structuring highly complex domain specific terminology and of making it machine-understandable, text mining community in the biomedical domain has constructed and developed a set of lexical, terminological, and ontological resources. The Unified Medical Language System (UMLS\footnote{UMLS: \url{https://www.nlm.nih.gov/research/umls/}}) \citep{lindberg1993unified,Bodenreider_2004}, a repository of biomedical vocabularies that is maintained by NLM, is the most important resource. The UMLS includes three knowledge resources: the Mtathesaurus, the Semantic Network, and the SPECIALIST Lexicon. The UMLS Metathesaurus is a comprehensive, multilingual biomedical vocabulary thesaurus that contains information on biomedical concepts, their various names, and the hierarchical and transversal relations between them. The latest version of the Metathesaurus (UMLS 2017AB) contains approximately 3.64 million concepts and 13.9 million unique concept names from 201 different source vocabularies including dictionaries, terminologies, and ontologies. The Semantic Network contains information about the set of semantic types, or broad subject categories, and set of useful and important relationships, or semantic relations that may hold between these semantic types. The semantic types provides a consistent categorization of all concepts in the Metathesaurus by grouping these concepts according to the semantic types that have been assigned to them. The Semantic Network in the UMLS 2017AB version contains 127 semantic types and 54 semantic relationships. The SPECIALIST Lexicon provides the lexical, syntactic, and morphological information needed for the SPECIALIST NLP system. It contains commonly English words and biomedical vocabulary.

Medical Subject Headings (MeSH\footnote{MeSH: \url{https://www.nlm.nih.gov/mesh/}}), the controlled vocabulary thesaurus that is maintained and manually updated every year by NLM, consists of medical subject headings in a hierarchical tree which allows search at several levels of specificity. The latest version of the MeSH thesaurus (MeSH 2018) contains 28.939 MeSH descriptors, 116.909 total descriptor terms, and 244.154 supplementary concept records. Mesh is the widely used for the purpose of indexing journal citations for databases in the biomedical domain. For example, it is used by the U.S. NLM to index biomedical articles for the MEDLINE/PubMed database as well as for the cataloging of books and documents acquired by the library. Overall, the U.S. NLM provides over 297 knowledge sources\footnote{NLM resources: \url{https://eresources.nlm.nih.gov/nlm_eresources/}} and tools supporting biomedical text mining applications.

Systematized Nomenclature of Medicine Clinical Terms (SNOMED CT) \citep{stearns2001snomed}, the most comprehensive and precise clinical terminology available in the world that is originally maintained by the College of American Pathologists, aims at encoding the meanings that are used in clinical information and to support the efficient recording of clinical records. It was formed by the merger of two large health care reference terminologies, SNOMED Reference Terminology (SNOMED RT) and the U.K. National Health Service Clinical Terms, and it is accessible through the U.S. NLM and the National Cancer Institute (NCI). SNOMED CT is one of a suit of specified standards for use in U.S. healthcare information systems for the electronic exchange of clinical health information. The latest version of SNOMED CT released in January 2017 contains over 326.734 active concepts.

Gene Ontology\footnote{Gene Ontology: \url{http://www.geneontology.org/}} \citep{ashburner2000gene}, a controlled vocabularies of concepts used to define gene function, and semantic relationships between concepts, is an ontology designed to structure the description of genes and genes products common to all species. It classifies gene functions along three aspects: (1) molecular function, the molecular activities of a gene product, such as binding or catalysis; (2) cellular component, where a gene product is active; and (3) biological process, pathways and operations made up of the elemental activities of multiple gene products.

Other sets of interoperable ontologies in the biomedical domain are developed and maintained by collaborative effort in the Open Biological and Biomedical Ontology (OBO) Foundry\footnote{Open Biomedical Ontologies: \url{http://www.obofoundry.org/}} and the National Center for Biomedical Ontology (NCBO\footnote{National Center for Biomedical Ontology: \url{https://www.bioontology.org/}}). The set of NCBO ontologies are accessed through BioPortal\footnote{NCBO BioPortal: \url{http://bioportal.bioontology.org/}}. Other important centers that develop specialized resources for text mining applications in the biomedical domain include the U.K. National Centre for Text Mining (NaCTeM\footnote{NaCTeM: \url{http://www.nactem.ac.uk/}}) and the European Bioinformatics Institute (EMBL-EBI\footnote{EMBL-EBI: \url{https://www.ebi.ac.uk/}}).

\subsection{Supporting tools}

Besides the various available terminological, lexical and ontological resources in the biomedical domain that provide the lexical semantic information needed for QA and other text mining applications, the biomedical domain also has supporting tools that are specifically designed and developed to help exploit the semantic information contained in those aforementioned resources, such as biomedical entities and semantic relations, etc. Here, we define the most widely used tools to identify biomedical entities and semantic relationships among them in the text.

MetaMap \citep{aronson2001effective} that is based upon UMLS, is the most widely used tool by researchers working on biomedical text mining, including biomedical QA researchers for the named entity recognition task. MetaMap is highly configurable NLP application developed at NLM that maps terms in free biomedical text to UMLS Metathesaurus or, equivalently, identifies UMLS Metathesaurus concepts referred to in text. MetaMap provides a wide variety of configuration options to give researchers the opportunity to choose the best configuration for a given task. MetaMap, which is an open source tool and available for download\footnote{MetaMap: \url{https://metamap.nlm.nih.gov/}}, it requires a UMLS Terminology Services (UTS) account.

SemRep \citep{Rindflesch_2003} that is also relied on UMLS, is often used by researchers and practitioners to extract UMLS Semantic Network semantic relationships between UMLS concepts from sentences in biomedical text. SempRep\footnote{SemRep: \url{https://semrep.nlm.nih.gov/}} developed at the U.S. NLM to map sentences in free biomedical text to UMLS Semantic Network in order to extract three-part propositions, called semantic predications. Each predication consists of a subject argument, an object argument, and the semantic relation that binds them. The subject and object arguments of each predication are UMLS Metathesaurus concepts and their binding relationship is a UMLS Semantic Network semantic relationship.

\section{Related Work on Biomedical Question Answering}
\label{Chapter3_6}

Due to the unique characteristics of biomedical as an application domain for QA, recently researchers have proposed several specific methods and techniques in the different processing stages of QA system so as to produce more precise responses. As shown previously in section~\ref{Chapter3_1}, QA systems typically incorporate several components, including question classification, document retrieval, passage retrieval, and answer extraction each of which has to deal with specific challenges and issues. For each QA component, its basic challenges and approaches are outlined and recent and influential methods are reviewed in the following subsections.

\subsection{Question classification}
\label{Chapter3_6_1}
Question classification is usually the first component in QA pipeline as the first step towards developing biomedical QA systems is processing and classifying the question in order to identify the question type and therefore the answer type to produce. Several researchers in the biomedical domain have investigated question classification as a means of analyzing and filtering biomedical questions \citep{Simpson_2012}. In general, question classification may include several tasks since the questions may be classified along many dimensions: question type \citep{cruchet2008supervised,roberts2014automatically}, answer type \citep{McRoy_2016}, topic \citep{kobayashi2006representing, yu2008automatically,Cao_2010}, user \citep{Liu_2011,Roberts_2016}, answerability \citep{yu2005being}, resource \citep{roberts2016resource}, etc. Question classification essentially covers all these classification systems, and thus one could have multiple question classifiers to classify a single question along many dimensions. The most common classification in the biomedical domain is question type and answer type, sometimes referred to as the expected answer format and semantic answer type, respectively. For example, the question type and the semantic type of the expected answer for the biomedical question ``What is the treatment of acute myocarditis?'' are ``factoid'' and ``treatment'', respectively.

The purpose of question type classification in biomedical QA is to determine the type of question and therefore identify the expected answer format, to check whether the answer should be a biomedical entity name, a short summary, ``yes'' or ``no'', etc. Importantly, to produce the answer to a given question, the QA systems which deal with more than two types of questions, should know in advance the expected answer format that allows using specific answer extraction strategies. The identification of question types in biomedical QA systems is a very important task as it may strongly affect positively or negatively further processing stages: if the question type is not identified correctly, further QA processing stages will inevitably fail too. This task is usually carried out in open-domain QA systems by taking into account the ``Wh-'' particle and predefined patterns. However, the complexity of natural language creates many challenges to this task in the biomedical domain. As some questions that appear, at first, to belong to a certain type can result to be of a different one. For instance, the biomedical question ``Where is the protein CLIC1 localized?'', from the BioASQ training questions, is a factoid question. Although the way in which it is constructed, to start with ``where'', may lead it to be a summary question which is incorrect.

On the other hand, the answer type is the semantic type of the expected answer, and is useful in filtering out irrelevant answer candidates. The answer type has been called commonly topic or semantic type of the expected answer. The identification of answer type to a given question in biomedical QA, which is one of the most common task of question classification studied in the literature, has a critical impact on the overall performance of biomedical QA systems as is useful in choosing the best resource from which the answer should be extracted and candidate answer selection. For example, the question ``What is the best way to catch up on the diphtheria-pertussistetanus vaccine (DPT) after a lapse in the schedule?'' from National Library of Medicine's data set represents a pharmacological question, and the QA system may therefore identify the Micromedex pharmacological database as the resource to produce the answers.

The approaches for question classification may be classified in rule-based approach and machine learning-based approach. The former tries to
match the question with some predefined rules, while the latter can automatically classify questions into categories, by extracting some features from them. Many studies in open-domain QA have used rule-based approach \citep{Khoury_2011,Biswas_2014}. Otherwise, machine learning approaches have improved solution for open-domain question classification. One of the biggest advantages of machine learning is that one can focus on designing insightful features, and rely on learning process to efficiently cope with the features. For example, \citet{li2002learning} have presented a hierarchical classifier based on the sparse network of winnows for open-domain question classification. Two classifiers were involved in this work: the first one classifies questions into coarse categories; and the other one into fine categories. \citet{yu2005modified} have improved the bayesian model by applying the TFIDF measure to deal with the weight of words for Chinese question classification. \citet{xu2006syntactic} have employed affiliated ingredients as features of the learned model and used the results obtained by the syntactic analysis for extracting the question word. \citet{li2008classifying} have classified open-domain ``what'' type of questions into proper semantic categories using conditional random fields (CRFs). CRFs has been used to label all words in a question, and then choose the label of head noun as question category. \citet{Xu_2012} have introduced a question classification method based on the SVM classifier for QA system in the tourism domain. Other methodologies take benefit of both rule-based and machine-leaning approaches such as the work described in \citep{Hao_2017}.

Although many solutions have been proposed for question classification tasks in open-domain QA, only a few works have been completed in the biomedical domain which have attempted to define the information needs of physicians. For example, for searching the best available evidences supporting responses to clinical questions, the evidence-based medicine (EBM) paradigm recommends questions be organized according to the PICO (Problem or Patient/Population, Intervention, Comparison, Outcome) format \citep{richardson1995well}. In addition to the specific frame, taxonomies of biomedical questions in the EBM framework have also been developed.  \citet{Ely429} have proposed a taxonomy of generic types of clinical questions asked by doctors about patient care so as to help answer such questions. The taxonomy contains the 10 most frequent question topics (e.g., diagnosis, treatment, management, etc.) among 1396 collected clinical questions (e.g., ``What is the drug of choice for condition x?''). \citet{Bergus_2000} proposed a taxonomy of medical questions according to the PICO representation of questions and the categories of clinical tasks involved in the questions. \cite{Ely710} developed an evidence taxonomy to deal with obstacles be faced when attempting to answer physicians' questions with evidence. The authors have classified the collected 1101 questions from 103 family doctors into clinical vs. non-clinical. The clinical questions are divided into general vs. specific. The general questions can be classified into evidence vs. no evidence. The evidence questions are categorised into intervention vs. no intervention categories. \citet{jacquemart2003towards} have presented a taxonomy that consists of 8 broad semantic models (e.g., [which X]–(r)–[B]) for categorizing clinical questions in a French-language medical QA system. In this context, the recent taxonomy of biomedical questions which is created by the BioASQ challenge \citep{tsatsaronis2012bioasq} consists of four types of questions that may cover all kinds of potential questions: yes/no questions, factoid questions, list questions, and summary questions. \citet{seol2004scenario} identified four types of questions including treatment, diagnosis, etiology, and prognosis.
\citet{huang2006evaluation} presented a manual classification of clinical questions asked in natural language as a mean to examine the adequacy and suitability of the PICO framework. The authors have reaffirmed the usefulness of the PICO framework for structuring clinical questions, but also they found that it is less suitable for clinical questions that do not implicate therapy elements. \citet{niu2003answering,niu2004analysis,niu2005analysis,niu2006using} proposed a PICO-based question analysis approach within the EpoCare project. Their methods which aim at automatically answering questions from clinical evidence, extract potential answers using the PICO framework to structure both the question and answer texts. Similarly, \citet{Demner_Fushman_2006, Demner_Fushman_2007} used the PICO framework in their proposal approach to clinical QA. Their methods use the semantic unification of the PICO frame of the query and that of candidate answers. \citet{abacha2015means} developed a medical QA system named MEANS which consists of (1) corpora annotation, (2) question analysis and classification, and (3) answer search. The question classification task in the MEANS system aims at classifying the given medical questions into three types: definition, yes/no, and factoid so as to determine the type of the expected answers (e.g., ``yes'' or ``no'', biomedical entity name, etc.). Their method is based on a set of patterns that were manually constructed by analysing the 100 medical questions. \citet{yu2005classifying} performed machine learning approaches to classify medical questions based on  Ely et al.' s taxonomy into topics. They have shown that using the question taxonomy with the SVM classifier leads to the highest performance. \citet{cruchet2008supervised} described a question classification method to identify the answer types of questions in a bilingual French/English QA system which is adopted to health domain. The authors who studied a small number of English and French medical questions, used support vector machine (SVM) for classifying these questions into several categories (e.g. symptoms, prevention, evolution, treatment, etc.). \citet{Cao_2010} proposed a SVM classifier-based method to classify clinical questions asked by physicians in natural language into predefined general topics (e.g., diagnosis, management, pharmacological, treatment, etc.). The authors incorporated several features of clincal questions for the SVM classifier such as words, bigrams, stemming, UMLS concepts and semantic types. Since one question can be assigned to multiple topics, a binary SVM classifier was adopted for each of the 12 topics. \citet{Patrick_2012} proposed a question classification method for answering clinical questions applied to electronic patient notes. They first collected a set of clinical questions from staff in an Intensive Care Unit. They then designed a clinical question taxonomy for question and answering purposes. Finally, they built a multilayer classification model to classify the clinical questions. \citet{roberts2014automatically} introduced a multi-class SVM classifier-based method to automatically classify consumer health questions into semantic types (e.g., anatomy, cause, diagnosis, etc.) for the purpose of supporting automatic retrieval of medical answers from consumer health resources. More recently, deep learning models such as recurrent neural Network (RNN) and LSTM have been emerging as state-of-the-art for sequence modeling, particularly for text classification. Deep learning classification methods are multi-layer networks, which have been introduced for both feature extraction and classification tasks. The most well-known deep learning methods are the convolutional neural networks (CNN), which have known a great success since their introduction by \citep{lecun1989backpropagation}. Many deep learning models have been proposed such as RNN, recursive neural network and LSTM. Such models produce a more robust predictor than traditional machine learning methods. However, they  require a large number of training instances for the training phase.

Beyond these types of question classification in the biomedical domain, other approaches have studied (a) question answerability to separate answerable questions from unanswerable ones  \citep{yu2005classifying,yu2005being}, (b) resource identification to determine the resource type of biomedical questions \citep{roberts2016resource}, and (c)  relation extraction \citep{hristovski2015biomedical} using SemRep \citep{Rindflesch_2003}.

Although several question classification methods have been proposed in biomedical QA, question classification still requires further efforts in order to improve its performance. For instance, existing solutions for biomedical question classification have so far focused on extracting syntactic and semantic features from questions and using machine learning algorithms so as to classify questions into different topics. However, they do not take into account the syntactic dependency relations in questions. Intuitively, the incorporation of syntactically related pairs into other features may provide the best description and representation of questions. The motivation to find alternative features for machine-learning algorithms is the fact that words by themselves cannot capture the gist of a clinical question.

Another challenging issue in question classification is the identification of the types and formats of potential questions and intended answers, respectively. Note only current biomedical QA systems have limitations in terms of the types and formats of questions and answers that they can process, but also in most such systems which dealt with more than one type of questions, the users have to give or select manually the question type to each given question. As the ultimate goal of biomedical QA systems is to be able to deal with a variety of natural language questions and to generate appropriate natural language answers, biomedical question type classification is a necessary task needs so as to automatically identify the type of question and therefore to see whether the answer should be a biomedical entity name, a short summary, ``yes'' or ``no'', etc.

In the next chapter~\ref{Chapter4} we will present in details the proposed machine learning based methods for question classification in biomedical QA. The first method consists at identifying the type (i.e., yes/no, factoid, list and summary questions) of a given biomedical question in order to determine the expected answer format. It is based on our predefined set of handcrafted lexico-syntactic patterns and machine learning algorithms. The second method, which is based on lexical, syntactic and semantic features for machine learning algorithms, allows classifying questions into topics in order to filter out irrelevant answer candidates.

\subsection{Document retrieval}
\label{Chapter3_6_2}

Document retrieval is usually the second component in a typical QA pipeline as the second step towards answering a biomedical question posed in natural language is retrieving the set of textual documents that are likely to contain the answer. Retrieval of a set of relevant documents to a given query that is constructed from the question is usually carried out based on an existing IR system. The retrieved textual document set is often narrowed down to a smaller set of most relevant documents which constitutes the input for further QA processing steps. In particular, document retrieval is one of the significant components that serves as the building block of efforts since the correct answers can be only found when the set of textual documents from which the QA system extracts the answers is retrieved correctly \citep{Monz_2003,Collins_Thompson_2004,athenikos2010biomedical,neves2015question}.

While many efforts and investigations have been made in this direction in the open domain, initiated by the QA Track in TREC evaluations which takes place regularly every year since 1999 \citep{voorhees1999trec,roberts2002information,Monz_2003,gaizauskas2004information,Collins_Thompson_2004,voorhees2005trec,Teufel}, it still remains a real challenge in the biomedical domain due to a variety of reasons. The most basic obstacle is that there exists a vast amount of textual data especially in the form of scientific articles that is constantly and rapidly increasing as new scientific articles are published every day in the biomedical domain. Furthermore, not only these data are generally expressed in natural language, which makes its automated processing more difficult and complex, but also terminological variations and synonyms make document retrieval difficult in general for the biomedical domain. Thus, an efficient access to relevant information is a challenging task. Although biomedical text mining is challenged by a prominent use of domain-specific terminology, document retrieval may benefit from the specificity and limited scope of textual documents and the potential queries that a domain-specific terminology provides. Due to the specific characteristics of biomedical as an application domain for document retrieval, recently, most proposed systems in some way make use of domain-specific semantic knowledge, such as the MeSH thesaurus, for document retrieval. For example, the NLM indexers use the MeSH descriptors to index MEDLINE citations for PubMed search engine, a well-known information retrieval system in the biomedical domain which comprises more than 24 million citations for biomedical literature from MEDLINE, life science journals, and online books. \cite{lee2006beyond,Yu_2007} described a semantic-based approach for the development of a medical QA system named MedQA. The MedQA system is built upon four stages, namely, question analysis, information retrieval, answer extraction, and summarization techniques to automatically generate paragraph-level answers to definitional questions from the MEDLINE citations and World-Wide-Web. For information retrieval, the authors first applied LT CHUNK \citep{Mikheev_1996} to identify noun phrases from medical questions and used them as the query terms to retrieve relevant documents from the MEDLINE collection. They then used the tool LUCENE\footnote{LUCENE: \url{http://lucene.apache.org/core/}} \citep{goetz2000lucene} to index the MEDLINE documents and applied the vector-space model (VSM), a TFIDF based cosine similarity model for computing the relevance of each document to a query. To retrieve definitions from World-Wide-Web, the authors used Google and the TFIDF metric. \citet{Cao_2011} integrated the latest version of the probabilistic relevance model BM25 \citep{Robertson_2004} within their developed clinical QA system AskHERMES for the document retrieval stage, as it proved to be the best performing retrieval model for tasks such as those at the recent TREC. \cite{abacha2015means} applied NLP methods to process the source documents used to extract the answers to
natural language questions in their developed medical QA system named MEANS. The authors exploited NLP techniques, such as named entity recognition and relation extraction, and domain-specific semantic knowledge (e.g., UMLS Methasaurus and UMLS Semantic Network) to build RDF annotations of the source documents and SPARQL queries representing the users questions.

More recently, there has been a growing interest among researchers in developing new methods and techniques to improve the performance of document retrieval in biomedical QA  with the introduction of the biomedical QA Track at the BioASQ\footnote{BioASQ Challenge: \url{http://www.bioasq.org/}} challenge \citep{tsatsaronis2012bioasq} which takes place regularly since 2013. The BioASQ challenge is an EU-funded project aiming at fostering research and solutions on both the biomedical QA and large-scale online semantic indexing areas. The BioASQ challenge comprised three tasks: (1) Task a: on large-scale online biomedical semantic indexing, (2) Task b: on biomedical semantic QA, and (3) Task c: on funding information extraction from biomedical literature. The goal of Task b is to assess the performance of QA systems in different stages of the QA process. It is sub-divided into two phases: phase A and phase B. In phase A participants had to respond with biomedical concepts, relevant documents, relevant passages, and RDF triples. In phase B participants were asked to answer with exact answers and ideal answers (paragraph-sized summaries). \cite{weissenborn2013answering} developed a biomedical QA system which is composed of three main stages, namely, question analysis, document retrieval, and answer extraction to answer factoid questions. Documents relevant to potential questions are retrieved by making keyword queries to the GoPubMed search engine, which searches the MEDLINE biomedical citations database. Similarly, \cite{neves2014hpi} proposed a biomedical QA system which consists of two components, namely, question processing and document processing. Documents relevant to biomedical questions are retrieved by making queries to the GoPubMed search engine. Queries were constructed using the Stanford CoreNLP tools\footnote{Stanford CoreNLP package: \url{https://nlp.stanford.edu/software/corenlp.shtml}} for sentence splitting, tokenization, part-of-speech tagging and chunking. Indeed, queries was generated from the question based on both the terms and the chunks. The authors also performed query expansion using synonyms obtained using services from BioPortal. \cite{mao2014ncbi} used PubMed search functions for retrieving relevant documents to a given question in biomedical QA. Given a search query, the authors used PubMed results-ranking options such as by date or by relevance. \cite{choi2014classification} described a biomedical document retrieval approach based on semantic concept-enriched dependence model and sequential dependence model. The concept-enriched dependence model incorporates UMLS Methasaurus concepts identified using MetMap, while the sequential dependence model incorporates sequential query term dependence into the retrieval model. The authors used the Indri IR system \citep{strohman2005indri} and 2014 MEDLINE citations which composed of roughly 22 million citations.

Although previous document retrieval methods have proven to be quite successful at retrieving relevant documents in biomedical QA, document retrieval still require further efforts in order to improve its performance. One of the main observations that can be made about existing systems is that the task of document retrieval often set a framework in which an existing biomedical IR system is used, and completely depended on its ranking of documents. Indeed, there are many cases where the search engine mistakenly returns irrelevant citations high in the set or relevant citations low in the set. This problem is certainly a challenging issue as a biomedical QA system usually extracts the answers from the top-ranked documents.

In the chapter~\ref{Chapter5} we will present in details the proposed document retrieval method in biomedical QA.

\subsection{Passage retrieval}
\label{Chapter3_6_3}
Passage retrieval is usually the third component in a typical QA pipeline as the third step towards answering a biomedical question is processing the set of relevant documents so as to identify a set of text segments (also known as passages or snippets) that are likely to contain the answer. In this stage, which is known as passage or snippet retrieval, a set of passages are extracted from the retrieved and selected documents. In particular, a passage retrieval system may be defined as a specialized type of IR application that retrieves a set of passages/snippets rather than providing a whole ranked set of documents \citep{Buscaldi_2009}. Its main purpose in a QA system is to retrieve and return top-ranked passages which serve as answer candidates and the QA system extracts and selects the answers from them. Although a challenging task itself, passage retrieval remains one of the most important modules for the development of a QA system as the overall performance of a QA system heavily depends on the effectiveness of the integrated passage retrieval component: if a passage retrieval system fails to find any relevant passage for a given question, further processing steps to extract an answer will inevitably fail too. In this context, several studies such as the one reported in \citep{Otterbacher_2009}, highlighted that the correct answer to a given question can be found with high precision when it already exists in one of the retrieved passages.

Although passage retrieval in open-domain QA is a well-studied research area \citep{Clarke_2003,Otterbacher_2009,Ryu_2014,Saneifar_2014,Othman_2016}, it still remains a real challenge in biomedical QA. As described in section~\ref{Chapter3_4}, QA in the biomedical domain has its own characteristics such as the presence of large-sized corpora, complex technical terms, compound words, and domain-specific format and typology of questions. Due to the unique characteristics of biomedical as an application domain for passage retrieval in QA, recently, researchers have increasingly sought to incorporate and use lexical, terminological, and ontological
resources throughout their proposed methods. For example, \citet{Zhou_2007} examined the influence of incorporating deep semantic knowledge (e.g., information about concepts and relationships between concepts) in a biomedical IR system. The authors showed that appropriate use of domain specific knowledge yields about 23\% improvement over the best reported results in the Genomics Track of TREC 2006 of TREC \citep{voorheestrec}. \citet{Chen_2011} explored the hidden connection from MEDLINE documents based on a passage retrieval method. Their method first uses Mesh concepts retrieved from the sentence-level windows, and then ranks them by z-score, TFIDF (Term Frequency Inverse Document Frequency) and PMI (Pointwise Mutual Information). In the passage retrieval experiments, the authors showed that the TFIDF and PMI methods can achieve much better performance than those in the concept retrieval experiment. \citet{Cao_2011} built a clinical QA system named AskHERMES to perform robust semantic analysis on complex clinical questions. The AskHERMES system consists of three main components: question analysis, document/passage retrieval, and answer extraction. In the passage retrieval component, the authors first used the BM25 model for document retrieval, and then extracted relevant passages based on both word-level and word sequence-level similarity between a question and a sentence in the candidate answer passage.

With the introduction of biomedical QA Track at the BioASQ challenge \citep{tsatsaronis2012bioasq} in 2013, passage retrieval in biomedical QA has received much attention from NLP/IR researchers. \citet{lingeman2014umass} applied the sequential dependence model to consecutive text segments inside the document in order to create a ranking on the subdocument level. They used a granularity of 50 words, which are shifted through the document in increments of 25 words. \citet{yenalaiiith} proposed a biomedical passage retrieval method based on cosine similarity and existence test score between the question and each sentence in the retrieved documents. Their method first uses the PubMed search engine to find relevant documents and then ranks them based on cosine similarity and existence test score. Their system finally extracts the sentences from the abstracts of the 10 top relevant documents and keeps only the 10 top sentences matching most with the biomedical question after finding similarity scores for all sentences. \citet{ligeneric} presented a biomedical passage retrieval method to exactly locate the passages for the biomedical questions from the users. In their method, the expansion query based on word embedding and the sequential dependence model are used to retrieve relevant passages from the retrieved documents using. The authors first used predefined rules to identify the candidate passages and then applied the sequential dependence
model to rank them. \citet{yang2015learning} applied the LETOR model to score and rank the obtained candidate passages from the abstracts of the top-ranked documents retrieved by their document retrieval system. \cite{peng2015fudan} used statistical language model in order to retrieve relevant documents and then query keywords are extracted from the retrieved documents by giving extra credit to terms that appear close to the query keywords for passages retrieval. \cite{neves2015hpi} proposed a passage retrieval approach based on the in-memory database in biomedical QA. The system first extracts candidate passages from the relevant documents based on the built-in information retrieval features available in the IMDB, which uses approximated string similarity to match terms from the question to the words in the documents. Then, the system proceeds the ranked candidate passages and retrieves only the top-ranked passages using TFIDF metrics. \citet{Morid_2016} proposed a method for extracting clinically useful sentences from synthesized online clinical resources that represent the most clinically useful information for directly answering clinicians's information needs. The authors developed a Kernel-based Bayesian Network classification model based on different domain-specific feature types (e.g., UMLS concepts) extracted from sentences in a gold standard composed of 18 UpToDate documents.

Though the existing passage retrieval methods have proven to be quite successful at extracting passages in biomedical QA, passage retrieval still requires further efforts in order to improve its performance. The passage retrieval component in biomedical QA, compared to that of document retrieval can benefit even more from incorporation of domain-specific semantic knowledge.

In the chapter~\ref{Chapter6} we will present in details the proposed passage retrieval method in biomedical QA.

\subsection{Answer extraction}
\label{Chapter3_6_4}

Answer processing and extraction is the last component in a typical QA pipeline as the last step towards answering a biomedical question posed in natural language is extracting and generating the final answer and presenting it to the user posing the question. Answer extraction is considered the most challenging task of a QA system as this is when the precise answer has to be extracted from the candidates answers retrieved and selected by the passage retrieval component. The answer depends directly to the question type since extracting the answer for example to the factoid question (e.g., ``Which is the gene most commonly mutated in Tay-Sachs disease?''), which is asking for a biomedical named entity as an answer, is not the same as extracting the answer to a yes/no question (e.g., ``Is imatinib an antidepressant drug?''), which is looking for ``yes'' or ``no'' as an answer, for instance. In general, if a QA system deals with more than two types of questions, then the appropriate answer extraction technique, which extracts the final answer from the candidate answers, is selected and chosen according to the question type that was automatically identified by the question classification component. An answer extraction technique ranks the candidate answers according to the degree to which they match the expected answer type.

Considering both the various types of questions that may a QA system deals with and their different intended types of answers, answer extraction within the biomedical domain is a challenging task. In particular, most approaches to biomedical QA in some way make use of domain-specific semantic knowledge for answer extraction. In MedQA system \citep{lee2006beyond,Yu_2007} which dealt only with definitional questions (i.e., questions with the format of ``What is X?''), the answer extraction component which extracts definitional sentences from the retrieved documents relevant sentences, uses lexico-syntactic patterns and UMLS Methasaurus concepts (UMLS 2005AA). \cite{Cao_2011} described an answer presentation approach based on clustering technique in their developed clinical QA system AskHERMES which returns summaries as answers to all potential questions.
The authors grouped all relevant text passages into different topics based on clustering technique to locate relevant information
of interest before delving into more detail. Topics are assigned to each cluster in the AskHERMES system using content-bearing query terms and expanded terms from the UMLS Methasaurus. \cite{abacha2015means} presented an answer search methodology based on semantic search and query relaxation in their proposed semantic medical QA system, called MEANS. In the answer search phase, the authors first associated to each initial SPARQL query one or several less precise queries according to the number of expected answers. Then, the constructed SPARQL queries are executed in order to interrogate RDF triples generated on the document processing step. The authors exploited named entity recognition and relation extraction techniques, and domain-specific semantic knowledge, such as UMLS Methasaurus and UMLS Semantic Network, to build RDF annotations of the source documents and SPARQL queries representing the users questions.

With the introduction of biomedical QA Track at the BioASQ challenge \citep{tsatsaronis2012bioasq} in 2013, various answer extraction techniques have been recently presented for different types of questions.  The BioASQ challenge released benchmark datasets of biomedical questions in English, along with gold standard answers. There were four types of questions: yes/no, factoid, list, and summary questions. Yes/no questions, these are questions that require either ``yes'' or ``no'' answer. Factoid questions, these are questions that require a particular entity name (e.g., of a disease, drug, or gene), a number, or a similar short expression as an answer. List questions, these are questions that expect a list of entity names (e.g., a list of gene names, a list of drug names), numbers, or similar short expressions as an answer. Summary questions, these are questions that expect short summaries as answer. In phase B of Task b of the BioASQ challenge, participants were asked to answer with exact answers and ideal answers (paragraph-sized summaries). Exact answers are only required in the case of yes/no, factoid, list, while ideal answers are expected to be returned for questions. The released questions were accompanied by their types and the correct answers for the required elements (documents and passages) of the first phase. In this context, \cite{yang2015learning} described the development of a biomedical QA system named OAQA that returns only the exact answers for factoid and list questions based on learning-based approaches. The authors trained three supervised models and several features (e.g., lemma, the semantic type of each concept, etc.) using factoid and list questions of BioASQ training questions. The first model which is an answer type detection model aims at identifying the semantic answer type of a given question, the second assigns a score to each candidate answer while the third is a collective re-ranking model. Similarly, \cite{peng2015fudan} described a biomedical QA system that retrieves solely the exact answers for factoid and list questions. The system first used PubTator \citep{Wei_2013} for generating the candidate answers and then ranked them using term frequency metrics. \cite{choi2015snumedinfo} presented an answer extraction method based on keyword terms to generate the ideal answers for biomedical questions. In their method, candidate passages are ranked based on number of keywords and then combined to form the ideal answer. On the other hand, \cite{neves2015hpi} described answer extraction approaches for generating both exact and ideal answers to yes/no, factoid, list and summary questions from the gold-standard passages provided by the BioASQ challenge. Her approaches are notables for the use of the IMDB database and its built-in text analysis features. For yes/no questions, decision on either the answers ``yes'' or ``no'' was based on the sentiment analysis predictions provided by the IMDB. For factoid and list questions, the author extracted exact answers based on the annotations of noun phrases and topics provided by IMDB. For ideal answers, the author built summaries based on the the phrases which contain sentiments. In other work, \cite{schulze2016hpi} used an algorithm that is based on LexRank \citep{erkan2004lexrank} and named entities for the generation of summaries as answers to biomedical questions. The authors first built a sentence graph and then calculated cosine similarity of each sentence with each other sentences using named entity as dimension for the vector. They finally used a LexRank-based method to calculate the sentences ranking.

While these systems have proven to be quite successful at answering biomedical questions, they provide a limited amount of question and answer types, for instance, some of them \citep{Cao_2011} handle only with definition questions or returns solely short summaries as answers for all types of questions, and the most of the other ones do not deal with yes/no questions which are one of the most complicated question types to answer as they are seeking for a clear ``yes'' or ``no'' answer. In addition, the biomedical QA systems still requires further efforts in order to improve their performance in terms of precision to currently supported question and answer types. Furthermore, in spite of the importance of answering yes/no questions in the biomedical domain, we observed that only few studies have been presented compared to other types of questions, such as factoid and summary questions. Even though there are only two possible answers, ``yes'' or ``no,'' such questions can be quite hard to answer due to the complicated sentiment analysis process of the candidate answers.

In the chapter~\ref{Chapter7} we will present in details the proposed answer extraction methods in biomedical QA.

\subsection{Integral biomedical question answering systems}
\label{Chapter3_6_5}
While many efforts have been made in the biomedical QA area, only few integral QA systems, which can automatically retrieve answers to biomedical natural language questions, have been presented up to now. In this section, we survey and discuss the main integral biomedical QA systems. However, techniques and methods used in different stages of such systems have been detailed in the previous sections.

In this context, \cite{lee2006beyond,Yu_2007} designed, implemented, and evaluated a medical definitional QA system (MedQA) which is composed of five components including (1) question classification, (2) query generation, (3) document retrieval, (4) answer extraction, and (5) text summarization. In MedQA, at first, the question classification component automatically classifies medical questions into categories of the taxonomy created by \citep{Ely429} based on supervised machine-learning approaches for which specific answer search is developed. Next, the document retrieval component uses the query terms to retrieve relevant documents from either the Web documents using Google or the locally-indexed MEDLINE corpora using both Lucene to index the MEDLINE collection and the vector-space model for computing the relevance of a document to a query. Then, the answer extraction component identifies from the retrieved documents relevant sentences that answer the questions based on lexico-syntactic patterns. Finally, the text summarization component removes the redundant sentences and condenses the sentences into a coherent summary which is considered as answer. Although the MedQA system returns short summaries that could potentially answer medical questions, current MedQA's capacity is limited: it only provides answers to definitional questions.

\cite{cruchet2009trust} built a biomedical QA system called HONQA which extracts sentences from Health On the Net Foundation (HON) certified websites and provides them as answers for biomedical questions. HONQA is based on a learning approach for identifying the question type and semantic resources such as UMLS to guide the system, particularly in the choice of answers, but no details are presented in the publication. In its current form, it is not able to provide exact answers to other question types, for instance,  yes/no and factoid questions \citep{Bauer_2012}.

\cite{gobeill2009question} developed a biomedical QA system, EAGLi, which aims at extracting answers to biomedical questions from MEDLINE documents. Given a natural language question, EAGLi first analyzes the question in order to find the question type and to build the query based on a set of patterns. Then, it retrieves a set of relevant documents from MEDLINE using either PubMed or EasyIR, a local search engine in MEDLINE. Finally, the system extracts and computes a score for each of concepts expressed in the most relevant documents, and finally outputs a ranked list of candidate answers.  The current EAGLi's capacity is limited to Wh-type questions since it only covers the definitional and factoid questions.

\cite{Cao_2011} described a clinical QA system named AskHERMES that returns short summaries as answers of ad-hoc clinical questions expressed in natural language. AskHERMES was developed through the main following steps: (1) question analysis, (2) document retrieval, (3) passage retrieval, and (4) summarization and answer presentation. In the question analysis step, the authors have first classified clinical questions into general topics (e.g., device, diagnosis, epidemiology, etc.) based on SVM classifier to facilitate information retrieval and then identified keywords that capture the most important content of the question using conditional random fields. In the document retrieval step, the BM25 model was used to retrieve relevant documents. After that, they have extracted candidate passages based on dynamically generates passage boundaries and scored them based on both word-level and word sequence-level similarity in the passage retrieval step. Finally, the answer was generated based on structural clustering using content-bearing terms. The AskHERMES system returns passages (short texts) that could potentially answer all types of clinical questions. However, it returns a large number of results, which tends to defeat the intent of a QA system in reducing the amount of information that must be read. Moreover, the system supports only a single answer type in form of multiple sentence passages for all questions types \citep{Bauer_2012}.

\cite{abacha2015means} developed a semantic medical QA system called MEANS based on NLP techniques to process medical natural language questions and documents used to find answers, and semantic Web technologies at both representation and interrogation levels. MEANS is composed of three main phases: (1) corpora annotation, (2) question analysis and classification, and (3) answer search. The authors have applied NLP methods, named entity recognition and relation extraction so as to build RDF annotations of the source documents and SPARQL queries representing the users questions. They further have defined the MESA ontology  to represent the concepts and relations between them in order to construct SPARQL translations of natural language questions. To extract answers, the SPARQL queries were executed in order to interrogate RDF triples constructed in the corpus-annotation step. Even so, the authors have dealt with four questions types, they have focused on factoid and yes/no questions since more specific processes are still required to deal with complex questions (e.g. why, when).

\cite{hristovski2015biomedical} introduced a biomedical QA system, SemBT, based on semantic relations extracted from the biomedical literature. SemBT consists of three main processing steps: (1) preprocessing, (2) question processing, and (3) answer processing. During the preprocessing step, the authors first have extracted semantic relations using the SemRep natural language processing system from sentences retrieved from MEDLINE citations, and then stored them in a database. In the question processing step, the authors have constructed a query for searching in the database of the extracted semantic relations.  Finally, in the answer processing phase, they have presented the resulting semantic relations as answers in a top-down fashion, first semantic relations with aggregated occurrence frequency, then particular sentences from which the semantic relations are extracted. The SemBT system returns answers in the form of semantic relations and particular sentences from which the semantic relations are extracted. However, in its current implementation, the questions must be in the form Subject-Relation-Object, and hence, it does not allow asking questions in natural language format, for example, the natural language question ``What drugs can be used to treat diabetes?'' can be asked in SemBT as ``phsu treats diabetes'' where ``phsu'' stands for ``pharmacological substance'' and ``treats'' is the name of the semantic relation \citep{hristovski2015biomedical}.

More recently, \cite{Kraus_2017} developed the Olelo system for intuitive exploration of
biomedical literature. The Olelo system consists of three main modules: (1) question processing, (2) document/passage retrieval, and (3) answer processing. In Olelo, the question processing module is based on a system described in \citep{schulze2016hpi}. In the second module, Olelo first uses the tokens and the matched terms to formulate a query to the database so as to retrieve abstracts of the retrieved documents and then ranks the obtained abstracts according to the occurrence and importance of the searched tokens. Finally, an answer is returned to the user depending on the type of the question by the third module. Although Olelo has proven to be quite successful at answering biomedical questions, currently, Olelo supports only three question types including factoid questions, list questions and summary questions. Indeed, it does not support yes/no questions which are one of the most complicated questions to answer as they are seeking for a clear ``yes'' or ``no'' answer.

In this thesis work, our goal is to go beyond the previous biomedical QA systems and develop a QA system with the ability to automatically handle with a large amount of question types including yes/no questions, factoid questions, list questions and summary questions that are commonly asked in the biomedical domain \citep{tsatsaronis2012bioasq}. We propose a fully automated system SemBioNLQA - Semantic Biomedical Natural Language Question Answering - which has the ability to handle the kinds of yes/no questions, factoid questions, list questions and summary questions that are commonly asked in the biomedical domain. SemBioNLQA is derived from our previously established methods in (1) question classification, (2) document retrieval, (3) passage retrieval, and (4) answer extraction systems. We develop the SemBioNLQA system based on the integration of these methods and techniques. These proposed methods are presented in details in the following chapters. Table~\ref{tab:3.1c} summarizes the dimensions and characteristics of the aforementioned systems and SemBioNLQA.

\begin{table}[h!]
\centering
\caption[Question answering systems comparison matrix of features between the aforementioned systems and our proposed system SemBioNLQA]{Question answering systems comparison matrix of features between the aforementioned systems and our proposed system SemBioNLQA. The ``-'' indicates that the system did not include target question types classification.}
\label{tab:3.1c}
\begin{tabular}{M{4.1cm}M{4cm}M{2.9cm}M{3.5cm}}
\hline\noalign{\smallskip}
QA systems &Question format&Question types&Answer types \\

\noalign{\smallskip}\hline\noalign{\smallskip}
MedQA\newline \citep{lee2006beyond}&	Natural language& Definition & Summaries\\
\cmidrule(l){1-4}
HONQA\newline \citep{cruchet2009trust}&	Natural language&Definition,\newline factoid& Sentence \\
\cmidrule(l){1-4}
EAGLi\newline \citep{gobeill2009question}&	Natural language&Definition,\newline factoid& Multi-phrase passages and
a list of single Sentities \\
\cmidrule(l){1-4}
AskHERMES\newline \citep{Cao_2011} &	Natural language&-& Multiple sentence passages\\
\cmidrule(l){1-4}

MEANS\newline \citep{abacha2015means}&	Natural language	& Definition,\newline yes/no,\newline factoid&
Sentence, named entity, ``yes'' or ``no''  \\
\cmidrule(l){1-4}

SemBT\newline \citep{hristovski2015biomedical}&Subject-Relation-Object&-	& Semantic relations, sentences  \\
\cmidrule(l){1-4}

Olelo\newline \citep{Kraus_2017}&	Natural language	& factoid,\newline list,\newline summary& MeSH term, list of Mesh terms, short summaries \\

\cmidrule(l){1-4}

SemBioNLQA &	Natural language	& Yes/no,\newline factoid,\newline list,\newline summary& ``Yes'' or ``no'',\newline UMLS entity,\newline list of UMLS entity, short summaries \\
\noalign{\smallskip}\hline
\end{tabular}
\end{table}

Compared with the aforementioned systems, SemBioNLQA is aimed to be able to accept a variety of natural language questions and to generate appropriate natural language answers by providing both exact and ideal answers. It provides exact answers ``yes'' or ``no'' for yes/no questions, biomedical named entities for factoid questions, and a list of biomedical named entities for list questions. In addition to exact answers for yes/no, factoid and list questions, SemBioNLQA also returns ideal answers, while it retrieves only ideal answers for summary questions.

\section{Synthesis and Positioning}
\label{Chapter3_8}

Although several methods have been proposed in biomedical QA over recent years, biomedical QA still requires further efforts in order to improve its performance. For instance, existing solutions for biomedical question classification have so far focused on extracting syntactic and semantic features from questions and using machine learning algorithms so as to classify questions into different topics. However, they do not take into account the syntactic dependency relations in questions. Intuitively, the incorporation of syntactically related pairs into other features may provide the best description and representation of questions. The motivation to find alternative features for machine-learning algorithms is the fact that words by themselves cannot capture the gist of a clinical question.

Another challenging issue in question classification is the identification of the types and formats of potential questions and intended answers, respectively. The BioASQ taxonomy of biomedical questions consists of four types of questions including yes/no questions, factoid questions, list questions and summary questions that may cover all kinds of potential questions. Note only current biomedical QA systems have limitations in terms of the types and formats of questions and answers that they can process, but also in most such systems which dealt with more than one type of questions, the users have to give or select manually the question type to each given question. As the ultimate goal of biomedical QA systems is to be able to deal with a variety of natural language questions and to generate appropriate natural language answers, biomedical question type classification is a necessary task needs so as to automatically identify the type of question and therefore to see whether the answer should be a biomedical entity name, a short summary, ``yes'' or ``no'', etc.

On the second level, one of the main observations that can be made about existing systems is that the task of document retrieval often set a framework in which an existing biomedical IR system is used, and completely depended on its ranking of documents. Indeed, there are many cases where the search engine mistakenly returns irrelevant citations high in the set or relevant citations low in the set. This problem is certainly a challenging issue as a biomedical QA system usually extracts the answers from the top-ranked documents. On the other hand, the passage retrieval stage, compared to that of document retrieval can benefit even more from incorporation of domain-specific semantic knowledge. Although the existing passage retrieval methods have proven to be quite successful at extracting passages in biomedical QA, passage retrieval still requires further efforts in order to improve its performance.

On the third level, although the importance of answering questions in the biomedical domain, until now there are only few integral systems such as the ones described in \citep{lee2006beyond,cruchet2009trust,gobeill2009question,Cao_2011,abacha2015means,Kraus_2017} that can retrieve answers to biomedical questions written in natural language. While these systems have proven to be quite successful at answering biomedical questions, they provide a limited amount of question and answer types, for instance, some of them \citep{lee2006beyond,cruchet2009trust,Cao_2011} handle only with definition questions or returns solely short summaries as answers for all types of questions, and the most of the other ones do not deal with yes/no questions which are one of the most complicated question types to answer as they are seeking for a clear ``yes'' or ``no'' answer. In addition, the biomedical QA systems still requires further efforts in order to improve their performance in terms of precision to currently supported question and answer types. Furthermore, in spite of the importance of answering yes/no questions in the biomedical domain, we observed that only few studies have been presented compared to other types of questions, such as factoid and summary questions. Even though there are only two possible answers, ``yes'' or ``no,'' such questions can be quite hard to answer due to the complicated sentiment analysis process of the candidate answers.

The biomedical QA approach we propose through this thesis work takes into account these aspects and implements innovative methods in question classification, document retrieval, passage retrieval and answer extraction components. We propose a fully automated biomedical QA system with the ability to handle the kinds of yes/no questions, factoid questions, list questions, and summary questions that are commonly asked in the biomedical domain through the use of NLP methods, machine-learning approaches, and UMLS Metathesaurus. The proposed system provides the exact answers (e.g., ``yes'', ``no'', a biomedical entity, etc.) and the ideal answers (i.e., paragraph-sized summaries of relevant information) for yes/no, factoid and list questions, while it retrieves only the ideal answers for summary questions.

In the next chapters (chapter~\ref{Chapter4}, chapter~\ref{Chapter5}, and chapter~\ref{Chapter6}), we will present in details:

\begin{itemize} 

\item The proposed machine learning based methods for question classification in biomedical QA. The first method consists at identifying the type (i.e., yes/no, factoid, list and summary questions) of a given biomedical question in order to determine the expected answer format. It is based on our predefined set of handcrafted lexico-syntactic patterns and machine learning algorithms. The second method, which is based on lexical, syntactic and semantic features for machine learning algorithms, allows classifying questions into topics in order to filter out irrelevant answer candidates.

\item The proposed document retrieval method which retrieves relevant citations to a given biomedical question from the MEDLINE database. The proposed method first builds the query by extracting biomedical concepts, then uses a specialized IR system that gives access to the MEDLINE database to retrieve relevant documents, and finally ranks them based on a semantic similarity. We have also proposed an alternative based on a probabilistic IR model and biomedical concepts to retrieve and extract a set of relevant passages (i.e., snippets) from the retrieved documents to given biomedical questions.

\item The proposed answer extraction methods for extracting natural language answers from passages that potentially containing answers through the use of semantic knowledge, NLP techniques and statistical techniques. The first answer extraction method, based on a sentiment lexicon, aims at generating the exact answers to yes/no questions. The second method makes uses a biomedical metathesaurus to provide the exact answers suited for factoid and list questions which require with respectively a biomedical entity and a list of them as answers. The third method, aiming at retrieving the ideal answers (i.e., short summaries of relevant information) to biomedical questions, is based on a probabilistic IR model and biomedical concepts.

\item The developed biomedical QA system named SemBioNLQA which is aimed to be able to accept a variety of natural language questions and to generate appropriate answers by providing both exact and ideal answers.  SemBioNLQA, which is fully automatic system, includes innovative methods previously proposed in question classification, document retrieval, passage retrieval and answer extraction components. It is derived from our previously established contributions in each of the aforementioned components.

\end{itemize}

\section{Evaluation of Biomedical Question Answering Systems}
\label{Chapter3_7}

We introduce in this section the experimental setup that is used in this thesis work for comparing different proposed methods in biomedical QA. The experimental setup is purposed to be as uniform and consistent across the different proposed methods of the thesis work reported here, so as to facilitate the interpretation and discussion of the experimental results. This experimental setup furthermore aims at making easier the reproducibility of our experiments and the comparison with our results, avoiding particular experimental setup decisions which might provide unfair advantages to our proposed methods. In subsection~\ref{Chapter3_7_1} and subsection~\ref{Chapter3_7_2} the evaluation datasets and evaluation measures will be introduced, respectively.

\subsection{Evaluation datasets}
\label{Chapter3_7_1}

Several fora such as BioASQ challenges, QA4MRE Alzheimer Disease, and TREC Genomics Track have been organized to promote research and benchmarking on biomedical QA from textual data.

The BioASQ challenge\footnote{The BioASQ challenge: \url{http://www.bioasq.org/}} \citep{tsatsaronis2012bioasq} which started in 2013, within 2017 edition \citep{Nentidis_2017}, comprised three tasks:
\begin{itemize}
  \item Task 5a on Large-Scale Online Biomedical Semantic Indexing
  \item Task 5b on Biomedical Semantic Question Answering
  \item Task c on Funding Information Extraction From Biomedical Literature
\end{itemize}

BioASQ which was part of a EU-funded project, was developed to boost state of the art performance for biomedical QA.  In Task 5b, BioASQ has provided the benchmark datasets which contain development and test biomedical questions, in English, along with golden standard answers. The challenge, within each edition, the organizers released a training set of biomedical questions-answers pairs and test sets of questions. There were four types of questions: yes/no questions, factoid questions, list questions, and summary questions. The benchmark datasets have been created by the BioASQ team of biomedical experts. The goal of Task 5b is to assess the performance of QA systems in different stages of the QA process. It is sub-divided into two phases: phase A and phase B. In phase A participants had to respond with biomedical concepts, relevant documents, relevant passages, and RDF triples. In phase B participants were asked to answer with exact answers and ideal answers (paragraph-sized summaries). Exact answers are only required in the case of yes/no, factoid, list, while ideal answers are expected to be returned for questions. The released questions were accompanied by their types and the correct answers for the required elements (documents and passages) of the first phase. Each batch of questions was released every two weeks and participants had 24 hours to submit results.

The QA4MRE (Question Answering for Machine Reading Evaluation) for biomedical text about Alzheimer's Disease took place in two editions (2012 \citep{morante2012machine} and 2013 \citep{morante2013machine}) and aimed to boost development of solutions in machine reading. Participant systems were asked to extract and generate the answers to a set of questions. A set of full text documents were provided along with ten questions, each of which had five candidate answers.  The test set consists of four reading tests where each test contains one document and ten questions related to that document as well as five possible answers per question. Questions are in the form of multiple choice, a particular type of factoid questions that are less complex in comparison to typical factoid questions, as participant systems can use the possible answers to query for candidate answers and do not need to find the final answer but just choose the most likely one.

The TREC Genomics challenge which is part of TREC, took place in 2006 \citep{voorheestrec} and 2007 \citep{voorheestrec2007} and aimed to foster development of solutions in passage retrieval as part of QA systems with focus on particular biomedical entity types, such as diseases, mutations, genes, proteins, and pathways. The challenge organizers have provided a set of 162.259 full text documents collected from about 49 journals related to genomics. Participants were asked to retrieve relevant passages to a given topic question from a collection of 162.259 documents. A set of 28 and 36 topic questions was constructed in 2006 and 2007, respectively. Evaluation for each participant system was carried out manually by TREC experts who were provided with the top scoring 1000 passages for each topic question.

Table~\ref{tab:3_7_1.1} shows different datasets provided by the BioASQ challenge, the QA4MRE challenge and the TREC Genomics challenge.

\begin{table}[h!]
\centering
\caption{Comparing biomedical QA datasets provided by the BioASQ challenge, TREC Genomics and QA4MRE Alzheimer Disease}
\label{tab:3_7_1.1}
\begin{tabular}{p{5cm}p{4.1cm}p{1.2cm}p{1.1cm}p{0.6cm}p{1.7cm}}
\hline\noalign{\smallskip}
Dataset &Questions (Train+Test)&Yes/No&Factoid&List&Summary\\

\noalign{\smallskip}\hline\noalign{\smallskip}
TREC 2006 Genomics&28&&&&\multicolumn{1}{c}{\ding{51}}\\
TREC 2007 Genomics&36&&&&\multicolumn{1}{c}{\ding{51}}\\
QA4MRE Alzheimer Disease&40&&\multicolumn{1}{c}{\ding{51}}& \multicolumn{1}{c}{\ding{51}}&\\
BioASQ 2013&29+282&\multicolumn{1}{c}{\ding{51}}&\multicolumn{1}{c}{\ding{51}}&\multicolumn{1}{c}{\ding{51}}&\multicolumn{1}{c}{\ding{51}}\\
BioASQ 2014&310+500&\multicolumn{1}{c}{\ding{51}}&\multicolumn{1}{c}{\ding{51}}&\multicolumn{1}{c}{\ding{51}}&\multicolumn{1}{c}{\ding{51}}\\
BioASQ 2015&810+500&\multicolumn{1}{c}{\ding{51}}&\multicolumn{1}{c}{\ding{51}}&\multicolumn{1}{c}{\ding{51}}&\multicolumn{1}{c}{\ding{51}}\\
BioASQ 2016&1307+500&\multicolumn{1}{c}{\ding{51}}&\multicolumn{1}{c}{\ding{51}}&\multicolumn{1}{c}{\ding{51}}&\multicolumn{1}{c}{\ding{51}}\\
BioASQ 2017&1799+500&\multicolumn{1}{c}{\ding{51}}&\multicolumn{1}{c}{\ding{51}}&\multicolumn{1}{c}{\ding{51}}&\multicolumn{1}{c}{\ding{51}}\\
\noalign{\smallskip}\hline
\end{tabular}
\end{table}

To evaluate the different proposed methods of the thesis work reported here, we have used the datasets provided by the BioASQ challenge since as shown in Table~\ref{tab:3_7_1.1} they include a variety of questions types and answers, and also a large number of training and test questions. For these reasons, recently the BioASQ datasets are the most widely used to test the effectiveness of different parts of biomedical QA systems \citep{balikas2014results,balikas2015results,krithara2016results,Nentidis_2017}. Table~\ref{tab:3_7_1.2} shows the BioASQ 3b training questions used in this thesis work. The dataset include 810 questions-answers where each question was assigned to one category (yes/no, factoid, list, or summary). Table~\ref{tab:3_7_1.3} shows the test sets used to evaluate and compare the proposed methods in biomedical QA with current state-of-the-art methods. Table~\ref{tab:3_7_1.4} presents some examples of biomedical questions and their categories from BioASQ training questions.

\begin{table}[!htbp]
\centering
\caption{The question types and the number of BioASQ training questions assigned to each category}
\label{tab:3_7_1.2}
\begin{tabular}{p{5.3cm}p{4.5cm}p{5.2cm}}
\hline\noalign{\smallskip}
Question type & \#Questions & Category percentage  \\
\noalign{\smallskip}\hline\noalign{\smallskip}
Yes/No	& 237 &	29.26\%\\
Factoid	& 192 &	23.70\% \\
List	& 213 & 26.30\% \\
Summary	& 168 &	20.74\% \\
\noalign{\smallskip}\hline
\end{tabular}
\end{table}

\begin{table}[h!]
\centering
\caption{Number of questions and their categories in test sets of biomedical questions provided in the 2015, 2016, and 2017 and BioASQ challenges}
\label{tab:3_7_1.3}
\begin{tabular}{p{4cm}p{1.6cm}p{1.3cm}p{1.3cm}p{0.8cm}p{2cm}p{2.3cm}}
\hline\noalign{\smallskip}
\multirow{2}{*}{BioASQ dataset} &\multirow{2}{*}{Batch}&  \multicolumn{4}{c}{Question type} &\multirow{2}{*}{\#Total}\\
\cmidrule(l){3-6} &&\#Yes/No	& \#Factoid& \#List& \#Summary&\\

\noalign{\smallskip}\hline\noalign{\smallskip}
\multirow{5}{*}{BioASQ Task 3b 2015}  & Batch 1&	33	&26 &22 &19& 100 \\
& Batch 2&16&32 &28 &24& 100 \\
& Batch 3&29&26 &17 &28& 100 \\
& Batch 4&25&29 &23 &20& 97 \\
& Batch 5&28&22 &24 &26& 100 \\
\cmidrule(l){1-7}
\multirow{5}{*}{BioASQ Task 4b 2016}  & Batch 1&	28	&39 &11 &22& 100 \\
& Batch 2&32&31 &21 &16& 100 \\
& Batch 3&25&26 &21 &28&  100\\
& Batch 4&21&31 &15 &33&  100\\
& Batch 5&27& 34& 20&16&  97\\
\cmidrule(l){1-7}
\multirow{5}{*}{BioASQ Task 5b 2017}  & Batch 1&	17	&25 &22 &36& 100 \\
& Batch 2&27&31 &15 &27& 100 \\
& Batch 3&31&26 &15 &28&  100\\
& Batch 4&29&33 &13 &25&  100\\
& Batch 5&26& 35& 22&17&  100\\
\noalign{\smallskip}\hline
\end{tabular}
\end{table}

\begin{table}[h!]
\centering
\caption{Question categories with some examples of biomedical questions collected from BioASQ training dataset}
\label{tab:3_7_1.4}
\begin{tabular}{M{2cm}M{13.4cm}}
\hline\noalign{\smallskip}
Category  & \hspace{0.5cm}Sample questions   \\
\noalign{\smallskip}\hline\noalign{\smallskip}
Yes/No &\begin{minipage}[t]{\linewidth}
 \begin{itemize} [nosep,nolistsep]
           \item Does SCRIB deregulation promote cancer?
           \item Is CADASIL syndrome a hereditary disease?
           \item Can PLN mutations lead to dilated cardiomyopathy?
         \end{itemize} \end{minipage}\\
         \cmidrule(l){1-2}
Factoid & \begin{minipage}[t]{\linewidth}\begin{itemize}[nosep,nolistsep]
            \item Which gene is involved in CADASIL?
            \item What is the inheritance pattern of Hunter disease or mu-copolysaccharidosis II?
            \item How many genes are in the gene signature screened by MammaPrint?
          \end{itemize}\end{minipage}\\
          \cmidrule(l){1-2}
List &\begin{minipage}[t]{\linewidth}\begin{itemize}[nosep,nolistsep]
        \item Which are the clinical characteristics of TSC?
        \item Which proteins induce inhibition of LINE-1 and Alu retrotransposition?
        \item What is being measured with an accelerometer in back pain patients
      \end{itemize}\end{minipage}\\
       \cmidrule(l){1-2}
Summary & \begin{minipage}[t]{\linewidth}\begin{itemize}[nosep,nolistsep]
            \item What is the treatment of acute pericarditis?
            \item How does trimetazidine affect intracellular kinase signal-ing in the heart?
            \item Why does the prodrug amifostine (ethyol) create hypoxia?
          \end{itemize}\end{minipage}
\\
\noalign{\smallskip}\hline
\end{tabular}
\end{table}

In particular, in a part of this thesis work (cf. section~\ref{Chapter4.3} of chapter~\ref{Chapter4}), we have used the publicly available data set\footnote{Set of clinical questions available at: \url{http://clinques.nlm.nih.gov/about.html}} of clinical questions maintained by the U.S. NLM in order to evaluate the effectiveness of the method we propose for question topic classification in biomedical QA. This benchmark collection of questions has been widely used in the literature to test and evaluate question topic classification methods such as in \citep{yu2008automatically,Cao_2010}. All these questions have been manually labelled and released by the U.S. NLM. This data set contains 4654 ad hoc clinical questions, collected from healthcare providers across the USA in four studies \citep{Ely_1999,elyjhon,ely1997lifelong,D_Alessandro_2003}. There are a total of 12 categories and each question is assigned to one or more categories. Some examples of clinical questions and their categories are shown in Table~\ref{tab:4.3.3}. The typology of those questions is illustrated in Table~\ref{tab:4.3.4}. Table~\ref{tab:4.3.5} presents the 12 categories and the number of the clinical questions assigned to each one. 3559 questions, 386 questions, 700 questions, four questions, five questions
are assigned to one category, two categories, three categories, four categories, and five categories, respectively.

\begin{table}[!h]
\centering
\caption{Examples of clinical questions and their topics maintained by the U.S. National Library of Medicine}
\label{tab:4.3.3}
\begin{tabular}{M{11cm}M{4.5cm}}
\hline\noalign{\smallskip}
Question & Topics \\
\noalign{\smallskip}\hline\noalign{\smallskip}
Mother is alcoholic and abuses tobacco. What are statistics regarding inheritance of tobacco abuse and relationship to social
situation?	& Epidemiology\\\cmidrule(l){1-2}

Does she have any underlying inflammation of her kidneys? Creatinine approximately 1.0, 3+ albumin on urinalysis, just over
500 milligrams protein/24 hrs, normal intravenous pyelogram. & Management \newline Diagnosis\\\cmidrule(l){1-2}

Coronary angioplasty and stent placed last week. Started on Ticlid, looks like she’s allergic to it. She’s supposed to be on
Ticlid one more week. Obviously we've got to stop it. Do they want her on something else or just stop it? & Management \newline Treatment \&
Prevention \newline Pharmacological\\

\noalign{\smallskip}\hline
\end{tabular}
\end{table}

\begin{table}[!h]
\centering
\caption[Typology of 4654 clinical questions and their representatives. The first column represents generic question proportions]{Typology of 4654 clinical questions and their representatives. The first column represents generic question proportions. The second column represents number of each question types and their percentages. Respectively, question examples are in the last column \citep{Cao_2010}}
\label{tab:4.3.4}
\begin{tabular}{M{3cm}M{4.3cm}M{7.8cm}}
\hline\noalign{\smallskip}
Question type &\#Question (percentage)& Question example \\
\noalign{\smallskip}\hline\noalign{\smallskip}
What& 2231 (48\%)& What is Endolimax nana and should you treat it?\\
Which& 62 (1\%)& Which dose of Premarin is green?\\
Why& 134 (3\%)& Why is she having pelvic pain?\\
How& 697 (15\%)& How do you inject the bicipital tendon?\\
Can& 187 (4\%) &Can Lorabid cause headaches?\\
Do& 320 (7\%) &Do we need to do a spinal tap to rule out meningitis?\\
Others& 1023 (22\%)& Is this respiratory plan for an extubated child okay?\\
\noalign{\smallskip}\hline
\end{tabular}
\end{table}

\begin{table}[!htbp]
\centering
\caption[The topics of 4654 clinical questions, the number of the clinical questions assigned and the percentage of the total questions]{The topics of 4654 clinical questions, the number of the clinical questions assigned and the percentage of the total questions \citep{Cao_2010}}
\label{tab:4.3.5}
\begin{tabular}{M{5.2cm}M{4.8cm}M{5cm}}
\hline\noalign{\smallskip}
Topics &\#Questions &Questions percentage\\
\noalign{\smallskip}\hline\noalign{\smallskip}
Device& 37& 0.8\%\\
Diagnosis& 994& 21.4\%\\
Epidemiology &104& 2.2\%\\
Etiology &173& 3.7\%\\
History& 42& 0.9\%\\
Management& 1403& 30.1\%\\
Pharmacological& 1594& 34.3\%\\
Physical Finding& 271& 5.8\%\\
Procedure &122& 2.6\%\\
Prognosis &53& 1.1\%\\
Test &746& 16.0\%\\
Treatment \& Prevention& 868& 18.7\%\\
Unspecified& 0& 0\%\\
\noalign{\smallskip}\hline
\end{tabular}
\end{table}

\subsection{Evaluation measures}
\label{Chapter3_7_2}

Several indicators have been used in this thesis work to evaluate the effectiveness of the proposed methods in biomedical QA. As shown previously, a typical QA system consists of four stages including question classification, document retrieval, passage retrieval and answer extraction, that can be studied and evaluated independently. Therefore, we present in this subsection the typical evaluation measures used for each stage.
\subsubsection{Evaluation measures for question classification}

The accuracy metric has been widely used in order to evaluate question type classification methods \cite{khoury2011question,li2002learning,xu2006syntactic,loni2011survey}. Accuracy, as it is defined in equation~\ref{eq:1qc}, is the number of correct made predictions divided by the total number of made predictions.
\begin{equation}\label{eq:1qc}
Accuracy= \frac {No.\; of \;{Correctly} \;{Classifed} \;{Questions}}{Total\; No.\;of\; Tested\; Questions}
\end{equation}

Additionally, precision, recall, and F1-measure are widely used to evaluate the effectiveness of question type classification methods at classifying questions into each of the predefined question types (multi-class classification). For each question type, precision, recall, and F1-measure are computed using the standard equation~\ref{eq:2qc}, equation~\ref{eq:3qc}, and equation~\ref{eq:4qc}, respectively. Precision is defined as the number of true positive over the number of true positive plus the number of false positive. Recall is defined as the number of true positive over the number of true positive plus the number of false negative. In other words, precision is the fraction of correct classification for a certain category, whereas recall is the fraction of instances of a category that were correctly classified. F1-measure also know as F1-score is defined as the harmonic mean of precision and recall.

\begin{equation}\label{eq:2qc}
Precision= \frac {True\; Positive}{True\; Positive+False\; Positive}
\end{equation}

\begin{equation}\label{eq:3qc}
Recall= \frac {True\; Positive}{True\; Positive+False\; Negative}
\end{equation}
\begin{equation}\label{eq:4qc}
{F1-measure}={F1-score}= \frac {2*Precision*Recall}{Precision+Recall}
\end{equation}

\subsubsection{Evaluation measures for document and passage retrieval}

As indicators of retrieval effectiveness, mean precision, mean recall, mean F1-measure and mean average precision (MAP) were used. Given a set of golden items (documents and passages), and a set of items (documents and passages) returned by document and passage retrieval system (for a particular biomedical question in our case), precision, recall, and F1-measure are calculated using the previously defined equation~\ref{eq:2qc}, equation~\ref{eq:3qc}, and equation~\ref{eq:4qc}, respectively, where true positives is the number of returned items that are also present in the golden set, false positives is the number of returned items that are not present in the golden set, and false negatives is the number of items of the golden set that were not returned by the system. Given a set of questions $q_1,q_2, \ldots, q_n$, mean precision, mean recall, and mean F1-measure of the document and passage retrieval system is obtained by averaging its precision, recall, and F1-measure, respectively, for all the questions \citep{Balikas_2013}.

As precision, recall, and F1-measure do not take into account the order of the items returned by IR system for each question,  it is common in IR
to compute MAP of the returned set of items \citep{Balikas_2013}. MAP is obtained by averaging the average precision (AP) of over a set of questions, defined in the following equation~\ref{eq:5ir}:

\begin{equation}\label{eq:5ir}
MAP= \sum_{i=1}^{n} AP_i
\end{equation}

where $AP_i$ is the average precision of the set returned for question $q_i$, as defined in equation~\ref{eq:6ir}, where $|L|$ is the number of items in the set, $|L_R|$ is the number of relevant items (documents or passages), $P(r)$ is the precision when the returned set is treated as containing only its first $r$ items, and $rel(r)$ equals 1 if the r-th item of the set is in the golden list (i.e., if the r-th item is relevant) and 0 otherwise returned for all the questions \citep{Balikas_2013}.

\begin{equation}\label{eq:6ir}
AP=  \frac {\sum_{r=1}^{|L|} P(r)*rel(r)}{|L_R|}
\end{equation}

\subsubsection{Evaluation measures for answer extraction}

The performance of the proposed answer extraction methods are evaluated using the evaluation measures described by the BioASQ challenge \citep{tsatsaronis2012bioasq}. In the case of yes/no questions, the exact answers had to be either ``yes'' or ``no''. Therefore, accuracy
is the most used evaluation metric to evaluate responses of yes/no questions. Let $n$ be, the number of yes/no questions, and $k$ the number of correctly answered yes/no questions, accuracy is computed using the following equation~\ref{eq:2ae}:

\begin{equation}\label{eq:2ae}
Accuracy= \frac {k}{n}
\end{equation}

For factoid questions, mean reciprocal rank (MRR) is the main measure used to evaluate the exact answers returned by the system. Assuming that there are $n$ factoid questions, MRR is computed using the equation~\ref{eq:3ae} where $r_i$ is the topmost position that contains the
golden entity name (or one of its synonyms) in the returned list of possible responses.

\begin{equation}\label{eq:3ae}
MRR= \frac {1}{n} * \sum_{i=1}^{n}\frac {1}{r_i}
\end{equation}

On the other hand, to evaluate the exact answers of list questions, the mean average precision, mean average recall, and mean average
F-measure metrics, which are computed by averaging precision, recall, and F1-measure over the list questions are used. These measures (i.e., precision, recall, and F1-measure) are computed using the standard equations defined in (\ref{eq:2qc}), (\ref{eq:3qc}) and (\ref{eq:4qc}), where true positives is the number of possible answers that are included both in the returned and the golden list; false positives is the number of possible answers that are mentioned in the returned, but not in the golden list; and false negatives is the number of possible answers that are included in the golden, but not in the returned list. Indeed, mean average F1-measure is the the official score used by the BioASQ challenge to rank the participant systems for list questions.

Finally, the ideal answers of questions (all questions types: yes/no, factoid, list and summary), were automatically evaluated using ROUGE-2 and ROUGE-SU4. Basically, ROUGE counts the overlap between an automatically constructed summary by the system and a set of golden summaries manually constructed by humans. More details of these evaluations metrics appear in \citep{Balikas_2013}.

\section{Summary of the Chapter}
\label{Chapter3_9}

In this chapter, we have presented a state of the art on QA systems which aim at answering natural language questions from textual documents.  We have started this chapter by presenting an introduction to question answering. We have described in details the generic architecture of QA systems. We then have presented specificities and characteristics of QA in the biomedical domain, respectively. We have also presented the main resources that can be exploited for QA in the biomedical domain. Next, we have reviewed current research efforts directed toward QA in the biomedical domain. We have described in details techniques and methods that have been proposed for each component of a biomedical QA system, including question classification, document retrieval, passage retrieval and answer extraction. Finally, we have presented a synthesis of the different methods, and described the metrics used for the evaluation of the methods we proposed in this Phd thesis.

The following chapters (chapter~\ref{Chapter4}, chapter~\ref{Chapter5}, and chapter~\ref{Chapter6}) are dedicated to the presentation of our methods for question classification, document retrieval, passage retrieval and answer extraction in biomedical QA. 

\chapter{Question Classification in Biomedical Question Answering} 

\label{Chapter4} 
\setcounter{secnumdepth}{4}
\minitoc
This chapter presents the methods we propose for question classification in biomedical QA, a key task that is studied and evaluated separately. Section~\ref{Chapter4.2} will be dedicated to our proposed method for question type classification in biomedical QA. We consecrate section~\ref{Chapter4.3} to our proposed question topic classification method in biomedical QA.

\section{Introduction}
\label{Chapter4.1}
As we have described previously in section~\ref{Chapter3_6_1}, question classification is usually the first component in QA pipeline as the first step towards developing biomedical QA systems is processing and classifying the question in order to identify the question type and therefore the answer type to produce. The need of biomedical QA systems to handle a variety of natural language questions and to extract appropriate natural language answers, has resulted in questions being classified along various dimensions, including question type, topic, user, resource, etc. The most common question classification in biomedical QA is question type, sometimes referred to as the expected answer format. The identification of the type of a given biomedical question type is useful in candidate answer extraction as it allows to a biomedical QA system to know in advance the expected answer format, and therefore to use the appropriate answer extraction technique. In other words, the identification of the expected answer format required by a natural language question is usually carried out based on the question type and its linguistic information. For instance, a question identified as yes/no requires one of the two answers ``yes'' or ``no''. In contrast, a factoid question expects a particular answer type, such as an entity name. In this context, the most recent taxonomy of biomedical questions that is created by the BioASQ challenge \citep{tsatsaronis2012bioasq} consists of four types of questions that may cover all kinds of potential questions:

\begin{enumerate}
  \item Yes/No questions: They require only one of the two possible answers: ``yes'' or ``no''. For example, ``Is calcium overload involved in the development of diabetic cardiomyopathy?'' is a yes/no question and the answer is ``yes''.
  \item Factoid questions: They require a particular entity name (e.g., of a disease, drug, or gene), a number, or a similar short expression as an answer. For example, ``Which enzyme is deficient in Krabbe disease?'' is a factoid question and the answer is a single entity name ``galactocerebrosidase''.
  \item List questions: They expect a list of entity names (e.g., a list of gene names,  list of drug names), numbers, or similar short expressions as an answer. For example, ``What are the effects of depleting protein km23–1 (DYNLRB1) in a cell?'' is a list question.
  \item Summary questions: They expect a summary or short passage in return. For example, the expected answer format for the question ``What is the function of the viral KP4 protein?'' should be short text summary.
\end{enumerate}

With a view to developing an automatic biomedical QA system, the types of biomedical questions should be automatically identified by the system. The task of a question type classifier is to assign a class label, a category name which specifies the type of question, to a given natural language question. This task is usually carried out by checking whether it contains auxiliary verbs or Wh-question particles.  However, the complexity of the natural language poses many issues and challenges to this task in the biomedical domain, as some question that appear, at first, to belong to certain type can result to be of a different one. For instance, the biomedical question ``Where is the protein CLIC1 localized?'', from the BioASQ training questions, is a factoid question. Although the way in which it is constructed, by starting with ``where'', may lead it to the wrong classification of it being a summary question.

Another important dimension of classifying questions in biomedical QA is topic, sometimes referred to as the semantic type of the expected answer, as previously discussed in section~\ref{Chapter3_6_1}. Question topic classification refers to the process by which one or more topics are automatically derived from the given biomedical questions. The answer type is the semantic type of the expected answer, and is useful in generating specific answer retrieval strategies. For example, the question ``What is the best way to catch up on the diphtheria-pertussistetanus vaccine (DPT) after a lapse in the schedule?'' from NLM's data set represents a pharmacological question, and a biomedical QA system can therefore use the Micromedex pharmacological database as the resource to extract the answers. Recently, various approaches have addressed clinical question topic classification. For instance, \citep{yu2008automatically} and \citep{Cao_2010} proposed a clinical question topic classification method based on the combination of words, stemming, bigrams, UMLS concepts and UMLS semantic types as features for machine-learning algorithms to automatically classify an ad hoc clinical question into general topics. However, the existing methods have not taken into account the syntactic dependency relations in questions. Therefore, this may impact negatively the performance of the clinical question topic classification system.

In this context, we propose in the first contribution of this thesis work two machine learning-based methods for question classification in biomedical QA. The first method aims at classifying biomedical questions into one of the four categories: (1) yes/no, (2) factoid, (3) list, and (4) summary. The second method automatically identifies general topics from ad hoc clinical questions.  In the remaining of the chapter, we go into details about proposed question classification methods in biomedical QA.

\section{Question Type Classification}
\label{Chapter4.2}

In this section we describe the machine learning-based method we propose for biomedical question type classification \citep{kdir15,Sarrouti_MIM_2017}. As stated previously, the purpose of this method is to classify biomedical questions into one of the four categories: yes/no, factoid, list and summary  questions.

\subsection{Method}
\label{Chapter4.2.1}
\begin{figure}[th]
\captionsetup{justification=justified}
\graphicspath{{Figures/}}
\centering
\includegraphics[width=16cm, height=12cm]{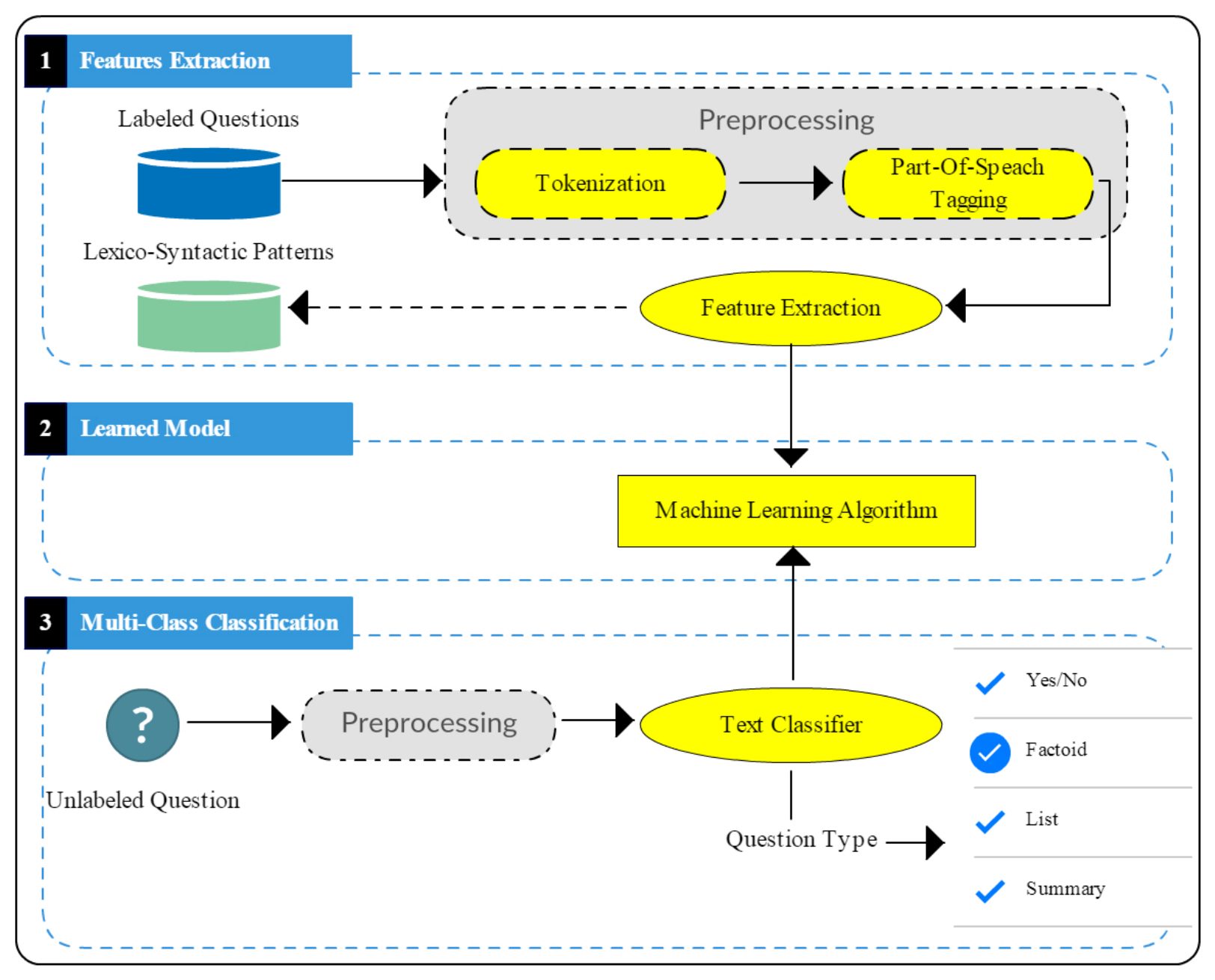}
\caption[Overview of the proposed biomedical question type classification method]{Overview of the proposed biomedical question type classification method.}
\label{fig:qc}
\end{figure}

With a view to achieving the goal of classifying the biomedical questions into the aforementioned categories, we first extract the appropriate features from BioASQ questions using our handcrafted lexico-syntactic patterns and then feed them to a machine-learning algorithm to conduct the classification task. The flowchart of the proposed method, as shown in Figure~\ref{fig:qc}, is constructed through the following main steps: (1) features extraction from biomedical questions using the set of our handcrafted lexico-syntactic patterns, (2) learned classifiers, and (3) predicting the class label using the trained classifiers.

\subsubsection{Machine-learning models}
\label{Chapter4.2.1.1}
We have experimented with several machine-learning algorithms including SVM, Naive Bayes, and Decision Tree. Since a question can be assigned to one of 4 classes (i.e., yes/no, factoid, list, and summary), a multi-class classification has been used in this study. We have found that SVM outperforms the others, followed by Naive Bayes. Therefore, we have decided to report the results of SVM. Mallet\footnote{Mallet: \url{http://mallet.cs.umass.edu/index.php}.} and libSVM\footnote{LibSVM: \url{https://www.csie.ntu.edu.tw/~cjlin/libsvm/}.} are the freely available platforms for machine-learning algorithms that have been used in this work. We have used Mallet for Naive Bayes and Decision Tree. For SVM, we have used LibSVM.
\subsubsection{Machine-learning features}
\label{Chapter4.2.1.2}

We explored different features for machine-learning systems including words, bigrams, part-of-speech (POS) tags and features obtained by our predefined handcrafted lexico-syntactic patterns.
We first have experimented with bag-of-words also known as unigrams (terms in the questions are extracted, all tokens which match stop words list\footnote{Stop words list: \url{http://www.textfixer.com/resources/common-english-words.txt}} were removed). Then, we have generated N-grams, proximity-based sequences of words obtained by sliding a window of size N over the biomedical question, and use them as features. In this work, the bigrams terms, i.e., N-grams with $N=2$, are used. We also experimented with part-of-speech tags as features for machine-learning algorithms. We have used the Stanford CoreNLP tools for tokenization and POS tagging, as research shows a good performance of the latter in the biomedical domain \citep{lin2007syntactic}. Finally, we
have used features that have been provided by our predefined handcrafted lexico-syntactic patterns. The patterns were derived by analyzing 810 BioASQ training questions. They were found by passing the questions one by one to the Stanford CoreNLP for tokenization and POS tagging so as to capture their syntactic structure. Indeed, we have extracted words and their POS tags from each of BioASQ training questions. We have found after analysing all these questions that the ones belonging to a particular class have some typical structure. In addition, we found that adding
WordNet synonyms led to an enhanced performance of question classification; we therefore have used WordNet to generate the synonyms of some words in some patterns (e.g., see patterns for factoid questions).

As already noted, in order to extract the best feature set, our predefined patterns were used. The patterns are regular expressions that are represented as a text string. For a given biomedical question, the appropriate pattern is selected from the set of patterns as follow: after preprocessing (tokenization and POS tagging) the biomedical question using Stanford CoreNLP, the pattern is set to the left end of the biomedical question, and matching process starts. After a mismatch is found, the pattern is shifted one place right and a new matching process starts, and so on. Below are the set of patterns that have been used for the four categories: yes/no, factoid, list, and summary. The ``+'', ``|'', and ``*'' signs indicate the concatenation, OR, and any terms, respectively. Note also that NN, JJ, VBZ, VBP, and TAG indicate noun, adjective, verb 3rd person singular present, verb non-3rd person singular present, and the obtained par-of-speech tag of the word, respectively.

\begin{enumerate}
  \item \textbf{Yes/No patterns}: We have proposed the following expression for yes/no questions.
\begin{enumerate}
\item {[Be verbs $\mid$  Modal verbs $\mid$ Auxiliary verbs] + [*] +?};  where be verbs = \{am, is, are, been, being, was, were\},
modal verbs=  \{can, could, shall, should, will, would, may, might\}, and auxiliary verbs=  \{do, did, does, have, had, has\}
\end{enumerate}
\item \textbf{Factoid patterns}: We have defined the following patterns for questions which can belong to factoid category.
\begin{enumerate}

\item {[What $\mid$ Which] + [VBZ] + [*] + [X] + [*] +?}; where X = \{number, name, indication, value, frequency, prevalence, or WordNet synonyms of these words\}
\item {[What $\mid$ Which] + [NN] + [*] +?}
\item {[What $\mid$ Which] + [does $\mid$ do] + [*] + [stand for $\mid$ bind to] +? }
\item {[Which] + [TAG] + [*]+?}
\item {[Where] + [*] + [NN] + [*] + [VBZ]+ [located] +?}
\item {[When] + [TAG] + [*] +?}
\item {[Why] + [TAG] + [*] +?}
\item {[How] + [TAG] + [*] +?}
\end{enumerate}

\item \textbf{List patterns}:  We have provided the following patterns for questions which can belong to list category.
\begin{enumerate}

 \item {[What $\mid$ Which] + [VBP] + [*] + [X] + [*] +?; where X = \{numbers, names, indications, values, frequencies, prevalence, or WordNet synonyms of these words\}}
\item {[What $\mid$ Which] + [VBP] + [*]+ [NN] + [*] +?}
\item {[Which] + [TAG] + [*] +?;}
\item {[Where] + [NN] + [*] + [NN] + [VBP] + [used] +? }
\item {[When] + [TAG] + [*] +?}
\item {[How] + [TAG] + [*] +?}
\item {[Why] + [TAG] + [*] +?}
\end{enumerate}
\item \textbf{Summary patterns}: For the biomedical questions which can belong to summary category,  we have used the following patterns.
\begin{enumerate}
\item {[What $\mid$ Which] + [VBZ] + [.*] + [X]+ [.*] +?; where X = \{definition, role, aim, effect, influence, mechanism, treatment, or WordNet synonyms of these words\} }
\item {[What] + [VBZ] + [NN] + [*] +?}
\item {[What] + [does] + [NN] + [*] + [do] +?}
\item {[Define $\mid$ Explain $\mid$ WordNet synonyms] + [*] + [NN] +?}
\item {[Why] + [TAG] + [*] +?}
\item {[How] + [TAG] + [*] +?}
\end{enumerate}
\end{enumerate}
\par
The key idea behind using the features provided by our set of patterns, is that only some words (e.g., Wh-question particles) in the biomedical question commonly represent the question type. Consider for example the biomedical question ``What is the definition of autophagy?'' from BioASQ training questions. The features vector (v) of this question is simply the patterns (see the first pattern of summary questions) that represent the structure of question. This question can be represented as follows: v= {{(what, 1), (VBZ, 1), (definition, 1)} where the pair is in the form (feature, frequency). Table~\ref{tab:4.2.1.2.1} shows the different feature space of the previous question.

\begin{table}[!h]
\centering
\caption{The different feature spaces of the biomedical question ``What is the definition of
autophagy?''}
\label{tab:4.2.1.2.1}
\begin{tabular}{M{4.6cm}M{10.7cm}}
\hline\noalign{\smallskip}
Feature space & Features  \\
\noalign{\smallskip}\hline\noalign{\smallskip}
Unigram& 	{(What, 1) (is, 1) (the, 1) (definition, 1) (of, 1) (autophagy, 1) (?, 1)}\\
\cmidrule(l){1-2}
Bigram &	{(What-is, 1) (is-the, 1) (the-definition, 1) (definition-of, 1) (of-autophagy, 1) (autophagy-?, 1)}\\
\cmidrule(l){1-2}
Part-of-speech&	{(WP, 1) (VBZ, 1) (DT, 1) (NN, 1) (IN, 1) (NN, 1)}\\
\cmidrule(l){1-2}
Part-of-speech + unigram &	{{(What, 1) (is, 1) (the, 1) (definition, 1) (of, 1) (autophagy, 1) (?, 1)}, {(WP, 1) (VBZ, 1) (DT, 1) (NN, 1) (IN, 1) (NN, 1)}}\\
\cmidrule(l){1-2}
\textbf{Set of patterns}	& \textbf{{(what, 1) (VBZ, 1) (definition, 1)}}\\

\noalign{\smallskip}\hline
\end{tabular}
\end{table}

We denote that biomedical questions that do not match any defined patterns are presented with their unigrams and part-of-speech tags.

The motivation to find alternative features for machine learning algorithms is the fact that words by themselves cannot capture the structure of a biomedical question. Intuitively, using our handcrafted lexico-syntactic patterns can capture the syntactic view of a biomedical question better than the other methods.
\subsection{Experimental results}

To validate the efficiency of our biomedical question type classification system, several experiments have been carried out using
the BioASQ training dataset and five batches of testing datasets that have been previously described in section~\ref{Chapter3_7_1} of chapter~\ref{Chapter3}. We have used a training set of 810 questions-answers, where one of the four types of questions: yes/no, factoid, list, summary was assigned to each question, as shown in Table~\ref{tab:3_7_1.2}. We have also used five batches of testing datasets provided in BioASQ Task 3b. Each of testing datasets is approximately comprised of 100 biomedical questions, as shown in Table~\ref{tab:3_7_1.3}. As indicators of classification effectiveness accuracy, precision, recall and F1-measure defined in equation~\ref{eq:1qc}, equation~\ref{eq:2qc}, equation~\ref{eq:3qc}, equation~\ref{eq:4qc}, respectively,  were used (cf. section~\ref{Chapter3_7_2}, chapter~\ref{Chapter3}).

For a machine-learning algorithm, we have used the multi-class SVM classifier. As a matter of fact, the linear kernel has been used for the SVM classifier since it is often recommended for text classification. Thus, the results have shown that linear kernel outperforms the other kernels such as RBF kernel, tree kernel, and composite kernel. As we have trained the SVM with a linear kernel, we only have needed to optimize the C parameter. The best value of C parameter is 1.01 which was fixed after 5-fold cross-validation.

Additionally, we have explored different features, including unigrams, bigrams, part-of-speech tags, the combination of part-of-speech tags and unigrams, and our set of handcrafted lexico-syntactic patterns. Table~\ref{tab:4.2.1.2.4} shows the SVM results in five feature models for automatically classifying a biomedical question into a category in terms of accuracy. The results show that the best method was trained on features extracted by our predefined patterns, which led to an accuracy of 89.40\%.

\begin{table}[h!]
\centering
\caption[The obtained results using SVM on five batches of testing datasets to automatically assign a category to biomedical questions]{The obtained results using SVM on five batches of testing datasets to automatically assign a category to biomedical questions. We explored different features, including unigram, bigram, part-of-speech, part-of-speech + unigram, and our set of predefined patterns.}
\label{tab:4.2.1.2.4}
\begin{tabular}{M{4.4cm}M{7cm}M{3.6cm}}
\hline\noalign{\smallskip}
Data sets&	Feature models &Accuracy (\%) \\
\noalign{\smallskip}\hline\noalign{\smallskip}
\multirow{5}{*}{Five batches} & Unigram & 79.48\\[1.5pt]
             & Bigram &  	65.18\\[1.5pt]
             &Part-of-speech&	77.08\\[1.5pt]
             &Part-of-speech+unigram&	80.08\\[1.5pt]
             &\textbf{Set of patterns}&	\textbf{89.40}\\[1.5pt]
\noalign{\smallskip}\hline
\end{tabular}
\end{table}

Using unigrams as features, the overall accuracy was 79.48\%. The results are meant to be a benchmark. We have found that other features have an impact on the performance of biomedical question type classification. Bigrams decreased the performance to 65.18\% in terms of accuracy
(an absolute decrease of 14\% accuracy). When part-of-speech tags were used as features, the overall performance decreased to 77.08\% accuracy, although the decrease was not statistically significant (an absolute decrease of 2.4\% accuracy). The incorporation of part-of-speech tags and unigrams as features improved the performance to 80.08\% accuracy.

We then experimented with our set of handcrafted lexico-syntactic patterns so as to show how well a system can perform the classification task by combining our predefined patterns and a machine-learning algorithm. We found that using our patterns features' provider of SVM leads to the highest accuracy of 89.40\%. Table~\ref{tab:4.2.1.2.5} shows the detailed results for each question category in two feature models, i.e., unigrams and set of our handcrafted lexico-syntactic patterns, on five batches of testing datasets using the SVM classifier. In two feature models, the category ``summary'' has the lowest classification performance, and ``yes/no'' has the highest one.

\begin{table}[h!]
\centering
\caption[The detailed results for each question category in two feature models (unigram and set of patterns) by applying SVM classifier]{The detailed results for each question category in two feature models (unigram and set of patterns) by applying SVM classifier. P, R, F, A indicate precision, recall, f1-measure, and accuracy, respectively.}
\label{tab:4.2.1.2.5}
\begin{tabular}{p{2.6cm}p{4cm}p{2.1cm}p{1.1cm}p{1.1cm}p{1.1cm}p{1.3cm}}
\hline\noalign{\smallskip}
Data sets &	Feature models &Class &	P (\%)	&R (\%)	&F (\%)	&A (\%) \\
\noalign{\smallskip}\hline\noalign{\smallskip}
\multirow{9}{*}{Five Batches}& \multirow{4}{*}{\parbox{3cm}{Unigram:\\Baseline model}} & Yes/No	& 95.00&	98.59&	96.70&\multirow{5}{*}{79.48}\\
                        & &Factoid  &	76.36&	66.70&	70.96\\
                         && List&	72.81&	92.28&	81.31\\
                         && Summary&	72.02&	59.91&	64.80\\
 \cmidrule(l){2-6}
                        &\multirow{4}{*}{\parbox{4cm}{Set of patterns:\\Proposed method} }& Yes/No	& 95.55&	100.0&	97.41&\multirow{5}{*}{89.40}\\
                         & &Factoid&	92.92&	80.14&	85.91\\
                         & &List&	83.27&	94.22&	88.37\\
                         & &Summary&	84.02&	80.49&	81.89\\
\noalign{\smallskip}\hline
\end{tabular}
\end{table}

\subsection{Discussion}
Overall, as shown in Table~\ref{tab:4.2.1.2.4}, the best system for automatically assigning a category to a biomedical question was trained on features extracted by our handcrafted lexico-syntactic patterns, which led to an accuracy of 89.40\%. Our results also show that feature selection impacts the biomedical question type classification performance. Using bag-of-words as features, the performance for automatically assigning a category to a question was an accuracy of 79.48\%. Thus, this feature set was chosen as a competitive baseline since it is widely used as baseline system in many question classification studies \citep{li2002learning,li2006learning,Patrick_2012,roberts2014automatically}. Bigrams decreased the performance to 65.18\% accuracy and the decrease was statistically significant (p < 0.01, t-test). Using the bigrams feature set by themselves could not capture the structure of a biomedical question in the context of this study, since the diversity in each category increased when using this feature set. For instance, for the biomedical question ``Why are insulators necessary in gene therapy vectors?'', the bigram ``why are'' is noisy, as the two words are not syntactically connected. Such noisy bigrams have a significant impact on the performance of question classification. Part-of-speech tags, on the other hand, slightly decreased the performance to 77.08\% accuracy. However, the decline was not statistically meaningful. In contrast, the combination of unigrams and part-of-speech tags as features improved the accuracy slightly (the absolute increase of 0.34\% accuracy), although the increase was not statistically significant. Meanwhile, using features extracted by our handcrafted lexico-syntactic patterns for SVM achieved the highest accuracy of 89.40\%. It outperforms the baseline system with a large margin 9.92\% in terms of accuracy and the increase is statistically significant. Inconsistency in category assignment may be responsible for the relation between the training size and the category classification performance. Typically, there is a strong positive relation between the training size and the classification performance: the larger the training size is the better a classifier performs. The Pearson correlation coefficient between F1-measure and the number of training size of a category shows an R-value of 0.83 (strong positive
correlation), which means that high F1 measure goes with high number of training data of a category (which confirms our hypothesis). We can see from Table~\ref{tab:4.2.1.2.5} and Table~\ref{tab:3_7_1.2} that the best performing category, yes/no, has the largest number of question instances (237), and the worst
performing category summary has the least number of question instances (168). Besides, \cite{metzler2005analysis} noticed that the ambiguity of labeled data has an impact on the category assignment. Unluckily, we have found that some biomedical questions in BioASQ training dataset were ambiguous. For instance, the biomedical question ``Which are the mutational hotspots of the human KRAS oncogene?'' is labeled with ``summary'' while it is also labeled by ``list'' category. Another example is the question ``Which are the newly identified DNA nucleases that can be used to treat thalassemia?'' that is labeled with ``factoid'' while it is also labeled by “list”. Biomedical question type classification can improve or decrease the performance
of an automatic biomedical QA system, because the answers extraction is based on the expected answer format of the questions. For example, extracting the answer to the question ``Which enzyme is deficient in Krabbe disease?'', which is asking for a
biomedical entity name, is not the same as extracting the answer to ``Is calcium overload involved in the development of diabetic cardiomyopathy?'', which is looking for ``yes'' or ``no'' as an answer. So, the class that can be assigned to a biomedical question affects greatly all the other steps of the QA process and therefore it is of vital importance to assign it properly. A study presented by \cite{Moldovan_2003} showed that
more than 36\% of the errors in a QA system are directly due to the question classification. In addition, the biomedical question
classification system can also improve the performance of IR systems \citep{Cao_2010}, because the question category can be used to choose the search strategy when the question is reformed to a query over IR systems. This is the case of the question ``What is the
definition of autophagy?'' from BioASQ datasets, identifying that the question category is ``summary'', the searching template
for locating the answer can be for example ``autophagy is a ...'' or ``definition of autophagy is ...'', which are much better than
simply searching by question words.
\section{Question Topic Classification}

In this section, we propose a machine learning-based method for question topic classification in biomedical QA \citep{Sarrouti_IBRA_2017}. The aim of this method is to classify clinical questions into general topics defined by the U.S. NLM. These topics are: device, diagnosis, epidemiology, etiology, history, management, pharmacological, physical, finding, procedure, prognosis, test, treatment and prevention.

\label{Chapter4.3}
\subsection{Method}

With a view to achieving the goal of classifying natural language questions into general topics and improving the performance of question topic classification in biomedical QA, we propose to incorporate the syntactic dependency relations into other features used in \citep{yu2008automatically,Cao_2010} for machine-learning approaches. The incorporation of syntactic and semantic features including words, bigrams, stemming, UMLS Metathesaurus concepts and semantic types introduced by \cite{yu2008automatically} and \cite{Cao_2010} seem to be quite enough to represent the questions. However, this method doesn't give the expected results. Intuitively, the incorporation of syntactically related pairs into these features may provide the best description of questions. A syntactic dependency relation \citep{nastase2007using} is a pair of grammatically related words in a phrase: the main verbs in two connected clauses, a verb and each of its arguments, a noun and each of its modifiers. It describes the syntactic structure of a sentence by using a typed dependency to establish relationships among words in terms of head and dependents.

The flowchart of the proposed method, as shown in Figure~\ref{fig:qc2}, is constructed through the following main steps: (1) features extraction from biomedical questions, (2) learned classifiers, and (3) predicting the class label using the trained classifiers.
\begin{figure}[th]
\captionsetup{justification=justified}
\graphicspath{{Figures/}}
\centering
\includegraphics[width=16cm, height=12cm]{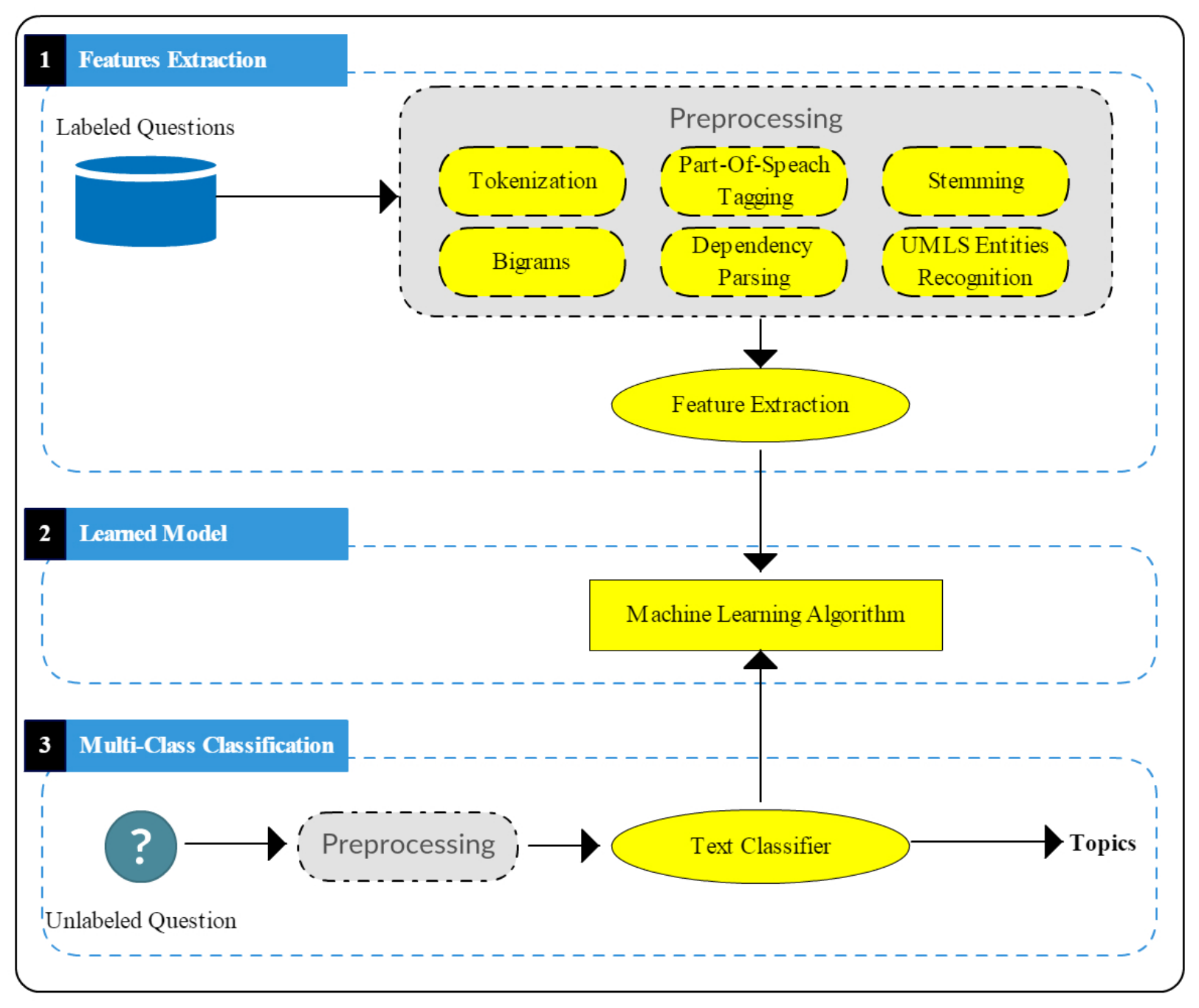}
\caption[Overview of the proposed biomedical question topic classification method]{Overview of the proposed biomedical question type classification method.}
\label{fig:qc2}
\end{figure}

\subsubsection{Machine-learning models}

In question classification, most of the existing methods have used SVM as it leads to the best results in comparison with other classifiers \citep{yu2005classifying,yu2008automatically,Cao_2010}. Accordingly, we have experimented with SVM and other machine-learning algorithms including Naive Bayes and Decision Tree. Since a question can be assigned to one or more general topics, a multi-label classification has been used in this study. We therefore developed a binary machine learning classifier for each of the predefined topics. We have found that SVM
outperforms the others, followed by Naïve Bayes. Therefore, we have decided to report and compare the results of both SVM and Naïve Bayes. Mallet and libSVM are the freely available platforms for machine-learning algorithms that have been used in this study. We have used Mallet for Naïve Bayes and Decision Tree. For SVM, we have used libSVM.

\subsubsection{Machine-learning features}
After preprocessing the clinical questions (terms in the questions were extracted and stop words were removed), we have explored bag-of-words as features. We then have applied \cite{Porter_1980} stemmer and \cite{Krovetz_1993} algorithms for term normalisation process and used them as additional features. Next, we have extracted the syntactic dependency relations between words in the question using the Stanford parser \citep{de2006generating}, and explored them as additional features. The overall process sequence is depicted in Table~\ref{tab:4.3.1}. After that, we have generated the N-grams, proximity-based sequences of words obtained by sliding a window of size N over the question, of each question and used them as additional features. In this work, the bigrams (N-grams with $N=2$) terms were used. Finally, we have mapped the terms in questions into the UMLS Metathesaurus in order to identify concepts and their semantic types, and explored them as additional features. To do so, we have used the MetaMap tool to identify appropriate UMLS Metathesaurus concepts and semantic types in questions. Table~\ref{tab:4.3.2} illustrates an example of mapping a sample clinical question to UMLS concepts and semantic types.

\begin{table}[!h]
\centering
\caption[Linguistic preprocessing of a sample clinical question]{Linguistic preprocessing of a sample clinical question, tokenisation and part-of-speech tagging is shown in ``Step 1'': ``WP'' corresponds to a Wh-pronoun, ``VBZ'' to a verb in the third person singular present, etc. The parse tree is shown in ``Step 2'' and includes the root of the tree (ROOT), the question phrase (WHNP), several noun phrases (NP), etc. The dependencies tree is shown in ``Step 3'': ``nsubj'' is the nominal subject relation, ``cop'' refers to the relation between a complement and the copular verb, etc.}
\label{tab:4.3.1}
\begin{tabular}{M{1.7cm}M{13.7cm}}
\hline\noalign{\smallskip}

Question&What is the dose of Zithromax for this 35-kilogram kid?\\
\cmidrule(l){1-2}

Step 1 & What/\textcolor{red}{WP} is/\textcolor{red}{VBZ} the/\textcolor{red}{DT} dose/\textcolor{red}{NN} of/\textcolor{red}{IN} Zithromax/\textcolor{red}{NNP} for/\textcolor{red}{IN} this/\textcolor{red}{DT} 35-kilogram/\textcolor{red}{JJ} kid/\textcolor{red}{NN} ?/\textcolor{red}{.}
\\
\cmidrule(l){1-2}
Step 2&
\resizebox{\linewidth}{!}{%
\Tree[.\textcolor{red}{ROOT}
  [.\textcolor{red}{.SBARQ}
    [.\textcolor{red}{WHNP} [.WP What ]]
    [.\textcolor{red}{SQ} [.VBZ is ]
      [.\textcolor{red}{NP}
        [.\textcolor{red}{NP} [.DT the ] [.NN dose ]]
        [.\textcolor{red}{PP} [.IN of ]
          [.\textcolor{red}{NP}
            [.\textcolor{red}{NP} [.NNP Zithromax ]]
            [.\textcolor{red}{PP} [.IN for ]
              [.\textcolor{red}{NP} [.DT this ] [.JJ 35-kilogram ] [.NN kid ]]]]]]]]]}
 \\
 \cmidrule(l){1-2}
 Step 3 &\textcolor{red}{root}(ROOT-0, What-1)
\textcolor{red}{cop}(What-1, is-2)
\textcolor{red}{det}(dose-4, the-3)
\textcolor{red}{nsubj}(What-1, dose-4)
\textcolor{red}{case}(Zithromax-6, of-5)
\textcolor{red}{nmod:of}(dose-4, Zithromax-6)
\textcolor{red}{case}(kid-10, for-7)
\textcolor{red}{det}(kid-10, this-8)
\textcolor{red}{amod}(kid-10, 35-kilogram-9)
\textcolor{red}{nmod:for}(Zithromax-6, kid-10)\\
\noalign{\smallskip}\hline
\end{tabular}
\end{table}





\begin{table}[!h]
\centering
\caption{Example of mapping the clinical question ``Mother is alcoholic and abuses tobacco. What are statistics regarding inheritance of tobacco abuse and relationship to social situation?'' to UMLS Metathesaurus concepts and semantic types. CUI and TUI indicate concept unique identifier and type unique identifier, respectively.}
\label{tab:4.3.2}
\begin{tabular}{M{4.1cm}M{2cm}M{6.4cm}M{2cm}}
\hline\noalign{\smallskip}
UMLS concepts & UMLS CUI & UMLS semantic types& UMLS TUI  \\
\noalign{\smallskip}\hline\noalign{\smallskip}
Mother (person)&C0026591& Family Group&T099\\
\cmidrule(l){1-4}
Tobacco&C0040329&Hazardous or Poisonous Substance\newline Organic Chemical&T131\newline T109\\
\cmidrule(l){1-4}
Tobacco Use Disorder&C0040336&Mental or Behavioral Dysfunction&T048\\\cmidrule(l){1-4}
Alcoholics&C0687725&Patient or Disabled Group&T101\\\cmidrule(l){1-4}
Statistics (publications)&C0600673&Intellectual Product&T170\\\cmidrule(l){1-4}
Drug abuse&C0013146&Mental or Behavioral Dysfunction&T048\\\cmidrule(l){1-4}
Concept Relationship&C1705630&Idea or Concept&T078\\\cmidrule(l){1-4}
Mode of inheritance&C1708511&Genetic Function&T045\\\cmidrule(l){1-4}
Social situation&C0748872&Social Behavior&T054\\
\noalign{\smallskip}\hline
\end{tabular}
\end{table}

\subsection{Experimental results}

\subsubsection{Results}
To validate the efficiency of the proposed method to question topic classification in biomedical QA, several experiments have been conducted using the set of 4654 clinical questions maintained by the U.S. NLM and previously described in section~\ref{Chapter3_7} of chapter~\ref{Chapter3}. As indicators of classification effectiveness, F1-score defined in equation~\ref{eq:1qc} is used (cf. section~\ref{Chapter3_7_2}, chapter~\ref{Chapter3}), where the recall is the number of correctly predicted clinical questions divided by the total number of annotated questions in the same category, and precision is the number of correctly predicted clinical questions divided by the total number of predicted questions in the same category. We randomly select negative data to repeat the classifications ten times. We then report the average F1-scores.

We have explored supervised machine-learning algorithms to automatically classify an ad hoc clinical question written in natural language into one or more topics predefined by NLM. We have used the freely available Mallet and libSVM for supervised machine learning systems. We have experimented with three machine-learning algorithms including Naïve Bayes, Decision Tree, and SVM. We have found that SVM outperforms other classifiers, followed by Naive Bayes. Therefore, we have decided to report the results of SVM and Naive Bayes.  Indeed, the linear kernel has been used for the SVM classifier since it is often recommended for text classification. Thus, the results have shown that linear kernel outperforms the other kernels such as RBF kernel, tree kernel, and composite kernel. On the other hand, we have also compared the performance of our proposed method with different common features for machine-learning algorithms, including bag-of-words, bag-of-stems, bag-of-bigrams, bag-of-syntactic dependency relations, bag-of-UMLS concepts, bag-of-UMLS semantic types and the combination of features used by \cite{yu2008automatically,Cao_2010}.

For evaluation, we have arranged that each classifier has a baseline of 50\%. In other words, each classifier is trained on the same number of positive and negative data. For example, when we trained a binary classifier of Pharmacological, we had 1594 questions that were assigned to this category (see Table~\ref{tab:4.3.5}). This set of 1594 questions represents the positive training data. To generate negative training data, we have randomly selected 1594 questions from the remaining categories.

Table~\ref{tab:4.3.6} and Table~\ref{tab:4.3.7} show the F1-score for each topic using Naïve Bayes and SVM classifiers, respectively. We have explored different combinations of features including bag-of-words (BOW), bag-of-bigrams (BOB), bag-of-stems (BOS) using both Porter and Krovetz stemmers, bag-of-biomedical named entities (BOBNE), bag-of-syntactic dependency relations (BOSDR), bag-of-UMLS concept and semantic types (BOCST), and our combination of features, the proposed method. The proposed combination of features consists of BOW, BOB, BOS$_{porter}$, BOSDR, and BOCST. The overall results show that the best system was trained on our combination of features, which led to F-scores of 77.18\% and 71.77\% using SVM and Naïve Bayes respectively. Table~\ref{tab:4.3.8} and Table~\ref{tab:4.3.9} show the comparison of the proposed method with other combination of features for classifying clinical questions into categories. Table~\ref{tab:4.3.8} shows the increasing performance of the proposed method in comparison with other representations using Naïve Bayes as a classifier, and Table~\ref{tab:4.3.9} shows the increasing performance using the SVM classifier. These results show that our method is more effective as compared with state-of-the-art methods and outperforms them by an average of 4.5\% F1-score using Naïve Bayes and 4.73\% F1-score using SVM.

Figure~\ref{fig:topics1} and Figure~\ref{fig:topics2} show the classification performance of topic assignment in terms of F1-score as a function of training size using Naive Bayes and SVM, respectively. In both systems, the topic pharmacology has the highest classification performance (79.53\% F1-score for Naive Bayes and 84.85\% F1-score for SVM), and history has the lowest classification performance (55.71\% F1-score for Naive Bayes and 68.75\% F1-score for SVM).

\begin{landscape} 
\centering
\begin{table}
\centering
\caption[The obtained results in terms of F1-score using Naïve Bayes to automatically assign topics to ad hoc clinical questions]{The obtained results in terms of F1-score using Naïve Bayes to automatically assign topics to ad hoc clinical questions. We explored different combinations of feature sets including bag-of-words (BOW), bag-of-bigrams (BOB), bag-of-stems (BOS) using both Porter and Krovetz stemmers, bag-of-biomedical named entities (BOBNE), BOW+bag-of-UMLS concept and semantic types (BOCST), BOW+bag-of-syntactic dependency relations (BOSDR), BOW+BOB+BOCTY+BOS$_{porter}$ used in \citep{yu2008automatically,Cao_2010}, BOW+BOB+BOCST+BOS$_{krovetz}$, and our proposed method, the combination of BOW+BOB+BOCST+BOSDR+BOS$_{porter}$.}
\label{tab:4.3.6}
\begin{tabular}{M{4.5cm}M{1.2cm}M{1.2cm}M{1.2cm}M{1.1cm}M{1.1cm}M{1.2cm}M{1.4cm}M{1.6cm}M{1.6cm}M{1.6cm}}
\hline\noalign{\smallskip}
\multirow{4}{*}{Topics}&\multicolumn{9}{c}{Features} \\\cmidrule(l){2-11}
&&&&&&&\multicolumn{2}{c}{\thead{\cite{yu2008automatically}\\ \cite{Cao_2010}}}&\multicolumn{2}{c}{\thead{Proposed method}}\\\cmidrule(l){8-9}\cmidrule(l){10-11}
 &\multirow{2}{*}{BOW}& \multicolumn{2}{c}{BOS}& \multirow{2}{*}{BOBNE} &\multirow{2}{*}{\thead{BOW+\\BOCST}}& \multirow{2}{*}{\thead{BOW+\\BOSDR}}& \multicolumn{2}{c}{\thead{BOW+BOB+BOCST+}}& \multicolumn{2}{c}{\thead{BOW+BOB+BOSDR+BOCST+}}\\
\cmidrule(l){3-4}\cmidrule(l){8-9}\cmidrule(l){10-11}
&  &Porter& Krovetz&&&&BOS$_P$& BOS$_K$&BOS$_P$& BOS$_K$    \\

\noalign{\smallskip}\hline\noalign{\smallskip}
Device&55.13\%&56.42\%&55.85\%&55.46\%&60.45\%&56.87\%&68.73\%&68.19\%&69.47\%&68.96\%\\
Diagnosis&69.38\%&70.43\%&68.53\%&67.45\%&71.15\%&71.54\%&72.20\%&71.17\%&73.98\%&72.08\%\\
Epidemiology&65.82\%&66.88\%&64.42\%&63.12\%&67.55\%&67.65\%&70.62\%&69.14\%&71.63\%&70.40\%\\
Etiology&71.22\%&73.03\%&72.00\%&67.05\%&74.08\%&72.60\%&75.38\%&74.17\%&76.54\%&75.51\%\\
History&46.40\%&47.72\%&47.07\%&44.57\%&52.34\%&48.35\%&54.16\%&53.50\%&55.71\%&55.02\%\\
Management&60.26\%&61.35\%&60.44\%&61.08\%&63.62\%&62.05\%&65.34\%&64.39\%&66.78\%&65.83\%\\
Pharmacological&74.86\%&75.35\%&74.96\%&71.75\%&77.26\%&76.67\%&78.45\%&77.80\%&79.53\%&78.89\%\\
Physical\&Finding&67.89\%&68.58\%&67.90\%&64.89\%&69.65\%&68.74\%&73.76\%&72.59\%&74.61\%&73.50\%\\
Procedure&67.36\%&69.03\%&67.80\%&66.43\%&71.06\%&69.01\%&72.38\%&71.13\%&74.04\%&72.77\%\\
Prognosis&69.44\%&70.42\%&69.24\%&64.10\%&71.94\%&70.46\%&72.46\%&70.89\%&73.52\%&72.07\%\\
Test&72.51\%&73.05\%&72.66\%&68.32\%&75.78\%&73.85\%&76.89\%&76.18\%&78.24\%&77.53\%\\
Treatment\&Prevention&63.30\%&64.67\%&63.92\%&59.11\%&65.82\%&64.25\%&66.31\%&64.69\%&67.16\%&65.53\%\\\cmidrule(l){1-11}
Average&65.30\%&66.41\%&65.40\%&62.78\%&68.39\%&66.84\%&70.56\%&69.49\%&71.77\%&70.67\%\\

\noalign{\smallskip}\hline
\end{tabular}
\end{table}

\end{landscape} 

\begin{landscape} 
\centering
\begin{table}
\centering
\caption[The obtained results in terms of F1-score using SVM to automatically assign topics to ad hoc clinical questions]{The obtained results in terms of F1-score using SVM to automatically assign topics to ad hoc clinical questions. We explored different combinations of feature sets including bag-of-words (BOW), bag-of-bigrams (BOB), bag-of-stems (BOS) using both Porter and Krovetz stemmers, bag-of-biomedical named entities (BOBNE), BOW+bag-of-UMLS concept and semantic types (BOCST), BOW+bag-of-syntactic dependency relations (BOSDR), BOW+BOB+BOCST+BOS$_{porter}$ used in \citep{yu2008automatically,Cao_2010}, BOW+BOB+BOCST+BOS$_{krovetz}$, and our proposed method, the combination of BOW+BOB+BOCST+BOSDR+BOS$_{porter}$.}
\label{tab:4.3.7}
\begin{tabular}{M{4.5cm}M{1.2cm}M{1.2cm}M{1.2cm}M{1.1cm}M{1.1cm}M{1.2cm}M{1.4cm}M{1.6cm}M{1.6cm}M{1.6cm}}
\hline\noalign{\smallskip}
 \multirow{4}{*}{Topics}&\multicolumn{9}{c}{Features} \\\cmidrule(l){2-11}
&&&&&&&\multicolumn{2}{c}{\thead{\cite{yu2008automatically}\\ \cite{Cao_2010}}}&\multicolumn{2}{c}{\thead{Proposed method}}\\\cmidrule(l){8-9}\cmidrule(l){10-11}
 &\multirow{2}{*}{BOW}& \multicolumn{2}{c}{BOS}& \multirow{2}{*}{BOBNE} &\multirow{2}{*}{\thead{BOW+\\BOCST}}& \multirow{2}{*}{\thead{BOW+\\BOSDR}}& \multicolumn{2}{c}{\thead{BOW+BOB+BOCST+}}& \multicolumn{2}{c}{\thead{BOW+BOB+BOSDR+BOCST+}}\\
\cmidrule(l){3-4}\cmidrule(l){8-9}\cmidrule(l){10-11}
&  &Porter& Krovetz&&&&BOS$_{porter}$& BOS$_{krovetz}$&BOS$_{porter}$& BOS$_{krovetz}$    \\

\noalign{\smallskip}\hline\noalign{\smallskip}
Device&57.89\%&58.23\%&57.69\%&56.31\%&65.60\%&60.10\%&74.01\%&73.50\%&74.99\%&74.44\%\\
Diagnosis&74.22\%&74.24\%&73.14\%&70.38\%&75.63\%&75.19\%&77.13\%&75.33\%&78.10\%&77.05\%\\
Epidemiology&71.15\%&70.58\%&68.10\%&54.96\%&72.03\%&71.97\%&74.74\%&72.31\%&75.93\%&73.53\%\\
Etiology&80.31\%&80.67\%&78.64\%&75.05\%&81.02\%&80.95\%&82.47\%&80.07\%&83.11\%&81.67\%\\
History&52.72\%&55.69\%&55.03\%&50.96\%&58.57\%&54.31\%&67.18\%&66.52\%&68.75\%&68.09\%\\
Management&69.70\%&69.48\%&68.51\%&65.07\%&70.13\%&70.02\%&71.07\%&70.16\%&71.49\%&70.10\%\\
Pharmacological&82.41\%&82.83\%&82.16\%&76.20\%&84.66\%&83.04\%&84.71\%&84.04\%&84.85\%&84.19\%\\
Physical\&Finding&72.10\%&72.27\%&71.08\%&70.62\%&76.09\%&73.14\%&78.82\%&77.01\%&79.35\%&78.38\%\\
Procedure&69.56\%&70.08\%&68.81\%&68.18\%&75.47\%&71.32\%&78.68\%&77.42\%&79.12\%&78.35\%\\
Prognosis&72.68\%&73.68\%&72.13\%&69.27\%&73.89\%&72.87\%&74.03\%&72.51\%&74.17\%&73.61\%\\
Test&79.97\%&80.14\%&76.40\%&75.02\%&81.15\%&80.52\%&83.22\%&82.48\%&83.64\%&81.90\%\\
Treatment\&Prevention&68.19\%&69.00\%&67.40\%&65.16\%&69.99\%&69.21\%&71.73\%&70.10\%&72.63\%&71.91\%\\\cmidrule(l){1-11}
Average&70.91\%&71.41\%&69.92\%&66.43\%&73.68\%&71.89\%&76.48\%&75.12\%&77.18\%&76.10\%\\

\noalign{\smallskip}\hline
\end{tabular}
\end{table}

\end{landscape} 

\begin{table}[h!]
\centering
\caption{Comparison between the proposed representation (the combination of various features: BOW+BOB+BOCST+BOSDR+BOS$_{porter}$) and state-of-the-art representations on 4654 natural language clinical questions using Naive Bayes as a classifier in terms of F-score.}
\label{tab:4.3.8}
\begin{tabular}{M{2cm}M{1.2cm}M{1.2cm}M{1.2cm}M{1.2cm}M{1.2cm}M{1.3cm}M{1.5cm}M{1.7cm}}
\hline\noalign{\smallskip}
\multirow{2}{*}{Features}& \multirow{2}{*}{BOW}& \multicolumn{2}{c}{BOS}& \multirow{2}{*}{BOBNE} &\multirow{2}{*}{\thead{BOW+\\BOCST}}& \multirow{2}{*}{\thead{BOW+\\BOSDR}}& \multicolumn{2}{c}{\thead{BOW+BOB+ \\ BOCST+}}\\
\cmidrule(l){3-4}\cmidrule(l){8-9}
&  &Porter& Krovetz&&&&BOS$_{porter}$& BOS$_{krovetz}$    \\

\noalign{\smallskip}\hline\noalign{\smallskip}
F-score&65.30\%&66.41\%&65.40\%&62.78\%&68.39\%&66.84\%&70.56\%&69.49\%\\
Increase\newline performance&+6.47\%&+5.36\%&+6.37\%&+8.99\%&+3.83\%&+4.93\%&+1.21\%&+2.28\%\\
\noalign{\smallskip}\hline
\end{tabular}
\end{table}

\begin{table}[h!]
\centering
\caption{Comparison between the proposed representation (the combination of various features: BOW+BOB+BOCST+BOSDR+BOS$_{porter}$) and state-of-the-art representations on 4654 natural language clinical questions using SVM as a classifier in terms of F-score.}
\label{tab:4.3.9}
\begin{tabular}{M{2cm}M{1.2cm}M{1.2cm}M{1.2cm}M{1.2cm}M{1.2cm}M{1.3cm}M{1.5cm}M{1.7cm}}
\hline\noalign{\smallskip}
\multirow{2}{*}{Features}& \multirow{2}{*}{BOW}& \multicolumn{2}{c}{BOS}& \multirow{2}{*}{BOBNE} &\multirow{2}{*}{\thead{BOW+\\BOCST}}& \multirow{2}{*}{\thead{BOW+\\BOSDR}}& \multicolumn{2}{c}{\thead{BOW+BOB+ \\ BOCST+}}\\
\cmidrule(l){3-4}\cmidrule(l){8-9}
&  &Porter& Krovetz&&&&BOS$_{porter}$& BOS$_{krovetz}$    \\
\noalign{\smallskip}\hline\noalign{\smallskip}
F-score&70.91\%&71.41\%&69.92\%&66.43\%&73.68\%&71.89\%&76.48\%&75.12\%\\
Increase\newline
performance&+6.26\%&+5.74\%&+7.26\%&+10.75\%&+3.5\%&+5.29\%&+0.7\%&+2.06\%\\
\noalign{\smallskip}\hline
\end{tabular}
\end{table}

\begin{figure}[h!]
\captionsetup{justification=justified}
\graphicspath{{Figures/}}
\centering
\includegraphics[width=15cm, height=9cm]{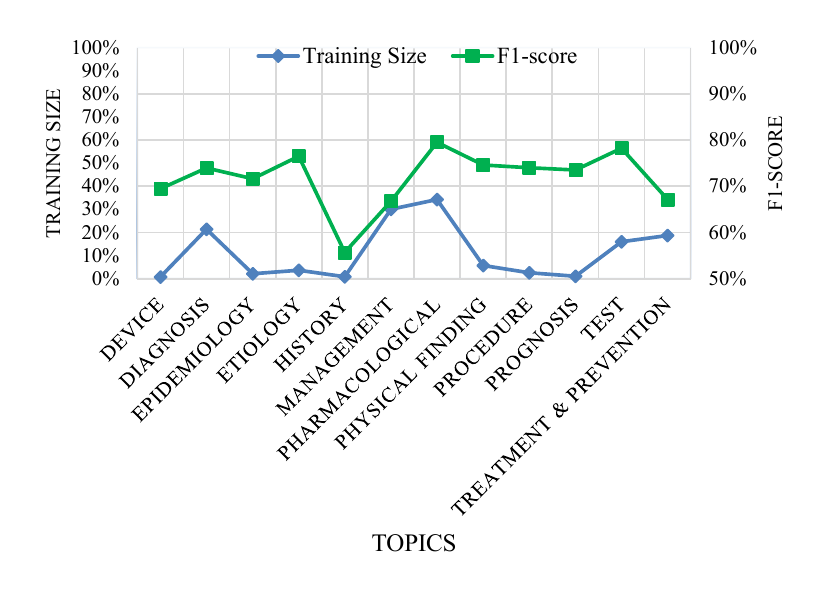}
\caption[The classification performance of topic assignment of the proposed method using Naive Bayes and the corresponding training size]{The classification performance of topic assignment of the proposed method using Naive Bayes and the corresponding training size.}
\label{fig:topics1}
\end{figure}

\begin{figure}[h!]
\captionsetup{justification=justified}
\graphicspath{{Figures/}}
\centering
\includegraphics[width=15cm, height=9cm]{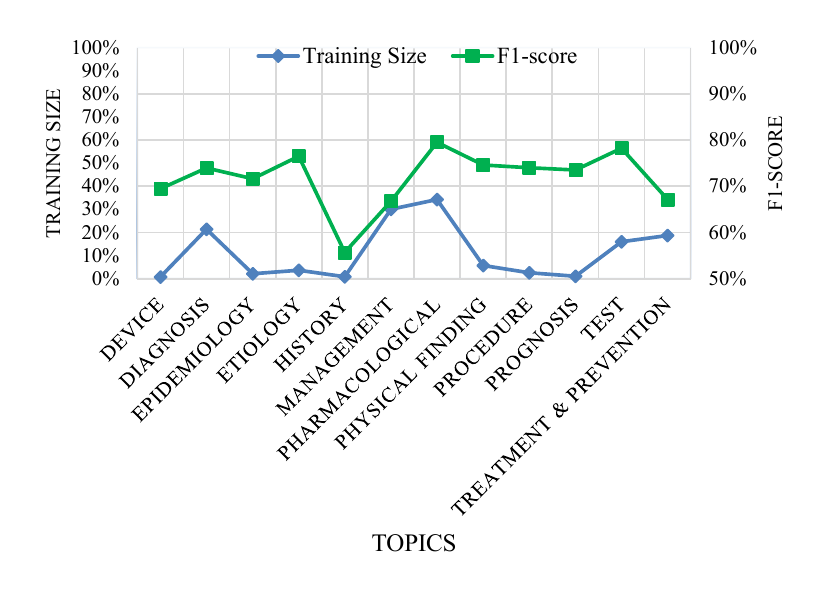}
\caption[The classification performance of topic assignment of the proposed method using SVM and the corresponding training size]{The classification performance of topic assignment of the proposed method using SVM and the corresponding training size.}
\label{fig:topics2}
\end{figure}
\subsection{Discussion}

While question classification has been widely investigated, few approaches are currently able to efficiently classify natural language questions into general topics, in particular for complex questions. This is a challenging task, particularly in a more specific domain such as the clinical domain.

Our experiments presented in Table~\ref{tab:4.3.6} and Table~\ref{tab:4.3.7} show that our SVM-based approach is promising for question topic classification in the context of biomedical QA. Our overall results confirm the findings presented in \cite{yu2008automatically,Cao_2010}, where among the multiple classification systems, the SVM-based one yielded the best results. If we compare our proposed method with the latter, we use the syntactic dependency relations as discriminative features to represent natural language clinical questions. The results
show that our representation which consists of BOW, BOB, BOCST, BOSDR, and BOS$_{porter}$ as features, is more effective as compared with state-of-the-art representations. As shown in Table~\ref{tab:4.3.6} and Table~\ref{tab:4.3.7} the average performance for assigning a topic to a natural language question was a 77.18\% F1-score using SVM and 71.77\% F1-score using Naïve Bayes, as opposed to the baseline of 50\% obtained by random guessing.

Table~\ref{tab:4.3.8} shows the increasing performance of the proposed method in comparison with state-of-the-art methods using Naïve Bayes as a classifier, and Table~\ref{tab:4.3.9} shows the increasing performance using SVM as a classifier. As shown in Table~\ref{tab:4.3.8}, using the Naïve Bayes classifier, our proposed representation outperforms state-of-the-art representations and leads to the highest F1-score of 71.77\%. It outperforms BOW with 6.47\%, BOS$_{porter}$ with 5.36\%, BOS$_{krovetz}$ with 6.37\%, BOBNE with 8.99\%, the combination of BOW and BOCST with 3.83\%, the combination of BOW, BOB, BOS$_{porter}$, BOCST that are used by \cite{yu2008automatically,Cao_2010} with 1.21\%. In the case of using SVM as a classifier, as we can see from Table~\ref{tab:4.3.9}, our proposed method still achieves higher F-score of 77.18\% as compared with the other representations. It generally outperforms BOW with a large margin 6.26\%, BOS$_{porter}$ with 5.74\%, BOS$_{krovetz}$ with 7.26\%, and the combination of BOW and BOCST with 3.5\%. Similarly, our proposed method has a better F-score performance than the method presented by \cite{yu2008automatically,Cao_2010} with an increase performance of 0.7\% F1-score.

As shown in Figure~\ref{fig:topics1} and Figure~\ref{fig:topics2}, inconsistency in topic attribution may be responsible for the topic training size of each topic and the topic classification performance. Typically, the classification performance depends directly on the training size: the larger the training size is the better a classifier performs. However, The Pearson correlation coefficient
between classification performance and the number of training size of a category shows an R-value of 0.2265 using SVM and 0.2396 using Naïve Bayes (weak correlation), which means that the relationship between classification performance and the number of training size is weak. Although we can see clearly from Figure~\ref{fig:topics1} and Figure~\ref{fig:topics2} that the best performing category, pharmacological category (84.85\% F-score using SVM and 79.75\% F1-score using Naïve Bayes), has the biggest number of question instances (1594) and the
worst performing category, history category (68.75\% F1-score using SVM and 55.71\% F1-score using Naïve Bayes), has the smallest number of question instances (43). Management category also has the highest number of question instances (1403), but it did not perform well (71.49\% F-score using SVM and 66.78\% F1-score using Naïve Bayes). Etiology category, on the other side, performs well (83.11\% F1-score using SVM
and 76.54\% F-score using Naïve Bayes) even though the number of instances available for training is small (173). We believe that Etiology is an unambiguous category for assignment. However, on the same data set, \cite{Cao_2010} have shown that the classification performance of clinical questions does not correlate with the number of categories assigned to the question.

Therefore, despite the noisy data, the obtained results show that our proposed method which is based on the incorporation of syntactic dependency relations with words, Porter stemmer, bigrams, UMLS Methasaurus concepts and semantic types, achieves good performance compared with the current state-of-the-art methods for clinical question topic classification.

\section{Summary of the Chapter}

In this chapter we have described the methodologies we proposed for question types and topic classification in biomedical QA. They were all based on machine learning approaches.

In section~\ref{Chapter4.2} we have presented in details the method we proposed for biomedical question type classification. This method which aims at assigning one of the four categories: yes/no, factoid, list, and summary to a natural language question in the biomedical domain, is based on combining both handcrafted lexico-syntactic patterns and machine-learning approaches. We have used the set of our handcrafted lexico-syntactic patterns to extract appropriate features for machine learning algorithms. We have experimented with several commonly used machine-learning approaches for question classification, including Naïve Bayes, Decision Tree, and SVM, and the obtained results have shown that SVM performed the best on the dataset made available as a part of the BioASQ challenge. We have also conducted experiments with different feature sets and best results were obtained using our handcrafted lexico-syntactic patterns. The predefined patterns yielding the best results are also made available which encourage replication of results (cf. section~\ref{Chapter4.2.1.2}).

In section~\ref{Chapter4.3} we have explained in details the methodology we proposed to question topic classification for the purpose of supporting automatic retrieval of clinical answers. This method aims at assigning one or more general topics to clinical questions written in natural language. It is based on various features including words, word stems, bigrams, UMLS concepts and semantic types, and syntactic dependency relations between pair words. We have explored several machine learning algorithms such as Naïve Bayes, Decision Tree, and SVM, showing SVM achieved the best results for this task on the annotated data that is released by NLM. We have also conducted experiments with different feature sets and best results were obtained using our combination of features which consists of bag-of-words, bag-of-Porter stemmer stems, bag-of-bigrams, bag-of-UMLS concepts and semantic types, and bag-of-syntactic dependency relations.


\chapter{Document and Passage Retrieval in Biomedical Question Answering} 

\label{Chapter5} 
\setcounter{secnumdepth}{4}
\minitoc

This chapter presents the methods we propose for document and passage retrieval in biomedical QA, key tasks that are also studied and evaluated separately. Section~\ref{Chapter5.2} will be dedicated to our proposed method for document retrieval in biomedical QA. We consecrate section~\ref{Chapter5.3} to our proposed passage retrieval method in biomedical QA.

\section{Introduction}
\label{Chapter5.1}

Document retrieval and passage retrieval are the most important components of any biomedical QA system, as shown in section~\ref{Chapter3_6_2} and section~\ref{Chapter3_6_3} of the chapter~\ref{Chapter3}, respectively. Document retrieval aims at retrieving the set of relevant documents that are likely to contain the answer to a given query constructed from the question, and passage retrieval consists in extracting relevant passages or snippets from the retrieved documents which serve as candidate answers and the biomedical QA system extracts and generates the answers from them. Undoubtedly, the overall performance of a biomedical QA system heavily depends on the effectiveness of the integrated document retrieval and passage retrieval components: if the document retrieval module and the passage retrieval module of a biomedical QA system fail to find the most relevant documents and passages, respectively, for the potential biomedical question, further processing steps to extract the answers will inevitably fail too. In other words, the correct answers to posted natural language questions can be found only when they already exist in one of the retrieved documents and passages. Additionally, it has been proved that the performance of the document retrieval and passage retrieval modules significantly affect the performance of the whole system \citep{Monz_2003,Tellex_2003}.

In the biomedical domain, the issues of retrieving relevant documents and passages to a given natural language question over a sizable textual document collection have been the lively topics of research in recent years, especially since the launch of the biomedical QA track at the BioASQ challenge \citep{tsatsaronis2012bioasq}. Although the most recent document retrieval systems presented in \citep{balikas2014results} and passage retrieval systems described in \citep{neves2015hpi,yenalaiiith,ligeneric,yang2015learning} have proven to be quite successful at extracting passages in biomedical QA, document retrieval and passage retrieval still require further efforts in order to improve their performance. In such systems the tasks of document retrieval and passage retrieval often set a framework in which an existing biomedical IR system is used, and completely depended on its ranking of documents and passages. Indeed, there are many cases where the IR system mistakenly returns irrelevant citations or passages high in the set or relevant citations or passages low in the set. This problem is certainly a challenging issue as a biomedical QA system usually extracts the answers from the top-ranked documents and passages. We assume that if we solve this issue, we could enhance the performance of  document retrieval and passage retrieval engines in biomedical QA and therefore increase the number of correct documents and passages retrieved containing the appropriate answers to the given natural language question.

In the second contribution of this thesis work, we propose two new methods to enhance the performance of document retrieval and passage retrieval in biomedical QA. The first method aims at retrieving most relevant documents to a given query from the MEDLINE database. The second method identifies and extracts relevant passages from the abstracts of the retrieved documents based on Stanford CoreNLP sentence length as passage length, stemmed words UMLS concepts as features for a probabilistic IR model called BM25. In the remaining of the chapter, we go into details about these proposed methods.

\section{Document Retrieval}
\label{Chapter5.2}

\subsection{Method}
In this section, we present in details the method we propose for document retrieval method in biomedical QA \citep{Sarrouti_2016}. As previously stated, this method aims at retrieving a set of relevant documents from the MEDLINE database to a given biomedical question in biomedical QA. To achieve that, we first create the query from the input question to be fed into a typical IR system. We then re-rank the retrieved documents based on the semantic UMLS similarity \citep{mcinnes2009umls} between biomedical concepts of a given question and each title of the returned documents. The proposed method takes as input a natural language question and a set of biomedical articles retrieved by an existing biomedical search engine (PubMed/MEDLINE). It then re-ranks the retrieved articles promoting to the top of the set the ones it considers most relevant to the given question. The flowchart of the proposed method, as shown in Figure~\ref{fig:dr}, is constructed through the following main steps: (1) query formulation, (2) querying the MEDLINE database using a typical information retrieval, and (3) re-ranking the retrieved documents in which we propose a method based on UMLS similarity.

\begin{figure}[H]
\captionsetup{justification=justified}
\graphicspath{{Figures/}}
\centering
\includegraphics[width=16cm, height=12.3cm]{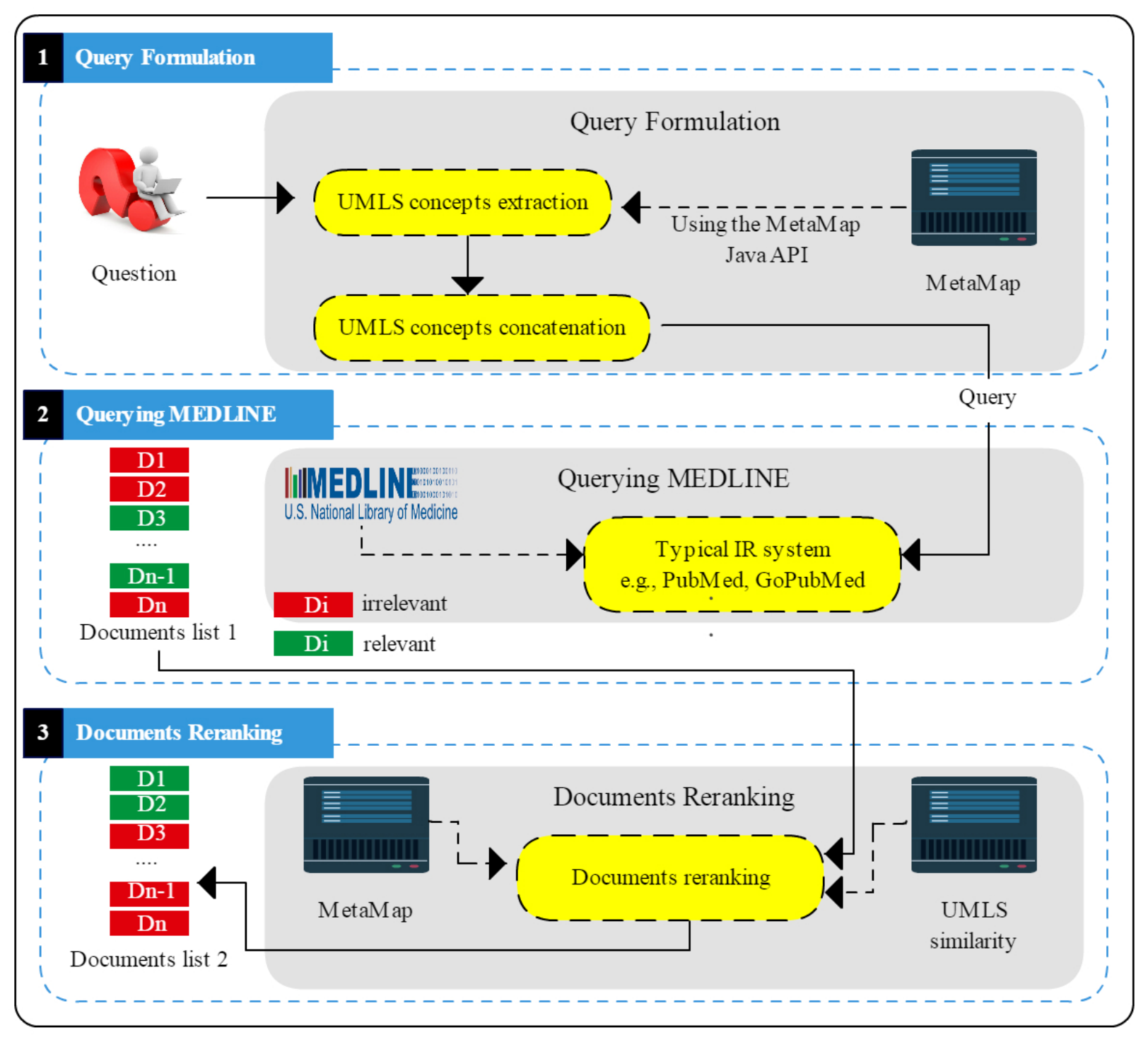}
\caption[Overview of the proposed document retrieval method in biomedical QA]{Overview of the proposed document retrieval method in biomedical QA.}
\label{fig:dr}
\end{figure}

\subsubsection{Query formulation}

Query formulation is the task of constructing from the natural language question a set of keywords or concepts that form a query that can be sent to an IR system. In this work, we construct the query to be fed into a domain-specific IR system that allows the use of ontological terms instead of keywords for a more precise and efficient retrieval of relevant documents. To do so, we first use the MetaMap tool \citep{aronson2001effective} for mapping terms of the question to UMLS Metathesaurus in order to identify biomedical entity names and then connect them with the ``AND'' operator. For example, the biomedical question ``Which hormone concentrations are altered in patients with the Allan Herndon Dudley syndrome`?'' might be reformulated as ``hormones AND allan herndon dudley syndrome (AHDS) AND patients AND Mental concentration''; the question ``Is imatinib an antidepressant drug?'' as ``antidepressive agents AND imatinib''.

\subsubsection{Querying the MEDLINE database}

At this stage, a typical search engine is used to return a set of biomedical documents from the MEDLINE database to a given query constructed by the query formulation task. A search engine takes as input a query and outputs a set of potentially relevant documents, assigning a relevance score
to each document. During this stage of this work, we use GoPubMed search by calling the BioASQ PubMed service\footnote{BioASQ PubMed service: \url{http://gopubmed.org/web/gopubmed/bioasq/pubmedmedline}}. We also use PubMed search engine by calling E-utilities\footnote{E-utilities: \url{https://eutils.ncbi.nlm.nih.gov/entrez/eutils/esearch.fcgi?}} Web service from PubMed. Figure~\ref{fig:pubmed} shows the list of documents returned by PubMed for the question ``Is imatinib an antidepressant drug?''. The number in the tag ``id'' represents PubMed Identifier (PMID) of a document.

\begin{figure}[h!]
\captionsetup{justification=justified}
\centering
\begin{mdframed}[style=MyFrame]
\lstset{
    language=xml,
    tabsize=3,
    label=code:sample,
    rulesepcolor=\color{white},
    xleftmargin=20pt,
    framexleftmargin=20pt,
    keywordstyle=\color{blue}\bf,
    stringstyle=\color{red},
    breaklines=true,
    showstringspaces=false,
    basicstyle=\footnotesize,
    emph={food,name,price},emphstyle={\color{white}}
    }
\begin{lstlisting}
<?xml version="1.0" encoding="UTF-8" ?> 
<!DOCTYPE eSearchResult PUBLIC "-//NLM//DTD esearch 20060628//EN" "https://eutils.ncbi.nlm.nih.gov/eutils/dtd/20060628/esearch.dtd">
<eSearchResult> <Count>10</Count><RetMax>10</RetMax><RetStart>0</RetStart><IdList>
<Id>26573171</Id>
<Id>23470622</Id>
<Id>23343697</Id>
<Id>21395523</Id>
<Id>18360710</Id>
<Id>17008009</Id>
<Id>16755209</Id>
<Id>15470331</Id>
<Id>12688840</Id>
<Id>11879090</Id>
</IdList> </eSearchResult>
\end{lstlisting}
\end{mdframed}
\caption[List of documents returned by PubMed for the question ``Is imatinib an antidepressant drug?'']{List of documents returned by PubMed for the question ``Is imatinib an antidepressant drug?''}
\label{fig:pubmed}       
\end{figure}

\subsubsection{Documents reranking}

Documents reranking, the re-ranking of the retrieved documents, is the important step of the proposed document retrieval system. In the proposed method, we do not completely depend on GoPubMed ranking of documents. So we re-rank the returned documents $(d_1, d_2, ... , d_{n})$ based on the semantic similarity between a given question and the title of each document. The idea is to assign a new relevance score to each of the returned documents. To do so, we first map both the given biomedical question and the titles of its set of the possibly relevant documents returned by the search engine to UMLS Metathesaurus using the MetaMap program so as to identify UMLS concepts. We then compute the sum of the semantic similarity scores between UMLS concepts of a given question and each title of the returned documents using UMLS similarity package\footnote{UMLS similarity: \url{http://maraca.d.umn.edu/cgi-bin/umls\_similarity/umls\_similarity.cgi?version=yes}} \cite{mcinnes2009umls}. Path length has been used as similarity measure where the similarity score is inversely proportional to the number of nodes along the shortest path between the concepts in Mesh\footnote{Mesh: \url{https://www.nlm.nih.gov/mesh/}} ontology. Figure~\ref{fig:sim} shows an example of simantic similarity scores between the biomedical question ``Is Tuberous Sclerosis a genetic disease?'' and the title ``Tuberous sclerosis complex diagnosed from oral lesions.'' of PubMed document (PMID=24310804).

\begin{figure}[h!]
\captionsetup{justification=justified}
\graphicspath{ {Figures/}}
\centering
\includegraphics [width=15cm, height=7cm]{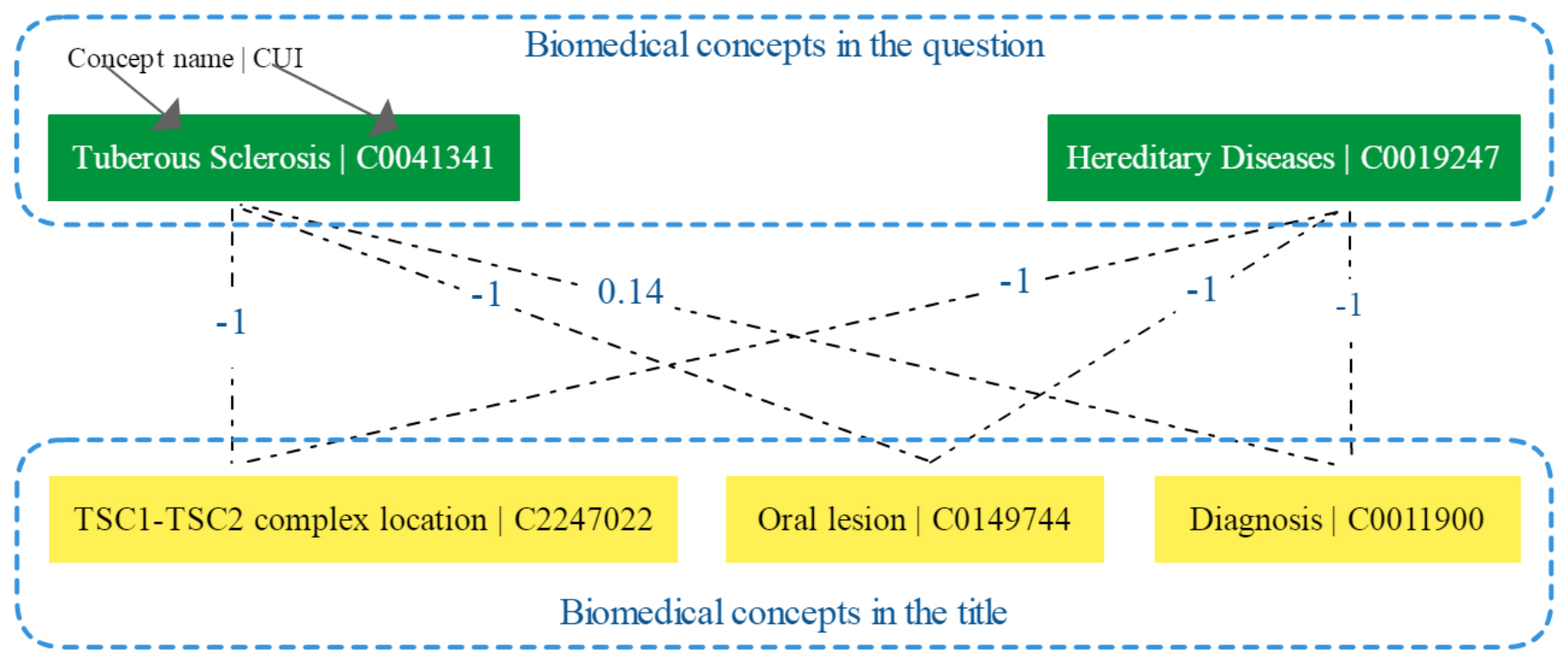}
\caption[An example of semantic similarity scores between biomedical concepts extracted using the MetaMap tool]{An example of semantic similarity scores between biomedical concepts extracted using the MetaMap tool. CUI indicates UMLS concept unique identifier. Semantic similarity scores between CUIs are computed using path length as the similarity measure. -1 means that there is not a semantic relationship between pair UMLS concepts.}
\label{fig:sim}       
\end{figure}

We finally rerank the set of relevant documents based on the semantic similarity scores. The higher semantic score, the higher the document appears in the set. Based on these steps, we design the algorithm~\ref{alg:1} to retrieve the high informative documents for a given biomedical question from PubMed articles.

\begin{algorithm}[h!]
\caption{Biomedical document re-ranking}
\label{alg:1}
\begin{algorithmic}[1]
\State $\textbf{Input} \leftarrow {Question \;Q\; and\; List\;of\;Documents\; R}$
\State $\textbf{Output} \leftarrow {List \;of \;Top\;Documents\; N}$
\Function{RankDocuments}{Q, R}
\State $scores \leftarrow \{\}$
\State $i \leftarrow {1}$
\Do
    \State $T[i]\leftarrow  R[i].title $
    \State $scores[i] \leftarrow \Call{ComputeSimilarity}{$Q, T[i]$}$ \Comment{call function to compute the semantic similarity between the question and document title}
    \State $i \gets i+1$
\doWhile{$i\leq n$} \Comment{n is the size of R}
\State $SortDocuments(R, scores)$ \Comment{sort R according the score of each document}
\State $N [1...m]\leftarrow R [1...m]$ \Comment{keep only the m top documents}
\State \Return $N$
\EndFunction

\Function{ComputeSimilarity}{Q, docTitle}
\State $Question\_concepts\_CUI[1...l]\leftarrow \Call{MappingQuestionToUMLS}{$Q$}$ \Comment{we used MetaMap tool for mapping text to UMLS}
\State $DocTitle\_concepts\_CUI[1...k]\leftarrow \Call{MappingDocTitleToUMLS}{$docTitle$}$
\State $similarity \leftarrow 0$, $sumSimilarity \leftarrow 0$
\State $i \leftarrow 1$, $j \leftarrow 1$
\Do
    \State $QCUI\leftarrow  Question\_Concepts\_CUI[i] $
\Do
    \State $TCUI\leftarrow  docTitle\_Concepts\_CUI[j] $
    \State $similarity \leftarrow UMLS::Similarity(QCUI, TCUI) $
    \If{$similarity \neq -1$} \Comment{there is no relationship between two concepts}
        \State $sumSimilarity \leftarrow sumSimilarity+similarity$
      \EndIf
    \State $j \gets j+1$
\doWhile{$j\leq k$}
    \State $i \gets i+1$
\doWhile{$i\leq l$}

\State \Return $sumSimilarity$
\EndFunction
\end{algorithmic}
\end{algorithm}

\subsection{Experimental results}

In order to evaluate and test the effectiveness of the proposed document retrieval method and our re-ranking technique, as well as to compare the results with state-of-the-art methods, we carried out experiments on benchmark test data provided by the BioASQ challenge. We used the typical evaluation measures used in IR which are: mean precision, mean recall, mean F-measure and mean average precision (MAP) described in section~\ref{Chapter3_7_2} (cf. chapter~\ref{Chapter3}). MAP is the most important measure for evaluating and comparing IR systems as it emphasises early precision and strongly takes into account recall. The proposed method takes as input a biomedical question and its set of possibly relevant documents that was retrieved by the GoPubMed search engine, and it reranks the retrieved documents promoting to the top the documents it considers most relevant to the biomedical question. In particular, we retrieved up to the 200 top ranked documents using the GoPubMed search engine, then reranked them using the proposed algorithm~\ref{alg:1}, and finally kept only the 100 top-ranked documents. Indeed, we have decided to go with the 100 first documents since only the 100 first ones from the resulting list are permitted to be submitted for the test in BioASQ 2014.

Table~\ref{tab:5.2.1} presents the experimental results of the proposed biomedical document retrieval method and the ones of the best state-of-the-art systems presented in \citep{balikas2014results} which were ranked within the 10 top tier systems on the standard test data provided by the BioASQ challenge \citep{tsatsaronis2012bioasq}. The experimental results showed Mean Precision of 0.2331, Mean Recall of 0.3644, Mean F-measure of 0.2253, and MAP of 0.2758. As can be seen from Table~\ref{tab:5.2.1}, the obtained results from the proposed document retrieval method in biomedical QA have an absolute competitiveness with the top 10 state-of-the-art methods results.

\begin{table}[h!]
\centering

\caption{The experimental results of the proposed document retrieval system and state-of-the-art systems presented in \citep{balikas2014results}, which were ranked within the 10 top tier systems on batch 1 of BioASQ 2014.}
\label{tab:5.2.1}
\begin{tabular}{p{4cm}p{3cm}p{2.8cm}p{3cm}p{1.3cm}}
\hline \noalign{\smallskip}
System &Mean Precision &Mean Recall &Mean F-measure &MAP  \\
\noalign{\smallskip}\hline\noalign{\smallskip}
\textbf{Our System }&\textbf{0.2331} &\textbf{0.3644} &\textbf{0.2253} &\textbf{0.2758} \\
SNUMedinfo1 &0.0457& 0.5958 &0.0826& 0.2612 \\
SNUMedinfo3 &0.0457 &0.5947 &0.0826 &0.2587 \\
SNUMedinfo2 &0.0451 &0.5862 &0.0815 &0.2547\\
SNUMedinfo4 &0.0457 &0.5941 &0.0826 &0.2493\\
SNUMedinfo5 &0.0459 &0.5947 &0.0829 &0.2410\\
Top 100 Baseline &0.2274 &0.4342 &0.2280 &0.1911 \\
Top 50 Baseline &0.2290 &0.3998 &0.2296 &0.1888 \\
Main system &0.0413 &0.2625 &0.0678 &0.1168 \\
Biomedical Text Ming &0.2279 &0.2068 &0.1665 &0.1101  \\
Wishart-S2 &0.1040 &0.1210 &0.0793 &0.0591 \\

\noalign{\smallskip}\hline

\end{tabular}
\end{table}

We have experimented with and without our document reranking method. Table~\ref{tab:5.2.2} shows the increase performance of the proposed document reranking method in comparison with GoPubMed document ranking in biomedical QA. In particular, this evaluation aims to answer the following question:

 \begin{itemize}
 \item Is the proposed reranking technique able to achieve improvement with respect to the original document ranking provided by a typical IR system?
 \end{itemize}

\begin{table}[h!]
\centering

\caption{Comparison of our document reranking method and GoPubMed document ranking for biomedical QA on batch 1 of BioASQ 2014.}
\label{tab:5.2.2}
\begin{tabular}{M{7.2cm}M{1.7cm}M{1.6cm}M{2.2cm}M{1.5cm}}
\hline \noalign{\smallskip}
System &Mean \newline  Precision &Mean  \newline Recall &Mean \newline F-measure &MAP  \\
\noalign{\smallskip}\hline\noalign{\smallskip}
 GuPubMed document ranking &0.2253	&0.3111&	0.1913&	0.1439 \\
\textbf{Proposed document reranking method}&\textbf{0.2331} &\textbf{0.3644} &\textbf{0.2253} &\textbf{0.2758} \\\cmidrule(l){1-5}
Increase performance&+0.0078&+0.0533&+0.034 &+0.1319 \\

\noalign{\smallskip}\hline

\end{tabular}
\end{table}

As shown in Table~\ref{tab:5.2.2}, the proposed document reranking method significantly outperformed the GoPubMed document ranking for placing the most relevant documents at the top of the set to a given biomedical question in biomedical QA. In terms of MAP, the proposed system achieved 0.2758, whereas the GoPuMed search engine achieved 0.1439.

\subsection{Discussion}

A typical IR system takes as input a query and outputs a set of potentially citations, assigning a relevance score to each citation. The higher the score is, the higher the citation appears in the set. However, there are many cases where the search engine mistakenly returns irrelevant citations high in the set or relevant citations low in the set. This poses a real problem for biomedical QA systems which usually extract the answer from the first retrieved documents to a given biomedical question. Taking an example from BioASQ training questions, assume we have the biomedical question ``Which are the cellular targets of imatinib mesylate?'' (identifier 53188992b166e2b806000019), PubMed returns 1115 articles. The BioASQ training dataset indicates that only two are relevant to the question, those with PMID (PubMed IDentifier) 15887238 and 15844661. The first one is ranked 154th and the second one 237th. This means that the first 10 or even 100 articles do not answer the question. In this work, we studied the problem of deciding if a PubMed document is relevant to a specific biomedical question written in natural language. We developed a document retrieval engine in which we proposed a new reranking method to re-rank the retrieved set of documents and move the most relevant documents to the top of the set.

The conducted experiments presented in Table~\ref{tab:5.2.1}, clearly demonstrate that the proposed method is more effective as compared with current state-of-the-art systems which were ranked within the 10 top tier systems presented in \citep{balikas2014results} in the BioASQ challenge. For instance, in terms of MAP, the proposed system outperformed SNUMedinfo \cite{choi2014classification}, the best performing system in the challenge, which uses both the semantic concept-enriched dependence model and the sequential dependence model.

Additionally, it is important to notice that in terms of Mean Precision, Mean F-measure and MAP, the proposed system achieved performance improvements over the state-of-the-art systems. In terms of Precision@100 which shows how useful the retrieval results is, the proposed system which achieved 0.2331, is more effective as compared to the state-of-the-art systems at retrieving relevant documents among the 100 retrieved documents.  The high precision of the proposed document retrieval system means that our system retrieved more relevant documents than irrelevant ones.

On the other hand, we have experimented with and without our document reranking method, as shown in Table~\ref{tab:5.2.2}, in order to show if the proposed reranking technique is able to achieve improvement with respect to the original document ranking provided by a typical IR system. Effectively, the experimental results have shown that the proposed document reranking method outperformed the GoPubMed document ranking by a statistically significant margin (0.1319 of MAP). On the other hand, as can be seen from Table~\ref{tab:5.2.2}, there is not a significant difference between the results in terms of Precssion@100 of the proposed method and GoPubMed. This means that both systems have returned the same number of relevant documents to the given biomedical questions. However, in terms of MAP, the proposed reranking system significantly outperformed the GoPubMed document ranking (0.2758 against 0.1439 of MAP). This means that the proposed system was very successful to move the relevant documents at the top of the set of retrieved documents. We attribute this to the UMLS similarity used in this work to calculate the similarity scores between the biomedical questions and the titles of the retrieved documents.

Placing and moving the relevant documents at the top of the set of retrieved documents to given biomedical questions is very useful for biomedical QA systems as they usually extract the answers from the first top-ranked documents. Furthermore, it is much better if the relevant documents are shown first for the users since by returning irrelevant citations high in the list or relevant citations low in the list to a given query, make it unlikely for a user to read them or the users have to read them one by one until they finally find a relevant document. Therefore, from the results and analysis, we can draw a conclusion that the proposed document retrieval system may improve the overall performance of biomedical QA systems, and allows a practical and competitive alternative to help information seekers find quickly the relevant documents to their queries.

Although the set of PubMed citations is generally ranked by relevance, the top-ranked citations is probably not the answers to the given biomedical questions. This is because citations are not an ideal unit to rank with respect to the ultimate goals of a biomedical QA system. A highly relevant citation that does not prominently answer a biomedical question is not an appropriate candidate for further QA processing. Therefore, the next step is to identify and extract a set of candidate answer passages from the retrieved and selected documents. Accordingly, we will present in the following section~\ref{Chapter5.3} an efficient method to extract a set of relevant passages to a given biomedical question.

\section{Passage Retrieval}
\label{Chapter5.3}
\subsection{Method}

In this section, we present in details the proposed passage retrieval method in biomedical QA \citep{Sarrouti_2017}, which is comprised of two stages: document retrieval and passage retrieval. During the document retrieval stage, the query that is constructed from the user question is handed over to the developed document retrieval system, which is similar to our proposed document retrieval approach, presented in previous section~\ref{Chapter5.2}, in order to find relevant documents. Then, the proposed passage retrieval method extracts relevant passages from the abstracts of the retrieved documents in the passage retrieval stage. The flowchart of the proposed method is shown in Figure~\ref{fig:pr}.

\begin{figure}[h!]
\captionsetup{justification=justified}
\graphicspath{{Figures/}}
\centering
\includegraphics[width=16cm, height=12cm]{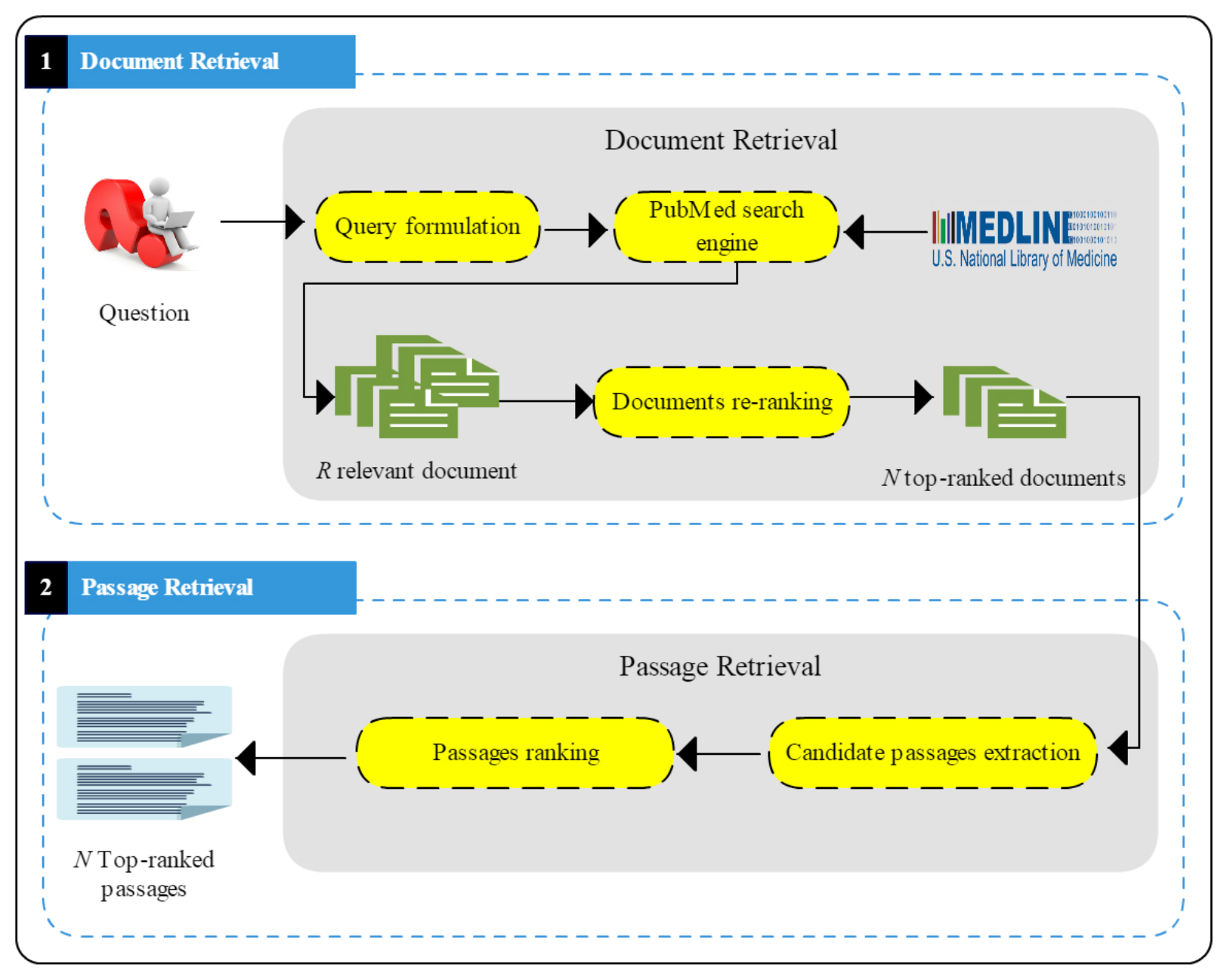}
\caption[Overview of the proposed passage retrieval method in biomedical QA]{Overview of the proposed passage retrieval method in biomedical QA.}
\label{fig:pr}
\end{figure}

The input to the proposed biomedical passage retrieval method is a natural language question. A document retrieval stage uses the MetaMap tool \citep{aronson2001effective} to identify the biomedical entities from the given question, and then connect them ``+'' operator to construct the query. Next, it sends the query to PubMed search engine by calling E-utilities\footnote{E-utilities: \url{https://eutils.ncbi.nlm.nih.gov/entrez/eutils/esearch.fcgi?}} web service from PubMed in order to retrieve the $R$ relevant documents. After that, it re-ranks the $R$ retrieved documents based on our proposed document re-ranking algorithm~\ref{alg:1}, and finally keeps only the $N$ top-ranked ones to be input for the next stage, passage extraction.

In the passage retrieval stage, once the $N$ top-ranked documents $(d_1, d_2, ... , d_{n})$ are retrieved by the document retrieval system for a given question, we take abstracts of all these documents and forward them one by one to Stanford CoreNLP \citep{Manning_2014} in order  to extract all sentences, i.e., a set of candidate passages. The passage length is similar to that of the Stanford CoreNLP sentence. Figure~\ref{fig:passages} shows a list of candidate passages extracted from the abstract of the PubMed document\footnote{\url{https://www.ncbi.nlm.nih.gov/pubmed/23044018}} using the Stanford CoreNLP tool.

\begin{figure}[h!]
\captionsetup{justification=justified}
\centering
\begin{mdframed}[style=MyFrame]
\begin{enumerate}
  \item Epilepsy, a neurologic disorder characterized by the predisposition to recurrent unprovoked seizures, is reported in more than 300 genetic syndromes.
  \item Muenke syndrome is an autosomal-dominant craniosynostosis syndrome characterized by unilateral or bilateral coronal craniosynostosis, hearing loss, intellectual disability, and relatively subtle limb findings such as carpal bone fusion and tarsal bone fusion.

  \item Muenke syndrome is caused by a single defining point mutation in the fibroblast growth factor receptor 3 (FGFR3) gene.

  \item Epilepsy rarely occurs in individuals with Muenke syndrome, and little detail is reported on types of epilepsy, patient characteristics, and long-term outcomes.

  \item We present seven patients with Muenke syndrome and seizures.

  \item A review of 789 published cases of Muenke syndrome, with a focus on epilepsy and intracranial anomalies in Muenke syndrome, revealed epilepsy in six patients, with intracranial anomalies in five.

  \item The occurrence of epilepsy in Muenke syndrome within our cohort of 58 patients, of whom seven manifested epilepsy, and the intracranial anomalies and epilepsy reported in the literature, suggest that patients with Muenke syndrome may be at risk for epilepsy and intracranial anomalies."

  \item Furthermore, the impact of Muenke syndrome on the central nervous system may be greater than previously thought.

\end{enumerate}

\end{mdframed}
\caption[List of candidate passages extracted from the abstract of a PubMed document (PIMD=23044018) using Stanford CoreNLP]{List of candidate passages extracted from the abstract of the PubMed document (PIMD=23044018) using Stanford CoreNLP}
\label{fig:passages}       
\end{figure}

Then, we compute the similarity scores based on BM25 as retrieval model between the biomedical question and each of the candidate passages. Before that, terms in both the question and passages are extracted, all tokens which match stop words list\footnote{Stop words:  \url{http://www.textfixer.com/resources/common-english-words.txt}} are removed and stemmed words are extracted using Porter' stemmer \citep{Porter_1980}. In addition, we have used the MetaMap\footnote{MetaMap: \url{http://metamap.nlm.nih.gov/}} program \citep{aronson2001effective} for mapping the biomedical question and passages to UMLS Metathesaurus (2016AA knowledge source) in order to extract UMLS concepts, and we then used them as additional features. Moreover, the MetaMap word sense disambiguation system \citep{Aronson_2010} has been used to resolve ambiguities in the texts by identifying the meaning of ambiguous terms. Therefore, stemmed words and UMLS concepts were then used as index terms. Indeed, each of the candidate passages is represented by the weighted vector $p_i={(stem_1^i, stem_2^i, ..., stem_n^i, concept_1^i, concept_2^i, ..., concept_n^i)}$ which contains stemmed words and UMLS concepts as features.

As it is stated, the similarity between a candidate passage and a biomedical question is defined by the BM25 model. Indeed, the BM25 function scores each passage in a set of passages according to the passage's relevance to a particular biomedical question. Formally, let a biomedical question $Q$ with terms and concepts $Q= (q_1, q_2, ... , q_n)$, the BM25 score for a passage $p$ is given by:
\begin{equation}\label{eq:1}
BM25\;(p)= \sum_{i=1}^n IDF(q_i)\frac {f(q_i)*(k_1+1)}{f(q_i)+k_1*(1-b+b*|P|/p_{avg})}\
\end{equation}
where:
\begin{itemize}
  \item $f(q_i)$ is the number of times term $q_i$ occurs in passage $p$;
  \item |P| is the number of words and concepts in passage $p$;
  \item $p_{avg}$ is the average number of words and concepts per passage;
  \item $k_1$ and $b$ are free parameters for Okapi BM25 \citep{Robertson96okapiat}. $k_1$ is a positive parameter
that calibrates the passage term frequency scaling and b is a parameter $(0 \leq b \leq 1)$ that determines the passage length scaling. By default, $k_1$ and $b$ are set to 1.2, 0.75, respectively.
\end{itemize}
The first quantity in the sum is the inverse document frequency (IDF). For a set of passages with $N$ passages, inverse document frequency for term and concept $q_i$ is defined in the following equation:
\begin{equation}\label{eq:2}
IDF(q_i)= \log {\frac {N-N(q_i)+0.5}{N(q_i)+0.5}}\
\end{equation}
where $N(q_i)$ is the number of passages in the set of passages that contain term or concept $q_i$. Note that terms with a lower $IDF$ are ignored since such terms are not informative.

Finally, after finding BM25 scores for all passages, we sort them and take the $M$ top-ranked passages matching most with the biomedical question.

\subsection{Experimental results}

In order to demonstrate the effectiveness of the proposed biomedical passage retrieval method and compare the results with the state-of-the-art methods, we carry out a comprehensive set of experiments on benchmark datasets provided by the bioASQ challenge. We have used the five test sets provided in BioASQ Task 3b.  Each set of testing datasets contains approximately 100 biomedical questions. As indicators of retrieval effectiveness, mean precision, mean recall, mean F1-measure and MAP described in section~\ref{Chapter3_7_2} were used (cf. chapter~\ref{Chapter3}). As it has already been stated before, the proposed method first retrieves the $N$ top-ranked documents based on PubMed and UMLS similarity, and then retrieves the $N$ top-ranked passages from the abstracts of the retrieved documents to a given biomedical question. To do so, the $N$ top-ranked documents are segmented into sentences using the Stanford CoreNLP tool in order to make a set of candidate passages. The passage length is similar to the Stanford CoreNLP sentence one. Then, the candidate passages are ranked using stemmed words and UMLS concepts as features for the BM25 model. The best values of parameters b and k1 of BM25 are respectively 0.85 and 1.2 which were fixed after 5-fold cross-validation. Finally, we kept the $N$ top-ranked passages from the set of candidate passages. In particular, we have decided to go with the 10 first passages ($N=10$) since only the 10 first ones from the resulting list are permitted to be submitted for the test in BioASQ 2015. MAP@10 gives an indication of how the passage retrieval system performs for QA systems that are primarily interested in the first set of retrieved documents. Figure~\ref{fig:docssrel} and Figure~\ref{fig:passagesrel} show an example of the retrieved documents and 10 top-ranked passages, respectively, for the biomedical question ``What symptoms characterize the Muenke syndrome?'' (52bf1d3c03868f1b0600000d) from the BioASQ datasets.

\begin{figure}[h!]
\captionsetup{justification=justified}
\centering
\begin{mdframed}[style=MyFrame]
\begin{enumerate}
\setlength\itemsep{0em}
  \item  \href{https://www.ncbi.nlm.nih.gov/pubmed/23378035}{https://www.ncbi.nlm.nih.gov/pubmed/23378035}
 \item  \href{https://www.ncbi.nlm.nih.gov/pubmed/23044018}{https://www.ncbi.nlm.nih.gov/pubmed/23044018}
 \item  \href{https://www.ncbi.nlm.nih.gov/pubmed/22872265}{https://www.ncbi.nlm.nih.gov/pubmed/22872265}
 \item  \href{https://www.ncbi.nlm.nih.gov/pubmed/22016144}{https://www.ncbi.nlm.nih.gov/pubmed/22016144}
 \item  \href{https://www.ncbi.nlm.nih.gov/pubmed/22446440}{https://www.ncbi.nlm.nih.gov/pubmed/22446440}
 \item  \href{https://www.ncbi.nlm.nih.gov/pubmed/22622662}{https://www.ncbi.nlm.nih.gov/pubmed/22622662}
 \item  \href{https://www.ncbi.nlm.nih.gov/pubmed/21233754}{https://www.ncbi.nlm.nih.gov/pubmed/21233754}
 \item  \href{https://www.ncbi.nlm.nih.gov/pubmed/21403557}{https://www.ncbi.nlm.nih.gov/pubmed/21403557}

\end{enumerate}

\end{mdframed}
\caption[List of the retrieved PubMed documents for the biomedical question ``What symptoms characterize the Muenke syndrome?'' (52bf1d3c03868f1b0600000d) from BioASQ datasets]{List of the retrieved PubMed documents for the biomedical question ``What symptoms characterize the Muenke syndrome?'' (52bf1d3c03868f1b0600000d) from BioASQ datasets}
\label{fig:docssrel}       
\end{figure}

\begin{figure}[h!]
\captionsetup{justification=justified}
\centering
\begin{mdframed}[style=MyFrame]
\begin{enumerate}
\setlength\itemsep{0em}
  \item She was treated with ibuprofen, shoe inserts, a CAM walker boot, and stretching exercises without much improvement in symptoms.
  (PMID \href{https://www.ncbi.nlm.nih.gov/pubmed/23378035}{23378035})
  \item Muenke syndrome is characterized by coronal suture synostosis, midface hypoplasia, subtle limb anomalies, and hearing loss.
   (PMID \href{https://www.ncbi.nlm.nih.gov/pubmed/22872265}{22872265})
  \item Muenke syndrome is characterized by various craniofacial deformities and is caused by an autosomal-dominant activating mutation in fibroblast growth factor receptor 3 (FGFR3(P250R)).
      (PMID \href{https://www.ncbi.nlm.nih.gov/pubmed/22622662}{22622662})
  \item Epilepsy, a neurologic disorder characterized by the predisposition to recurrent unprovoked seizures, is reported in more than 300 genetic syndromes. (PMID \href{https://www.ncbi.nlm.nih.gov/pubmed/23044018}{23044018})

  \item Muenke syndrome is characterized by coronal craniosynostosis (bilateral more often than unilateral), hearing loss, developmental delay, and carpal and/or tarsal bone coalition. (PMID \href{https://www.ncbi.nlm.nih.gov/pubmed/23378035}{23378035})

  \item Muenke syndrome is an autosomal-dominant craniosynostosis syndrome characterized by unilateral or bilateral coronal craniosynostosis, hearing loss, intellectual disability, and relatively subtle limb findings such as carpal bone fusion and tarsal bone fusion.
(PMID \href{https://www.ncbi.nlm.nih.gov/pubmed/23044018}{23044018})
  \item We show in this study that knock-in mice harboring the mutation responsible for the Muenke syndrome (FgfR3(P244R)) display postnatal shortening of the cranial base along with synchondrosis growth plate dysfunction characterized by loss of resting, proliferating and hypertrophic chondrocyte zones and decreased Ihh expression. (PMID \href{https://www.ncbi.nlm.nih.gov/pubmed/22016144}{22016144})

 \item To better understand the pathophysiology of the Muenke syndrome, we present collective findings from several recent studies that have characterized a genetically equivalent mouse model for Muenke syndrome (FgfR3 (P244R)) and compare them with human phenotypes.CONCLUSIONS: FgfR3 (P244R) mutant mice show premature fusion of facial sutures, premaxillary and/or zygomatic sutures, but rarely the coronal suture. (PMID \href{https://www.ncbi.nlm.nih.gov/pubmed/22872265}{22872265})

  \item We present seven patients with Muenke syndrome and seizures. (PMID \href{https://www.ncbi.nlm.nih.gov/pubmed/23044018}{23044018})

  \item The mouse model can also be further explored to discover previously unreported yet potentially significant phenotypes of Muenke syndrome. (PMID \href{https://www.ncbi.nlm.nih.gov/pubmed/22872265}{22872265})

\end{enumerate}

\end{mdframed}
\caption[List of the retrieved passages for the biomedical question ``What symptoms characterize the Muenke syndrome?'' (52bf1d3c03868f1b0600000d) from BioASQ datasets]{List of the retrieved passages for the biomedical question ``What symptoms characterize the Muenke syndrome?'' (52bf1d3c03868f1b0600000d) from BioASQ datasets}
\label{fig:passagesrel}       
\end{figure}

Table~\ref{tab:5.3.1} presents the experimental results of the proposed biomedical passage retrieval method on five batches of testing datasets provided by the BioASQ challenge \citep{tsatsaronis2012bioasq}. In terms of MAP, our system achieves 0.1178, 0.1180, 0.1491, 0.1633, and 0.1437 in Batch 1, Batch 2, Batch 3, Batch 4, and Batch 5, respectively.

\begin{table}[h!]
\centering

\caption{
The overall results of the proposed method on five batches of testing datasets provided by the BioASQ challenge.}
\label{tab:5.3.1}
\begin{tabular}{p{2.3cm} p{3.2cm}p{3.2cm}p{3.8cm}p{1.6cm}}
\hline \noalign{\smallskip}
Batches  &Mean Precision &Mean Recall &Mean F1-measure &MAP  \\
\noalign{\smallskip}\hline\noalign{\smallskip}
{Batch 1} &0.1340&	0.1153&	0.1069&	0.1178 \\

{Batch 2} &0.1465	&0.1084&	0.1082&	0.1180\\

{Batch 3} &0.1688	&0.1018	&0.1138&	0.1491\\

{Batch 4}&        0.1928&	0.1254&	0.1321&	0.1633\\

{Batch 5}&       0.1816	&0.1401&	0.1370&	0.1437\\
\noalign{\smallskip}\hline

\end{tabular}
\end{table}

Table~\ref{tab:5.3.2}, on the other hand, shows a comparison in terms of MAP between the proposed biomedical passage retrieval method and the current state-of-the-art methods presented in \citep{neves2015hpi,yenalaiiith,ligeneric,yang2015learning} on five batches of testing datasets. The ``+'' sign and number in parentheses indicate the statistically significant improvements of the proposed method over the state-of-the-art methods. The conducted experiments clearly demonstrate that the proposed method outperformed the current state-of-the-art methods by a statistically significant margin. The proposed method outperforms the current state-of-the-art methods by an average of 6.84\% in terms of MAP.
\begin{table}[h!]
\centering

\caption[Comparison in terms of MAP of the proposed biomedical passage retrieval method with the state-of-the-art methods on five batches of testing datasets provided by the BioASQ challenge]{Comparison in terms of MAP of the proposed biomedical passage retrieval method with the state-of-the-art methods on five batches of testing datasets provided by the BioASQ challenge. The ``-'' replace the scores of systems that did not evaluated on some batches.}
\label{tab:5.3.2}
\begin{tabular}{M{4.2cm}M{1.9cm} M{1.9cm}M{1.9cm}M{1.9cm}M{1.9cm}}
\hline \noalign{\smallskip}
\multirow{2}{*}{Models}& \multicolumn{5}{c}{MAP@10} \\
\cmidrule(l){2-6}
& Batch 1 & Batch 2 &Batch 3 &Batch 4&Batch 5 \\
\noalign{\smallskip}\hline\noalign{\smallskip}
\textbf{Proposed Method}&\textbf{0.1178}&\textbf{0.1180}&\textbf{0.1491}&\textbf{0.1633}&\textbf{0.1437}\\\cmidrule(l){1-6}
\cite{neves2015hpi}&0.0347 (+8.31\%)& 0.0355 (+8.3\%)& 0.0452 (+10.39\%)& 0.0624 (+10.09\%)& 0.0572 (+8.65\%)\\\cmidrule(l){1-6}
\cite{yenalaiiith}&0.0545 (+6.33\%)& 0.0709 (+4.71\%)& -& 0.0913 (+7.2\%)& -\\\cmidrule(l){1-6}
\cite{ligeneric}&0.0570 (+6.08\%)& 0.0521 (+6.59\%)& 0.0832 (+6.59\%)& 0.0936 (+6.97\%)& 0.1201 (+2.36\%)\\\cmidrule(l){1-6}
\cite{yang2015learning}&-&-&0.0892 (+5.99\%)&0.0957 (+6.76\%)&0.1027 (+4.1\%)\\
\noalign{\smallskip}\hline

\end{tabular}
\end{table}

Moreover, we present in Table~\ref{tab:5.3.3} the obtained results of the different implementation variations of the BM25 model, i.e., with or without stemming or UMLS concepts, and the two passage lengths: OpenNLP\footnote{OpenNLP: \url{https://opennlp.apache.org/}} sentence length and Stanford CoreNLP\footnote{Stanford CoreNLP: \url{http://stanfordnlp.github.io/CoreNLP/}}  sentence length. Concretely, these experiments aim to answer the following questions:

\begin{enumerate}
  \item What is the effect of stemming and UMLS concepts on the performance of passage retrieval systems?
  \item What is the effect of the passage length on the overall performance of passage retrieval systems?
\end{enumerate}

\begin{table}[h!]
\caption{The obtained results of the different implementation variations of the BM25 model, i.e., with or without stemming or UMLS concepts, and the two passage lengths: OpenNLP sentence length and Stanford CoreNLP sentence length}
\label{tab:5.3.3}
\begin{tabular}{p{6.2cm}p{1.5cm} p{1.5cm}p{1.5cm}p{1.5cm}p{1.5cm}}
\hline \noalign{\smallskip}
\multirow{2}{*}{Models}& \multicolumn{5}{c}{MAP@10} \\
\cmidrule(l){2-6}
& Batch1 & Batch2 &Batch3 &Batch4&Batch5 \\
\noalign{\smallskip}\hline\noalign{\smallskip}
$BM25_{OpenNLP, stem}$&0.1000& 0.1118 &0.1212  & 0.1289 &0.1208  \\
$BM25_{StanfordCoreNLP, stem}$&0.1128 & 0.1106 & 0.1354& 0.1367& 0.1349\\
$BM25_{StanfordCoreNLP, UMLS}$&0.1051 & 0.1014 & 0.1197 & 0.1626 & 0.1266 \\
$BM25_{StanfordCoreNLP, stem+UMLS}$&0.1178 & 0.1180 & 0.1491 & 0.1633 & 0.1437\\
\noalign{\smallskip}\hline

\end{tabular}
\end{table}

Additionally, we have experimented with TFIDF metrics used in \citep{neves2015hpi} to ordering the candidate passages obtained by Stanford CoreNLP sentence length so as to confirm how well passage length and retrieval models impact the performance of the passage retrieval systems. Table~\ref{tab:5.3.4} presents the results of TFIDF metric and Stanford CoreNLP sentence/passage length. The results clearly demonstrate that using the same retrieval model used in \citep{neves2015hpi}, i.e., TFIDF metric, and Stanford CoreNLP passage length outperformed the passage retrieval system presented by Neves \cite{neves2015hpi}, which uses the built-in information retrieval features available in the IMDB by a statistically significant margin.

\begin{table}[h!]
\caption{The detailed results using TFIDF metrics used in \citep{neves2015hpi} and Stanford CoreNLP passage length on five batches of testing datasets provided by the BioASQ challenge.}
\label{tab:5.3.4}
\begin{tabular}{p{2.6cm} p{3.2cm}p{3.1cm}p{3.6cm}p{1.6cm}}
\hline \noalign{\smallskip}
Batches  &Mean Precision &Mean Recall &Mean F1-measure &MAP  \\
\noalign{\smallskip}\hline\noalign{\smallskip}
{Batch 1} &0.1135	&0.1105&	0.0972&	0.0908 \\

{Batch 2} &0.1383	&0.1021	&0.1015&	0.0963\\

{Batch 3} &0.1431&	0.1046&	0.1042&	0.1010\\

{Batch 4}&0.1607	&0.0992&	0.1062&	0.1210\\

{Batch 5}& 0.1235&	0.1042&	0.0952&	0.1014\\
\noalign{\smallskip}\hline

\end{tabular}
\end{table}

\subsection{Discussion}

Though passage retrieval has been widely investigated, few approaches are currently able to efficiently extract passages from large collections of documents. This is a challenging task in any QA system, particularly in the biomedical domain. The definition of a passage (also known as a snippet) is necessarily system dependent, but the typical units include sentences, paragraphs, and sections. In this work, our goal was to develop an efficient passage retrieval method in biomedical QA. In our method, the length of a passage is similar to that of the sentence. The proposed system retrieves top-ranked documents which are broken up into suitable snippets/passages and reranked. Experimental results presented in Table~\ref{tab:5.3.1} and Table~\ref{tab:5.3.2} show that the proposed method is more effective as compared with the state-of-the-art methods proposed in \citep{neves2015hpi,yenalaiiith,ligeneric,yang2015learning}. The proposed method significantly outperforms these methods by an average of 6.84\% in terms of MAP. Furthermore, it is important to notice that in all batches of testing datasets, the proposed biomedical passage retrieval method achieves performance improvements over the state-of-the-art methods. For instance, compared with the approach proposed in \citep{neves2015hpi}, which employs TFIDF weighting metrics to compute similarity between questions and the set of candidate passages obtained by the built-in information retrieval features available in the IMDB database, our method gives better results (an average improvement of 9.15\% in terms of MAP). The increased performance was statistically significant (the P-value is 0.0357, the result is significant at p < 0.05). Our method outperforms also the method based on some specific rules used to define passage length and the sequential dependence model as retrieval model proposed and evaluated in \citep{ligeneric}. It outperforms the latter by an average of 5.72\% in terms of MAP.

Passage length and retrieval models are two factors that impact the performance of passage retrieval systems \citep{levow2013uwcl}. After deep analyzing passages in the BioASQ training dataset, we found that the distribution of passage length is approximately similar to that of a sentence. Indeed, if the set of the candidate passages retrieved from relevant documents is not identified correctly, the retrieval model used to rank these candidate passages and select the best ranked ones will inevitably fail too. Therefore, we have experimented with two sentence splitter tools, i.e., two passage lengths, in order to show how well a system can perform the passage retrieval task from PubMed documents by using the Stanford CoreNLP sentence length. Table~\ref{tab:5.3.3} presents the effect of the two passage lengths, OpenNLP\footnote{OpenNLP: \url{https://opennlp.apache.org/}} sentence length and Stanford CoreNLP\footnote{Stanford CoreNLP: \url{http://stanfordnlp.github.io/CoreNLP/}} sentence length, on passage retrieval using the BM25 model with or without stemming or UMLS concepts. The results show that passage length has a great impact on passage retrieval using the BM25 model. The highest MAP is obtained using the Stanford CoreNLP sentence/passage length. Moreover, the results also show that features impact the passage retrieval performance. Using stemmed words as features for the BM25 model, the performance in terms of MAP was 0.1128, 0.1106, 0.1354, 0.1367, and 0.1349 in Batch 1, Batch 2, Batch 3, Batch 4, and Batch 5, respectively. When UMLS concepts were used as additional features, the overall performance in terms of MAP decreased to 0.1051, 0.1014, 0.1197, 0.1626, and 0.1266 in Batch 1, Batch 2, Batch 3, Batch 4, and Batch 5, respectively. In contrast, the proposed method, i.e., the incorporation of stemmed words and UMLS concepts as features improved the performance to the highest MAP in all batches of testing datasets. Moreover, the increased performance was statistically significant.

In summary, from the results and analysis we can draw a conclusion that our proposed biomedical passage retrieval method allows extracting passages in biomedical QA with high mean average precision. We attribute this to the Stanford CoreNLP passage length, stemmed words and UMLS concepts features used in this work, and the utilization of the BM25 model. The biomedical passage retrieval module plays a vital role in QA systems since it allows retrieving the top-ranked passages which serve as candidate answers and the biomedical QA system selects the answers from them. It has a significant impact on the overall performance of a biomedical QA system. Therefore, by improving the performance of passage retrieval, the overall performance of biomedical QA systems will inevitably improve too.

\section{Summary of the Chapter}

In this chapter we have presented the methods we propose for document and passage retrieval in biomedical QA which aim at retrieving a set of relevant documents and passages to given biomedical questions, respectively.

In section~\ref{Chapter5.2} we described in details the proposed document retrieval method in biomedical QA. The proposed method consists of three stages, namely, query formulation, querying the MEDLINE database, and document reranking. The query formulation stage uses MetaMap to identify UMLS concepts so as to form the query from the input biomedical question. In the second stage, querying the MEDLINE database, we have used a typical IR system which retrieves a set of possibly relevant PubMed documents to the query. As there are many cases where the IR system mistakenly returns irrelevant PubMed documents high in the set or relevant PubMed documents low in the set, we have developed in the last stage a document reranking system based on UMLS similarity to re-rank the retrieved set of documents and move the most relevant documents to the top of the set. Experimental evaluations performed on large standard dataset, provided by the BioASQ challenge, have shown that the proposed method is more effective as compared with the state-of-the-art methods.

In section~\ref{Chapter5.3} we described in details the proposed passage retrieval method in biomedical QA which employs the Stanford CoreNLP sentence boundary as a passage length, stemmed words and UMLS concepts as features for the BM25 model to retrieve relevant passages in biomedical QA. For a given biomedical question,  we first have used PubMed search engine to retrieve relevant documents and reranked them based on UMLS similarity between concepts of the biomedical question and each title of the retrieved documents. We then have taken the abstracts of top-ranked documents and used Stanford CoreNLP for sentence splitter to generate a set of candidate passages. The passage length is similar to that of the Stanford CoreNLP sentence. We finally have built a new ranking of the set of candidate passages and kept the $N$ top-ranked ones using stemmed words and UMLS concepts as features for the BM25 model. Experiments conducted on large standard datasets provided by the BioASQ challenge have demonstrated the effectiveness of our method. The evaluation results show that the proposed method can achieve 6.84\% improvement over the current state-of-the-art methods in terms of MAP. 

\chapter{Answer Extraction and End-to-End Biomedical Question Answering System SemBioNLQA} 

\label{Chapter6} 
\setcounter{secnumdepth}{4}
\minitoc

This chapter presents the methods we propose for answer extraction in biomedical QA, a key task that is studied and evaluated separately. Section~\ref{Chapter6.2} will present the proposed methods for answer extraction in biomedical QA. We consecrate section~\ref{Chapter6.3} to our proposed biomedical QA system named SemBioNLQA.

\section{Introduction}
\label{Chapter6.1}

Answer extraction is usually the last component in a typical QA pipeline as the final step towards developing biomedical QA systems is processing the set of relevant passages so as to extract the final answers. Answer extraction is the most challenging task of a biomedical QA system since this is when the precise answer has to be extracted from the candidate answers retrieved and selected by the passage retrieval component. The output from the answer extraction component is a specific answer like ``Pthirus pubis'' to the biomedical question ``What is the cause of Phthiriasis Palpebrarum?''. In general, the appropriate answers to the users questions should be extracted according to the type of the given question. A biomedical question like ``Does the histidine-rich Ca-binding protein (HRC) interact with triadin?'' expects an answer of type ``yes'' or ``no''. A biomedical question like ``What is the role of edaravone in traumatic brain injury?'' expects an answer of type ``summary''. In this context, the most recent taxonomy of biomedical questions that is created by the BioASQ challenge \citep{tsatsaronis2012bioasq} consists of four types of questions which may cover all kinds of potential questions:

\begin{enumerate}
  \item Yes/No questions: They require only one of the two possible answers: ``yes'' or ``no''. For example, ``Is calcium overload involved in the development of diabetic cardiomyopathy?'' is a yes/no question and the answer is ``yes''.
  \item Factoid questions: They require a particular entity name (e.g., of a disease, drug, or gene), a number, or a similar short expression as an answer. For example, ``Which enzyme is deficient in Krabbe disease?'' is a factoid question and the answer is a single entity name ``galactocerebrosidase''.
  \item List questions: They expect a list of entity names (e.g., a list of gene names,  list of drug names), numbers, or similar short expressions as an answer. For example, ``What are the effects of depleting protein km23–1 (DYNLRB1) in a cell?'' is a list question.
  \item Summary questions: They expect a summary or short passage in return. For example, the expected answer format for the question ``What is the function of the viral KP4 protein?'' should be a short text summary.
\end{enumerate}

Since the launch of the biomedical QA track at the BioASQ challenge, theories and methods in biomedical QA continue to evolve to better meet the needs of users questions, thanks to the many editions of the BioASQ challenge. In Phase B, Task b of the BioASQ challenge, participants were asked to answer with the exact answers and the ideal answers (i.e, paragraph-sized summaries). Exact answers are only required in the case of yes/no, factoid, list, while ideal answers are expected to be returned for all biomedical questions. In this context, \cite{yang2015learning} have described a learning-based method for biomedical QA that returns only the exact answers for factoid and list questions. Similarly, \cite{peng2015fudan}} have developed a biomedical QA system that retrieves solely the exact answers for factoid and list questions. The system first used PubTator \citep{Wei_2013} for generating the candidate answers and then ranked them based on their frequency in relevant documents and snippets. \cite{choi2015snumedinfo}, on the other hand, has presented a biomedical QA system which retrieves only the ideal answers for each given question. Meanwhile, \cite{neves2015hpi} has proposed a biomedical QA system named HPI based on the IMDB database and its built-in text analysis features to generate both the exact and the ideal answers for biomedical questions. \cite{schulze2016hpi}, the winning team of the 2016 BioASQ challenge, have presented a biomedical QA system based on the LexRank algorithm \citep{erkan2004lexrank} to retrieve only the ideal answers to biomedical questions. Although the aforementioned systems have proven to be quite successful at answering biomedical questions, biomedical answer extraction still requires further efforts in order to improve its performance as the most of the aforementioned systems do not deal with all question and answer types. For instance, only few answer extraction methods for biomedical yes/no questions have been presented, compared to other question types such as factoid, list, and summary. The most recent yes/no biomedical answer extraction method is the one which was presented by \cite{neves2015hpi}. The author has made a decision on either the answer is ``yes'' or ``no'' based on the sentiment analysis predictions provided by the IMDB technology.

On the other hand, despite the importance of answering biomedical questions, until recently there are only few integral
biomedical QA systems such as the ones described in \citep{lee2006beyond,cruchet2009trust,gobeill2009question,Cao_2011,abacha2015means,Kraus_2017} that can retrieve answers to biomedical questions written in natural language. While these systems have proven to be quite successful at answering biomedical questions, they provide a limited amount of question and answer types (cf. Table~\ref{tab:3.1c}), for instance, most of them  \citep{lee2006beyond,cruchet2009trust,Cao_2011,Kraus_2017} only handle definition questions or returns solely short summaries as answers for all types of questions, and the other ones do not deal with yes/no questions which are one of the most complicated question types to answer as they are seeking for a clear ``yes'' or ``no'' answer. Furthermore, such systems still require further efforts in order to improve their performance in terms of precision to currently supported question and answer types.

In this thesis work, compared to the aforementioned systems, our ultimate goal is to develop a biomedical QA system that is able to accept a variety of natural language questions and to generate appropriate natural language answers. We will present in section~\ref{Chapter6.2} the methods we propose for the extraction of the answers to biomedical questions including yes/no questions, factoid questions, list questions, and summary questions. In section~\ref{Chapter6.3} we will present our fully automated system SemBioNLQA - Semantic Biomedical Natural Language Question Answering - which has the ability to handle the aforementioned questions that are commonly asked in the biomedical domain. SemBioNLQA is derived from our previously established methods in (1) question classification (2) document retrieval, (3) passage retrieval, and (4) answer extraction components.

\section{Answer Extraction}
\label{Chapter6.2}
\subsection{Methods}

As our goal is to develop a biomedical QA system which has the ability to deal with four types of questions (i.e., yes/no questions,
factoid questions, list questions, and summary questions), therefore, we developed novel answer extraction methods for each question type. The proposed biomedical QA system will be able to retrieve quickly user' information needs by returning exact answers (e.g., ``yes'', ``no'', a biomedical entity name, etc.) and ideal answers (i.e., paragraph-sized summaries of relevant information) for yes/no, factoid and list questions,
whereas it only provides the ideal answers for summary questions. In this section, we describe in details the methods we propose for each of the aforementioned questions \citep{Sarrouti_yes_2017,Sarrouti_bioasq_2017}.

\subsubsection{Yes/no questions}

Yes/no questions, these are questions that require either ``yes'' or ``no'' as exact answer like ``yes'' to the biomedical yes/no question ``Does the CTCF protein co-localize with cohesin?''. Even though there are only two possible answers, ``yes'' or ``no,'' such questions can be quite hard to answer due to the complicated sentiment analysis process of the candidate answer passages. In this thesis work, in order to answer yes/no questions, we first used the Stanford CoreNLP tools for tokenization and part-of-speech tagging one by one the $N$ retrieved candidate passages answers $(p_1, p_2, ... , p_{n})$. Then, each word of the candidate answer passages is assigned its SENTIWORDNET score. SENTIWORDNET \footnote{SENTIWORDNET: \url{http://sentiwordnet.isti.cnr.it/}} which is a lexical resource for sentiment analysis and opinion mining \citep{baccianella2010sentiwordnet}, assigns to each synset of WordNet three sentiment scores: ``positivity'', ``negativity'', ``objectivity''. Assuming that there are $k$ words (w) in a candidate answer passage $p$, the final sentiment score (SC) of the candidate answer passage is defined by the following equation:

\begin{equation}\label{eq:1}
SC \;(p)= \sum_{i=1}^{k} SentiWordNet \; (w_i)
\end{equation}

The key idea behind using the SentiWordNet 3.0 lexical resource is that it is the result of the automatic annotation of all the synsets of WordNet 3.0 according to the notions of ``positivity,'' ``negativity,'' and ``neutrality''. The WordNet which is a large lexical database of English, is comprised of 155,287 words and 117,659 synsets, also called synonyms \citep{Miller_1995}. Furthermore, \cite{marchand2013lvic} have shown that SentiWordNet 3.0 outperforms other sentiment lexicons in the determination of the polarity, such as Bing Liu's Opinion Lexicon \citep{Hu_2004} and MPQA Subjectivity Lexicon \citep{Wilson_2005}.

Finally, the decision to output ``yes'' or ``no'' depends on the number of positive or negative candidate answers: ``yes'' for a positive final sentiment candidate answers score and ``no'' for a negative one. Algorithm~\ref{alg:2} further illustrates how the proposed yes/no answer extraction method works.

\begin{algorithm}[h!]
\caption{Biomedical yes/no answer generator}
\label{alg:2}
\begin{algorithmic}[1]
\State $\textbf{Input} : {yes/no\; question \;Q\; and\; set\;of\;candidate\;answers\; P}$
\State $\textbf{Output} : {answer: \;``yes"\;or\; ``no"}$
\State $postive \leftarrow {0}$ \Comment{number of positive candidate answer passages}
\State $negative \leftarrow {0}$ \Comment{number of negative candidate answer passages}
\State $i \leftarrow {1}$
\Function{PreProcessing}{$p: condidate\;answer\;passage$}
\State $TOKEN [1...m]\leftarrow \Call{TokenizationAndPOSTagging}{$p$}$
\State \Return $TOKEN$
\EndFunction

\Do
  \State $W[1...m] \leftarrow  \Call{PreProcessing} {P[i]}$ \Comment{get a set of words and their POS tags of a candidate answer}
  \State $ score \leftarrow 0.0 $
  \State $j \leftarrow {1}$
  \Do
    \State $score \gets score+ \Call{SentiWordNet} {$W[j]$}$
    \State $j \gets j+1$

  \doWhile{($j\leq m$)} \Comment{m is the size of the set of words W}
   \If{($score \geq 0$)}
        \State $positive \gets positive+1$
   \Else
   \State $negative \gets negative+1$
   \EndIf
\State $i \gets i+1$
\doWhile{($i\leq np$)} \Comment{np is number of candidate answers}

\If{($positive \geq negative$)} \Comment{the decision for the answers ``yes'' or ``no'' is based on the number of positive and negative candidate answers}
        \State $output \gets ``yes"$
\Else
 \State $output \gets ``no"$
\EndIf
\end{algorithmic}
\end{algorithm}

\begin{figure}[h!]
\graphicspath{ {Figures/}}
\centering
\includegraphics [width=16cm, height=12cm]{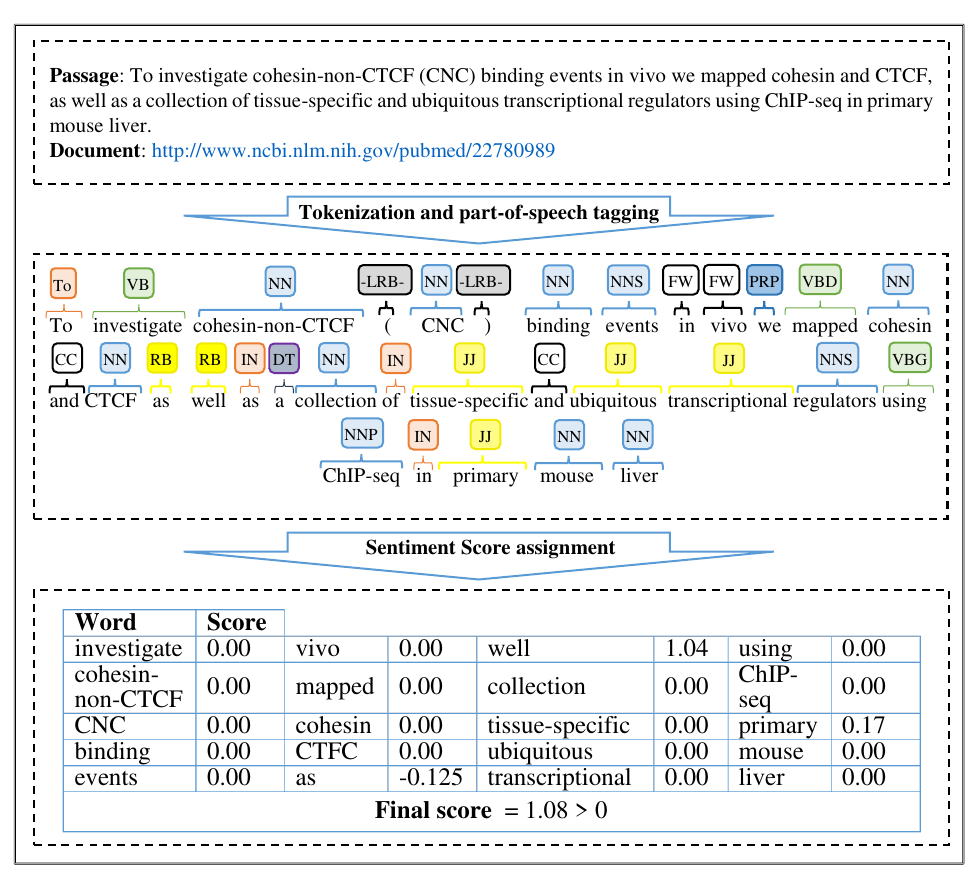}
\caption{An example of tokenization, part-of-speech tagging and sentiment score assignment.}
\label{fig:sc}       
\end{figure}

Figure~\ref{fig:sc} shows an example of the whole process, i.e, tokenization, part-of-speech tagging, and sentiment score assignment for a candidate answer passage of the biomedical yes/no question ``Does the CTCF protein co-localize with cohesin?'' from the BioASQ training datasets.

\subsubsection{Factoid questions}

Factoid questions are the questions that expect a particular entity name (e.g., of a disease, drug, or gene), a number, or a similar short expression as an answer like ``Cysteine'' to the biomedical factoid question ``Which amino acid residue appears mutated in most of the cases reported with  cadasil syndrome?''.

\begin{figure}[h!]

\graphicspath{ {Figures/}}
\centering
\includegraphics [width=16cm, height=14cm]{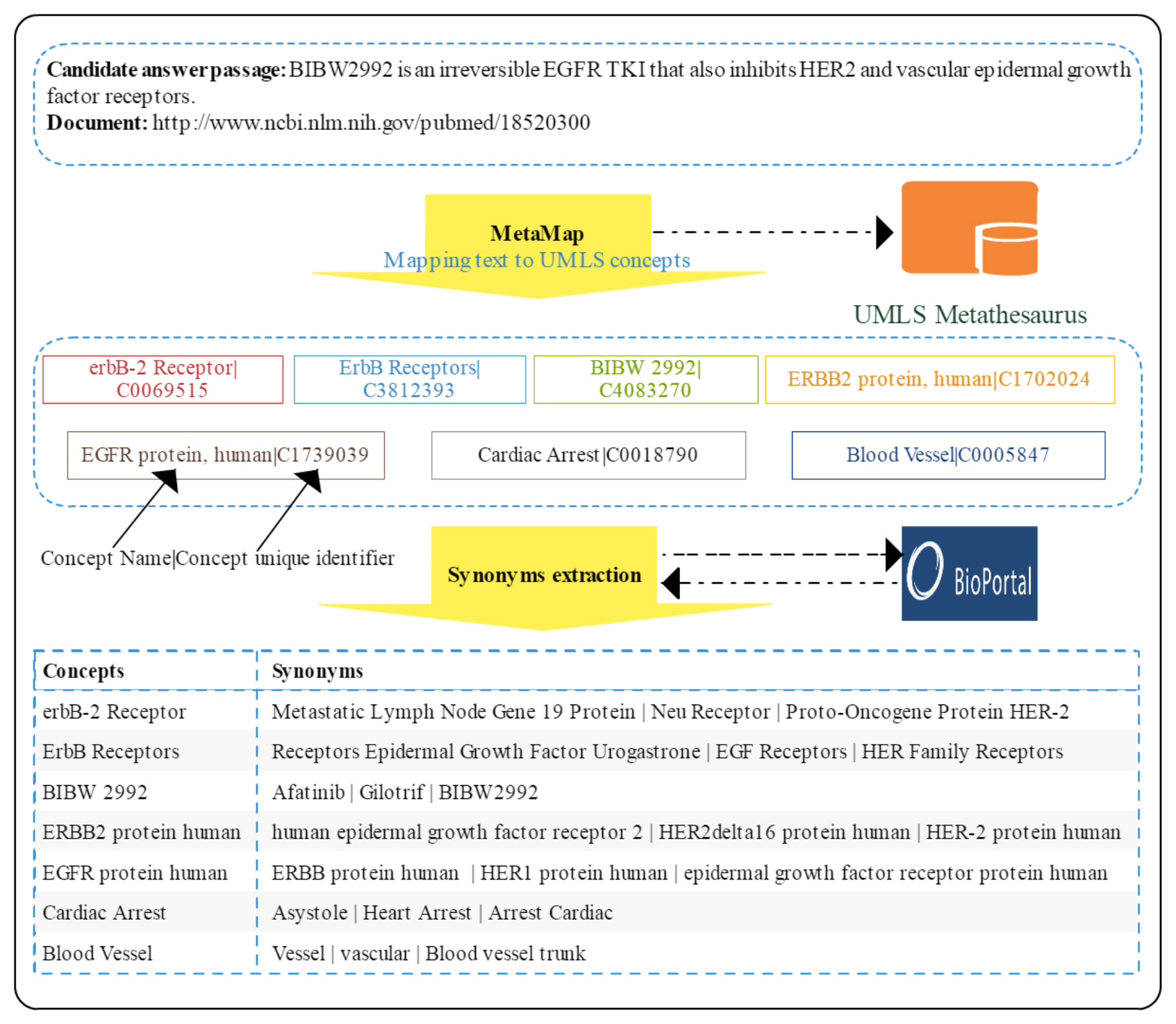}
\caption{An example of mapping to UMLS Metathesaurus and synonyms extraction.}
\label{fig:mapping}       

\end{figure}

To achieve the goal of answering factoid questions in our proposed biomedical QA system, we have proposed a factoid answer extraction method based on UMLS metathesaurus, BioPortal synonyms and the term frequency metric. We first have mapped the $N$ candidate answers $(p_1, p_2, ... , p_{n})$ of a given biomedical factoid question to the UMLS metathesaurus (2016AA knowledge source) using the MetaMap program so as to extract the set of biomedical entity names $Es$. We then have ranked the obtained set of biomedical entity names based on the term frequency metric $TF(e_i, Es)$, the number of times entity name $e_i$ appeared in the set of biomedical entity names $Es$. We have explored several term weighting methods such as TFIDF and BM25, showing term frequency achieved the best result for this task. We speculate that the answers are located in the first and second candidate answers. Next, synonyms for each of the $T$ top-ranked entity names are retrieved using Web services from BioPortal\footnote{\url{http://data.bioontology.org/documentation}}. Finally, the $T$ top-ranked biomedical entity names and their $T$ top synonyms are displayed as answers, excluding entities also mentioned in the question. The idea behind excluding entities mentioned in the question is that after analysing the training set of questions and answers released by the BioASQ organizers, we found that the most entities that appear in questions are not part of the answers. For example, the answer of the factoid question ``What is the name of Bruton's tyrosine kinase inhibitor that can be used for treatment of chronic lymphocytic leukemia?'' which contains several entities (e.g., ``Chronic Lymphocytic Leukemia''), is ``Ibrutinib'' which is not part of the question entities. As described by the BioASQ challenge, a factoid question has one correct answer, but up to five candidate answers and their synonyms are allowed. Figure~\ref{fig:mapping} shows an example of the whole process, i.e, mapping to UMLS metathesaurus, and synonyms extraction for a candidate answer of the factoid question ``Which type of lung cancer is afatinib used for?''.

\subsubsection{List questions}

List questions are the questions which require a list of entity names (e.g., a list of gene names,  list of drug names), numbers, or similar short expressions as an answer like ``bortezomib'', ``vincristine'', ``doxorubicin'', ``etoposide'', ``cisplatin'', ``fludarabine'', and ``SD-1029 Stat3 inhibitor'' to the biomedical question ``Which drugs have been found effective for the treatment of chordoma?''. The main difference between factoid and list questions is that the former require a single list of answers while the latter expect a list of lists of entity names, numbers, or similar short expressions. As it is shown by the BioASQ challenge, each entity may be accompanied by a list of synonyms. Therefore, a list of entities should be provided for each question by the proposed biomedical QA system. In other words, the exact answer is the same of factoid questions, but the interpretation is different for list questions: All $T$ top-ranked entities are considered part of the same answer for the list question, not as candidates. The proposed method used to answer list questions in our system is similar to the one described for factoid questions.

\subsubsection{Summary questions}

Summary questions are the questions which expect a summary or short passage in return like ``\emph{The histidine-rich Ca-binding protein (HRC), a 165 kDa sarcoplasmic reticulum (SR) protein, regulates SR Ca cycling during excitation contraction coupling.  HRC mutations or polymorphisms lead to cardiac dysfunction.  The Ser96Ala genetic variant of HRC is associated with life-threatening ventricular arrhythmias and sudden death in idiopathic dilated cardiomyopathy (DCM)}'' to the biomedical question ``What is the role of the histidine rich calcium binding protein (HRC) in cardiomyopathy?''. As summary questions do not require exact answers, therefore, they are simply answered in this thesis work by formulating short summaries, i.e., ideal answers, of relevant information. For the given biomedical questions, the ideal answers are formed by concatenating the two top-ranked candidate answers which were retrieved by the proposed passage retrieval approach (cf. section~\ref{Chapter5.3}). We first have forwarded abstracts of the $N$ relevant documents of a given biomedical question to Stanford CoreNLP sentence splitter so as to segment them into sentences. We then have preprocessed the obtained set of candidate answers including tokenization, removing stop words, and applying Porter' stemmer to extract stemmed words. Additionally, we have used the MetaMap program for mapping both biomedical questions and candidate passages to UMLS concepts in order to extract biomedical concepts. Moreover, the MetaMap word sense disambiguation system has been used to resolve ambiguities in the texts by identifying the meaning of ambiguous terms. Using stemmed words and UMLS concepts as features, we finally ranked the set of candidate answers using BM25 as retrieval model, and concatenated the two top-ranked candidate answer passages.

In particular, in addition to the \emph{exact answers} returned for previous questions, i.e., yes/no questions, factoid questions and list questions, we also provide \emph{ideal answers}. Therefore, our proposed biomedical QA system provides both \emph{exact answers} and the \emph{ideal answers} for yes/no, factoid and list questions, whereas it only provides \emph{ideal answers} for summary questions.

\subsection{Experimental results}

In order to assess the effectiveness of the methods we proposed for biomedical answer extraction and compare with the current state-of-the-art methods, we performed several experiments on large standard datasets provided by the BioASQ challenge. We have used the test sets of biomedical questions provided in BioASQ Task 3b 2015, BioASQ Task 4b 2016, and BioASQ Task 5b 2017 described in subsection~\ref{Chapter3_7_1}, section~\ref{Chapter3_7}, chapter~\ref{Chapter3}. Moreover, we have participated in the 2017 BioASQ challenge (BioASQ Task 5b 2017).

The BioASQ challenge in phase B of Task 3b/4b/5b provides the test set of biomedical questions along with their golden documents, golden snippets, and questions types, i.e., whether yes/no, factoid, list or summary in order to evaluate biomedical answer extraction approaches in their best way. Given biomedical question and its golden passages, each participating system may return an ideal answer, i.e., a paragraph-sized summary of relevant information. In the case of yes/no, factoid, and list questions, the systems may also return exact answers; for summary questions, no exact answers will be returned. Accordingly, we have relied on the gold-standard passages provided by the BioASQ challenge, instead of the ones retrieved by the system to evaluate our answer extraction methods.

As indicators of answer extraction effectiveness: Accuracy was used for exact answers of yes/questions; mean reciprocal rank (MRR) was used for exact answer of factoid questions; mean average precision, mean average recall, and mean average f1-measure were used for exact answers of list questions; ROUGE-2 and ROUGE-SU4 were used for ideal answers. These evaluation metrics are described in details in subsection~\ref{Chapter3_7_2}, section~\ref{Chapter3_7}, chapter~\ref{Chapter3}. Additionally, the BioASQ challenge have also developed an online evaluation system\footnote{BioASQ evaluation system: \url{http://participants-area.bioasq.org/oracle/}} that allows uploading JSON result files and obtaining evaluations results at any time. Table~\ref{tab:6.1} and Table~\ref{tab:6.2} show the experimental results of the
proposed answer extraction methods and comparison with the state-of-the-art studies presented in \citep{zhang2015fudan,neves2015hpi,choi2015snumedinfo,yang2015learning,schulze2016hpi} on five batches of testing datasets provided by the BioASQ challenge in 2015 and 2016, respectively. In addition, Table~\ref{tab:6.3} and Table~\ref{tab:6.4} present the results of our participation in Phase B, Task 5b of the 2017 BioASQ challenge using our biomedical answer extraction system. The values inside parameters indicate our current rank, the total number of submissions, and the total number of participated teams for the task. Our system name for submission was ``sarrouti''.

\begin{table}[h!]
\centering
\caption[The overall results of the proposed biomedical answer extraction methods and comparison with the current state-of-the-art
methods on five batches of testing datasets provided by BioASQ 3b 2015.]{The overall results of the proposed biomedical answer extraction methods and comparison with the current state-of-the-art methods on five batches of testing datasets provided by BioASQ 3b 2015. The ``-'' replace the scores of systems that did not evaluate on this batch or did not deal with this task, while ``nr'' indicated that the results are not reported for this evaluation measure. Acc, P, R, F, R-2, R-SU4 indicate accuracy, precision, recall, and f-measure, rouge-2, rouge-SU2, respectively.}
\label{tab:6.1}
\begin{tabular}{p{1.6cm}p{3.6cm}p{1.1cm}p{1cm}p{1cm}p{1cm}p{1cm}p{1cm}p{1.2cm}}
\hline\noalign{\smallskip}
 \multirow{3}{*}{Datasets} &\multirow{3}{*}{System name}& \multicolumn{5}{c}{Exact answers} &  \multicolumn{2}{c}{\multirow{2}{*}{Idial answers}}\\
\cmidrule(l){3-7}

 & & Yes/No & Factoid & \multicolumn{3}{c}{ List}  &\multicolumn{2}{c}{}   \\
 \cmidrule(l){3-3}\cmidrule(l){4-4} \cmidrule(l){5-7} \cmidrule(l){8-9}
  & &  Acc &  MRR& P& R &F1  & R-2& R-SU4\\

\noalign{\smallskip}\hline\noalign{\smallskip}
\multirow{4}{*}{Batch 1} &	Our system & 0.6970&	0.1692	& 0.1545&	0.2409&	0.1830& 0.2716&	0.2860\\
            &\cite{zhang2015fudan}&-&	0.1423	& nr & nr & 0.0756& -& -\\
            &\cite{neves2015hpi}&0.6667&	-	& 0.0292& 0.0603& 0.0364& 0.1884& 0.2008\\
            &\cite{choi2015snumedinfo}&-&	-	& -& -& -& -& 0.3071\\
\cmidrule(l){1-9}
\multirow{4}{*}{Batch 2} &	Our system &0.6250&	0.1776	& 0.1929&	0.2714&	0.2127& 0.3123&	0.3364\\
           &\cite{zhang2015fudan}&-&	0.0859	& nr&nr& 0.1160& -& -\\
            &\cite{neves2015hpi}&0.5625&	-	& 0.0714& 0.0161& 0.0262& 0.2026& 0.2227\\
            &\cite{choi2015snumedinfo}&-&	-	& -& -& -& -& 0.3710\\
            \cmidrule(l){1-9}
\multirow{4}{*}{Batch 3} &	Our system &0.8621&	0.1840	& 0.2353&	0.2927&	0.2524& 0.3879&	0.4078\\
            &\cite{zhang2015fudan}&-&	0.0846	& nr& nr& 0.1319& -& -\\

            &\cite{yang2015learning}&-&	0.1615&	0.0539&  0.6933 & 0.0969 & -& -\\

            &\cite{neves2015hpi}&0.6207&	-	& -& -& -& 0.1934& 0.2189\\
            &\cite{choi2015snumedinfo}&-&	-	& -& -& -& -& 0.3941\\
            \cmidrule(l){1-9}
\multirow{4}{*}{Batch 4} &	Our system &0.7600&	0.2960	& 0.2783&	0.2713&	0.2588& 0.3917&	0.4108\\
           &\cite{zhang2015fudan}&-&	0.2299	& nr& nr& 0.2192& -& -\\

           &\cite{yang2015learning}&-&	0.5155&	0.3836&  0.3480 & 0.3168 & -& -\\

            &\cite{neves2015hpi}&0.5600&	0.0345	& 0.1522& 0.0473& 0.0689& 0.2504& 0.2724\\
            &\cite{choi2015snumedinfo}&-&	-	& -& -& -& -& 0.3906\\
            \cmidrule(l){1-9}
\multirow{4}{*}{Batch 5} &	Our system &0.6071&	0.1568	& 0.0583& 0.0736& 0.0625& 0.3440&	0.3533\\
            &\cite{zhang2015fudan}&-&	0.2500	& nr& nr& 0.1340& -& -\\

            &\cite{yang2015learning}&-&	0.2727 &	0.1704&  0.2573 & 0.1875 & -& -\\

            &\cite{neves2015hpi}&0.3571&	0.0909	& 0.0625& 0.0292& 0.0397& 0.1694& 0.1790\\
            &\cite{choi2015snumedinfo}&-&	-	& -& -& -& -& 0.3665\\
\noalign{\smallskip}\hline
\end{tabular}
\end{table}

\begin{table}[h!]
\centering
\caption[The overall results of the proposed biomedical answer extraction methods and comparison with the current state-of-the-art
methods on five batches of testing datasets provided by BioASQ 4b 2016]{The overall results of the proposed biomedical answer extraction methods and comparison with the current state-of-the-art methods on five batches of testing datasets provided by BioASQ 4b 2016. The ``-'' replace the scores of systems that did not deal with this task, while ``nr'' indicated that the results are not reported for this evaluation measure. Acc, P, R, F, R-2, R-SU4 indicate accuracy, precision, recall, and f-measure, rouge-2, rouge-SU2, respectively.}
\label{tab:6.2}
\begin{tabular}{p{1.6cm}p{3.6cm}p{1.1cm}p{1cm}p{1cm}p{1cm}p{1cm}p{1cm}p{1.2cm}}
\hline\noalign{\smallskip}
 \multirow{3}{*}{Datasets} &\multirow{3}{*}{System name}& \multicolumn{5}{c}{Exact answers} &  \multicolumn{2}{c}{\multirow{2}{*}{Idial answers}}\\
\cmidrule(l){3-7}

 & & Yes/No & Factoid & \multicolumn{3}{c}{ List}  &\multicolumn{2}{c}{}   \\
 \cmidrule(l){3-3}\cmidrule(l){4-4} \cmidrule(l){5-7} \cmidrule(l){8-9}
  & &  Acc &  MRR& P& R &F1  & R-2& R-SU4\\

\noalign{\smallskip}\hline\noalign{\smallskip}
\multirow{2}{*}{Batch 1} &	Our system &0.8214&	0.0726	& 0.2182&	0.3939&	0.2756& 0.4772&	0.4918\\
            &\cite{schulze2016hpi}&-&	-	& -& -& -& nr& 0.2231\\

\cmidrule(l){1-9}
\multirow{2}{*}{Batch 2} &	Our system &0.8750&	0.1452	& 0.2381&	0.2505&	0.2349& 0.5021&	0.5115\\
             &\cite{schulze2016hpi}&-&	-	& -& -& -& nr&0.2240 \\

            \cmidrule(l){1-9}
\multirow{2}{*}{Batch 3} &	Our system &0.8400&	0.1218	& 0.2381&	0.3627&	0.2812& 0.4978&	0.5061\\
            &\cite{schulze2016hpi}&-&	-	& -& -& -& nr& 0.2559\\

            \cmidrule(l){1-9}
\multirow{2}{*}{Batch 4} &	Our system &0.8095&0.1129&	0.1467&	0.2231&	0.1702& 0.5192&	0.5231\\
            &\cite{schulze2016hpi}&-&	-	& -& -& -& nr& 0.2280\\

            \cmidrule(l){1-9}
\multirow{3}{*}{Batch 5} &	Our system &0.8148& 0.1136&	0.1900&	0.2353&	0.1963& 0.4979&	0.5027\\
            &\cite{schulze2016hpi}&-&	-	& -& -& -& nr& 0.3233\\

\noalign{\smallskip}\hline
\end{tabular}
\end{table}

\begin{table}[h!]
\centering
\caption[The obtained results of our participation in ``Exact Answers'', Phase B, Task 5b of the 2017 BioASQ challenge using the proposed answer extraction methods]{The obtained results of our participation in ``Exact Answers'', Phase B, Task 5b of the 2017 BioASQ challenge using the proposed answer extraction methods. The first value inside parameters indicates our current rank and the total number
of submissions for the task, while the second indicates our current rank and the total number of participated teams.}
\label{tab:6.3}
\begin{tabular}{p{2cm}p{2.6cm}p{2.6cm}p{2cm}p{2cm}p{2.5cm}}
\hline\noalign{\smallskip}
 \multirow{2}{*}{Datasets} &  Yes/No & Factoid & \multicolumn{3}{c}{ List}   \\
 \cmidrule(l){2-2}\cmidrule(l){3-3} \cmidrule(l){4-6}
  &   Accuracy &  MRR& Precision& Recall &F-measure \\

\noalign{\smallskip}\hline\noalign{\smallskip}
\multirow{2}{*}{Batch 1}  &0.7647&	0.2033 (5/15)\newline (3/9)&	0.1909&	0.2658&	0.2129 (3/15)\newline (2/9) \\

\cmidrule(l){1-6}
\multirow{2}{*}{Batch 2}  &0.7778& 0.0887 (10/21)\newline (5/9)&	0.2400&	0.3922&	0.2920 (6/21)\newline (2/9)\\

            \cmidrule(l){1-6}
\multirow{2}{*}{Batch 3}  &0.8387 (1/21)\newline (1/10)& 0.2212 (9/21)\newline (4/10)&	0.2000&	0.4151&	0.2640 (6/21)\newline (3/10)\\

            \cmidrule(l){1-6}
\multirow{2}{*}{Batch 4} &0.6207 (2/27)\newline (2/11)& 0.0970 (13/27)\newline (5/11)&	0.1077&	0.2013&	0.1369 (12/27)\newline (5/11)\\

            \cmidrule(l){1-6}
\multirow{3}{*}{Batch 5} &0.4615 & 0.2071 (9/25)\newline (3/11)&	0.2091&	0.3087&	0.2438 (11/25)\newline (6/11)\\

\noalign{\smallskip}\hline
\end{tabular}
\end{table}

\begin{table}[h!]
\centering
\caption[The obtained results of our participation in ``Ideal Answers'', Phase B, Task 5b of the 2017 BioASQ challenge using the proposed answer extraction methods]{The obtained results of our participation in ``Ideal Answers'', Phase B, Task 5b of the 2017 BioASQ challenge using the proposed answer extraction methods. The first value inside parameters indicates our current rank and the total number
of submissions for the task, while the second indicates our current rank and the total number
of participated teams.}
\label{tab:6.4}
\begin{tabular}{p{2.1cm}p{1.7cm}p{2.2cm}p{2.2cm}p{1.3cm}p{1.9cm}p{1.9cm}}
\hline\noalign{\smallskip}
 \multirow{2}{*}{Datasets} & \multicolumn{2}{c}{Automatic scores} &  \multicolumn{4}{c}{Manual scores}\\
\cmidrule(l){2-3}\cmidrule(l){4-7}

 & Rouge-2& Rouge-SU4 & Readability&  Recall& Precision& Repetition \\

\noalign{\smallskip}\hline\noalign{\smallskip}
\multirow{2}{*}{Batch 1}  &0.4943 \newline(4/15)\newline(2/9)&0.5108\newline (3/15)\newline(2/9)&3.65 \newline(3/15)\newline(2/9)&	4.42 \newline (3/15)\newline(2/9)&	3.90 \newline(3/15)\newline(2/9)&	3.89\newline (3/15)\newline(2/9)\\

\cmidrule(l){1-7}
\multirow{2}{*}{Batch 2}  & 0.4579 \newline(4/21)\newline (2/9)& 0.4583\newline (4/21)\newline (2/9)&3.68\newline (6/21)\newline (2/9)&	4.59 \newline (2/21)\newline (2/9)&	4.01\newline (7/21)\newline (2/9)&	3.91 \newline(7/21)\newline (2/9)\\

            \cmidrule(l){1-7}
\multirow{2}{*}{Batch 3}  & 0.5566 \newline(4/21)\newline (2/10)&	0.5656 \newline(4/21)\newline (2/10)&3.91\newline (6/21)\newline (2/10)&	4.64 \newline (1/21)\newline (2/10)&	4.07 \newline(7/21)\newline (2/10)&	4.00 \newline(7/21)\newline (2/10)\\

            \cmidrule(l){1-7}
\multirow{2}{*}{Batch 4} &0.5895 \newline(4/27)\newline (2/11)&	0.5832 \newline(4/27)\newline (2/11)&3.86\newline (7/27)\newline (3/11)&	4.51 \newline (6/27)\newline (3/11)&	4.02 (3/27)\newline (2/11)&	3.95 \newline(6/27)\newline (2/11)\\

            \cmidrule(l){1-7}
\multirow{3}{*}{Batch 5} & 0.5772 \newline(7/25)\newline (3/11)&	0.5756\newline (7/25)\newline (3/11)&3.82\newline (8/25)\newline (3/11)&	4.53 \newline(5/25)\newline (2/11)	&3.91\newline (7/25)\newline (3/11)&	3.90 \newline(7/25)\newline (2/11)\\

\noalign{\smallskip}\hline
\end{tabular}
\end{table}

The proposed answer extraction methods provide both the exact and ideal answer to biomedical questions. For yes/no questions, the exact answer (``yes'' or ``no'') is formed by using SentiWordNet, a sentiment lexicon. Each word of the relevant snippets is assigned its SentiWordNet score, and the decision to output ``yes'' or ``no'' depends on the number of positive or negative snippets. For factoid questions, the exact answer is produced by identifying the biomedical entities that occur in the given top relevant snippets of the question, and reporting the five most frequent biomedical entities and their synonyms of the top snippets, excluding entities also mentioned in the question. For list questions, the exact answer is produced in the same manner, except that the most frequent entities and their synonyms of the top relevant snippets are now returned as the single list that answers the question. Note that in contrast to the 2015 and 2016 BioASQ challenges, biomedical QA systems are no longer allowed to provide an own list of synonyms in the 2017 challenge. On the other hand, to generate ideal answers (summaries),
we have applied MetaMap to the relevant snippets and the question, in order to obtain the UMLS concepts they refer to. We then have ranked the snippets by their BM25 similarity to the question (using Porter stems and UMLS concepts as features), and return the top two most highly ranked snippets (concatenated) as the ideal answer.

From the overall results, it can be seen that in each of the BioASQ challenges, the proposed biomedical answer extraction methods were very competitive and performed well in all challenges ranking within the top tier teams. Moreover, our system was one of the winners\footnote{Fifth BioASQ challenge winners:\url{http://www.bioasq.org/participate/fifth-challenge-winners}} in the 2017 edition of the BioASQ challenge.


\subsection{Discussion}

Although open-domain QA is a longstanding challenge widely studied over the last decades, few systems are currently able to handle a variety of natural language questions and to generate the appropriate answers.  In this thesis work, we have proposed answer extraction methods in biomedical QA to handle the kinds of yes/no questions, factoid questions, list questions, and summary questions. Our methods are able to provide exact answers and paragraph-sized ideal answers (summaries of relevant information) for yes/no, factoid and list questions, whereas they only retrieve ideal answers for summary questions.

The experimental results detailed in Table~\ref{tab:6.1} and Table~\ref{tab:6.2} have shown that the proposed biomedical answer extraction methods are more competitive as compared with the current state-of-the-art methods. As can be seen from Table~\ref{tab:6.1}, compared with the two methods presented in \citep{neves2015hpi,zhang2015fudan}, one based on the in-memory database and another using the PubTator tool, our methods significantly outperformed the aforementioned methods in extracting both the exact answers and the ideal answers for yes/no, factoid, list and summary biomedical questions. Moreover, the increased performance was statistically significant (the P-value is 7.6e-05, the result is significant at p < 0.01). In particular, it can be seen clearly from Table~\ref{tab:6.1} that in all batches of testing datasets, the proposed
yes/no answer extraction method achieves performance improvements over the state-of-the-art method
presented in \citep{neves2015hpi}. Compared with the latter, which employs the sentiment analysis predictions
provided by the IMDB database, the proposed method gives better results (an average improvement of 15.68\% in terms of accuracy). Moreover, the increase performance is statistically significant (the p-value is < 0.00001, the result is significant at p < 0.01).

Additionally, the proposed systems is also competitive compared with the system proposed in \citep{choi2015snumedinfo}, which dealt only with the ideal answers (0.2860 against 0.3071, 0.3364 against 0.3710, 0.4078 against 0.3941, 0.4108 against 0.3906, and 0.3533 against 0.3665 of Rouge-SU4 in batch 1, batch 2, batch 3, batch 4, and batch 5 respectively).

In the 2016 BioASQ challenge, as shown in Table~\ref{tab:6.2}, the proposed answer extraction system still achieves good performance compared with the 2016 winning system developed by \citep{schulze2016hpi} which dealt solely with the ideal answers of questions. The latter is based on the LexRank algorithm \citep{erkan2004lexrank}, but that solely used the named entities for the similarity function. The important thing to note here is that our system significantly outperforms the \citep{schulze2016hpi} system in all batches of testing datasets. The largest difference in ROUGE-SU4 between SemBioNLQA and the aforementioned system was 0.2951 (0.5231 - 0.2280 in batch 4) which clearly indicates that our method is not only effective but robust in extracting the ideal answers.

On the other hand, as part of our participation in Phase B, Task 5B of the 2017 BioASQ challenge, the proposed answer extraction methods performed well in the challenge ranking within the top tier teams as shown in Table~\ref{tab:6.3} and Table~\ref{tab:6.4}. More details on the results can be found in the official web site of BioASQ\footnote{BioASQ Task5b phase b\url{http://participants-area.bioasq.org/results/5b/phaseB/}}. A total of 16, 22, 21, 27, and 25 runs were submitted for batch 1, batch 2, batch 3, batch 4, batch 5 respectively. Note that many runs were submitted by the same teams. As shown, our team is ranked among the top three. In batch 1, it achieved the third and the fifth
position within the 15 participating systems in extracting the exact answers of list and factoid questions respectively. More specifically, our system obtained the second and the third position when considering results by teams, instead of each individual run. In batch 2, considering results by teams, our system obtained the second and the fourth position in extracting the exact answers of list and factoid questions respectively. It can also be seen that in batch 3 and batch 4, our system, achieved the first and the second position respectively for answering
yes/no questions. Because of a very skewed class distribution in other batches, we have not compared our yes/no answer extraction method with
the participant systems which always answering ``yes''. For the ideal answers, our system in terms of ROUGE-2 achieved the fourth position compared to the 15, 21, and 21 participating systems in batch 1, batch 2 and batch 3 respectively, while in terms of ROUGE-SU4, the proposed system obtained the third position in batch 1 and the fourth position in batch 2. Besides, considering results by teams, instead of each individual run, our systems achieved the second position in terms of ROUGE-2 and ROUGE-SU4 in batches 1-4, whereas it achieved the third position in batch 5. Based on the manual scores in terms of readability, recall, precision, and repetition calculated by the BioASQ experts for each participating systems, our system achieved the second position in batch 1, batch 2, batch 3, batch 5, while it achieved the third position in batch 4. Overall, our system was one of the fifth BioASQ challenge winners\footnote{BioASQ challenge winners: \url{http://www.bioasq.org/participate/fifth-challenge-winners}}. This proves that the proposed answer extraction system could effectively identify the ideal answers to a given biomedical question.

Although the proposed answer extraction system could effectively answer a variety of biomedical questions, we found that there are still some mistakes that the proposed system cannot fix. An example is that, when the question is ``Does HER2 under-expression lead to favorable response to trastuzumab?,'' (identifier 51542eacd24251bc05000084) from BioASQ training questions, a positive answer might be ``Trastuzumab is a monoclonal antibody targeted to the Her2 receptor and approved for treatment of Her2 positive breast cancer,'' the sentiment score considers this answer to be a positive as there are both ``positive'' and ``approved'' in the passage. However, this example should be counted as a ``no'' answer because trastuzumab is effective only in cancers where Her2 is over-expressed. Dealing with such problems requires more complex semantic analysis, and we may need to parse the passage to get a grammar tree. Nevertheless, parsing and semantic analysis will bring in new errors and make this challenge even more complicated. We also found that the current form of the proposed system was not able to provide answers to some questions especially for these which expect a number as answer instead of biomedical entities. For instance, the answers for the biomedical question ``What is the prevalence of short QT syndrome?'' (identifier 52fb78572059c6d71c000067)  and ``What is the number of protein coding genes in the human genome?'' (identifier 535d3c069a4572de6f000006) collected from BioASQ training questions,  are ``0.01\% -0.1\%'', ``Between 20,000 and 25,000'', respectively. Such questions seem to be quite complicate and need more specific information extraction methods.

\section{A Semantic Biomedical Question Answering System SemBioNLQA}
\label{Chapter6.3}
\subsection{Methods}

\begin{figure}[h!]
\captionsetup{justification=justified}
\graphicspath{{Figures/}}
\centering

\includegraphics [width=16cm, height=14cm]{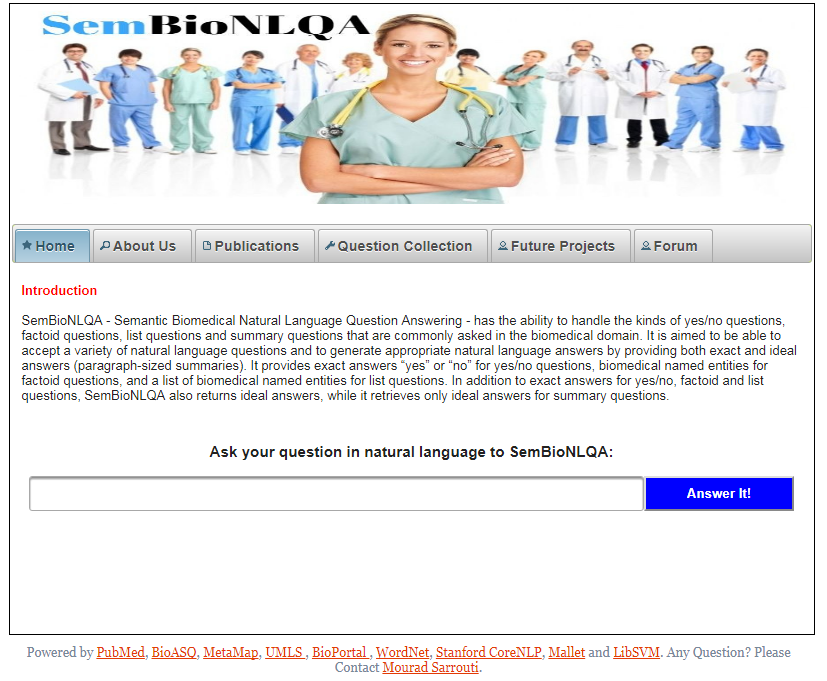}
\caption{The proposed semantic biomedical question answering system SemBioNLQA}
\label{fig:interface}
\end{figure}

In this section, we present the development, generic architecture and integrated components of our fully automated system SemBioNLQA - Semantic Biomedical Natural Language Question Answering - which has the ability to handle the kinds of yes/no questions, factoid questions, list questions and summary questions that are commonly asked in the biomedical domain. Figure~\ref{fig:interface} presents the SemBioNLQA Web system.

\begin{figure}[h!]
\captionsetup{justification=justified}
\graphicspath{{Figures/}}
\centering
\includegraphics[width=16cm, height=18cm]{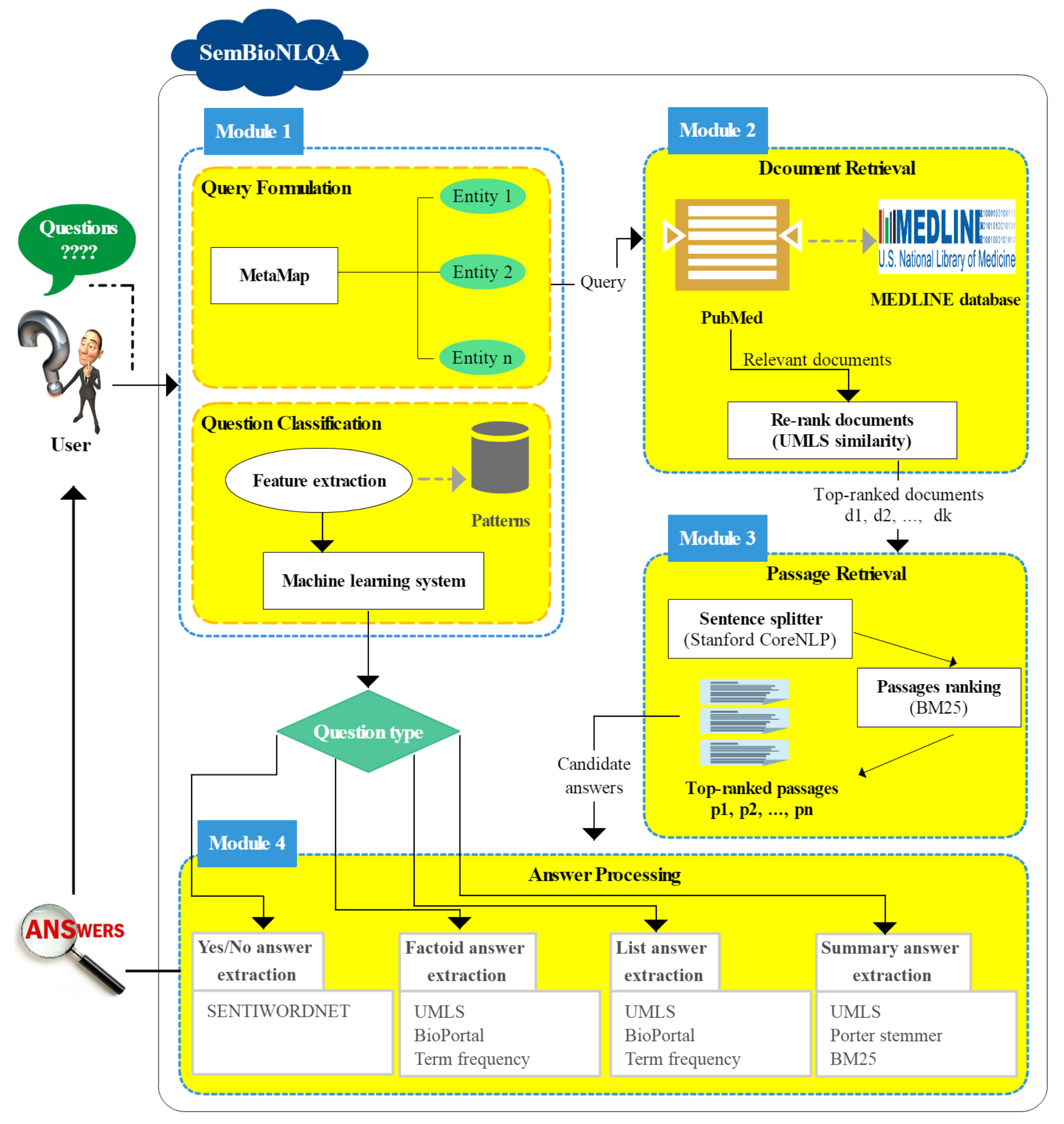}
\caption[Generic architecture of the proposed biomedical QA system SemBioNLQA]{Generic architecture of the proposed biomedical QA system SemBioNLQA}
\label{fig:QAA}
\end{figure}

The SemBioNLQA system, which consists of question classification, document retrieval, passage retrieval and answer extraction components, takes natural language questions as input, and outputs both \emph{exact answers} and \emph{ideal answers} as results.  SemBioNLQA is able to accept a variety of natural language questions and to generate appropriate natural language answers by providing both exact and ideal answers. It provides exact answers of type ``yes'' or ``no'' for yes/no questions, biomedical named entities for factoid questions, and a list of biomedical named entities for list questions. In addition to exact answers for yes/no, factoid and list questions, SemBioNLQA also returns ideal answers, while it retrieves only the ideal answers for summary questions. SemBioNLQA is derived from our previously established methods  in (1) question classification (cf. section~\ref{Chapter4.2}, chapter~\ref{Chapter4})  (2) document retrieval  (cf. section~\ref{Chapter5.2}, chapter~\ref{Chapter5}), (3) passage retrieval (cf. section~\ref{Chapter5.3}, chapter~\ref{Chapter5}), and (4) answer extraction system (cf. section~\ref{Chapter6.2}) which was one of the winners in the 2017 BioASQ challenge. Indeed, we developed the SemBioNLQA system based on the integration of these methods and techniques. Figure~\ref{fig:QAA} shows the architecture of SemBioNLQA and its main components.

SemBioNLQA first takes as its input a natural language biomedical question and includes preprocessing of the question, identification of the question type and the expected answer format to be required based on handcrafted lexico-syntactic patterns and support vector machine, as well as building a query from the question using UMLS entities to be fed into our document retrieval system based on PubMed and UMLS similarity. A document retrieval system is used to retrieve documents satisfying the query from the MEDLINE database. Then, it extracts relevant passages from top-ranked documents based on the BM25 model, stemmed words and UMLS concepts. Finally, it generates and returns both ``exact'' (depending on the expected answer for each question type) and paragraph-sized ``ideal'' answers from these passages based on the UMLS metathesaurus, BioPortal synonyms, SENTIWORDNET, term frequency metric and BM25 model. According to the QA classification approach presented by \cite{athenikos2010biomedical}, the SemBioNLQA can be classified as semantics-based biomedical QA.

\subsection{Experimental results}

In order to assess the effectiveness of the SemBioNLQA system and compare it with the current integral biomedical QA systems presented in \citep{gobeill2009question,Cao_2011,Kraus_2017}, we present two different evaluations: (1) a systematic/automatic evaluation on benchmark datasets provided by the BioASQ challenges, and (2) a manual evaluation in terms of quality of the answers using BioASQ training questions and answers.

\subsubsection{Systematic evaluation}
In this experiment, we present a systematic evaluation on real biomedical questions provided by the BioASQ challenge in its 2017 edition so as to compare with Olelo, the most current biomedical QA system. Fortunately, the Olelo system has participated in Task 5b, phase B of the 2017 BioASQ challenge which enabled us to compare with them as we also have participated in the challenge. Please note that in this experiment, both systems SemBioNLQA and Olelo relied only on the gold-standard passages provided by BioASQ, instead of the ones retrieved by the systems.

As indicators of answer extraction effectiveness: Accuracy was used for exact answers of yes/questions; mean reciprocal rank (MRR) was used for exact answer of factoid questions; mean average precision, mean average recall, and mean average f-measure were used for exact answers of list questions; ROUGE-2 and ROUGE-SU4 were used for ideal answers. These evaluation metrics are described in details in subsection~\ref{Chapter3_7_2}, section~\ref{Chapter3_7}, chapter~\ref{Chapter3}. Table~\ref{tab:6.3.1} shows the experimental results of SemBioNLQA and comparison with Olelo on five batches of testing datasets provided by the 2017 BioASQ organizers during our participation. Please also note that we do not compare with other BioASQ participants in this work since they have not published the integral systems yet. As we previously noted, the BioASQ challenges in phase B of Task b provide the test set of biomedical questions along with their golden documents, golden snippets, and questions types. Therefore, participants do not require a fully QA to participate in the challenge. More details on the results can be found in the BioASQ web site\footnote{\url{http://participants-area.bioasq.org/results/5b/phaseB/}}. Our system name for submission was ``sarrouti''.

\begin{table}[h!]
\centering
\caption[The overall results of SemBioNLQA and comparison with Olelo on five batches of testing datasets provided by BioASQ 5b 2017]{The overall results of SemBioNLQA and comparison with Olelo on five batches of testing datasets provided by BioASQ 5b 2017. The ``-'' and ``nr'' indicate that the system did not deal with this task and the results are not reported for this evaluation measure, respectively. Acc, P, R, F, R-2, R-SU4 indicate accuracy, precision, recall, and f-measure, rouge-2, rouge-SU2, respectively.}
\label{tab:6.3.1}
\begin{tabular}{p{1.6cm}p{3.6cm}p{1.1cm}p{1cm}p{1cm}p{1cm}p{1cm}p{1cm}p{1.2cm}}
\hline\noalign{\smallskip}
 \multirow{3}{*}{Datasets} &\multirow{3}{*}{System name}& \multicolumn{5}{c}{Exact answers} &  \multicolumn{2}{c}{\multirow{2}{*}{Idial answers}}\\
\cmidrule(l){3-7}

 & & Yes/No & Factoid & \multicolumn{3}{c}{ List}  &\multicolumn{2}{c}{}   \\
 \cmidrule(l){3-3}\cmidrule(l){4-4} \cmidrule(l){5-7} \cmidrule(l){8-9}
  & &  Acc &  MRR& P& R &F1  & R-2& R-SU4\\
\noalign{\smallskip}\hline\noalign{\smallskip}
\multirow{2}{*}{Batch 1} &	SemBioNLQA & 0.7647&	0.2033	& 0.1909& 0.2658& 0.2129&0.4943& 0.5108 \\
                         &	Olelo & -&	0.0400&nr&nr&	 0.0477 & 0.2958& 0.3243\\

\cmidrule(l){1-9}
\multirow{2}{*}{Batch 2} &	SemBioNLQA &0.7778& 0.0887&	0.2400&	0.3922&	0.2920& 0.4579& 0.4583\\
                         &	Olelo & -&	0.0323&nr&nr&	 0.0287& 0.2048& 0.2500\\

            \cmidrule(l){1-9}
\multirow{2}{*}{Batch 3} &	SemBioNLQA &0.8387& 0.2212&	0.2000&	0.4151&	0.2640 & 0.5566 &	0.5656 \\
                         &	Olelo & -&	 0.0192 &nr&nr&	 0.0549& 0.2891& 0.3262\\
            \cmidrule(l){1-9}
\multirow{2}{*}{Batch 4} &	SemBioNLQA &0.6207& 0.0970&	0.1077&	0.2013&	0.1369& 0.5895&	0.5832 \\
                         &	Olelo & -&	0.0513 &nr&nr&0.0513& 0.3460& 0.3516\\

            \cmidrule(l){1-9}
\multirow{2}{*}{Batch 5} &	SemBioNLQA &0.4615 & 0.2071&	0.2091&	0.3087&	0.2438& 0.5772 &	0.5756\\
                         &	Olelo & -&	-&nr&nr&	 0.0379& 0.2117& 0.2626\\

\noalign{\smallskip}\hline
\end{tabular}
\end{table}

In particular, we also report the overall end-to-end evaluation results of the SemBioNLQA system on BioASQ datasets provided by the challenge in 2015 and 2016 editions in order to demonstrate the effectiveness of the SemBioNLQA and also to make new comparisons easier. Table~\ref{tab:6.3.2} highlights the obtained results on BioASQ 3b 2015 and BioASQ 4b 2016 datasets. All answers returned by the SemBioNLQA system on either BioASQ 3b 2015 and BioASQ 4b 2016 datasets are available for download\footnote{\url{https://sites.google.com/site/mouradsarrouti/datasets}}. Concretely, this experiment aims to answer the following questions:

\begin{enumerate}
  \item Is SemBioNLQA able to achieve improvements over the existing biomedical QA systems?
  \item Do question classification, document retrieval and passage retrieval components have an impact on the overall performance of the SemBioNLQA system?
\end{enumerate}

\begin{figure}[h!]
\captionsetup{justification=justified}
\graphicspath{{Figures/}}
\centering

\includegraphics [width=16cm, height=22cm]{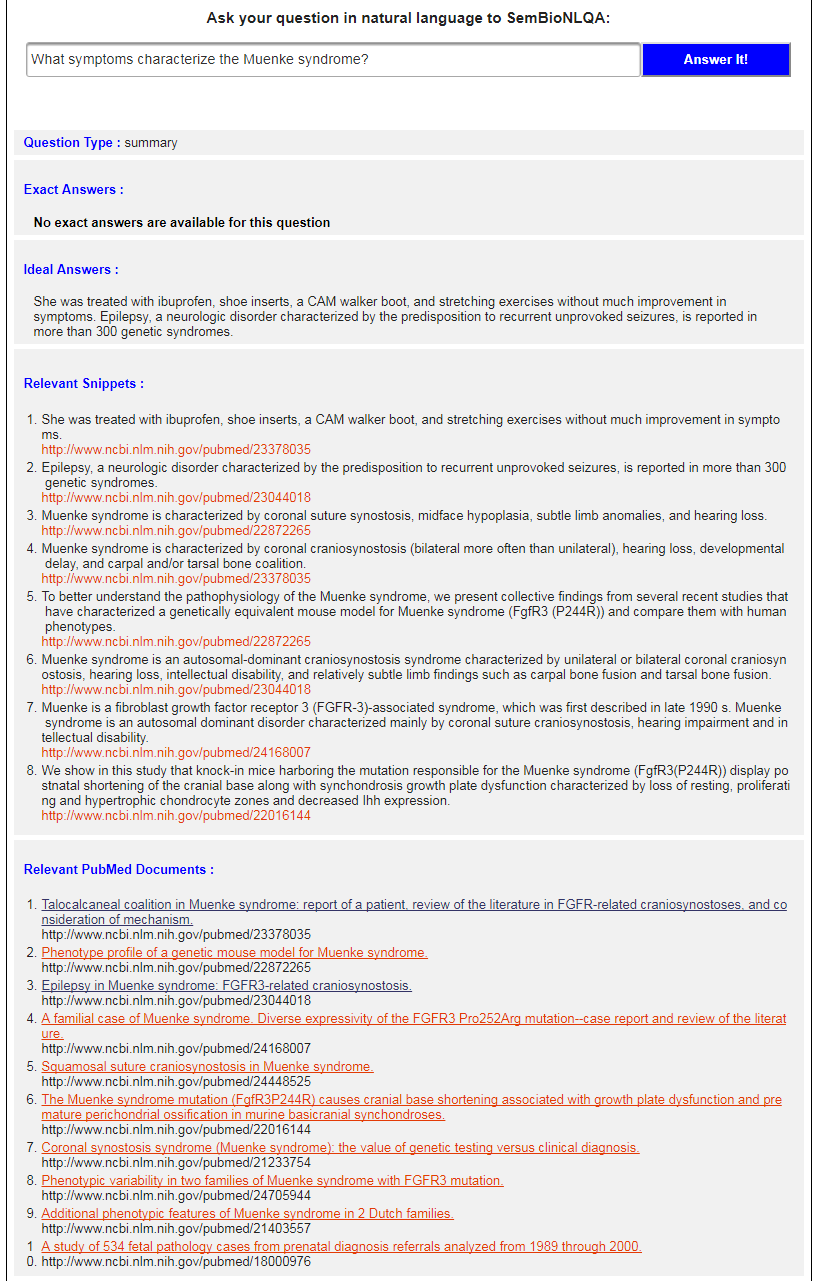}
\caption{The SemBioNLQA output for the biomedical question ``What symptoms characterize the Muenke syndrome?'' (identifier 52bf1d3c03868f1b0600000d)}
\label{fig:r1}
\end{figure}

\begin{figure}[h!]
\captionsetup{justification=justified}
\graphicspath{{Figures/}}
\centering

\includegraphics [width=16cm, height=12cm]{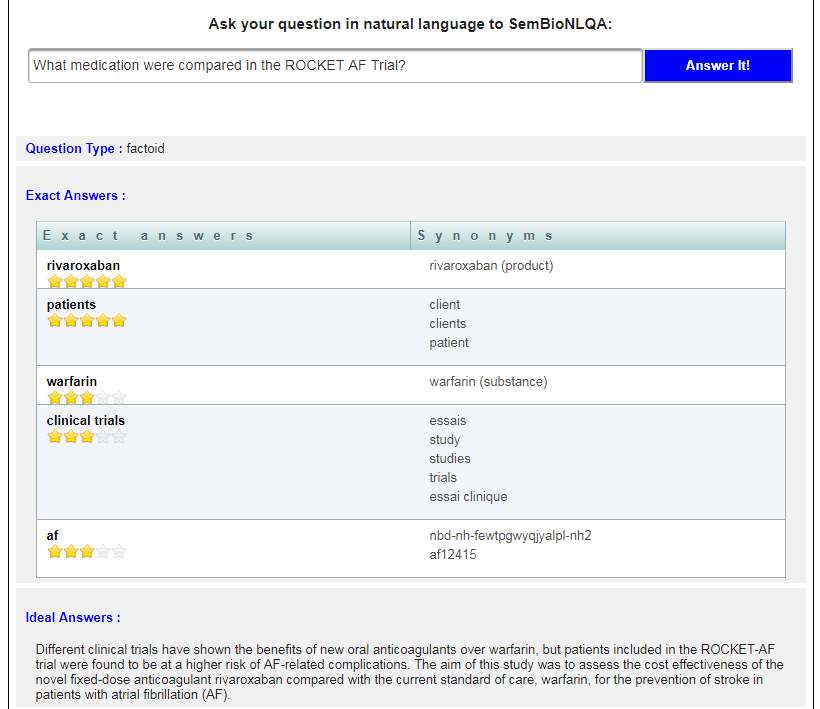}
\caption{The SemBioNLQA output for the biomedical question ``What medication were compared in the ROCKET AF Trial?'' (identifier 56bb616dac7ad10019000008)}
\label{fig:r2}
\end{figure}

\begin{figure}[h!]
\captionsetup{justification=justified}
\graphicspath{{Figures/}}
\centering

\includegraphics [width=16cm, height=8cm]{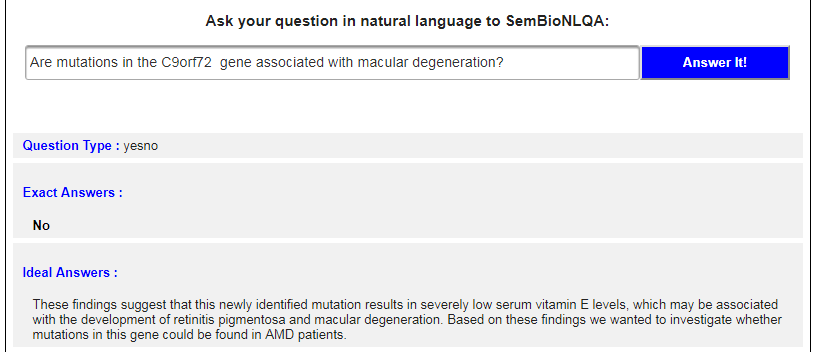}
\caption{The SemBioNLQA output for the biomedical question ``Are mutations in the C9orf72  gene associated with macular degeneration?'' (identifier 58e11bf76fddd3e83e00000c)}
\label{fig:r3}
\end{figure}

A direct comparison with the systems presented in \citep{neves2015hpi, zhang2015fudan,schulze2016hpi,yang2015learning,choi2015snumedinfo} and evaluated on either the 2015 or 2016 BioASQ challenges is not simple since the authors used the test set of biomedical questions along with their golden documents, golden snippets, and questions types released by the BioASQ challenges. While in the end-to-end evaluation of SemBioNLQA, the relevant documents, relevant passages, and the question type for a given biomedical question are obtained using its document retrieval, passage retrieval, and question classification systems, respectively.
\begin{table}[h!]
\centering
\caption[The overall evaluation results of the SemBioNLQA system on five batches of biomedical questions provided by BioASQ 3b 2015 and BioASQ 4b 2016]{The overall evaluation results of the SemBioNLQA system on five batches of biomedical questions provided by BioASQ 3b 2015 and BioASQ 4b 2016. Acc, P, R, F, R-2, R-SU4 indicate accuracy, precision, recall, and f-measure, rouge-2, rouge-SU2, respectively.}
\label{tab:6.3.2}
\begin{tabular}{p{3.1cm}p{1.3cm}p{1.2cm}p{1.1cm}p{1.1cm}p{1.1cm}p{1.1cm}p{1.1cm}p{1.3cm}}
\hline\noalign{\smallskip}
 \multirow{3}{*}{Dataset} &\multirow{3}{*}{Batch}& \multicolumn{5}{c}{Exact answers} &  \multicolumn{2}{c}{\multirow{2}{*}{Idial answers}}\\
\cmidrule(l){3-7}

 & & Yes/No & Factoid & \multicolumn{3}{c}{ List}  &\multicolumn{2}{c}{}   \\
 \cmidrule(l){3-3}\cmidrule(l){4-4} \cmidrule(l){5-7} \cmidrule(l){8-9}
  & &  Acc &  MRR& P& R &F1  & R-2& R-SU4\\

\noalign{\smallskip}\hline\noalign{\smallskip}
\multirow{5}{*}{BioASQ 3b 2015} &	Batch 1 &0.7273&	0.0128	& 0.0364&	0.0682&	0.0462& 0.1039&	0.1397\\
            &	Batch 2 &0.6875&	0.0339	& 0.0714&	0.1232&	0.0891& 0.1044&	0.1413\\
            &	Batch 3 &0.8621&	0.0641	& 0.0353&	0.0368&	0.0352& 0.1269&	0.1557\\
            &	Batch 4 &0.6800&	0.0586	& 0.0696&	0.0880&	0.0751& 0.1467&	0.1702\\
            &	Batch 5 &0.6786&	0.0795	& 0.0083&	0.0208&	0.0119& 0.0915&	0.1205\\

\cmidrule(l){1-9}
\multirow{5}{*}{BioASQ 4b 2016} &	Batch 1 &0.8214&	0.0534	& 0.1091&	0.1545&	0.1268&0.1552&	0.1855\\
            &	Batch 2 &0.7188&	0.0495	& 0.0476&	0.0477&	0.0462& 0.1378&	0.1720\\
            &	Batch 3 &0.8800&	0.1186	& 0.0857&	0.1667&	0.1122& 0.1430&	0.1730\\
            &	Batch 4 &0.8095&	0.0253	& 0.0400&	0.0667&	0.0500& 0.1097&	0.1289\\
            &	Batch 5 &0.8519&	0.0687	& 0.0300&	0.0517&	0.0368& 0.1609&	0.1836\\

\noalign{\smallskip}\hline
\end{tabular}
\end{table}
As it has already been stated before, to a given biomedical question, SemBioNLQA first retrieves the $N$ top-ranked documents, then finds the $N$ top-ranked passages and finally applies the appropriate answer extraction according to the question type detected by the question classification module. In particular, we have decided to go with the $N=10$ top-ranked documents and $N=10$ top-ranked passages since only the 10 first ones from the resulting list are permitted for the test in the 2015 and 2016 BioASQ challenges. After that, for yes/no questions, the exact answer (``yes'' or ``no'') is formed by using SENTIWORDNET, a sentiment lexicon. Each word of the relevant snippets is assigned its SENTIWORDNET score, and the decision to output ``yes'' or ``no'' depends on the number of positive or negative snippets. For factoid questions, the exact answer is produced by identifying the biomedical entities that occur in the given top relevant snippets of the question, and reporting the five most frequent biomedical entities and their synonyms of the top snippets, excluding entities also mentioned in the question. For list questions, the exact answer is produced in the same manner, except that the most frequent entities and their synonyms of the top relevant snippets are now returned as the single list that answers the question. On the other hand, to generate ideal answers (summaries), we have applied MetaMap to the relevant snippets and question in order to obtain the UMLS concepts they refer to. We then have ranked the snippets by their BM25 similarity to the question (using Porter stems and UMLS concepts as features), and return the top two most highly ranked snippets (concatenated) as the ideal answer. Figure~\ref{fig:r1}, Figure~\ref{fig:r2} and Figure~\ref{fig:r3}  show the SemBioNLQA output for three biomedical questions which come from BioASQ training questions.

\subsubsection{Manual evaluation}

In this experiment, we randomly selected 30 questions from the BioASQ training dataset and posed these to the four systems - AskHermes, EAGLi, Olelo and SemBioNLQA. This evaluation was carried out manually, and therefore, we needed to limit the number of questions and types. We decided to limit it to factoid, list and yes/no questions given that these types of answers are easier to check manually than summaries. This sequence of 30 questions, which are listed in Appendix A, contains 10 factoid questions, 11 list questions and 9 yes/no questions. In our evaluation, an answer is considered as correct if the first returned biomedical entity (for factoid questions), at least one of the first five returned biomedical entities (for list questions) or the Boolean value, i.e., ``yes'' or ``no'', (for yes/no questions) is correct. Indeed, we manually checked the results returned by each system to look for the correct standard answers as provided by the BioASQ challenge. Table~\ref{tab:6.3.4} presents and compares the results of the aforementioned systems and SemBioNLQA. All answers returned by the systems are available for download \footnote{\url{https://sites.google.com/site/mouradsarrouti/datasets}}.

\begin{table}[h!]
\centering
\caption{Comparison of the obtained results by SemBioNLQA, EAGLi, AskHERMES and Olelo in terms of number of recognized questions and correct answers}
\label{tab:6.3.4}
\begin{tabular}{M{5.5cm}M{4.9cm}M{4.7cm}}
\hline\noalign{\smallskip}
Systems& Number of recognized questions& Number of correct answers\\
\noalign{\smallskip}\hline\noalign{\smallskip}
EAGLi \citep{gobeill2009question}&7/30&3/30\\
AskHERMES \citep{Cao_2011}&12/30&2/30\\
Olelo \citep{Kraus_2017}&30/30&6/30\\
\textbf{SemBioNLQA}&\textbf{30/30}&\textbf{18}/\textbf{30}\\
\noalign{\smallskip}\hline
\end{tabular}
\end{table}

\subsection{Discussion}

In contrast to traditional IR systems which identify a simple list of relevant documents for the user's query usually expressed in terms of some keywords, QA systems aims at providing precise short answers to user questions written in natural language. It is the goal of such systems to move the burden of browsing and filtering the numerous results, which can be quite time consuming, from the information seekers to the computer. While open-domain QA has been widely studied, few integral systems such as the ones described in \citep{gobeill2009question,Cao_2011,Kraus_2017} are currently able to automatically answer questions from the ever-increasing volume of peer-reviewed scientific articles in the biomedical domain. In this work, we addressed shortcomings of these systems, such as limited usability and performance in terms of the precision for the currently supported question and answer types. As results, unlike these systems, our developed biomedical QA system SemBioNLQA has the ability to handle a large amount of questions and answers types such as yes/no, factoid, list and summary questions that may cover all types of questions. It returns exact answers in the form of ``yes'' or ``no'' for yes/no questions, biomedical entities for factoid questions and a list of biomedical entities for list questions. In addition to exact answers for the aforementioned questions types, SemBioNLQA also formulates and returns ideal answers. For summary questions, the system retrieves only ideal answers since such questions do not have precise answers.

The results of the systematic evaluation, detailed in Table~\ref{tab:6.3.1}, have shown that SemBioNLQA gets better results compared with Olelo, the most current biomedical QA system. The presented system significantly outperformed the Olelo system in extracting both exact and ideal answers for the currently supported questions. The important thing to note here is that our system significantly outperforms the Olelo system in all batches of testing datasets. There is a large difference in the results, a thing that clearly indicates that our system is not only effective but also robust.

From another side, as shown in Table~\ref{tab:6.3.4} which presents and compares the results of the manual evaluation of SemBioNLQA, EAGLi, AskHERMES and also Olelo in terms of number of recognized questions and correct answers, SemBioNLQA gets better results and succeeded at answering the majority of randomly selected BioASQ questions. We have manually analyzed the answers provided for the biomedical questions by each system. In contrast to SemBioNLQA, which has proven to be quite successful at extracting the exact answers depending on the expected answer for each question type, Olelo returned a summary as the answer for the most questions and AskHERMES returned a multiple sentence passage as answer for all questions. Indeed, SemBioNLQA was able to detect the questions type were of the factoid, list or yes/no  types, and thus generated exact answers depending on the expected answer for each question type. This indicates that its integration of our question classification method offers SemBioNLQA the ability to understand and correctly recognize the information needs of users. In contrast, even though Olelo was developed to handle with factoid, list and summary questions, it was not able to detect the types for given questions, and thus generated summaries for all questions, and therefore, the users have to read these summaries so as to find the precise answers. In particular, it only returns exact answers when both the headword and semantic types are detected, in addition to the candidate answers being of this same semantic type. On the other hand, as shown in Table~\ref{tab:6.3.4}, both SemBioNLQA and Olelo have succeeded in returning answers for all questions, while AskHERMES and EAGLi could not provide answers for the majority of the questions, instead, only the following messages ``Nothing found! Please refine your question'' in the former and ``EAGLi did not understand your question. Try a popular example, or go to the manual mode.'' in the latter.

On the other hand, it is clear from Table~\ref{tab:6.3.2} that the different components of the SemBioNLQA system have a significant impact on the answer extraction task and therefore on the overall performance of SemBioNLQA since if the set of retrieved documents, passages and the type of a given question are not identified correctly, further processing steps to extract the answers will inevitably fail too.  For instance, for the question ``What is the association of spermidine with $\alpha$-synuclein neurotoxicity?'' (identifier 56c073fcef6e394741000020) from the batch 1 of test  set of the 2016 BioASQ challenge, the returned type of question is ``summary'' whereas in the corpus the type of question is ``factoid''. Therefore, the SemBioNLQA system will inevitably fail to extract and output the correct answer since extracting the answer to a factoid question, which is asking for a biomedical entity, is not the same as extracting the answer to a summary question which is looking only for an ideal answer.

Overall, SemBioNLQA holds a number of advantages over the state-of-the-art systems. First, the integration of our question types classification method it offers have a clear advantage over Olelo in that it returns exact answers depending on the expected answer of each question type. Second, SemBioNLQA which is aimed to be able to accept a variety of natural language questions and to generate appropriate natural language answers, provides an unbeatable advantage over AskHERMES, EAGli and Olelo in that it handles with a large amount of questions types including yes/no, factoid, list and summary questions. Third, the systematic and manual evaluations results demonstrated that SemBioNLQA is more effective as compared with the aforementioned systems.

In summary, biomedical QA is a very challenging task since it accepts questions written in natural language and provides precise answers instead of only presenting potentially relevant documents by integrating various resources. Therefore, no current system can always perform well on the myriad questions that can be asked of it. SemBioNLQA provides a practical and competitive alternative to help users find exact and ideal answers.

\section{Summary of the Chapter}

Starting from the aim of answering a variety of natural language questions, we presented in this thesis work the proposed methods for the extraction of the answers to given biomedical questions including yes/no questions, factoid questions, list questions, and summary questions. These types of questions that are commonly asked in the biomedical domain, may cover all types of questions that can be posed by the users.

In section~\ref{Chapter6.2} we presented in details the answer extraction system that we proposed for extracting the answer for each of the aforementioned question types. The proposed system provides exact answers (e.g., ``yes'', ``no'', a biomedical entity name, etc.) and ideal answers (i.e., paragraph-sized summaries of relevant information) for yes/no, factoid and list questions, whereas it provides only the ideal answers for summary questions. Thanks to an evaluation on standard benchmarks provided by the 2015 and 2016 BioASQ challenges, we noted that SemBioNLQA achieved promising performances compared with the best existing methods. Moreover, as part of our participation in Phase B, Task 5b of the 2017 BioASQ challenge, our submission was placed within the top tier submissions out of all participants. Furthermore, our system was one of the fifth BioASQ challenge winners.

In section~\ref{Chapter6.3} we tackled a fully automatic QA system in the biomedical domain, SemBioNLQA, which has the ability to deal with four types of biomedical questions including yes/no questions, factoid questions, list questions, and summary questions. SemBioNLQA is currently able to provide exact answers and paragraph-sized ideal answers for yes/no, factoid and list questions, whereas it only retrieves ideal answers for summary questions. The system relied on (1) handcrafted lexico-syntactic patterns and a machine learning approach for question classification, (2) PubMed search engine and UMLS similarity for document retrieval, (3) the BM25 model, stemmed words and UMLS concepts for passage retrieval, and (4) UMLS metathesaurus, BioPortal synonyms, sentiment analysis and term frequency metric for answer retrieval. Compared with the existing biomedical QA systems, SemBioNLQA has the potential to deal with a large amount of question and answer types. Moreover, experimental evaluations performed
on biomedical questions and answers provided by the BioASQ challenge especially in 2017 (as part of our
participation), show that SemBioNLQA achieves good performances compared with the most current state-of-the-art system and allows a practical and competitive alternative to help information seekers find exact and ideal answers to their biomedical questions.

\chapter*{Conclusion and Future Work} 
\addcontentsline{toc}{chapter}{Conclusion and Future Work}
\label{Chapter7} 
\markboth{Conclusion and Future Work}{Conclusion and Future Work}


In this PhD thesis, we focused on investigating and improving question answering in the biomedical domain where several specific challenges are addressed, such as specialized lexicons and terminologies, the types of treated questions and the characteristics of targeted documents. This improvement concerns the different components of a QA system, i.e., (1) question analysis and classification, (2) document retrieval, (3) passage retrieval, and (4) answer extraction. We proposed several methods for enabling users (e.g., researchers and health care professionals) to find precise and short answers to their natural language questions from the incessantly growing information in the biomedical domain. In addition, we developed a semantic biomedical QA system named SemBioNLQA by integrating the different proposed methods. The experimental results on two well-known standard corpus provided by the BioASQ challenge and U.S. NLM  have shown the effectiveness of the proposed methods. Furthermore, a subsystem of our SemBioNLQA system was one of the 2017 BioASQ challenge winners.

In this chapter we present the main conclusions derived from our contributions, and perspectives for future work. In section \hyperref[Chapter7.2]{Summary and Contributions} we present a summary of this work and its resulting contributions. In section \hyperref[Chapter7.3]{Future Work} we provide the future research directions for expanding our work.

\section*{Contributions}
\label{Chapter7.2}
\addcontentsline{toc}{section}{Contributions}

First, we proposed a machine learning-based method to identify the type of a given biomedical question (i.e., yes/no questions, factoid questions, list questions, and summary questions) which enables to a biomedical QA system to use the appropriate answer extraction. Then, we proposed another machine learning-based method to assign one or more topics to given biomedical questions in order to determine the semantic type of the expected answer which is very useful in generating specific answer retrieval strategies. Next, we proposed a document retrieval method in biomedical QA to retrieve the relevant documents that are likely to contain the answers to biomedical questions from the MEDLINE database. After that, we proposed a passage retrieval method in biomedical QA to retrieve the relevant passages/snippets to given biomedical questions. Finally, we proposed specific answer extraction methods to extract both exact and ideal answers from the retrieved passages, and presented our fully automated system SemBioNLQA - Semantic Biomedical Natural Language Question Answering - which is aimed to accept a variety of natural language questions and to generate appropriate natural language answers by providing both exact and ideal answers.

We summarize and highlight the main contributions and findings of this thesis work, addressing the research goals described in the \hyperref[Chapter1]{General Introduction}.

\subsection*{A question type classification method in biomedical QA}

We proposed a machine learning-based method for question type classification in biomedical QA. This methods aims at automatically classifying biomedical natural language questions into one of the following four categories which specify the types of questions: yes/no questions, factoid questions, list questions and summary questions defined by the BioASQ challenge. The task of assigning one of the aforementioned categories to a given question in biomedical QA is very useful in candidate answer extraction as it allows to a biomedical QA system to know in advance the expected answer format, and therefore to use the appropriate answer extraction strategy. For example, the biomedical question ``Does nimotuzumab improve survival of glioblastoma patients?'' expects an answer of type ``yes'' or ``no'', therefore, a yes/no answer extraction method should be used in this case to extract the final answer .

In the proposed method, we first extracted appropriate features from biomedical questions using our predefined handcrafted lexico-syntactic patterns which were constructed by analysing the BioASQ training questions. We then have fed these features for a machine-learning system. Finally, for a given unlabeled question, the class label is predicted using the trained classifier. Experimental evaluations performed on large standard annotated datasets of biomedical questions, provided by the BioASQ challenge, demonstrated that our method exhibits significant improved performance when compared to baseline systems. The proposed method achieved a roughly 10-point increase over the best baseline in terms of accuracy. Moreover, the obtained results have shown that using handcrafted lexico-syntactic patterns as features' provider of SVM lead to the highest accuracy of 89.40\%. The predefined patterns yielding the best results are also made available which encourage replication of results.

\subsection*{A question topic classification method in biomedical QA}

We proposed a machine learning-based method for question topic classification in biomedical QA. This method aims at automatically assigning one or more general topics (e.g., pharmacological, treatment, test, etiology, etc.) to clinical questions written in natural language. The task of question topic classification or answer type recognition is to determine the answer type, the semantic type of the expected answer, which is very useful in generating specific answer retrieval strategies and choosing the best resource from which the answer should be extracted. If we know the semantic type of the expected answer for a given question, we can avoid looking at every passage or biomedical entity name in the entire suite of candidate answer passages for the answer.

The proposed method first extracts a set of syntactic and semantic features from the annotated questions, including words, word stems, bigrams, UMLS concepts and semantic types, and syntactic dependency relations between pair words. It then feed these features for a machine learning model. Finally, it assigns one or more topics to the given unlabeled question using the trained classifier. We have explored several machine learning algorithms such as Naïve Bayes, Decision Tree, and SVM, showing SVM achieved the best results for this task on the annotated data that is released by NLM. A set of experiments have shown that the proposed method is more effective as compared with the state-of-the-art methods. Furthermore, The proposed method significantly outperforms the current state-of-the-art methods by an average of 4.5\% in terms of F1-score.

\subsection*{A document retrieval method in biomedical QA}

We studied the problem of deciding if a PubMed document is relevant to a specific natural language biomedical question in the context of biomedical QA. We developed a document retrieval system in which we proposed a new document reranking system. The idea behind this proposal is that there are many cases where the search engine mistakenly returns irrelevant citations high in the set or relevant citations low in the set for a given question, which poses a real problem for biomedical QA systems since they usually extract the answers from the documents ranked in the top of the set.

In the proposed system which aims at retrieving relevant documents to given question in biomedical QA, we first constructed the query from the input question using UMLS concepts identified by the MetMap tool. We then retrieved a set of possibly relevant documents to the query using a typical IR system which searches the MEDLINE database. We finally reranked the retrieved set of documents promoting to the top the documents it considers the most relevant to the biomedical question based on the semantic similarity between the question and the titles of the returned documents. To compute the semantic similarity scores, we first mapped both the given biomedical question and the titles of its set of the possibly relevant to UMLS Metathesaurus using MetaMap to identify UMLS concepts. We then computed the semantic similarity scores between UMLS concepts of the question and each title of the returned documents using UMLS similarity which uses path length as a similarity measure. The conducted experiments demonstrated that the proposed document retrieval system is more effective as compared with current state-of-the-art systems which were ranked within the 10 top tier systems in the BioASQ challenge. Furthermore, the experimental results have shown that the proposed document reranking method is able to achieve a significant improvement with respect to the original document ranking provided by a typical IR system.

\subsection*{A passage retrieval method in biomedical QA}

We proposed a passage retrieval system in biomedical QA to retrieve the relevant passages/snippets that are likely to contain the answer for a given biomedical question. In a typical biomedical QA system, the highly relevant documents that do not prominently answer a biomedical question are not the ideal candidate answers for further processing. Therefore, passage retrieval remains one of the most important components of any biomedical QA system as it allows to extract the set of potential answer passages from  the retrieved set of documents which serve as answer candidates from which the biomedical QA system extracts the answers. We have shown that the overall performance of a biomedical QA system heavily depends on the effectiveness of the integrated passage retrieval system.

The proposed biomedical passage retrieval consists of two stages, namely (1) document retrieval and (2) passage retrieval. During the first stage, we have used the PubMed search engine to retrieve the top-ranked documents to a given query constructed from the question using UMLS concepts. We then reranked the retrieved set of documents promoting to the top the documents it considers most relevant to the biomedical question based on UMLS similarity. At the second stage, passage identification, we have first taken the abstracts from the top ranked documents retrieved at the first stage, and used Stanford CoreNLP's tokenizer to split these abstracts into sentences. Thus, we treated the obtained set of sentences as a set of candidate passages, where the passage length is similar to that of the sentence. We then re-ranked the set of candidate passages using the BM25 model as a scoring function, and stemmed words and UMLS as features for text passage representation. Experimental evaluations performed on large standard datasets, provided by the BioASQ challenge, have shown that the proposed method achieves good performances compared with the current state-of-the-art methods. Furthermore, the proposed method significantly outperforms the current state-of-the-art methods by an average of 6.84\% in terms of MAP. In particular, we have found that passage length has a great impact on passage retrieval performance. We also anticipated improved passage retrieval if consistent passage length can be achieved.

\subsection*{SemBioNLQA: A semantic biomedical QA system}

We developed a novel answer extraction system which has the ability to deal with four types of questions including yes/no questions, factoid questions, list questions, and summary questions. In this system, we proposed for each question type its specific answer extraction method. For yes/no questions, the exact answer (i.e., ``yes'' or ``no'') is formed by using SentiWordNet, a sentiment lexicon. Each word of the relevant passages is assigned its SentiWordNet score, and the decision to output ``yes'' or ``no'' depends on the number of positive or negative candidate answer passages. For factoid questions, the exact answer is produced by identifying the biomedical entities that occur in the given top relevant passages of the biomedical question, and reporting the five most frequent biomedical entities and their synonyms of the top passages, excluding entities also mentioned in the question. For list questions, the exact answer is produced in the same manner, except that the most frequent entities and their synonyms of the top relevant passages are now returned as the single list that answers the question. To generate ideal answers (i.e., short summaries), we have applied MetaMap to the relevant passages and the question, in order to obtain the UMLS concepts they refer to. We then have ranked the passages by their BM25 similarity to the question (using Porter stems and UMLS concepts as features), and return the top two most highly ranked snippets (concatenated) as the ideal answer. A set of experiments on BioASQ datasets have shown that the proposed system is more competitive as compared with current state-of-the-art systems. Furthermore, our system was one of the winners in the 2017 BioASQ challenge.

On the other hand, we presented the fourth contribution of this thesis work which develop a semantic biomedical QA system named SemBioNLQA. SemBioNLQA is aimed to accept a variety of natural language questions and to generate appropriate natural language answers by providing both exact and ideal answers. The SemBioNLQA system, which consists of question classification, document retrieval, passage retrieval and answer extraction components, takes natural language questions as input, and outputs both short exact answers and ideal answers as results. SemBioNLQA is derived from our previously established methods  in (1) question classification (2) document retrieval, (3) passage retrieval , and (4) answer extraction systems. Compared with the current state-of-the-art biomedical QA systems, SemBioNLQA, a fully automated system, has the potential to deal with a large amount of question and answer types. SemBioNLQA retrieves quickly users' information needs by returning exact answers (e.g., ``yes'', ``no'', a biomedical entity name, etc.) and ideal answers (i.e., paragraph-sized summaries of relevant information) for yes/no, factoid and list questions, whereas it provides only the ideal answers for summary questions. Furthermore, experimental evaluations performed on real biomedical questions and answers provided by the BioASQ challenge shows that SemBioNLQA is more effective as compared with the existing systems and allows a practical and competitive alternative to help information seekers find exact and ideal answers to their biomedical questions.

\section*{Future Work}
\label{Chapter7.3}
\addcontentsline{toc}{section}{Future Work}

This thesis work opens several directions for further perspectives at different levels. In the following, we will shed light on the main perspectives related to each of the four components of a QA system, i.e., (1) question analysis and classification, (2) document retrieval, (3) passage retrieval, and (4) answer extraction.

\subsection*{Question classification}

The first step towards answering a biomedical question is analyzing and classifying the question in order to determine the type of question and the type of answer to produce. It is a crucial step of a biomedical QA system as the performance of such system depends directly on the performance of its question classification component. At the level of this task:

\begin{itemize}
 \item We intend to extensively study the training questions we used in this thesis work in order to develop additional patterns for improving the question type classification performance.

 \item We would consider further improvements of our question type classification method and question topic applying deep learning models which have been emerging as state-of-the-art for text classification.
\end{itemize}

\subsection*{Document and passage retrieval}

Due to the immense and increasing volume of biomedical literature, not only document retrieval and passage retrieval systems are becoming important for end-users searching for relevant documents and passages to their queries, but are also very important in many other text mining applications, particularly in QA systems. The identification of relevant documents and passages that may contain the answer for given questions is a challenging and important step for a biomedical QA system. In this context, we identified the following directions for future work:

\begin{itemize}

\item We intend to apply deep learning models to estimate the probability of the document's relevance to the question. More recently, deep learning models have been used in many IR systems and applied to various types of text matching problems. Such models produce a more robust predictor compared to computing similarities between vector representations of the query and document. They have shown significant improvements over traditional IR factors \citep{mohan2017deep}. An overview of neural retrieval models can be found in \citep{zhang2016neural,mitra2017neural}.

\item We intend to improve the performance of biomedical passage retrieval by exploring synonyms, semantic relationships between UMLS concepts and the word embedding.

\item We found in this thesis work that passage length has a great impact on the overall performance of passage retrieval systems, therefore, we anticipate improved passage retrieval if consistent passage length can be achieved.

\item We also intend to explore deep learning models to estimate the probability of the passage's relevance to the question.

\end{itemize}

\subsection*{Answer extraction}

Answer extraction is considered as the most challenging task of a QA system as  it allows extracting the precise answer, then presenting it to the user posing the question. The answer depends directly on the type of the given question (e.g., ``yes'' or ``no'' for yes/no question, biomedical entity names for factoid questions, etc.). At the level of this task, based on this thesis work, we identified the following directions for future work:

\begin{itemize} 

\item We intend further improvements regarding answering yes/no questions by proposing more complex semantic and sentiment analysis methods.

\item Participating in the 2017 BioASQ challenge also provided us with a renewed interest in deep learning models. However, deep learning models are usually applied for factoid-type QA. Such approaches require a large number of question-answer pairs for the training phase. One of the 2017 wining systems of the 2017 BioASQ challenge developed by \cite{wiese2017neural} was a purely deep learning system. This system focused only on answering factoid and list questions. In particular, we are currently studying the different ways for incorporating deep learning models into our biomedical QA system in order to further improve its effectiveness particularly in answering factoid and list questions.

\item We envisage improving the effectiveness of the proposed QA system by searching in structured data (e.g., databases, ontologies, RDF triples, and the LOD cloud)

\item We also envisage working with full text (e.g., from PubMed Central) to see if we can improve answer extraction performance with a focused look at the full text. MEDLINE which indexes more than 24 million references, contains a link to the free full text of the article archived in PubMed Central that contains 4.4 million articles.

\end{itemize}


\appendix 

\chapter{List of BioASQ questions used for the manual evaluation} 

\label{AppendixA} 

\begin{enumerate}
  \item Which is the gene most commonly mutated in Tay-Sachs disease? \newline (identifier 536e46f27d100faa09000012)
  \item What medication were compared in the ROCKET AF Trial? \newline (identifier 56bb616dac7ad10019000008)
  \item Which enzyme does MLN4924 inhibit? (id 56ed03862ac5ed1459000004)
  \item Where is the protein Pannexin1 located? \newline (identifier 56af9f130a360a5e45000015)
  \item RTS S AS01 vaccine was developed to prevent which disease? \newline (identifier 56bc77a3ac7ad10019000015)
  \item Which type of myeloma is ixazomib being evaluated for? \newline (identifier 56ed0ffe2ac5ed1459000008)
  \item What enzyme is inhibied by Opicapone? (id 56c1d857ef6e394741000033)
  \item Which gene has been implicated in Majeed Syndrome? (id 56f7c15a09dd18d46b000012)
  \item Which gene is most commonly associated with severe congenital and cyclic neutropenia? \newline (identifier 5503133ae9bde6963400001d)
  \item Which is the receptor for the immunosuppressive drug cyclosporin A (CsA)? \newline (identifier 56f6c11109dd18d46b00000e)
  \item Which proteins participate in the formation of the Notch transcriptional activation complex? \newline (identifier 54fb6fb5d176fff445000004)
  \item What is being measured with an accelerometer in back pain patients? \newline (identifier 533f9df0c45e133714000016)
  \item List inhibtors targeting the mitochondrial permeability transition pore. \newline (identifier 5717cdd2070aa3d072000001)
  \item List symptoms of the IFAP syndrome. \newline (identifier 56c1f038ef6e394741000051)
  \item What is the functional role of the protein Drp1? \newline (identifier 5717dbfe7de986d80d000001)
  \item Which receptors are bound by Tasimelteon? \newline (identifier 56c1f043ef6e394741000057)
  \item Which disease phenotypes are associated to PRPS1 mutations? \newline (identifier 5713b0a51174fb175500000e)
  \item Which genes are thought to be involved in medulloblastoma development? \newline (identifier 5539029cbc4f83e828000012)
  \item Which miRNAs could be used as potential biomarkers for epithelial ovarian cancer? \newline (identifier 553fa78b1d53b76422000007)
  \item Which interleukins are inhibited by Dupilumab? \newline (identifier 56c1f005ef6e39474100003a)
  \item Which are the genes thought to be regulated by EWS/FLI? \newline (identifier 552faa43bc4f83e828000004)
  \item Is the ACE inhibitor indicated for lung cancer treatment? \newline (identifier 530cf4fe960c95ad0c000005)
  \item Is PTEN involved in follicular thyroid carcinoma? \newline (identifier 55031650e9bde69634000026)
  \item Is Fanconi anemia presented as a genetically and clinically heterogeneous disease entity? \newline (identifier 54ede5c394afd61504000006)
  \item Can the iPS cell technology be used in Fanconi anemia therapy? \newline (identifier 54edef0594afd6150400000d)
  \item Does surgery for ovarian endometriomas improve fertility? \newline (identifier 54f088ee94afd61504000015)
  \item Is irritable bowel syndrome more common in women with endometriosis? \newline (identifier 54f08d4a94afd61504000016)
  \item Is the regulation of Vsr endonuclease independent of the growth phase of bacteria? \newline (identifier 553fbe9fe00431e071000001
  \item Does TRIM37 gene mutation causes Mulibrey nanism? \newline (identifier 56c1f03cef6e394741000054)
  \item Is the gene MAOA epigenetically modified by methylation? \newline (identifier 56cf50253975bb303a00000b)

\end{enumerate}


\chapter{Résumé détaillé de la thèse en Français} 

\label{Chapter8} 
\setcounter{secnumdepth}{4}
\begin{spacing}{2}
\begin{flushleft}
{\raggedright \Huge\bfseries Contributions à l'amélioration des systèmes de questions-réponses en domaine biomédical}
\end{flushleft}
\vspace*{20pt}

\end{spacing}

\minitoc

\section{Introduction}
\label{Chapter8.1}

Au cours des dernières décennies, la recherche dans le domaine biomédical a suscité un intérêt croissant de la part de la communauté de recherche, reflétée par une croissance exponentielle de la littérature scientifique. En effet, des milliers d'articles sont publiés et ajoutés chaque année à la base de données MEDLINE. Par exemple, plus de 806 000 et 869 000 citations ont été ajoutées en 2015 et 2016 respectivement. MEDLINE est la première base de données de la littérature biomédicale développée par la bibliothèque nationale de médecine des États-Unis (NLM). Elle contient plus que 24 millions de références d'articles de journaux en sciences de la vie axés sur la biomédecine. Avec cette augmentation explosive de la quantité de connaissances scientifiques, l'absorption automatique de toutes les informations pertinentes est devenue un défi, même pour les experts dans ce domaine. Dans ce contexte, des recherches récentes telles que celles de \cite{hristovski2015biomedical}, ont souligné que les moteurs de recherche spécialisés (e.g., PubMed) qui donnent l'accès à la base de données MEDLINE sont largement utilisés. Pour une requête constituée essentiellement de mots-clés ou de concepts, ces moteurs retournent un ensemble de documents potentiellement pertinents et délèguent à l’utilisateur la tâche de trouver l'information recherchée. En effet, l'ensemble des documents récupérés représente une taille de réponse encore trop grande pour pouvoir identifier facilement l'information recherchée. De plus, l'utilisateur doit souvent faire face à la charge d'étude et au filtrage des documents renvoyés à leur requête afin de trouver l’information recherchée si elle existe dans ces documents. Cette démarche, souvent laborieuse, nécessite un effort supplémentaire de la part de l’utilisateur et entraîne une perte de temps considérable, sans pour autant avoir la garantie de trouver la réponse correcte. Dans ce sens, une étude d'évaluation présentée par \cite{ely2000taxonomy} a montré que les médecins consacraient en moyenne moins de deux minutes à chercher l'information pertinente pour répondre à des questions cliniques, bien que bon nombre de leurs questions restent sans réponse. Dans une autre étude réalisée par \cite{Hersh283}, au moins 30 minutes sont nécessaires en moyenne pour les étudiants en médecine et les infirmières praticiennes pour répondre aux questions cliniques en utilisant MEDLINE.

Pour faciliter, accélérer et améliorer la recherche d’information, des systèmes de questions-réponses (SQR) sont mis en œuvre. Contrairement aux systèmes de recherche d’information traditionnels qui retournent un ensemble de documents potentiellement pertinents à une requête donnée, les SQR visent à fournir des réponses directes et précises à partir d’une collection de documents, du Web ou d’une base de données à des questions posées en langue naturelle. En effet, les SQR nécessitent l’utilisation des méthodes complexes d’extraction d’information et de traitement automatique de la langue.

\begin{figure}[th]
\selectlanguage{french}
\captionsetup{justification=justified}
\graphicspath{{Figures/}}
\centering
\includegraphics[width=16cm, height=11cm]{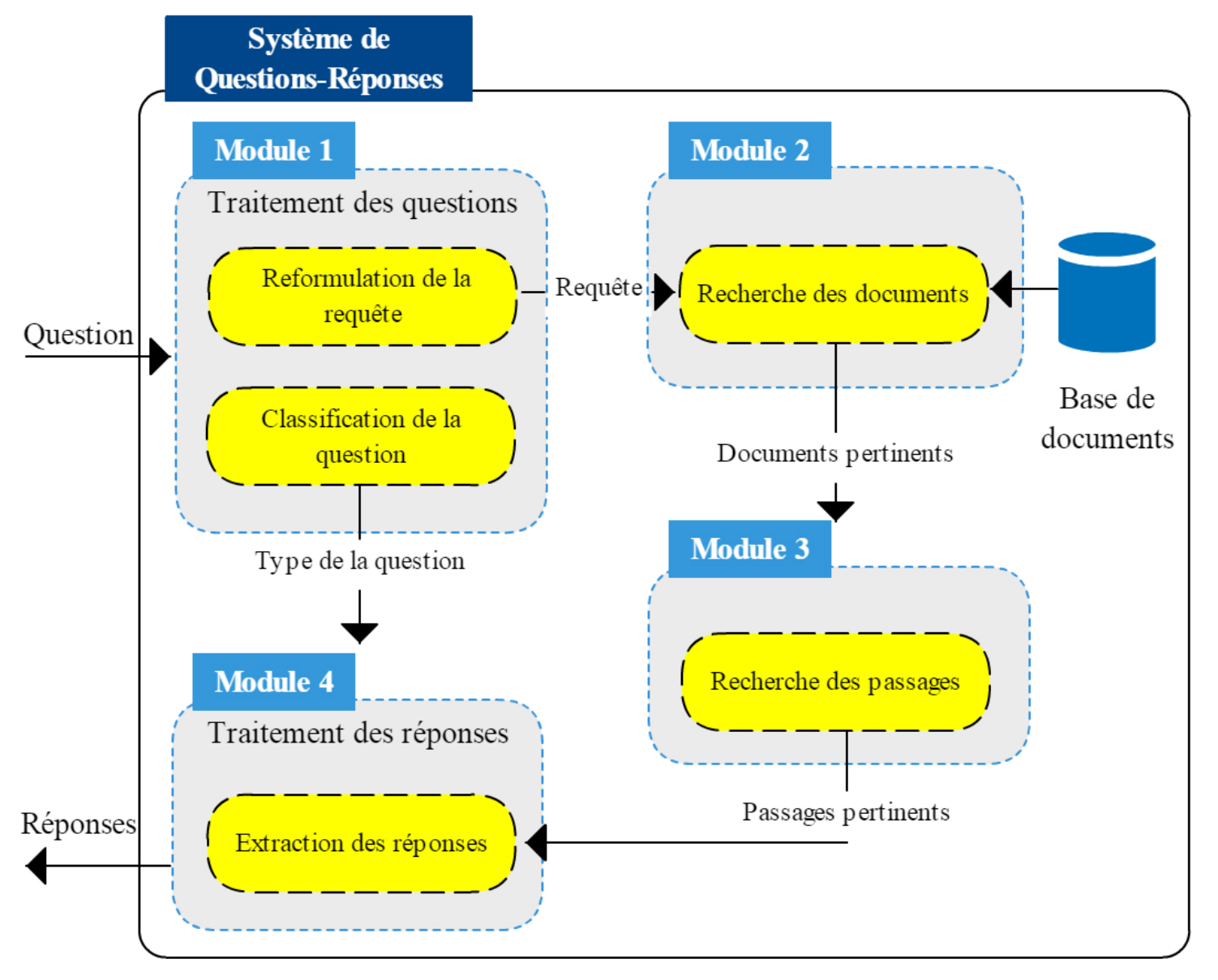}
\caption[Architecture générale d'un système de questions-réponses]{Architecture générale d'un système de questions-réponses}
\label{fig:QR}
\end{figure}

Typiquement, les systèmes de questions-réponses se composent de quatre composants comme le montre la figure~\ref{fig:QR}: (1) analyse et classification des questions, (2) sélection des documents pertinents, (3) recherche des passages pertinents, et (4) extraction des réponses.

L’entrée d’un SQR est une question formulée en langue naturelle. Le composant \emph{d'analyse et classification des questions} exploite des techniques d’analyse linguistique et de classification des questions pour déterminer à la fois le type de la question posée au système et le type de la réponse qu’il doit générer. Souvent, il peut y avoir plus de processus impliqués dans cette phase, tels que la reconnaissance des entités nommées et l'extraction des relations sémantiques. Le résultat de ce composant est une ou plusieurs caractéristiques de la question à utiliser dans les phases ultérieures. Le SQR construit ensuite à partir de la question une requête qui constitue l'entrée de deuxième module de \emph{sélection des documents pertinents}. Celui-ci fait appel à un moteur de recherche afin de trouver un ensemble de documents pertinents à la requête. Ensuite, à partir de ces documents, le processus de \emph{recherche des passages} procède à l'extraction des passages pertinents susceptibles de contenir les réponses candidates. Enfin, le module de \emph{traitement des réponses} ordonne les réponses candidates selon un score calculé en fonction du type de la réponse attendue qui a été déterminé par le module \emph{d'analyse et classification des questions}. La réponse la mieux classée parmi la liste des réponses candidates constitue la sortie du système de questions-réponses. Cette réponse finale est souvent accompagnée du texte brut à partir duquel la réponse a été extraite.

Bien que la recherche automatique des réponses aux questions en domaine ouvert est un défi largement étudié \citep{Green_1961,Woods_1973,Katz_2002,Kaisser_2008}, le domaine biomédical nécessite encore des efforts supplémentaires. En effet, les systèmes de questions-réponses dans le domaine ouvert concernent des questions qui ne sont pas limitées à un domaine spécifique, alors que les systèmes de questions-réponses dans un domaine spécifique, tel que le domaine biomédical, se rapporte à un contexte particulier et des problèmes plus complexes pour le traitement des questions et des réponses. \cite{athenikos2010biomedical} décrivent les caractéristiques particulières des systèmes de questions-réponses en domaine biomédical comme suit:
\begin{itemize}
  \item Corpus textuel de grande taille.
  \item Terminologie très complexe spécifique au domaine.
  \item Ressources lexicales, terminologiques et ontologiques spécifiques au domaine.
  \item Outils et méthodes pour exploiter les informations sémantiques incorporées dans les ressources lexicales, terminologiques et ontologiques.
  \item Format et typologie des questions spécifiques au domaine.
\end{itemize}

À la lumière de toutes ces caractéristiques, plusieurs travaux se sont intéressés à la problématique de recherche automatique de réponses aux questions dans le domaine biomédical. Certains ont proposé des systèmes pour répondre uniquement aux questions de type résumé ou générer des résumés pour tous les types de questions (e.g., \cite{lee2006beyond,cruchet2009trust,gobeill2009question,Cao_2011,Kraus_2017}). D’autres travaux se sont penchés sur les questions factuelles (e.g., \cite{abacha2015means}). Un nombre limité de systèmes s’est intéressé au traitement de plusieurs types de questions (e.g., \cite{neves2015hpi,zhang2015fudan,yang2015learning}). Cependant, ces systèmes exigent aux utilisateurs la spécification préalable du type de la question. De plus, malgré l'importance des questions booléennes dans le domaine biomédical, peu d'études (e.g., \cite{neves2015hpi}) ont été réalisées par rapport aux autres types de questions, comme les questions factuelles et les questions de type résumé. Même s'il n'y a que deux réponses possibles, «oui» ou «non», de telles questions sont parfois difficiles à aborder puisque leur traitement nécessite généralement des techniques complexes tel que l’analyse de sentiments, etc. Les systèmes de questions-réponses devraient prendre en considération les différents types de questions et leurs réponses appropriées.

Ce travail de thèse s'inscrit dans le cadre des systèmes de questions-réponses dans le domaine biomédical où plusieurs défis spécifiques sont relevés. Nous nous intéressons à l'étude et l’amélioration des méthodes permettant la recherche des réponses précises à des questions biomédicales dans une base de documents biomédicaux en langue anglaise. L'amélioration de la performance des systèmes de questions-réponses dépend de celle de chacune de leurs composantes. Notre objectif est la proposition d’un nouveau système de questions-réponses capable de traiter une variété de types de questions, y compris les questions booléennes, les questions factuelles, les questions de type liste et les questions de type résumé qui sont largement abordées en domaine biomédical. Nous avons apporté quatre contributions dans le cadre de ce travail de thèse:

\begin{enumerate}
\item Dans la première contribution, nous avons proposé une méthode de classification des questions permettant de déterminer le type de la question formulée en langue naturelle \citep{kdir15,Sarrouti_MIM_2017}. Etant basée à la fois sur les patrons lexico-syntaxiques et l'apprentissage automatique, celle-ci est exploitée par le SQR dans la phase d'extraction de la réponse finale. Dans le but de déterminer le type sémantique de la réponse attendue (i.e., un ou plusieurs sujets), nous avons proposé une variante de cette méthode \citep{Sarrouti_IBRA_2017} qui s’appuie sur d’autres caractéristiques de type lexical, morpho-syntaxique et sémantique. Le type sémantique de la réponse attendue d'une question biomédicale représente le sujet de la question tel que: «pharmacologie», «analyse», «traitement», etc. L'objectif est de permettre la réduction du nombre de documents parcourus lors de la recherche des réponses candidates.

\item La deuxième contribution consiste à suggérer une méthode de recherche des documents pertinents susceptibles de contenir les réponses aux questions biomédicales à partir de la base de données MEDLINE \citep{Sarrouti_2016}. Nous avons également proposé une alternative permettant la recherche des passages (i.e., extraits des documents) pertinents à des questions biomédicales données \citep{Sarrouti_2017}.
\item La troisième contribution propose des méthodes d’extraction des réponses appropriées \citep{Sarrouti_yes_2017,Sarrouti_bioasq_2017} permettant d'extraire à la fois les réponses \emph{exactes} et \emph{idéales} aux différents types de questions biomédicales, à savoir (1) les questions booléennes, (2) les questions factuelles, (3) les questions de type liste, et (4) les questions de type résumé.
\item La quatrième contribution consiste à développer un nouveau système de questions-réponses pour le domaine biomédical. En effet, l'ensemble des méthodes proposées sont développées et intégrées au sein d'un système global de questions-réponses, appelé SemBioNLQA. Il accepte en entrée une variété de questions et retourne des réponses appropriées \emph{exactes} et \emph{idéales}.
\end{enumerate}

\section{Méthodes proposées}
\label{Chapter8.3}
Nous proposons dans ce travail de thèse un ensemble de méthodes que nous avons développées et intégrées dans un système de questions-réponses pour le domaine biomédical afin de permettre aux utilisateurs (e.g., chercheurs et professionnels de la santé) de trouver des réponses précises et directes à leurs questions exprimées en langue naturelle. Dans ce qui suit, nous résumons et soulignons les principales contributions et résultats de ce travail de thèse.

\subsection{Méthodes proposées pour la classification des questions biomédicales}

Dans le chapitre~\ref{Chapter4}, nous avons proposé deux méthodes de classification de questions biomédicales à base de l'apprentissage automatique pour déterminer le type de la question ainsi que la catégorie sémantique de la réponse attendue à une question biomédicale donnée.

\subsubsection{Identification des types des questions biomédicales}
\label{Chapter8.3.1}
Dans la section~\ref{Chapter4.2} du chapitre~\ref{Chapter4}, nous avons décrit notre méthode basée sur l'apprentissage automatique pour l'identification des types des questions biomédicales. Celle-ci vise à classifier les questions biomédicales données dans l'une des quatre catégories définies par le challenge BioASQ. Ces types sont: (1) questions booléennes, (2) questions factuelles, (3) questions de type liste, et (4) questions de type résumé.

\begin{itemize}
\item	Questions booléennes: nécessitent seulement une des deux réponses possibles: «oui» ou «non». Par exemple, «Is calcium overload involved in the development of diabetic cardiomyopathy?» est une question booléenne et sa réponse est «oui».
\item	Questions factuelles: nécessitent une entité nommée biomédicale (par exemple, le nom d'une maladie, d'un médicament ou d'un gène), un nombre ou une expression courte comme réponse. Par exemple, «Which enzyme is deficient in Krabbe disease?» est une question factuelle et la réponse attendue est une entité nommée biomédicale «galactocérébrosidase».
\item	Questions de type liste: le type de la réponse attendue pour ce type de questions est une liste d’entités nommées biomédicales (par exemple, une liste de noms de gènes, une liste de noms de médicaments), des nombres ou des expressions courtes. Par exemple, «What are the effects of depleting protein km23–1 (DYNLRB1) in a cell?» est une question de type liste.
\item	Questions de type résumé: attendent un résumé ou un passage court comme réponse. Par exemple, le format de la réponse attendue pour la question «What is the function of the viral KP4 protein?» devrait être un résumé.

\end{itemize}

Le but de l'identification des types des questions dans les systèmes de questions-réponses est de déterminer le type de la question posée au système, et donc d’identifier le type de la réponse attendue pour vérifier si la réponse doit être une entité nommée biomédicale, un résumé, oui ou non, etc. Pour produire la réponse à une question donnée, le système de questions-réponses devrait connaître à l'avance le type de la réponse attendue pour sélectionner une méthode spécifique pour l'extraction de la réponse. Par exemple, pour la question biomédicale «Does nimotuzumab improve survival of glioblastoma patients?» qui admet «oui» ou «non» comme réponse, une méthode spécifique à l’extraction des réponses aux questions booléennes doit être utilisée. En effet, l'identification des types de questions est une tâche très importante car elle peut fortement affecter positivement ou négativement les étapes de traitement ultérieures: si le type de question n'est pas identifié correctement, les étapes du traitement de la réponse échoueront inévitablement.

\begin{figure}[th]
\captionsetup{justification=justified}
\graphicspath{{Figures/}}
\centering
\includegraphics[width=16cm, height=12cm]{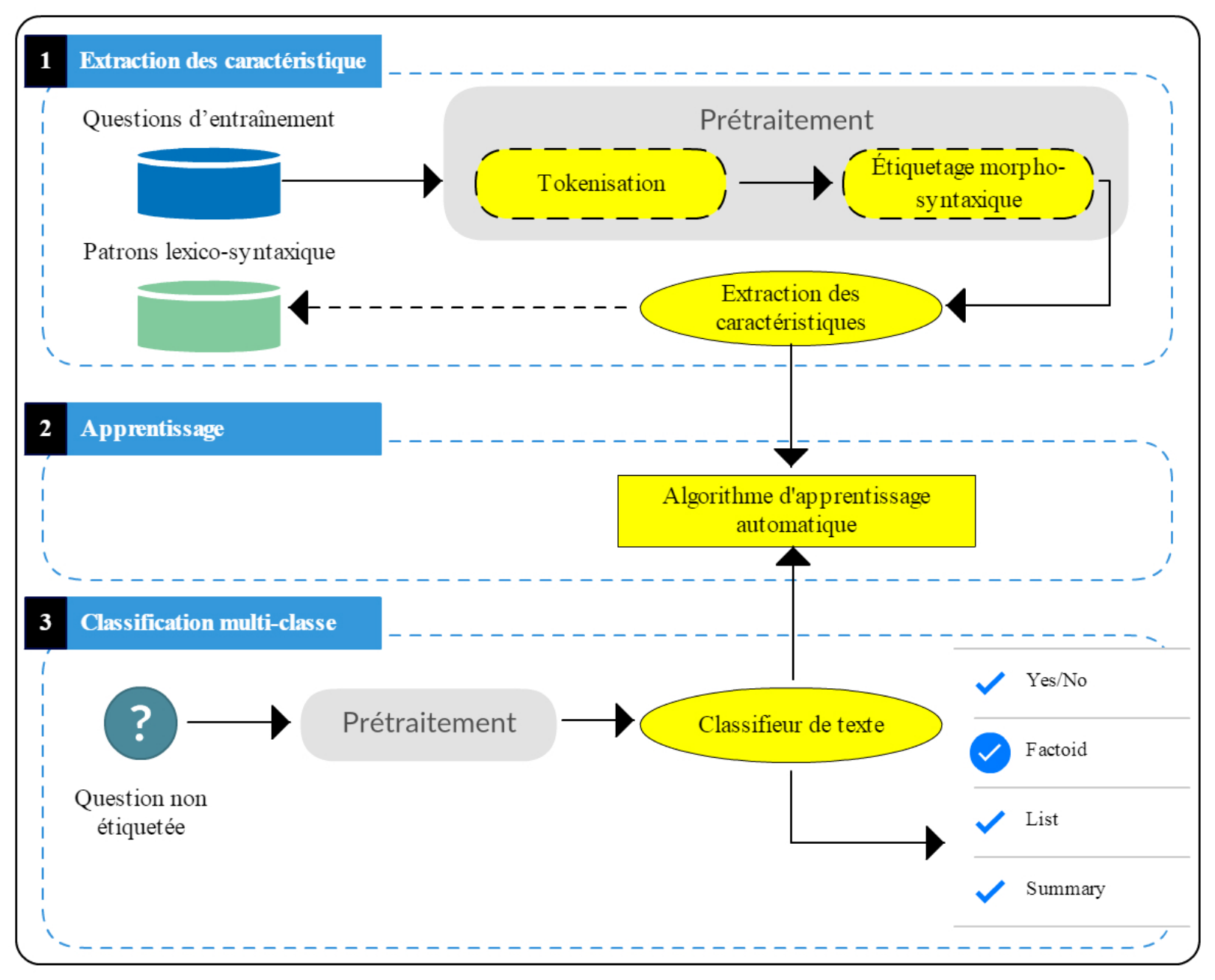}
\caption[Architecture de notre méthode de classification des types des questions biomédicales]{Architecture de notre méthode de classification des types des questions biomédicales}
\label{fig:QTC}
\end{figure}

Dans la méthode proposée, comme le montre la figure~\ref{fig:QTC}, nous avons d'abord extrait les caractéristiques appropriées à partir des questions biomédicales en utilisant nos patrons lexico-syntaxiques prédéfinis et formulés manuellement. Ces patrons ont été construits en analysant les questions d’entrainement fournies par le challenge BioASQ. Nous avons élaboré ces patrons manuellement après l'étape de la tokenisation et l'étiquetage morpho-syntaxique des questions d’entrainement effectuées en utilisant Stanford CoreNLP afin de capturer leur structure syntaxique. En effet, nous avons extrait les mots et leurs classes morphosyntaxiques de chacune des questions d’entrainement BioASQ. Après avoir analysé toutes ces questions appartenant à une catégorie particulière, nous avons proposé les structures syntaxiques possibles. Un patron est une expression régulière décrivant un modèle de question. Par exemple, le patron «{[What $\mid$ Which] + [VBZ] + [*] + [X]+ [*] +?; où X = \{definition, role, aim, effect, influence, mechanism, treatment, ou les synonymes WordNet de ces mots\}}» est l'un des patrons qui permettent de représenter les questions de type résumé. L'idée principale derrière la sélection des caractéristiques employées dans les patrons est que seuls certains mots de la question biomédicale (par exemple, les pronoms interrogatifs) permettent de représenter généralement le type de question.
Considérons par exemple la question biomédicale «What is the definition of autophagy?». Le vecteur de caractéristiques ($v$) de cette question est simplement les patrons qui représentent sa structure syntaxique. Cette question peut être représentée par un ensemble de paires sous forme (caractéristique, fréquence) comme suit: $v= \{(what, 1), (VBZ, 1), (definition, 1)\}$. Pour une question biomédicale donnée, le patron approprié est sélectionné parmi l'ensemble des patrons comme suit: après prétraitement (la tokenisation et l'étiquetage morpho-syntaxique) de la question biomédicale en utilisant Stanford CoreNLP, le patron est apparié à la question biomédicale de gouche à droite. En cas d'échec, nous examinons le patron suivant et ainsi de suite. Nous avons ensuite entraîné un modèle d’apprentissage automatique sur les caractéristiques extraites à partir du corpus d’entraînement. En effet, nous avons exploré plusieurs algorithmes d'apprentissage automatique tels que l’arbre de décision, naïve bayésien et SVM (machine à vecteurs de support). Nous avons retenu le classificateur SVM à cause de sa performance épreuvée en catégorisation des textes. Comme une question peut être attribuée à l'un des quatre types de questions déjà définis, la classification multi-classe a été adoptée dans ce travail. Enfin, nous avons utilisé le modèle de classification appris à l'aide du corpus d'apprentissage pour classifier les questions biomédicales non étiquetées à partir d’autres corpus, dits corpus de test.

Le tableau~\ref{tab:r1f} présente les principaux résultats obtenus sur le corpus standard de questions biomédicales fourni par BioASQ.

\begin{table}[h!]
\selectlanguage{french}

\centering
\caption[Résultats obtenus en utilisant SVM sur cinq lots de questions de tests pour attribuer automatiquement une catégorie aux questions biomédicales]{Résultats obtenus en utilisant SVM sur cinq lots de questions de tests pour attribuer automatiquement une catégorie aux questions biomédicales. Nous avons exploré différentes caractéristiques à savoir unigramme, bigramme, étiquetage morpho-syntaxique, étiquetage morpho-syntaxique + unigramme et notre ensemble de patrons lexico-syntaxiques.}
\label{tab:r1f}
\begin{tabular}{M{3cm}M{9cm}M{3cm}}
\hline\noalign{\smallskip}
Corpus de test&	Caractéristiques &Accuracy (\%) \\
\noalign{\smallskip}\hline\noalign{\smallskip}
\multirow{5}{*}{Cinq lots} & Unigramme & 79.48\\[1.5pt]
             & Bigramme &  	65.18\\[1.5pt]
             &Étiquetage morpho-syntaxique&	77.08\\[1.5pt]
             &Étiquetage morpho-syntaxique+unigramme&	80.08\\[1.5pt]
             &\textbf{Patrons lexico-syntaxiques proposés}&	\textbf{89.40}\\[1.5pt]
\noalign{\smallskip}\hline
\end{tabular}
\end{table}

Les résultats obtenus ont montré que la méthode proposée améliore significativement trois systèmes de l'état de l'art. La méthode proposée a atteint une augmentation d'environ 10 points par rapport au meilleur système de base en termes de précision. De plus, la précision obtenue est de 89.40\% qui représente des améliorations statistiquement significatives par rapport aux systèmes de base.

En général, la performance des approches fondées sur les modèles supervisés dépend à la fois de la présence d’un corpus d’apprentissage bien annoté et la sélection d’un ensemble pertinent d’attributs. Dans le corpus BioASQ utilisé, nous avons identifié certains problèmes d'ambiguïtés des questions. Par exemple, la question biomédicale «Which are the mutational hotspots of the human KRAS oncogene?» qui figure deux fois dans le corpus d'apprentissage est étiquetée avec la catégorie «résumé» et aussi avec la catégorie «liste». Ce problème impacte négativement la performance de la classification des types de questions.

\subsubsection{Identification des sujets des questions biomédicales}
\label{Chapter8.3.2}

Dans la section~\ref{Chapter4.3} du chapitre~\ref{Chapter4}, nous avons proposé une variante basée sur l’apprentissage automatique pour classifier des questions biomédicales en plusieurs sujets. Celle-ci vise à assigner automatiquement un ou plusieurs sujets appelés aussi catégories sémantiques (e.g., pharmacologique, thérapeutique, étiologique, etc.) à des questions biomédicales. La tâche de classification des questions en plusieurs sujets dans les systèmes de questions-réponses permet de déterminer le type sémantique de la réponse attendue. Cette information permet de réduire le nombre de documents parcourus lors de la recherche des réponses candidates et d'éviter d'examiner tous les passages ou entités nommées biomédicales dans la liste complète de réponses candidates, en se restreignant à une catégorie unique de réponses telle que traitement, test, etc. Par exemple, la question «What is the best way to catch up on the diphtheria pertussis tetanus vaccine (DPT) after a lapse in the schedule?» appartient à la catégorie pharmacologique, et par la suite le système de questions-réponses peut se limiter à la base de données pharmacologiques «Micromedex» comme ressource pour chercher la réponse.

\begin{figure}[h!]
\captionsetup{justification=justified}
\graphicspath{{Figures/}}
\centering
\includegraphics[width=16cm, height=12cm]{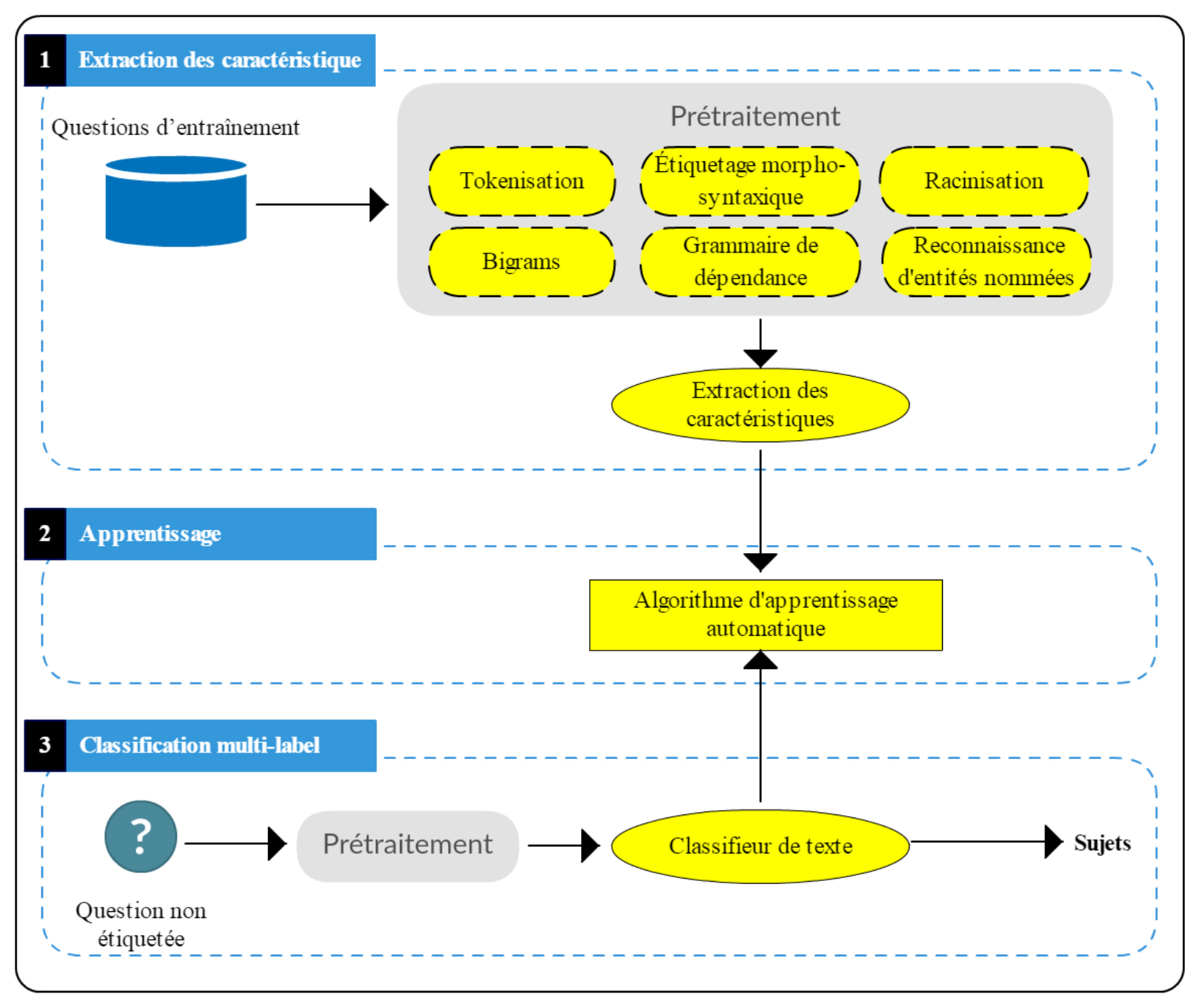}
\caption[Architecture de notre méthode de classification des sujets des questions biomédicales]{Architecture de notre méthode de classification des sujets des questions biomédicales.}
\label{fig:QSC}
\end{figure}

Comme le montre la figure~\ref{fig:QSC}, la méthode proposée extrait d'abord un ensemble de caractéristiques lexicales, syntaxiques et sémantiques à partir des questions cliniques, y compris les mots bruts, les bigrammes, les racines des mots, les concepts biomédicaux et leurs types sémantiques, et les relations syntaxiques de dépendance. Nous avons utilisé Porter stemmer \citep{Porter_1980}, le parseur Stanford \citep{de2006generating} et MetaMap \citep{aronson2001effective} pour extraire respectivement les racines des mots, les relations de dépendance syntaxique, et les concepts biomédicaux et leurs types sémantiques. Ensuite, nous avons entraîné un modèle d’apprentissage automatique sur les caractéristiques extraites à partir du corpus d’entraînement. En ce qui concerne les modèles d'apprentissage automatique, nous avons exploré plusieurs classificateurs tels que SVM, naïve bayésien et les arbres de décision. Nous avons adopté la classification multi-étiquettes dans cette étude puisqu'une question peut correspondre à un ou plusieurs sujets. Nous avons développé un classificateur d'apprentissage automatique binaire pour chacun des sujets prédéfinis dans le corpus fourni par NLM. Ces sujets sont: \emph{device}, \emph{diagnosis}, \emph{epidemiology}, etiology, \emph{history}, \emph{management}, \emph{ pharmacological}, \emph{physical}, \emph{finding}, \emph{procedure}, \emph{prognosis}, \emph{test}, \emph{treatment} et \emph{prevention}. Nous avons opté pour le classificateur SVM avec le noyau linéaire puisqu'il assure de meilleures performances en catégorisation de textes par rapport aux autres classificateurs \citep{yu2005classifying,yu2008automatically,Cao_2010}. Enfin, pour classifier les questions cliniques non étiquetées, nous avons adopté le modèle de classification appris précédemment en utilisant le corpus des questions cliniques fournies par NLM.

Pour tester notre méthode de classification des sujets de questions, nous avons mené plusieurs expérimentations sur le corpus standard de questions cliniques fourni par NLM. Ce corpus contient 4654 questions cliniques, recueillies auprès des experts en domaine de santé à travers les États-Unis. Les résultats obtenus, rapportés dans le tableau~\ref{tab:r2f}, montrent que la méthode proposée atteint une meilleure performance de 77.18 en termes de F1-score pour la classification des sujets de questions. Ces résultats prouvent que notre méthode est plus efficace que les méthodes actuelles présentées dans \citep{yu2008automatically,Cao_2010} et les surpasse d'une moyenne de F1-score de 4.5\% et 4.73\% respectivement en utilisant naïve bayésien et SVM. Les détails sur les expérimentations menées ainsi que les résultats obtenus sont présentés à la section~\ref{Chapter4.3} du chapitre~\ref{Chapter4}.

\begin{landscape} 
\centering
\begin{table}
\selectlanguage{french}
\centering
\caption[Résultats obtenus en utilisant SVM pour attribuer automatiquement des sujets à des questions cliniques]{Résultats obtenus en utilisant SVM pour attribuer automatiquement des sujets à des questions cliniques. Nous avons exploré différentes caractéristiques à savoir unigrame (BOW), bigramme (BOB), racines des mots (BOS) en utilisant Porter et Krovetz, entités nommées biomédicales (BOBNE), BOW+les concepts biomédicaux et leurs types sémantiques (BOCST), BOW+les relations syntaxiques de dépendance (BOSDR), BOW+BOB+BOCST+BOS$_{porter}$ utilisés dans \citep{yu2008automatically,Cao_2010}, BOW+BOB+BOCST+BOS$_{krovetz}$, et la représentation proposée BOW+BOB+BOCST+BOSDR+BOS$_{porter}$.}
\label{tab:r2f}
\begin{tabular}{M{4.5cm}M{1.2cm}M{1.2cm}M{1.2cm}M{1.1cm}M{1.1cm}M{1.2cm}M{1.4cm}M{1.6cm}M{1.6cm}M{1.6cm}}
\hline\noalign{\smallskip}
 \multirow{4}{*}{Topics}&\multicolumn{9}{c}{Features} \\\cmidrule(l){2-11}
&&&&&&&\multicolumn{2}{c}{\thead{\cite{yu2008automatically}\\ \cite{Cao_2010}}}&\multicolumn{2}{c}{\thead{Proposed method}}\\\cmidrule(l){8-9}\cmidrule(l){10-11}
 &\multirow{2}{*}{BOW}& \multicolumn{2}{c}{BOS}& \multirow{2}{*}{BOBNE} &\multirow{2}{*}{\thead{BOW+\\BOCST}}& \multirow{2}{*}{\thead{BOW+\\BOSDR}}& \multicolumn{2}{c}{\thead{BOW+BOB+BOCST+}}& \multicolumn{2}{c}{\thead{BOW+BOB+BOSDR+BOCST+}}\\
\cmidrule(l){3-4}\cmidrule(l){8-9}\cmidrule(l){10-11}
&  &Porter& Krovetz&&&&BOS$_{porter}$& BOS$_{krovetz}$&BOS$_{porter}$& BOS$_{krovetz}$    \\

\noalign{\smallskip}\hline\noalign{\smallskip}
Device&57.89\%&58.23\%&57.69\%&56.31\%&65.60\%&60.10\%&74.01\%&73.50\%&74.99\%&74.44\%\\
Diagnosis&74.22\%&74.24\%&73.14\%&70.38\%&75.63\%&75.19\%&77.13\%&75.33\%&78.10\%&77.05\%\\
Epidemiology&71.15\%&70.58\%&68.10\%&54.96\%&72.03\%&71.97\%&74.74\%&72.31\%&75.93\%&73.53\%\\
Etiology&80.31\%&80.67\%&78.64\%&75.05\%&81.02\%&80.95\%&82.47\%&80.07\%&83.11\%&81.67\%\\
History&52.72\%&55.69\%&55.03\%&50.96\%&58.57\%&54.31\%&67.18\%&66.52\%&68.75\%&68.09\%\\
Management&69.70\%&69.48\%&68.51\%&65.07\%&70.13\%&70.02\%&71.07\%&70.16\%&71.49\%&70.10\%\\
Pharmacological&82.41\%&82.83\%&82.16\%&76.20\%&84.66\%&83.04\%&84.71\%&84.04\%&84.85\%&84.19\%\\
Physical\&Finding&72.10\%&72.27\%&71.08\%&70.62\%&76.09\%&73.14\%&78.82\%&77.01\%&79.35\%&78.38\%\\
Procedure&69.56\%&70.08\%&68.81\%&68.18\%&75.47\%&71.32\%&78.68\%&77.42\%&79.12\%&78.35\%\\
Prognosis&72.68\%&73.68\%&72.13\%&69.27\%&73.89\%&72.87\%&74.03\%&72.51\%&74.17\%&73.61\%\\
Test&79.97\%&80.14\%&76.40\%&75.02\%&81.15\%&80.52\%&83.22\%&82.48\%&83.64\%&81.90\%\\
Treatment\&Prevention&68.19\%&69.00\%&67.40\%&65.16\%&69.99\%&69.21\%&71.73\%&70.10\%&72.63\%&71.91\%\\\cmidrule(l){1-11}
Average&70.91\%&71.41\%&69.92\%&66.43\%&73.68\%&71.89\%&76.48\%&75.12\%&77.18\%&76.10\%\\

\noalign{\smallskip}\hline
\end{tabular}
\end{table}

\end{landscape}

\subsection{Méthode proposée pour la sélection des documents pertinents}

Dans la section~\ref{Chapter5.2} du chapitre~\ref{Chapter5}, nous avons proposé une méthode de la recherche, à partir de la base de données MEDLINE, des documents susceptibles de contenir la réponse à une question biomédicale donnée.

La recherche de documents pertinents pour une requête se fait en utilisant un moteur de recherche existant. À une requête donnée, un moteur de recherche retourne un ensemble de documents, en attribuant un score d’appariement à chaque document. Plus le score est élevé, plus le document est pertinent. Cependant, dans certains cas le moteur de recherche renvoie des documents non pertinents à l'entête de la liste des documents retournés (avec un score élevé), ou aussi des documents pertinents mais à la fin de la liste (avec un score réduit). Ce qui se répercute négativement sur la performance des SQR puisque l'extraction des réponses se fait à partir des documents qui figurent à l'entête de la liste.

\begin{figure}[th]
\captionsetup{justification=justified}
\graphicspath{{Figures/}}
\centering
\includegraphics[width=16cm, height=12cm]{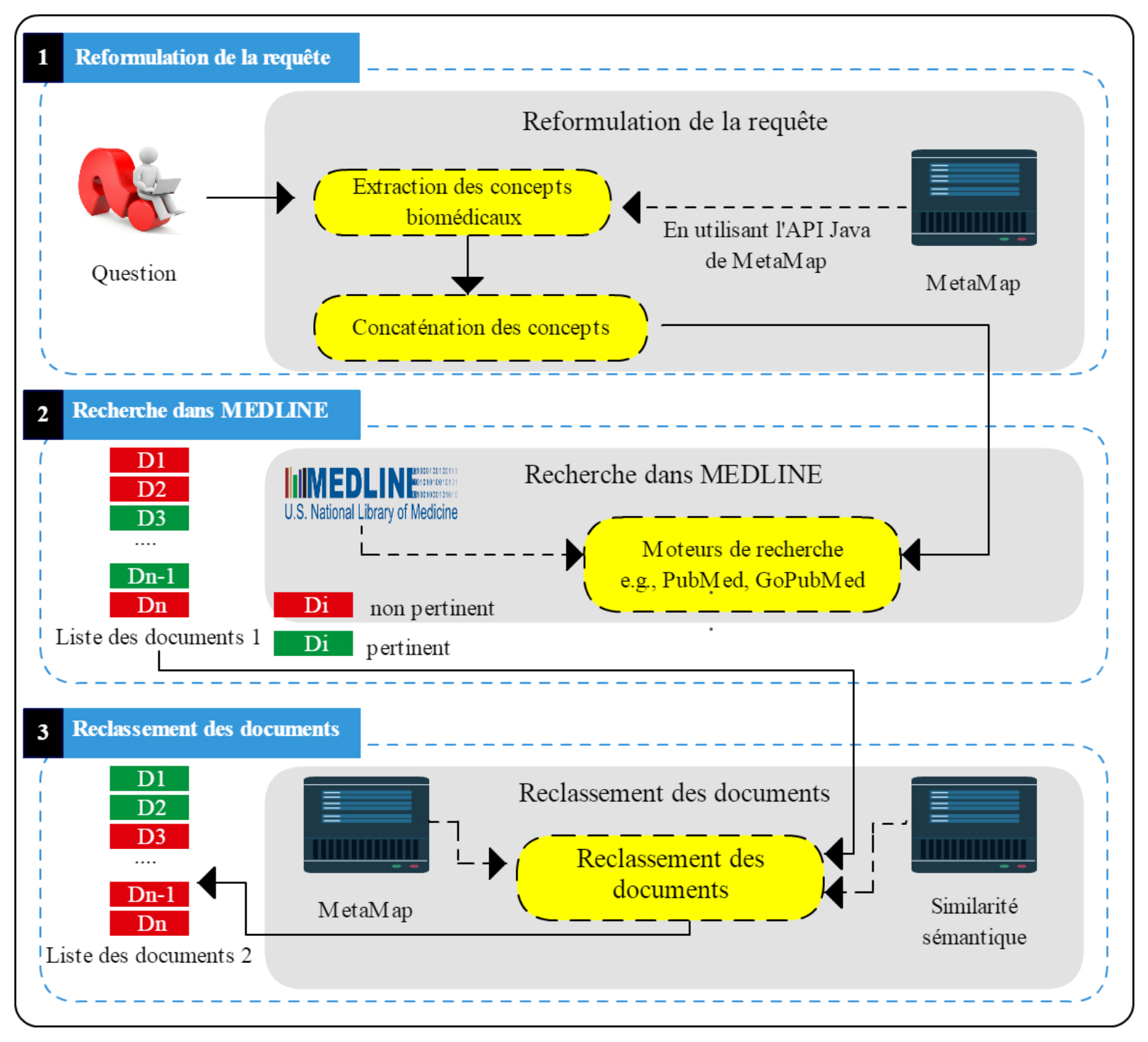}
\caption[Architecture de notre méthode de recherche des documents pertinents]{Architecture de notre méthode de recherche des documents pertinents.}
\label{fig:RD}
\end{figure}

Pour remédier à ce problème, nous avons présenté une méthode de recherche des documents pertinents qui se base sur le reclassement de documents. La figure~\ref{fig:RD} illustre les différentes étapes de la méthode proposée.

Notre méthode de recherche des documents pertinents à une question biomédicale est constituée de plusieurs étapes. D'abord, nous avons construit la requête à partir de la question en utilisant les concepts UMLS correspondant aux termes de la question extraits par l’outil MetaMap. Ensuite, nous procédons à la recherche des documents potentiellement pertinents à la requête, que nous nous avons construite, en utilisant le moteur de recherche GoPubMed qui accède à la base de données MEDLINE. Enfin, une technique de reclassement de documents est effectuée pour reclasser la liste des documents préalablement obtenue. Notre technique de reclassement repose sur la similarité sémantique entre la question et les titres des documents retournés dans le but d'attribuer un score pour chaque document. Pour calculer les scores de similarité sémantique, nous avons d’abord identifié les concepts biomédicaux contenus à la fois dans la question et les titres des documents déjà obtenus en utilisant l'outil MetaMap qui se réfère au metathesaurus l’UMLS. Ensuite, les scores entre les concepts de la question et ceux des titres des documents sont calculés sur la base de la distance de similarité sémantique qui correspond à la longueur du chemin (Path Length) dans l’ontologie Mesh.

Pour évaluer notre méthode, nous avons mené plusieurs expérimentations sur le corpus standard de questions biomédicales fourni par BioASQ. Comme mesures d'évaluation, nous avons utilisé le rappel, précision, F1-mesure et précision moyenne (MAP). Le tableau~\ref{tab:r4f} présente les résultats obtenus en comparaison aux principales méthodes de l'état de l'art. Nous pouvons ramarquer que notre méthode est plus efficace que les autres méthodes présentées dans \citep{balikas2014results} qui ont été classées parmi les 10 premiers systèmes lors du challenge BioASQ. Les détails concernant les expérimentations menées ainsi que les résultats obtenus sont présentés à la section~\ref{Chapter5.2} du chapitre~\ref{Chapter5}.

\begin{table}[h!]
\centering
\selectlanguage{french}
\caption{Comparaison des résultats expérimentaux du système de recherche de documents proposé et des systèmes de l'état de l'art présentés dans \citep{balikas2014results}.}
\label{tab:r4f}
\begin{tabular}{p{4cm}p{2.7cm}p{2.5cm}p{3.1cm}p{1.9cm}}
\hline \noalign{\smallskip}
Système &Précision &Rappel & F1-mesure &MAP  \\
\noalign{\smallskip}\hline\noalign{\smallskip}
\textbf{Notre système}&\textbf{0.2331} &\textbf{0.3644} &\textbf{0.2253} &\textbf{0.2758} \\
SNUMedinfo1 &0.0457& 0.5958 &0.0826& 0.2612 \\
SNUMedinfo3 &0.0457 &0.5947 &0.0826 &0.2587 \\
SNUMedinfo2 &0.0451 &0.5862 &0.0815 &0.2547\\
SNUMedinfo4 &0.0457 &0.5941 &0.0826 &0.2493\\
SNUMedinfo5 &0.0459 &0.5947 &0.0829 &0.2410\\
Top 100 Baseline &0.2274 &0.4342 &0.2280 &0.1911 \\
Top 50 Baseline &0.2290 &0.3998 &0.2296 &0.1888 \\
Main system &0.0413 &0.2625 &0.0678 &0.1168 \\
Biomedical Text Ming &0.2279 &0.2068 &0.1665 &0.1101  \\
Wishart-S2 &0.1040 &0.1210 &0.0793 &0.0591 \\

\noalign{\smallskip}\hline

\end{tabular}
\end{table}

\subsection{Méthode proposée pour la recherche des passages pertinents}
\label{Chapter8.3.3}

Dans la section~\ref{Chapter5.3} du chapitre~\ref{Chapter5}, nous avons proposé une méthode pour la recherche des passages susceptibles de contenir la réponse à une question biomédicale donnée. La définition d'un passage dépend nécessairement du système de recherche des passages, généralement un passage pourrait être une section, un paragraphe ou une phrase. En particulier, un système de recherche des passages peut être défini comme étant un type d’application spécifique de la recherche d’information qui permet de récupérer un ensemble de passages plutôt que de fournir un ensemble de documents. Son but principal est d’extraire et de retourner les passages les plus pertinents qui servent comme des réponses candidates à partir desquelles le système de questions-réponses extraira et sélectionnera la réponse.

Bien que la recherche des passages soit une tâche difficile, elle reste l'une des composantes les plus importantes pour le développement d'un système de questions-réponses. Cela s'explique par le fait que la performance globale d'un SQR dépend fortement de l'efficacité de son processus de recherche des passages. En effet, si le module de recherche des passages ne parvient pas à trouver au moins un passage pertinent pour une question donnée, l’étape de traitement ultérieure pour l’extraction de la réponse finale échouera inévitablement.

\begin{figure}[th]
\captionsetup{justification=justified}
\graphicspath{{Figures/}}
\centering
\includegraphics[width=16cm, height=12cm]{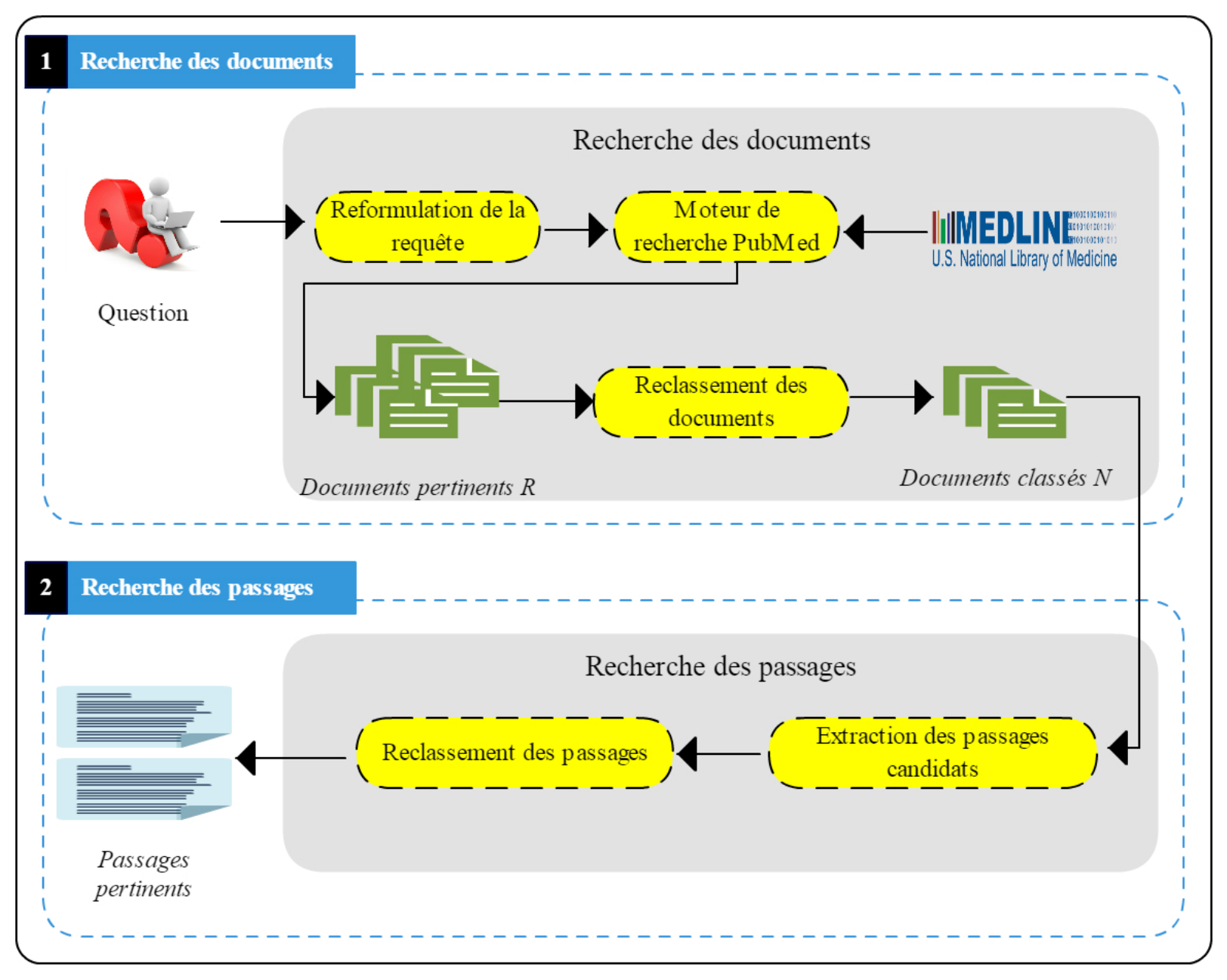}
\caption[Architecture de notre méthode de recherche des passages pertinents]{Architecture de notre méthode de recherche des passages pertinents.}
\label{fig:RP}
\end{figure}

Pour sélectionner les passages pertinents dans un SQR, nous avons proposé une nouvelle méthode de recherche des passages. Comme le montre la figure~\ref{fig:RP}, notre méthode comporte deux étapes, : (1) la recherche des documents, et (2) l’extraction des passages. La recherche des documents pertinents à une question se fait conformément à notre module précédent de recherche des documents. Ensuite, les passages pertinents sont extraits à partir des résumés de ces documents les plus pertinents. Pour ce faire, nous avons segmenté les résumés des documents sélectionnés en des phrases en utilisant l’outil Stanford CoreNLP \citep{Manning_2014}. L'ensemble des phrases obtenues constitue la liste des passages candidats. La longueur des passages est similaire à celle des phrases. Les scores sont calculés pour tous les passages candidats en fonction de leur degré d’appariement (ou de pertinence) avec la question en se basant sur le modèle BM25 introduit par \cite{Robertson96okapiat}. Ce modèle est le plus connu en recherche d’information. Comme caractéristiques représentatives des passages, nous avons combiné les racines des termes établies en utilisant le stemmer Porter \citep{Porter_1980}, et les concepts biomédicaux établis en utilisant MetaMap qui se réfère au metathesaurus de l’UMLS (version 2016AA). De plus, les mots vides sont éliminés au préalable et la désambiguïsation sémantique des concepts avec MetaMap \citep{Aronson_2010} est appliquée lors de l'extraction des concepts UMLS. Enfin, après avoir calculé les scores de pertinence pour tous les passages candidats, nous avons retenu les passages ayant les scores les plus élevés constituant ainsi les passages les plus pertinents.

Pour montrer l’efficacité de notre méthode, nous avons effectué un ensemble d’expérimentations sur un corpus d’évaluation standard fourni par le challenge BioASQ. Comme indicateurs d'efficacité, nous avons utilisé les mesures d'évaluation des systèmes de recherche d'information les plus connues à savoir le rappel, la précision, F1-mesure et la précision moyenne (MAP). Le tableau~\ref{tab:r5f} présente les résultats obtenus en termes de MAP en comparaison aux principaux systèmes de l'état de l'art. Ces résultats montrent clairement que la méthode proposée a abouti à des meilleures performances par rapport aux autres méthodes. De plus, notre méthode apporte une amélioration de la performance statistiquement significative qui atteint 6.84\% en moyenne par rapport aux autres méthodes de la littérature présentées dans \citep{neves2015hpi,yenalaiiith,ligeneric,yang2015learning}. D'autres expérimentations détaillées sont présentés en section~\ref{Chapter5.3} du chapitre~\ref{Chapter5}.

\begin{table}[h!]
\centering
\selectlanguage{french}
\caption[Comparaison en termes de MAP de la méthode de recherche des passages proposée avec celles de l'état de l'art sur cinq lots de jeux de données de test fournis par le challenge BioASQ]{Comparaison en termes de MAP de la méthode de recherche des passages proposée avec celles de l'état de l'art sur cinq lots de jeux de données de test fournis par le challenge BioASQ. Le signe «-» remplace les scores des systèmes qui n'ont pas été évalués sur certains lots.}
\label{tab:r5f}
\begin{tabular}{M{4.2cm}M{1.9cm} M{1.9cm}M{1.9cm}M{1.9cm}M{1.9cm}}
\hline \noalign{\smallskip}
\multirow{2}{*}{Méthodes}& \multicolumn{5}{c}{MAP@10} \\
\cmidrule(l){2-6}
& Lot 1 & Lot 2 &Lot 3 &Lot 4&Lot 5 \\
\noalign{\smallskip}\hline\noalign{\smallskip}
\textbf{Méthode proposée}&\textbf{0.1178}&\textbf{0.1180}&\textbf{0.1491}&\textbf{0.1633}&\textbf{0.1437}\\\cmidrule(l){1-6}
\cite{neves2015hpi}&0.0347 (+8.31\%)& 0.0355 (+8.3\%)& 0.0452 (+10.39\%)& 0.0624 (+10.09\%)& 0.0572 (+8.65\%)\\\cmidrule(l){1-6}
\cite{yenalaiiith}&0.0545 (+6.33\%)& 0.0709 (+4.71\%)& -& 0.0913 (+7.2\%)& -\\\cmidrule(l){1-6}
\cite{ligeneric}&0.0570 (+6.08\%)& 0.0521 (+6.59\%)& 0.0832 (+6.59\%)& 0.0936 (+6.97\%)& 0.1201 (+2.36\%)\\\cmidrule(l){1-6}
\cite{yang2015learning}&-&-&0.0892 (+5.99\%)&0.0957 (+6.76\%)&0.1027 (+4.1\%)\\
\noalign{\smallskip}\hline

\end{tabular}
\end{table}

\subsection{Méthodes proposées pour l'extraction de la réponse}
\label{Chapter8.3.4}

Dans la section~\ref{Chapter6.2} du chapitre~\ref{Chapter6}, nous avons proposé des nouvelles méthodes d'extraction des réponses. Celles-ci permettent d'extraire des réponses \emph{exactes} et \emph{idéales} (résumés) aux questions biomédicales exprimées en langage naturel.

L’extraction des réponses est la dernière étape pour répondre à une question visant à traiter l’ensemble des réponses candidates afin d’en extraire ou générer la (les) réponse (s) finale (s). Ce processus constitue l’étape la plus difficile d’un système de questions-réponses puisqu'il s'agit d'extraire la réponse finale qui devrait être une réponse courte et précise. A titre d'exemple, la question biomédicale «What is the cause of Phthiriasis Palpebrarum?» admet comme réponse «Pthirus pubis». En général, les réponses appropriées aux questions des utilisateurs devraient être extraites selon le type de la question donnée. Autrement dit, la nature de la réponse varie suivant le type de la question. Par exemple, une question biomédicale comme «Does the histidine-rich Ca-binding protein (HRC) interact with triadin?» s’attend à une réponse de type «oui» ou «non», alors qu'une question biomédicale comme «What is the role of edaravone in traumatic brain injury?» s’attend à une réponse de type «résumé».

Dans ce travail de thèse, notre but final est de développer un nouveau système de questions-réponses qui a la capacité de traiter quatre types de questions à savoir les questions booléennes, les questions factuelles, les questions de type liste et les questions de type résumé. Ces types de questions biomédicales devraient couvrir toute sorte de questions potentielles qui peuvent être posées par les utilisateurs. Dans la section~\ref{Chapter8.3.5}, nous avons développé des nouvelles méthodes d’extraction des réponses pour chaque type de question défini précédemment.

Pour les questions booléennes, nous nous sommes basés sur le lexique de sentiment SentiWordNet \citep{baccianella2010sentiwordnet} pour extraire la réponse exacte («oui» ou «non»). SENTIWORDNET qui est une ressource lexicale pour l'analyse des sentiments et l'exploration de l'opinion, assigne à chaque synset de WordNet trois scores de sentiment: «positivité», «négativité» ou «objectivité». L'idée clé derrière l'utilisation de la ressource lexicale SentiWordNet 3.0 est qu'elle est le résultat de l'annotation automatique de tous les synsets de WordNet 3.0 selon les notions de «positivité», «négativité» et «neutralité». WordNet, qui est une grande base de données lexicale, comprend 155 287 mots et 117 659 synsets, appelés également synonymes \citep{Miller_1995}. En effet, dans la méthode proposée, un score SentiWordNet est attribué à chaque mot des passages pertinents, et la décision de retourner «oui» ou «non» dépend du nombre de réponses candidates positives ou négatives. Pour les questions factuelles, la réponse précise est produite en identifiant les entités nommées biomédicales qui figurent dans les réponses candidates, et en rapportant les cinq entités nommées biomédicales les plus fréquentes et leurs synonymes. En effet, nous avons d'abord identifié les entités nommées biomédicales en utilisant MetaMap qui se réfère au metathesaurus de l’UMLS (version 2016AA). Nous avons ensuite classé l'ensemble obtenu des entités nommées biomédicales en utilisant la fréquence du terme. La fréquence d'une entité est simplement le nombre d'occurrences de cette entité dans la liste des entités considérée. Nous avons exploré plusieurs méthodes de pondération telles que TF-IDF et BM25, en montrant que la fréquence du terme a atteint le meilleur résultat pour cette tâche. Enfin, les entités nommées biomédicales les mieux classées et leurs synonymes sont considérés comme réponses à la question factuelle. En effet, nous avons exclu les entités nommées mentionnées dans la question. Les synonymes sont récupérés en utilisant les services Web de BioPortal \footnote{\url{http://data.bioontology.org/documentation}}. Comme décrit par le challenge BioASQ, une question factuelle a une seule réponse correcte, mais cinq réponses possibles au maximum et leurs synonymes sont autorisés. Pour les questions de type liste, la réponse exacte est produite de la même manière, sauf que les entités nommées les plus fréquentes et leurs synonymes les plus pertinents sont cette fois renvoyées sous forme de liste qui répond à la question. En effet, la réponse exacte est la même pour les questions factuelles, mais l'interprétation est différente pour les questions de type liste: toutes les entités les mieux classées sont considérées comme faisant partie de la même réponse à une question de type liste et non pas comme candidates. Pour générer des réponses idéales, nous avons d’abord appliqué MetaMap aux passages pertinents et à la question biomédicale afin d’obtenir les concepts biomédicaux auxquels ils se réfèrent. Nous avons ensuite classé ces passages par leur pertinence à la question en utilisant les racines des mots et les concepts biomédicaux comme termes d’index et le modèle BM25 pour calculer les scores d’appariements entre la question et les passages. Enfin, nous avons retourné les deux meilleurs passages (concaténés) comme réponse idéale. Les détails sur ces méthodes d'extraction des réponses sont présentés dans la section~\ref{Chapter6.2} du chapitre~\ref{Chapter6}.

Nous avons testé les méthodes d'extraction des réponses proposées dans le cadre du challenge BioASQ 2017. La tâche demandée consistait à extraire des réponses exactes et idéales aux questions biomédicales posées en langage naturel. Plusieurs systèmes ont participé
en proposant différentes approches. Nous avons également évalué nos méthodes sur un ensemble de questions fournies par BioASQ dans leurs éditions 2015 et 2016. Comme indicateurs d’efficacité de l'extraction des réponses: l'accuracy a été utilisée pour les réponses exactes des questions booléennes; la moyenne des réciproques du rang (MRR) a été utilisée pour les réponses exactes des questions factuelles; la précision, le rappel et F1-mesure moyenne ont été utilisés pour les réponses exactes des questions de type liste; ROUGE-2 et ROUGE-SU4 ont été utilisés pour des réponses idéales. De plus, pour ces dernières, les experts du challenge BioASQ ont évalué manuellement les réponses retournées par les systèmes participants en termes de la simplicité de leur lecture, de rappel, de précision et de répétition. Les tableaux~\ref{tab:ex1} et~\ref{tab:ex2} présentent les résultats obtenus lors de la participation au challenge BioASQ 2017. Ces résultats sont publiés par le site officiel du challenge BioASQ \footnote{\url{http://participants-area.bioasq.org/results/5b/phaseB/}}.

\begin{table}[h!]
\centering
\selectlanguage{french}
\caption[Résultats obtenus lors de notre participation dans la phase B «réponses exactes» de la tâche 5b du challenge BioASQ 2017 en utilisant les méthodes proposées l'extraction des réponses]{Résultats obtenus lors de notre participation dans la phase B «réponses exactes» de la tâche 5b du challenge BioASQ 2017 en utilisant les méthodes proposées l'extraction des réponses. Les deux premières valeurs mises entre parenthèses indiquent notre classement et le nombre total de soumissions, les deux seconds valeurs entre parenthèses représentent notre classement et le nombre total d'équipes participantes.}
\label{tab:ex1}
\begin{tabular}{p{3cm}p{2.2cm}p{2.2cm}p{2.1cm}p{2.1cm}p{2.1cm}}
\hline\noalign{\smallskip}
 \multirow{2}{*}{Jeux de données} &  Booléennes & Factuelle & \multicolumn{3}{c}{ Liste}   \\
 \cmidrule(l){2-2}\cmidrule(l){3-3} \cmidrule(l){4-6}
  &   Accuracy &  MRR& Précision& Rappel &F1-mesure \\
\noalign{\smallskip}\hline\noalign{\smallskip}
\multirow{2}{*}{Lot 1}  &0.7647&	0.2033 \newline(5/15)\newline (3/9)&	0.1909&	0.2658&	0.2129 (3/15)\newline (2/9) \\
\cmidrule(l){1-6}
\multirow{2}{*}{Lot 2}  &0.7778& 0.0887 \newline(10/21)\newline (5/9)&	0.2400&	0.3922&	0.2920 (6/21)\newline (2/9)\\
\cmidrule(l){1-6}
\multirow{2}{*}{Lot 3}  &0.8387 \newline(1/21)\newline (1/10)& 0.2212 \newline(9/21)\newline (4/10)&	0.2000&	0.4151&	0.2640 (6/21)\newline (3/10)\\
\cmidrule(l){1-6}
\multirow{2}{*}{Lot 4} &0.6207 \newline(2/27)\newline (2/11)& 0.0970 \newline(13/27)\newline (5/11)&	0.1077&	0.2013&	0.1369 (12/27)\newline (5/11)\\
\cmidrule(l){1-6}
\multirow{3}{*}{Lot 5} &0.4615 & 0.2071 \newline(9/25)\newline (3/11)&	0.2091&	0.3087&	0.2438 (11/25)\newline (6/11)\\

\noalign{\smallskip}\hline
\end{tabular}
\end{table}

\begin{table}[h!]
\centering
\selectlanguage{french}
\caption[Résultats obtenus lors de notre participation dans la phase B «Réponses idéales» de la tâche 5b du challenge BioASQ 2017 en utilisant les méthodes proposées pour l'extraction des réponses]{Résultats obtenus lors de notre participation dans la phase B «Réponses idéales» de la tâche 5b du challenge BioASQ 2017 en utilisant les méthodes proposées pour l'extraction des réponses. Les deux premières valeurs mises entre parenthèses indiquent notre classement et le nombre total de soumissions, les deux seconds valeurs entre parenthèses représentent notre classement et le nombre total d'équipes participantes.}
\label{tab:ex2}
\begin{tabular}{p{2.8cm}p{1.9cm}p{2.5cm}p{1.9cm}p{1.2cm}p{1.4cm}p{1.6cm}}
\hline\noalign{\smallskip}
 \multirow{2}{*}{Jeux de données} & \multicolumn{2}{c}{Scores automatiques} &  \multicolumn{4}{c}{Scores manuels}\\
\cmidrule(l){2-3}\cmidrule(l){4-7}
 & ROUGE-2& ROUGE-SU4 & Lisibilité&  Rappel& Précision& Répétition \\
\noalign{\smallskip}\hline\noalign{\smallskip}
\multirow{2}{*}{Lot 1}  &0.4943 \newline(4/15)\newline(2/9)&0.5108 \newline(3/15)\newline(2/9)&3.65 \newline(3/15)\newline(2/9)&	4.42 \newline (3/15)\newline(2/9)&	3.90 \newline(3/15)\newline(2/9)&	3.89 (3/15)\newline(2/9)\\
\cmidrule(l){1-7}
\multirow{2}{*}{Lot 2}  & 0.4579 \newline(4/21)\newline (2/9)& 0.4583\newline (4/21)\newline (2/9)&3.68\newline (6/21)\newline (2/9)&	4.59 (2/21)\newline (2/9)&	4.01 (7/21)\newline (2/9)&	3.91 (7/21)\newline (2/9)\\
 \cmidrule(l){1-7}
\multirow{2}{*}{Lot 3}  & 0.5566 \newline(4/21)\newline (2/10)&	0.5656 \newline(4/21)\newline (2/10)&3.91\newline (6/21)\newline (2/10)&	4.64 \newline (1/21)\newline (2/10)&	4.07 (7/21)\newline (2/10)&	4.00 (7/21)\newline (2/10)\\
 \cmidrule(l){1-7}
\multirow{2}{*}{Lot 4} &0.5895\newline (4/27)\newline (2/11)&	0.5832 \newline(4/27)\newline (2/11)&3.86\newline (7/27)\newline (3/11)&	4.51 \newline (6/27)\newline (3/11)&	4.02 \newline(3/27)\newline (2/11)&	3.95 (6/27)\newline (2/11)\\
\cmidrule(l){1-7}
\multirow{3}{*}{Lot 5} & 0.5772 (7/25)\newline (3/11)&	0.5756\newline (7/25)\newline (3/11)&3.82\newline (8/25)\newline (3/11)&	4.53 (5/25)\newline (2/11)	&3.91 (7/25)\newline (3/11)&	3.90 (7/25)\newline (2/11)\\

\noalign{\smallskip}\hline
\end{tabular}
\end{table}

Les résultats obtenus montrent que les méthodes d'extraction des réponses proposées sont plus compétitives par rapport à celles des systèmes participants. De plus, notre système a été l'un des systèmes gagnants\footnote{\url{http://www.bioasq.org/participate/fifth-challenge-winners}} lors du challenge BioASQ 2017. Les détails concernant les expérimentations menées ainsi que les résultats obtenus sur d'autres jeux de données sont présentés en section~\ref{Chapter6.2} du chapitre~\ref{Chapter6}.

\subsection{SemBioNLQA : un système de questions-réponses sémantique pour le domaine biomédical}
\label{Chapter8.3.5}

Nous avons développé dans ce travail de thèse un système intégrant l'ensemble des méthodes proposées permettant de répondre aux différents types des questions biomédicales posées en langage naturel. Selon la classification des systèmes de questions-réponses présentée par \cite{athenikos2010biomedical}, l’approche adoptée par notre système est une approche sémantique. Notre système de questions-réponses appelé SemBioNLQA est présentée en détail dans la section~\ref{Chapter6.3} du chapitre~\ref{Chapter6}.

Le système SemBioNLQA accepte en entrée une variété de questions posées en langage naturel et génère des réponses appropriées exactes et idéales. Il permet de traiter plusieurs types de questions dans le domaine biomédical: les questions booléennes, les questions factuelles, les questions de type liste et les questions de type résumé. Il fournit des réponses exactes «oui»/«non» pour les questions booléennes, les entités nommées biomédicales pour les questions factuelles et une liste d’entités nommées biomédicales pour les questions de type liste. De plus, il retourne également des réponses idéales pour les questions de type résumé.

Comme nous l'avons déjà mentionné, le système SemBioNLQA comporte quatre composants principaux: (1) analyse et classification des questions, (2) sélection des documents pertinents, (3) recherche des passages pertinents, et (4) extraction des réponses. Nous avons développé le système SemBioNLQA en intégrant les méthodes proposées pour ces quatre composantes. En effet, SemBioNLQA prend en entrée une question biomédicale exprimée en langage naturel et inclut le prétraitement de la question, l’identification du type de la question et le type de la réponse attendue en utilisant notre méthode proposée pour la classification des types des questions biomédicales, ainsi que la construction d’une requête. Ensuite, il se fonde sur nos méthodes de recherche des documents et des passages pour effectuer l’appariement requête-documents, et sélectionner les passages pertinents. Enfin, à partir des passages les mieux classés, SemBioNLQA génère et renvoie à la fois la réponse exacte et idéale en fonction du type de la question et la réponse attendue en utilisant notre composant d’extraction des réponses qui a été l’un des premiers vainqueurs lors de la compétition internationale BioASQ 2017.

Afin d’évaluer l’efficacité de notre système SemBioNLQA et de le comparer à d'autres SQR existants, nous avons effectué deux évaluations différentes: (1) une évaluation automatique sur un corpus d’évaluation standard fourni par le challenge BioASQ, et (2) une évaluation manuelle en terme de qualité des réponses en utilisant un ensemble de questions et leurs réponses à partir du corpus d’apprentissage fourni par le challenge BioASQ pour le comparer aux principaux systèmes existants tels que AskHermes \citep{Cao_2011}, EAGLi \citep{gobeill2009question} et Olelo \citep{Kraus_2017}. Dans l’évaluation automatique, nous avons utilisé l'accuracy comme indicateur d'efficacité pour les réponses exactes aux questions booléennes; la moyenne des réciproques du rang (MRR) a été utilisée pour les réponses exactes aux questions factuelles. De plus, la précision, le rappel et F1-measure moyenne ont été utilisés pour les réponses exactes aux questions de type listes; et ROUGE-2 et ROUGE-SU4 ont été utilisés pour des réponses idéales.

Pour l’évaluation manuelle, nous avons sélectionné au hasard une séquence de 30 questions de l’ensemble des questions d’apprentissage fourni par BioASQ et nous les avons soumis aux quatre systèmes à évaluer: AskHermes \citep{Cao_2011}, EAGLi \citep{gobeill2009question}, Olelo \citep{Kraus_2017} et SemBioNLQA. Cette séquence correspond à 10 questions factuelles, 11 questions de type liste, et 9 questions booléennes. Comme cette évaluation a été effectuée manuellement, nous avons dû limiter le nombre des questions. Pour cela, nous nous sommes limités aux questions factuelles, questions de type liste et les questions booléennes étant donné que les réponses exactes à ces types
de questions sont plus faciles à vérifier manuellement que les questions résumés. Lors cette évaluation, une réponse est considérée correcte si la première entité nommée biomédicale retournée est correcte pour les questions factuelles, au moins l’une des cinq premières entités nommées biomédicales retournées est correcte pour les questions de type liste, ou la valeur booléenne, soit «oui» ou «non», pour les questions booléennes est correcte. En effet, nous avons vérifié manuellement les résultats retournés par chaque système en les comparant aux réponses attendues fournies par le challenge BioASQ. Toutes les réponses retournées par les systèmes sont disponibles pour téléchargement\footnote{\url{https://sites.google.com/site/mouradsarrouti/datasets}}. Le tableau~\ref{tab:ex3} présente les résultats de notre système SemBioNLQA comparé aux trois systèmes de l'état de l'art.

\begin{table}[h!]
\centering
\selectlanguage{french}
\caption{Comparaison des résultats obtenus par notre système SemBioNLQA, EAGLi, AskHERMES et Olelo en terme du nombre de questions reconnues et de réponses correctes.}
\label{tab:ex3}
\begin{tabular}{M{5.5cm}M{4.9cm}M{4.7cm}}
\hline\noalign{\smallskip}
Systèmes& Nombre de questions reconnues&Nombre de réponses correctes\\
\noalign{\smallskip}\hline\noalign{\smallskip}
EAGLi \citep{gobeill2009question}&7/30&3/30\\
AskHERMES \citep{Cao_2011}&12/30&2/30\\
Olelo \citep{Kraus_2017}&30/30&6/30\\
SemBioNLQA&30/30&18/30\\
\noalign{\smallskip}\hline
\end{tabular}
\end{table}

Les résultats obtenus montrent que SemBioNLQA assure des meilleurs résultats par rapport aux systèmes évalués et permet de répondre correctement à la majorité des questions sélectionnées aléatoirement à partir du corpus BioASQ. Contrairement au système Olelo qui a retourné des résumés comme réponses à la plupart des questions, et AskHERMES qui a renvoyé des passages comme réponses à toutes les questions, notre système SemBioNLQA a réussi à fournir les réponses exactes en fonction de la réponse attendue pour chaque type de question. En effet, SemBioNLQA a pu identifier les types de questions données et généré des réponses exactes en fonction de la réponse attendue pour chaque type de question. Cela indique que l’intégration de notre méthode de classification des questions permet à SemBioNLQA de mieux analyser et de reconnaître correctement les questions posées par les utilisateurs. En revanche, même si Olelo a été développé pour traiter les questions factuelles, les questions de type liste et les questions de type résumé, il n’était pas capable d'identifier les types des questions, et donc il a généré des résumés pour tous les types de questions. Cela oblige les utilisateurs à lire ces résumés pour trouver les réponses précises.

Par ailleurs, les résultats obtenus ont montré que les différents composants de notre système SemBioNLQA améliorent significativement la tâche d’extraction des réponses et donc la performance globale du système SemBioNLQA. En effet, si l’ensemble des documents récupérés, les passages et le type d’une question donnée ne sont pas identifiés correctement, les étapes de traitement ultérieures pour extraire les réponses échoueront inévitablement. Par exemple, pour la question «What is the association of spermidine with $\alpha$-synuclein neurotoxicity?» du lot 1 du test du BioASQ 2016, le type de question retourné par le système est «résumé» alors que dans le corpus le type de cette question est «factuelle». Par conséquent, le système SemBioNLQA échouera inévitablement à extraire la bonne réponse puisque l’extraction de la réponse à une question factuelle qui nécessite une entité nommée biomédicale comme réponse, ne se base pas sur les mêmes techniques d'extraction des réponses aux questions de type résumé.

Selon les résultats obtenus, SemBioNLQA présente un certain nombre d’avantages par rapport aux systèmes existants tels que AskHERMES, EAGli et Olelo. Cela se manifeste en trois aspects : (1) l’intégration de notre méthode de classification des types de questions offre un net avantage sur Olelo en ce qui concerne les réponses exactes qui dépendent de chaque type de question, (2) SemBioNLQA offre un avantage par rapport aux systèmes AskHERMES, EAGli et Olelo qui réside dans le fait de traiter plusieurs types de questions: les questions booléennes, les questions factuelles, les questions de type liste et les questions de type résumé, et (3) la performance de notre système SemBioNLQA par rapport aux autres systèmes est démontré par les résultats obtenus à la fois par l’évaluation automatique et l’évaluation manuelle. Les détails sur les expérimentations menées ainsi que les résultats obtenus sont présentés en section~\ref{Chapter6.3} du chapitre~\ref{Chapter6}.

\section{Conclusion et Perspectives}
Dans ce travail de thèse, nous nous sommes penchés sur l’étude et l’amélioration des méthodes permettant la recherche automatique des réponses précises et idéales à des questions biomédicales dans une base de documents biomédicaux (en langue anglaise). Les améliorations que nous avons apportées concernent le traitement de plusieurs types de questions, la génération des réponses appropriées et la performance de SQR développé. En effet, nous avons proposé un ensemble de méthodes pour les systèmes de questions-réponses en domaine biomédical pour permettre aux utilisateurs (chercheurs et professionnels de la santé, etc.) la recherche des réponses précises à leurs questions formulées en langage naturel.

Notre première contribution consiste en la proposition de deux méthodes de classification des questions permettant de déterminer le type de la question et le type sémantique de la réponse attendue à une question biomédicale. La première est basée à la fois sur les patrons lexico-syntaxiques et l’apprentissage automatique pour déterminer la catégorie de la question posée par l'utilisateur. Celle-ci permet de classifier automatiquement les questions biomédicales dans l’une des quatre types de questions définies par le challenge BioASQ : (1) les questions booléennes, (2) les questions factuelles, (3) les questions de type liste, et (4) les questions de type résumé. La tâche d’assigner l’une des catégories précitées à une question donnée dans les SQR est très utile dans l’extraction de réponses car elle permet à un SQR d'identifier à l’avance le format de la réponse attendue. Les résultats expérimentaux ont démontré que notre méthode assure une amélioration significative en terme de performance par rapport à d'autres systèmes existants. La deuxième méthode qui constitue une variante de la première consiste à déterminer le type sémantique de la réponse attendue (i.e., un ou plusieurs sujets) pour réduire le nombre des documents examinés lors de la recherche des réponses candidates. Pour cela, nous nous somme basés sur d’autres caractéristiques de type lexical, morphosyntaxique et sémantique. L'ensemble d’expérimentations que nous avons menées a montré que notre méthode est plus efficace que les méthodes de l’état de l’art.

La deuxième contribution consiste à suggérer une méthode de recherche des documents pertinents susceptibles de contenir les réponses candidates à partir de la base de données MEDLINE. Notre méthode est basée sur le moteur de recherche GoPubMed spécialisé dans le domaine biomédical pour sélectionner la liste des documents pertinents, et la similarité sémantique entre la question et les titres des documents retournés. Le calcul de cette similarité sémantique est utilisé pour reclasser les documents obtenus en fonction de leur pertinence à la question. Ce processus est justifié par le fait que le moteur de recherche renvoie souvent les documents dans un ordre qui ne correspond pas à leur degré de pertinence vis-à-vis de la question alors que l'extraction des réponses candidates se fait à partir des documents les mieux classés. Les résultats expérimentaux ont montré que la méthode proposée pour la sélection des documents est plus efficace que les méthodes qui ont été classées parmi les 10 meilleures lors du challenge BioASQ. Nous avons également proposé une alternative permettant la recherche des passages (i.e. extraits de documents) pertinents aux questions biomédicales. Celle-ci est fondée sur le moteur de recherche PubMed, la similarité sémantique UMLS, les racines des mots, les concepts UMLS et le modèle BM25. Les résultats expérimentaux ont montré que la méthode proposée atteint de bonnes performances par rapport aux méthodes de l’état de l’art les plus récentes.

Notre troisième contribution propose des méthodes d’extraction des réponses finales permettant d’extraire à la fois les réponses exactes et idéales aux questions biomédicales. Dans ce travail de thèse, notre objectif final est de construire un système complet de questions-réponses en mesure de traiter quatre types de questions qui sont les questions booléennes, les questions factuelles, les questions de type liste et les questions de type résumé. Par conséquent, nous avons développé des nouvelles méthodes d’extraction des réponses pour chaque type de questions traitées. Celles-ci sont basées sur des approches statistiques et sémantiques. Les résultats obtenus ont montré l’efficacité des méthodes proposées. De plus, nous avons participé au challenge BioASQ 2017 avec un sous-système d'extraction des réponses qui a été classé parmi les premiers vainqueurs.

Notre dernière contribution dans le cadre de ce travail de thèse consiste à intégrer l'ensemble des méthodes proposées au sein d’un système global de questions-réponses, appelé SemBioNLQA. Celui-ci accepte en entrée une variété de questions et retourne des réponses appropriées exactes et idéales. Il fournit des réponses exactes «oui» ou «non» pour les questions booléennes, les entités nommées biomédicales pour les questions factuelles et une liste d’entités nommées biomédicales pour les questions de type liste. De plus, il retourne des réponses idéales pour ces types de questions précédentes et les questions de type résumé. Les résultats des expérimentations que nous avons effectuées sur une collection standard de questions biomédicales fournie par le challenge BioASQ ont montré que le système proposé est plus efficace que les systèmes existants.

Les contributions apportées dans ce travail de thèse permettent d'améliorer la recherche automatique des réponses à des questions exprimées en langue naturelle. De plus, elles ouvrent plusieurs directions pour des futurs travaux de recherche. Ces perspectives concernent différents volets:

\begin{itemize}
\item Pour améliorer davantage la performance du processus de classification des questions, des études lexicals, morpho-syntaxiques et sémantiques  plus approfondies sur les questions d’apprentissage utilisées au cours ce travail de thèse sont nécessaires.
\item D'autres améliorations de nos méthodes de classification des questions sont également possibles par l'usage des modèles d’apprentissage en profondeur \emph{deep learning} qui ont émergé dans la littérature en domaine de classification des textes.

\item Nous envisageons d'appliquer des modèles d’apprentissage en profondeur pour estimer la probabilité de pertinence du document par rapport à la question. Nous envisageons également d’explorer des modèles d’apprentissage en profondeur pour estimer la probabilité de la pertinence du passage à la question. Une vue d’ensemble des modèles apprentissage en profondeur peut être trouvée dans \citep{zhang2016neural,mitra2017neural}.

\item La participation au chellenge BioASQ 2017 nous a également motivé pour l'utilisation des modèles d’apprentissage profond dans la phase d’extraction de la réponse. Cependant, ces modèles sont généralement appliqués uniquement pour les SQR traitant des questions factuelles. De telles approches nécessitent un très grand nombre de paires questions-réponses pour la phase d’apprentissage. L’un des systèmes gagnants du challenge BioASQ 2017 développé par \cite{wiese2017neural} était un système d’apprentissage purement profond qui est restreint au traitement des questions factuelles et les questions de type liste. L'incorporation des modèles d’apprentissage profonds dans notre SQR permettrait d’améliorer son efficacité pour répondre aux questions factuelles et aux questions de type liste.

\item Nous envisageons étendre notre SQR pour les données structurées (par exemple, des bases de données, des ontologies, des triplets RDF, etc.) afin d'améliorer sa performance en terme de précision.

\item Au niveau de l'étape de l'extraction des passages, nous projetons de considérer le texte intégral des documents (par exemple, à partir de PubMed Central) au lieu de leurs résumés pour améliorer les performances des méthodes d’extraction de réponses. Nous pourrions exploiter la base de données MEDLINE qui contient plus de 24 millions de références complètes et PubMed Central qui archive 4.4 millions d’articles.
\end{itemize}


\chapter*{Publications of the Author} 
\label{AppendixB} 
\addcontentsline{toc}{chapter}{Publications of the Author}
\markboth{Publications of the Author}{Publications of the Author}

The research works carried out for the completion of this thesis work have resulted in several publications in national conferences, international conferences, book chapter, and specialized journals. We present these publications next, sorting them by publication type.

\section*{International journals (\#5)}

\textbf{M. Sarrouti}, S. O. E. Alaoui, A machine learning-based method for question type classification in biomedical question answering, Methods of Information in Medicine 56 (3) (2017) 209-216, \url{https://doi.org/10.3414/ME16-01-0116}.

\textbf{M. Sarrouti}, A. Lachkar, A new and efficient method based on syntactic dependency relations features for ad hoc clinical question classification, International Journal of Bioinformatics Research and Applications 13 (2) (2017) 161–177, \url{https://doi.org/10.1504/IJBRA.2017.10003490}.

\textbf{M. Sarrouti}, S. O. E. Alaoui, A passage retrieval method based on probabilistic information retrieval and UMLS concepts in biomedical question answering, Journal of Biomedical Informatics 68 (2017) 96–103, \url{https://doi.org/10.1016/j.jbi.2017.03.001}.

\textbf{M. Sarrouti}, S. O. E. Alaoui, A yes/no answer generator based on sentiment-word scores in biomedical question answering, International Journal of Healthcare Information Systems and Informatics 12 (3) (2017) 62–74, \url{https://doi.org/10.4018/ijhisi.2017070104}.

\textbf{M. Sarrouti}, S. O. E. Alaoui, SemBioNLQA: A semantic biomedical question answering system for retrieving exact and ideal answers to natural language questions, Artificial Intelligence in Medicine (\textbf{submitted}).

\section*{Book chapter (\#1)}

\textbf{M. Sarrouti}, S. O. E. Alaoui, A generic document retrieval framework based on UMLS similarity for biomedical question answering system, In: Czarnowski I., Caballero A., Howlett R., Jain L. (eds) Intelligent Decision Technologies 2016, Smart Innovation, Systems and Technologies 57 (2016) 207–216, \url{https://doi.org/10.1007/978-3-319-39627-9_18}. Springer, Cham.

\section*{International conferences and workshops (\#3)}

\textbf{M. Sarrouti}, S. O. E. Alaoui, A biomedical question answering system in BioASQ 2017, in: BioNLP 2017, Association for Computational Linguistics, 2017, \url{https://doi.org10.18653/v1/w17-2337}, July 30-August 4, 2017 Vancouver, Canada.

\textbf{M. Sarrouti,} S. O. E. Alaoui, A generic document retrieval framework based on UMLS similarity for biomedical question answering system, In: The 8th KES International Conference on Intelligent Decision Technologies (KES-IDT 2016), 2016, Volume 57, pp. 207–216; doi: \url{https://doi.org/10.1007/978-3-319-39627-9\_18}, June 15-17, 2016, Tenerife, Spain.

\textbf{M. Sarrouti}, A. Lachkar, S. O. E. Alaoui, Biomedical question types classification using syntactic and rule based approach, in: Proceedings of the 7th International Joint Conference on Knowledge Discovery, Knowledge Engineering and Knowledge Management, 2015, Volume 1: KDIR, pp. 265–272. ISBN 978-989-758-158-8; doi: \url{http://dx.doi.org/10.5220/0005598002650272}, November 12-14, 2015, Lisbon, Portugal.

\section*{National conferences (\#2)}

\textbf{M. Sarrouti}, S. O. E. Alaoui, Recherche de réponses précises à des questions biomédicales : système de questions-réponses, In: Signal, ImaGe, Multimédia et Applications (SIGMA), December 21-22, 2016, Fez, Morocco.

\textbf{M. Sarrouti}, A. A. Hassani, I. E. Bellili, A. Lachkar, S. O. E. Alaoui, Machine learning methods evaluation for biomedical named entities recognition, In: Les Journes Doctorales en Technologies de l'Information et de la Communication (JDTIC’14), June 19-20, 2014, Rabat, Morocco.


\chapter*{Bibliography}
\markboth{Bibliography}{Bibliography}

\addcontentsline{toc}{chapter}{Bibliography}
\bibliographystyle{model4-names}
\small
\bibliography{refbib}

\begin{thebibliography}{188}
\expandafter\ifx\csname natexlab\endcsname\relax\def\natexlab#1{#1}\fi
\providecommand{\url}[1]{\texttt{#1}}
\providecommand{\href}[2]{#2}
\providecommand{\path}[1]{#1}
\providecommand{\DOIprefix}{doi:}
\providecommand{\ArXivprefix}{arXiv:}
\providecommand{\URLprefix}{URL: }
\providecommand{\Pubmedprefix}{pmid:}
\providecommand{\doi}[1]{\href{http://dx.doi.org/#1}{\path{#1}}}
\providecommand{\Pubmed}[1]{\href{pmid:#1}{\path{#1}}}
\providecommand{\bibinfo}[2]{#2}
\ifx\xfnm\undefined \def\xfnm[#1]{\unskip,\space#1}\fi
\bibitem[{Abacha and Zweigenbaum(2015)}]{abacha2015means}
\bibinfo{author}{Abacha\xfnm[ A.B.]}, \bibinfo{author}{Zweigenbaum\xfnm[ P.]}.
\newblock \bibinfo{title}{{MEANS}: A medical question-answering system
  combining {NLP} techniques and semantic web technologies}.
\newblock \bibinfo{journal}{Information Processing {\&} Management}
  \bibinfo{year}{2015};\bibinfo{volume}{51}(\bibinfo{number}{5}):\bibinfo{pages}{570--594}.
\newblock \DOIprefix\doi{10.1016/j.ipm.2015.04.006}.
\bibitem[{Aronson(2001)}]{aronson2001effective}
\bibinfo{author}{Aronson\xfnm[ A.R.]}.
\newblock \bibinfo{title}{Effective mapping of biomedical text to the {UMLS}
  {Metathesaurus}: the {MetaMap} program.}
\newblock In: \bibinfo{booktitle}{Proceedings of the AMIA Symposium}.
  \bibinfo{organization}{American Medical Informatics Association};
  \bibinfo{year}{2001}. p.~\bibinfo{pages}{17}.
\bibitem[{Aronson and Lang(2010)}]{Aronson_2010}
\bibinfo{author}{Aronson\xfnm[ A.R.]}, \bibinfo{author}{Lang\xfnm[ F.M.]}.
\newblock \bibinfo{title}{An overview of {MetaMap}: historical perspective and
  recent advances}.
\newblock \bibinfo{journal}{Journal of the American Medical Informatics
  Association}
  \bibinfo{year}{2010};\bibinfo{volume}{17}(\bibinfo{number}{3}):\bibinfo{pages}{229--236}.
\newblock \URLprefix \url{https://doi.org/10.1136%2Fjamia.2009.002733}.
  \DOIprefix\doi{10.1136/jamia.2009.002733}.
\bibitem[{Ashburner et~al.(2000)Ashburner, Ball, Blake, Botstein, Butler,
  Cherry, Davis, Dolinski, Dwight, Eppig et~al.}]{ashburner2000gene}
\bibinfo{author}{Ashburner\xfnm[ M.]}, \bibinfo{author}{Ball\xfnm[ C.A.]},
  \bibinfo{author}{Blake\xfnm[ J.A.]}, \bibinfo{author}{Botstein\xfnm[ D.]},
  \bibinfo{author}{Butler\xfnm[ H.]}, \bibinfo{author}{Cherry\xfnm[ J.M.]},
  \bibinfo{author}{Davis\xfnm[ A.P.]}, \bibinfo{author}{Dolinski\xfnm[ K.]},
  \bibinfo{author}{Dwight\xfnm[ S.S.]}, \bibinfo{author}{Eppig\xfnm[ J.T.]},
  et~al.
\newblock \bibinfo{title}{Gene ontology: tool for the unification of biology}.
\newblock \bibinfo{journal}{Nature genetics}
  \bibinfo{year}{2000};\bibinfo{volume}{25}(\bibinfo{number}{1}):\bibinfo{pages}{25--29}.
\bibitem[{Athenikos and Han(2010)}]{athenikos2010biomedical}
\bibinfo{author}{Athenikos\xfnm[ S.J.]}, \bibinfo{author}{Han\xfnm[ H.]}.
\newblock \bibinfo{title}{Biomedical question answering: A survey}.
\newblock \bibinfo{journal}{Computer methods and programs in biomedicine}
  \bibinfo{year}{2010};\bibinfo{volume}{99}(\bibinfo{number}{1}):\bibinfo{pages}{1--24}.
\newblock \DOIprefix\doi{10.1016/j.cmpb.2009.10.003}.
\bibitem[{Baccianella et~al.(2010)Baccianella, Esuli and
  Sebastiani}]{baccianella2010sentiwordnet}
\bibinfo{author}{Baccianella\xfnm[ S.]}, \bibinfo{author}{Esuli\xfnm[ A.]},
  \bibinfo{author}{Sebastiani\xfnm[ F.]}.
\newblock \bibinfo{title}{Sentiwordnet 3.0: An enhanced lexical resource for
  sentiment analysis and opinion mining.}
\newblock In: \bibinfo{booktitle}{LREC}. volume~\bibinfo{volume}{10};
  \bibinfo{year}{2010}. p. \bibinfo{pages}{2200--2204}.
\bibitem[{Balikas et~al.(2015)Balikas, Kosmopoulos, Krithara, Paliouras and
  Kakadiaris}]{balikas2015results}
\bibinfo{author}{Balikas\xfnm[ G.]}, \bibinfo{author}{Kosmopoulos\xfnm[ A.]},
  \bibinfo{author}{Krithara\xfnm[ A.]}, \bibinfo{author}{Paliouras\xfnm[ G.]},
  \bibinfo{author}{Kakadiaris\xfnm[ I.]}.
\newblock \bibinfo{title}{Results of the {{BioASQ}} tasks of the question
  answering lab at {CLEF} 2015}.
\newblock In: \bibinfo{booktitle}{{CLEF} 2015}. \bibinfo{year}{2015}. .
\bibitem[{Balikas et~al.(2013)Balikas, Partalas, Kosmopoulos, Petridis,
  Malakasiotis, Pavlopoulos, Androutsopoulos, Baskiotis, Gaussier, Artieres and
  Gallinari}]{Balikas_2013}
\bibinfo{author}{Balikas\xfnm[ G.]}, \bibinfo{author}{Partalas\xfnm[ I.]},
  \bibinfo{author}{Kosmopoulos\xfnm[ A.]}, \bibinfo{author}{Petridis\xfnm[
  S.]}, \bibinfo{author}{Malakasiotis\xfnm[ P.]},
  \bibinfo{author}{Pavlopoulos\xfnm[ I.]},
  \bibinfo{author}{Androutsopoulos\xfnm[ I.]}, \bibinfo{author}{Baskiotis\xfnm[
  N.]}, \bibinfo{author}{Gaussier\xfnm[ E.]}, \bibinfo{author}{Artieres\xfnm[
  T.]}, \bibinfo{author}{Gallinari\xfnm[ P.]}.
\newblock \bibinfo{title}{Evaluation Framework Specifications}.
\newblock \bibinfo{type}{Project deliverable} \bibinfo{number}{D4.1};
  \bibinfo{year}{2013}.
\newblock \URLprefix
  \url{sites/default/files/PublicDocuments/BioASQ_D4.1-EvaluationFrameworkSpecification_final.pdf}.
\bibitem[{Balikas et~al.(2014)Balikas, Partalas, Ngomo, Krithara, Gaussier and
  Paliouras}]{balikas2014results}
\bibinfo{author}{Balikas\xfnm[ G.]}, \bibinfo{author}{Partalas\xfnm[ I.]},
  \bibinfo{author}{Ngomo\xfnm[ A.C.N.]}, \bibinfo{author}{Krithara\xfnm[ A.]},
  \bibinfo{author}{Gaussier\xfnm[ E.]}, \bibinfo{author}{Paliouras\xfnm[ G.]}.
\newblock \bibinfo{title}{Results of the {BioASQ} track of the question
  answering lab at {CLEF} 2014}.
\newblock \bibinfo{journal}{Proceedings of {Question} {Answering} {Lab} at
  {CLEF}}
  \bibinfo{year}{2014};\bibinfo{volume}{2014}:\bibinfo{pages}{1181--1193}.
\bibitem[{Bauer and Berleant(2012)}]{Bauer_2012}
\bibinfo{author}{Bauer\xfnm[ M.A.]}, \bibinfo{author}{Berleant\xfnm[ D.]}.
\newblock \bibinfo{title}{Usability survey of biomedical question answering
  systems}.
\newblock \bibinfo{journal}{Human Genomics}
  \bibinfo{year}{2012};\bibinfo{volume}{6}(\bibinfo{number}{1}):\bibinfo{pages}{17}.
\newblock \URLprefix \url{https://doi.org/10.1186%2F1479-7364-6-17}.
  \DOIprefix\doi{10.1186/1479-7364-6-17}.
\bibitem[{Bergus(2000)}]{Bergus_2000}
\bibinfo{author}{Bergus\xfnm[ G.R.]}.
\newblock \bibinfo{title}{Does the structure of clinical questions affect the
  outcome of curbside consultations with specialty colleagues?}
\newblock \bibinfo{journal}{Archives of Family Medicine}
  \bibinfo{year}{2000};\bibinfo{volume}{9}(\bibinfo{number}{6}):\bibinfo{pages}{541--547}.
\newblock \URLprefix \url{https://doi.org/10.1001%2Farchfami.9.6.541}.
  \DOIprefix\doi{10.1001/archfami.9.6.541}.
\bibitem[{Biswas et~al.(2014)Biswas, Sharan and Kumar}]{Biswas_2014}
\bibinfo{author}{Biswas\xfnm[ P.]}, \bibinfo{author}{Sharan\xfnm[ A.]},
  \bibinfo{author}{Kumar\xfnm[ R.]}.
\newblock \bibinfo{title}{Question classification using syntactic and rule
  based approach}.
\newblock In: \bibinfo{booktitle}{2014 International Conference on Advances in
  Computing, Communications and Informatics ({ICACCI})}.
  \bibinfo{publisher}{{IEEE}}; \bibinfo{year}{2014}. \URLprefix
  \url{https://doi.org/10.1109%2Ficacci.2014.6968434}.
  \DOIprefix\doi{10.1109/icacci.2014.6968434}.
\bibitem[{Bodenreider(2004)}]{Bodenreider_2004}
\bibinfo{author}{Bodenreider\xfnm[ O.]}.
\newblock \bibinfo{title}{The unified medical language system ({UMLS}):
  integrating biomedical terminology}.
\newblock \bibinfo{journal}{Nucleic Acids Research}
  \bibinfo{year}{2004};\bibinfo{volume}{32}(\bibinfo{number}{90001}):\bibinfo{pages}{267D--270}.
\newblock \URLprefix \url{https://doi.org/10.1093%2Fnar%2Fgkh061}.
  \DOIprefix\doi{10.1093/nar/gkh061}.
\bibitem[{Boyack et~al.(2011)Boyack, Newman, Duhon, Klavans, Patek, Biberstine,
  Schijvenaars, Skupin, Ma and Börner}]{Boyack_2011}
\bibinfo{author}{Boyack\xfnm[ K.W.]}, \bibinfo{author}{Newman\xfnm[ D.]},
  \bibinfo{author}{Duhon\xfnm[ R.J.]}, \bibinfo{author}{Klavans\xfnm[ R.]},
  \bibinfo{author}{Patek\xfnm[ M.]}, \bibinfo{author}{Biberstine\xfnm[ J.R.]},
  \bibinfo{author}{Schijvenaars\xfnm[ B.]}, \bibinfo{author}{Skupin\xfnm[ A.]},
  \bibinfo{author}{Ma\xfnm[ N.]}, \bibinfo{author}{Börner\xfnm[ K.]}.
\newblock \bibinfo{title}{Clustering more than two million biomedical
  publications: Comparing the accuracies of nine text-based similarity
  approaches}.
\newblock \bibinfo{journal}{{PLoS} {ONE}}
  \bibinfo{year}{2011};\bibinfo{volume}{6}(\bibinfo{number}{3}):\bibinfo{pages}{e18029}.
\newblock \URLprefix \url{https://doi.org/10.1371%2Fjournal.pone.0018029}.
  \DOIprefix\doi{10.1371/journal.pone.0018029}.
\bibitem[{de~Bruijn and Martin(2002)}]{de_Bruijn_2002}
\bibinfo{author}{de~Bruijn\xfnm[ B.]}, \bibinfo{author}{Martin\xfnm[ J.]}.
\newblock \bibinfo{title}{Getting to the (c)ore of knowledge: mining biomedical
  literature}.
\newblock \bibinfo{journal}{International Journal of Medical Informatics}
  \bibinfo{year}{2002};\bibinfo{volume}{67}(\bibinfo{number}{1-3}):\bibinfo{pages}{7--18}.
\newblock \URLprefix \url{https://doi.org/10.1016%2Fs1386-5056%2802%2900050-3}.
  \DOIprefix\doi{10.1016/s1386-5056(02)00050-3}.
\bibitem[{Buscaldi et~al.(2009)Buscaldi, Rosso, G{\'{o}}mez-Soriano and
  Sanchis}]{Buscaldi_2009}
\bibinfo{author}{Buscaldi\xfnm[ D.]}, \bibinfo{author}{Rosso\xfnm[ P.]},
  \bibinfo{author}{G{\'{o}}mez-Soriano\xfnm[ J.M.]},
  \bibinfo{author}{Sanchis\xfnm[ E.]}.
\newblock \bibinfo{title}{Answering questions with an n-gram based passage
  retrieval engine}.
\newblock \bibinfo{journal}{Journal of Intelligent Information Systems}
  \bibinfo{year}{2009};\bibinfo{volume}{34}(\bibinfo{number}{2}):\bibinfo{pages}{113--134}.
\newblock \URLprefix \url{https://doi.org/10.1007%2Fs10844-009-0082-y}.
  \DOIprefix\doi{10.1007/s10844-009-0082-y}.
\bibitem[{Cannataro and Guzzi(2012)}]{jhon_2012}
\bibinfo{author}{Cannataro\xfnm[ M.]}, \bibinfo{author}{Guzzi\xfnm[ P.H.]}.
\newblock \bibinfo{title}{Protein-to-protein interaction {DATABASES}}.
\newblock \bibinfo{journal}{Data Management of Protein Interaction Networks}
  \bibinfo{year}{2012};:\bibinfo{pages}{43--70}\URLprefix
  \url{https://doi.org/10.1002%2F9781118103746.ch5}.
  \DOIprefix\doi{10.1002/9781118103746.ch5}.
\bibitem[{Cao et~al.(2011)Cao, Liu, Simpson, Antieau, Bennett, Cimino, Ely and
  Yu}]{Cao_2011}
\bibinfo{author}{Cao\xfnm[ Y.]}, \bibinfo{author}{Liu\xfnm[ F.]},
  \bibinfo{author}{Simpson\xfnm[ P.]}, \bibinfo{author}{Antieau\xfnm[ L.]},
  \bibinfo{author}{Bennett\xfnm[ A.]}, \bibinfo{author}{Cimino\xfnm[ J.J.]},
  \bibinfo{author}{Ely\xfnm[ J.]}, \bibinfo{author}{Yu\xfnm[ H.]}.
\newblock \bibinfo{title}{{AskHERMES}: An online question answering system for
  complex clinical questions}.
\newblock \bibinfo{journal}{Journal of Biomedical Informatics}
  \bibinfo{year}{2011};\bibinfo{volume}{44}(\bibinfo{number}{2}):\bibinfo{pages}{277--288}.
\newblock \URLprefix \url{http://dx.doi.org/10.1016/j.jbi.2011.01.004}.
  \DOIprefix\doi{10.1016/j.jbi.2011.01.004}.
\bibitem[{Cao et~al.(2010)Cao, Cimino, Ely and Yu}]{Cao_2010}
\bibinfo{author}{Cao\xfnm[ Y.g.]}, \bibinfo{author}{Cimino\xfnm[ J.J.]},
  \bibinfo{author}{Ely\xfnm[ J.]}, \bibinfo{author}{Yu\xfnm[ H.]}.
\newblock \bibinfo{title}{Automatically extracting information needs from
  complex clinical questions}.
\newblock \bibinfo{journal}{Journal of Biomedical Informatics}
  \bibinfo{year}{2010};\bibinfo{volume}{43}(\bibinfo{number}{6}):\bibinfo{pages}{962--971}.
\newblock \URLprefix \url{https://doi.org/10.1016%2Fj.jbi.2010.07.007}.
  \DOIprefix\doi{10.1016/j.jbi.2010.07.007}.
\bibitem[{Chen et~al.(2011)Chen, Lin and Yang}]{Chen_2011}
\bibinfo{author}{Chen\xfnm[ R.]}, \bibinfo{author}{Lin\xfnm[ H.]},
  \bibinfo{author}{Yang\xfnm[ Z.]}.
\newblock \bibinfo{title}{Passage retrieval based hidden knowledge discovery
  from biomedical literature}.
\newblock \bibinfo{journal}{Expert Systems with Applications}
  \bibinfo{year}{2011};\bibinfo{volume}{38}(\bibinfo{number}{8}):\bibinfo{pages}{9958--9964}.
\newblock \URLprefix \url{https://doi.org/10.1016%2Fj.eswa.2011.02.034}.
  \DOIprefix\doi{10.1016/j.eswa.2011.02.034}.
\bibitem[{Choi(2015)}]{choi2015snumedinfo}
\bibinfo{author}{Choi\xfnm[ S.]}.
\newblock \bibinfo{title}{Snumedinfo at {CLEF QA} track {BioASQ} 2015}.
\newblock In: \bibinfo{booktitle}{Working Notes for the Conference and Labs of
  the Evaluation Forum (CLEF), Toulouse, France}. \bibinfo{year}{2015}. .
\bibitem[{Choi and Choi(2014)}]{choi2014classification}
\bibinfo{author}{Choi\xfnm[ S.]}, \bibinfo{author}{Choi\xfnm[ J.]}.
\newblock \bibinfo{title}{Classification and retrieval of biomedical
  literatures: {SNUMedinfo} at {CLEF} {QA} {Track} {BioASQ} 2014}.
\newblock \bibinfo{journal}{Proceedings of {Question} {Answering} {Lab} at
  {CLEF}} \bibinfo{year}{2014};.
\bibitem[{Chowdhury(2010)}]{chowdhury2010introduction}
\bibinfo{author}{Chowdhury\xfnm[ G.G.]}.
\newblock \bibinfo{title}{Introduction to modern information retrieval}.
\newblock \bibinfo{publisher}{Facet publishing}, \bibinfo{year}{2010}.
\bibitem[{Clarke and Terra(2003)}]{Clarke_2003}
\bibinfo{author}{Clarke\xfnm[ C.L.A.]}, \bibinfo{author}{Terra\xfnm[ E.L.]}.
\newblock \bibinfo{title}{Passage retrieval vs. document retrieval for factoid
  question answering}.
\newblock In: \bibinfo{booktitle}{Proceedings of the 26th annual international
  {ACM} {SIGIR} conference on Research and development in informaion retrieval
  - {SIGIR}}. \bibinfo{publisher}{{ACM} Press}; \bibinfo{year}{2003}.
  \URLprefix \url{https://doi.org/10.1145%2F860435.860534}.
  \DOIprefix\doi{10.1145/860435.860534}.
\bibitem[{Cohen(2005)}]{Cohen_2005}
\bibinfo{author}{Cohen\xfnm[ A.M.]}.
\newblock \bibinfo{title}{A survey of current work in biomedical text mining}.
\newblock \bibinfo{journal}{Briefings in Bioinformatics}
  \bibinfo{year}{2005};\bibinfo{volume}{6}(\bibinfo{number}{1}):\bibinfo{pages}{57--71}.
\newblock \URLprefix \url{https://doi.org/10.1093%2Fbib%2F6.1.57}.
  \DOIprefix\doi{10.1093/bib/6.1.57}.
\bibitem[{Collins-Thompson et~al.(2004)Collins-Thompson, Callan, Terra and
  Clarke}]{Collins_Thompson_2004}
\bibinfo{author}{Collins-Thompson\xfnm[ K.]}, \bibinfo{author}{Callan\xfnm[
  J.]}, \bibinfo{author}{Terra\xfnm[ E.]}, \bibinfo{author}{Clarke\xfnm[
  C.L.]}.
\newblock \bibinfo{title}{The effect of document retrieval quality on factoid
  question answering performance}.
\newblock In: \bibinfo{booktitle}{Proceedings of the 27th annual international
  conference on Research and development in information retrieval - {SIGIR}
  04}. \bibinfo{publisher}{{ACM} Press}; \bibinfo{year}{2004}. \URLprefix
  \url{https://doi.org/10.1145%2F1008992.1009127}.
  \DOIprefix\doi{10.1145/1008992.1009127}.
\bibitem[{Cordell(2009)}]{Cordell_2009}
\bibinfo{author}{Cordell\xfnm[ H.J.]}.
\newblock \bibinfo{title}{Detecting gene{\textendash}gene interactions that
  underlie human diseases}.
\newblock \bibinfo{journal}{Nature Reviews Genetics}
  \bibinfo{year}{2009};\bibinfo{volume}{10}(\bibinfo{number}{6}):\bibinfo{pages}{392--404}.
\newblock \URLprefix \url{https://doi.org/10.1038%2Fnrg2579}.
  \DOIprefix\doi{10.1038/nrg2579}.
\bibitem[{Cruchet et~al.(2008)Cruchet, Gaudinat and
  Boyer}]{cruchet2008supervised}
\bibinfo{author}{Cruchet\xfnm[ S.]}, \bibinfo{author}{Gaudinat\xfnm[ A.]},
  \bibinfo{author}{Boyer\xfnm[ C.]}.
\newblock \bibinfo{title}{Supervised approach to recognize question type in a
  qa system for health}.
\newblock \bibinfo{journal}{Studies in health technology and informatics}
  \bibinfo{year}{2008};\bibinfo{volume}{136}:\bibinfo{pages}{407}.
\bibitem[{Cruchet et~al.(2009)Cruchet, Gaudinat, Rindflesch and
  Boyer}]{cruchet2009trust}
\bibinfo{author}{Cruchet\xfnm[ S.]}, \bibinfo{author}{Gaudinat\xfnm[ A.]},
  \bibinfo{author}{Rindflesch\xfnm[ T.]}, \bibinfo{author}{Boyer\xfnm[ C.]}.
\newblock \bibinfo{title}{What about trust in the question answering world}.
\newblock In: \bibinfo{booktitle}{AMIA 2009 annual symposium}.
  \bibinfo{address}{San Francisco, USA}; \bibinfo{year}{2009}. p.
  \bibinfo{pages}{1--5}.
\bibitem[{D'Alessandro et~al.(2004)D'Alessandro, Kreiter and
  Peterson}]{D_Alessandro_2003}
\bibinfo{author}{D'Alessandro\xfnm[ D.M.]}, \bibinfo{author}{Kreiter\xfnm[
  C.D.]}, \bibinfo{author}{Peterson\xfnm[ M.W.]}.
\newblock \bibinfo{title}{An evaluation of information-seeking behaviors of
  general pediatricians}.
\newblock \bibinfo{journal}{{PEDIATRICS}}
  \bibinfo{year}{2004};\bibinfo{volume}{113}(\bibinfo{number}{1}):\bibinfo{pages}{64--69}.
\newblock \URLprefix \url{https://doi.org/10.1542%2Fpeds.113.1.64}.
  \DOIprefix\doi{10.1542/peds.113.1.64}.
\bibitem[{De~Marneffe et~al.(2006)De~Marneffe, MacCartney, Manning
  et~al.}]{de2006generating}
\bibinfo{author}{De~Marneffe\xfnm[ M.C.]}, \bibinfo{author}{MacCartney\xfnm[
  B.]}, \bibinfo{author}{Manning\xfnm[ C.D.]}, et~al.
\newblock \bibinfo{title}{Generating typed dependency parses from phrase
  structure parses}.
\newblock In: \bibinfo{booktitle}{Proceedings of LREC}.
  \bibinfo{organization}{Genoa Italy}; volume~\bibinfo{volume}{6};
  \bibinfo{year}{2006}. p. \bibinfo{pages}{449--454}.
\bibitem[{Demner-Fushman et~al.(2006)Demner-Fushman, Few, Hauser and
  Thoma}]{Demner_Fushman_2006}
\bibinfo{author}{Demner-Fushman\xfnm[ D.]}, \bibinfo{author}{Few\xfnm[ B.]},
  \bibinfo{author}{Hauser\xfnm[ S.E.]}, \bibinfo{author}{Thoma\xfnm[ G.]}.
\newblock \bibinfo{title}{Automatically identifying health outcome information
  in {MEDLINE} records}.
\newblock \bibinfo{journal}{Journal of the American Medical Informatics
  Association}
  \bibinfo{year}{2006};\bibinfo{volume}{13}(\bibinfo{number}{1}):\bibinfo{pages}{52--60}.
\newblock \URLprefix \url{https://doi.org/10.1197%2Fjamia.m1911}.
  \DOIprefix\doi{10.1197/jamia.m1911}.
\bibitem[{Demner-Fushman and Lin(2007)}]{Demner_Fushman_2007}
\bibinfo{author}{Demner-Fushman\xfnm[ D.]}, \bibinfo{author}{Lin\xfnm[ J.]}.
\newblock \bibinfo{title}{Answering clinical questions with knowledge-based and
  statistical techniques}.
\newblock \bibinfo{journal}{Computational Linguistics}
  \bibinfo{year}{2007};\bibinfo{volume}{33}(\bibinfo{number}{1}):\bibinfo{pages}{63--103}.
\newblock \URLprefix \url{https://doi.org/10.1162%2Fcoli.2007.33.1.63}.
  \DOIprefix\doi{10.1162/coli.2007.33.1.63}.
\bibitem[{Dogan et~al.(2017)Dogan, Chatr-aryamontri, Kim, Wei, Peng, Comeau and
  Lu}]{Islamaj_Dogan_2017}
\bibinfo{author}{Dogan\xfnm[ R.I.]}, \bibinfo{author}{Chatr-aryamontri\xfnm[
  A.]}, \bibinfo{author}{Kim\xfnm[ S.]}, \bibinfo{author}{Wei\xfnm[ C.H.]},
  \bibinfo{author}{Peng\xfnm[ Y.]}, \bibinfo{author}{Comeau\xfnm[ D.]},
  \bibinfo{author}{Lu\xfnm[ Z.]}.
\newblock \bibinfo{title}{{BioCreative} {VI} precision medicine track: creating
  a training corpus for mining protein-protein interactions affected by
  mutations}.
\newblock In: \bibinfo{booktitle}{{BioNLP} 2017}.
  \bibinfo{publisher}{Association for Computational Linguistics};
  \bibinfo{year}{2017}. \URLprefix
  \url{https://doi.org/10.18653%2Fv1%2Fw17-2321}.
  \DOIprefix\doi{10.18653/v1/w17-2321}.
\bibitem[{Dram{\'{e}} et~al.(2016)Dram{\'{e}}, Mougin and Diallo}]{Dram__2016}
\bibinfo{author}{Dram{\'{e}}\xfnm[ K.]}, \bibinfo{author}{Mougin\xfnm[ F.]},
  \bibinfo{author}{Diallo\xfnm[ G.]}.
\newblock \bibinfo{title}{Large scale biomedical texts classification: a {kNN}
  and an {ESA}-based approaches}.
\newblock \bibinfo{journal}{Journal of Biomedical Semantics}
  \bibinfo{year}{2016};\bibinfo{volume}{7}(\bibinfo{number}{1}).
\newblock \URLprefix \url{https://doi.org/10.1186%2Fs13326-016-0073-1}.
  \DOIprefix\doi{10.1186/s13326-016-0073-1}.
\bibitem[{Ely et~al.(2005)Ely, Osheroff, Chambliss, Ebell and
  Rosenbaum}]{elyjhon}
\bibinfo{author}{Ely\xfnm[ J.W.]}, \bibinfo{author}{Osheroff\xfnm[ J.A.]},
  \bibinfo{author}{Chambliss\xfnm[ M.L.]}, \bibinfo{author}{Ebell\xfnm[ M.H.]},
  \bibinfo{author}{Rosenbaum\xfnm[ M.E.]}.
\newblock \bibinfo{title}{Answering physicians' clinical questions: Obstacles
  and potential solutions}.
\newblock \bibinfo{journal}{Journal of the American Medical Informatics
  Association}
  \bibinfo{year}{2005};\bibinfo{volume}{12}(\bibinfo{number}{2}):\bibinfo{pages}{217--224}.
\newblock \URLprefix \url{http://dx.doi.org/10.1197/jamia.M1608}.
  \DOIprefix\doi{10.1197/jamia.M1608}.
\bibitem[{Ely et~al.(1999)Ely, Osheroff, Ebell, Bergus, Levy, Chambliss and
  Evans}]{Ely_1999}
\bibinfo{author}{Ely\xfnm[ J.W.]}, \bibinfo{author}{Osheroff\xfnm[ J.A.]},
  \bibinfo{author}{Ebell\xfnm[ M.H.]}, \bibinfo{author}{Bergus\xfnm[ G.R.]},
  \bibinfo{author}{Levy\xfnm[ B.T.]}, \bibinfo{author}{Chambliss\xfnm[ M.L.]},
  \bibinfo{author}{Evans\xfnm[ E.R.]}.
\newblock \bibinfo{title}{Analysis of questions asked by family doctors
  regarding patient care}.
\newblock \bibinfo{journal}{{BMJ}}
  \bibinfo{year}{1999};\bibinfo{volume}{319}(\bibinfo{number}{7206}):\bibinfo{pages}{358--361}.
\newblock \URLprefix \url{https://doi.org/10.1136%2Fbmj.319.7206.358}.
  \DOIprefix\doi{10.1136/bmj.319.7206.358}.
\bibitem[{Ely et~al.(2002{\natexlab{a}})Ely, Osheroff, Ebell, Chambliss,
  Vinson, Stevermer and Pifer}]{ely2002obstacles}
\bibinfo{author}{Ely\xfnm[ J.W.]}, \bibinfo{author}{Osheroff\xfnm[ J.A.]},
  \bibinfo{author}{Ebell\xfnm[ M.H.]}, \bibinfo{author}{Chambliss\xfnm[ M.L.]},
  \bibinfo{author}{Vinson\xfnm[ D.C.]}, \bibinfo{author}{Stevermer\xfnm[
  J.J.]}, \bibinfo{author}{Pifer\xfnm[ E.A.]}.
\newblock \bibinfo{title}{Obstacles to answering doctors' questions about
  patient care with evidence: qualitative study}.
\newblock \bibinfo{journal}{BMJ}
  \bibinfo{year}{2002}{\natexlab{a}};\bibinfo{volume}{324}(\bibinfo{number}{7339}):\bibinfo{pages}{710}.
\newblock \DOIprefix\doi{10.1136/bmj.324.7339.710}.
\bibitem[{Ely et~al.(2002{\natexlab{b}})Ely, Osheroff, Ebell, Chambliss,
  Vinson, Stevermer and Pifer}]{Ely710}
\bibinfo{author}{Ely\xfnm[ J.W.]}, \bibinfo{author}{Osheroff\xfnm[ J.A.]},
  \bibinfo{author}{Ebell\xfnm[ M.H.]}, \bibinfo{author}{Chambliss\xfnm[ M.L.]},
  \bibinfo{author}{Vinson\xfnm[ D.C.]}, \bibinfo{author}{Stevermer\xfnm[
  J.J.]}, \bibinfo{author}{Pifer\xfnm[ E.A.]}.
\newblock \bibinfo{title}{Obstacles to answering doctors{\textquoteright}
  questions about patient care with evidence: qualitative study}.
\newblock \bibinfo{journal}{BMJ}
  \bibinfo{year}{2002}{\natexlab{b}};\bibinfo{volume}{324}(\bibinfo{number}{7339}):\bibinfo{pages}{710}.
\newblock \URLprefix \url{http://www.bmj.com/content/324/7339/710}.
  \DOIprefix\doi{10.1136/bmj.324.7339.710}.
  \href{http://arxiv.org/abs/http://www.bmj.com/content/324/7339/710.full.pdf}{\tt
  arXiv:http://www.bmj.com/content/324/7339/710.full.pdf}.
\bibitem[{Ely et~al.(1997)Ely, Osheroff, Ferguson, Chambliss, Vinson and
  Moore}]{ely1997lifelong}
\bibinfo{author}{Ely\xfnm[ J.W.]}, \bibinfo{author}{Osheroff\xfnm[ J.A.]},
  \bibinfo{author}{Ferguson\xfnm[ K.J.]}, \bibinfo{author}{Chambliss\xfnm[
  M.L.]}, \bibinfo{author}{Vinson\xfnm[ D.C.]}, \bibinfo{author}{Moore\xfnm[
  J.L.]}.
\newblock \bibinfo{title}{Lifelong self-directed learning using a computer
  database of clinical questions}.
\newblock \bibinfo{journal}{Journal of family practice}
  \bibinfo{year}{1997};\bibinfo{volume}{45}(\bibinfo{number}{5}):\bibinfo{pages}{382--389}.
\bibitem[{Ely et~al.(2000{\natexlab{a}})Ely, Osheroff, Gorman, Ebell,
  Chambliss, Pifer and Stavri}]{ely2000taxonomy}
\bibinfo{author}{Ely\xfnm[ J.W.]}, \bibinfo{author}{Osheroff\xfnm[ J.A.]},
  \bibinfo{author}{Gorman\xfnm[ P.N.]}, \bibinfo{author}{Ebell\xfnm[ M.H.]},
  \bibinfo{author}{Chambliss\xfnm[ M.L.]}, \bibinfo{author}{Pifer\xfnm[ E.A.]},
  \bibinfo{author}{Stavri\xfnm[ P.Z.]}.
\newblock \bibinfo{title}{A taxonomy of generic clinical questions:
  classification study}.
\newblock \bibinfo{journal}{BMJ}
  \bibinfo{year}{2000}{\natexlab{a}};\bibinfo{volume}{321}(\bibinfo{number}{7258}):\bibinfo{pages}{429--432}.
\newblock \DOIprefix\doi{10.1136/bmj.321.7258.429}.
\bibitem[{Ely et~al.(2000{\natexlab{b}})Ely, Osheroff, Gorman, Ebell,
  Chambliss, Pifer and Stavri}]{Ely429}
\bibinfo{author}{Ely\xfnm[ J.W.]}, \bibinfo{author}{Osheroff\xfnm[ J.A.]},
  \bibinfo{author}{Gorman\xfnm[ P.N.]}, \bibinfo{author}{Ebell\xfnm[ M.H.]},
  \bibinfo{author}{Chambliss\xfnm[ M.L.]}, \bibinfo{author}{Pifer\xfnm[ E.A.]},
  \bibinfo{author}{Stavri\xfnm[ P.Z.]}.
\newblock \bibinfo{title}{A taxonomy of generic clinical questions:
  classification study}.
\newblock \bibinfo{journal}{BMJ}
  \bibinfo{year}{2000}{\natexlab{b}};\bibinfo{volume}{321}(\bibinfo{number}{7258}):\bibinfo{pages}{429--432}.
\newblock \URLprefix \url{http://www.bmj.com/content/321/7258/429.1}.
  \DOIprefix\doi{10.1136/bmj.321.7258.429}.
  \href{http://arxiv.org/abs/http://www.bmj.com/content/321/7258/429.1.full.pdf}{\tt
  arXiv:http://www.bmj.com/content/321/7258/429.1.full.pdf}.
\bibitem[{Erkan and Radev(2004)}]{erkan2004lexrank}
\bibinfo{author}{Erkan\xfnm[ G.]}, \bibinfo{author}{Radev\xfnm[ D.R.]}.
\newblock \bibinfo{title}{Lexrank: Graph-based lexical centrality as salience
  in text summarization}.
\newblock \bibinfo{journal}{Journal of Artificial Intelligence Research}
  \bibinfo{year}{2004};\bibinfo{volume}{22}:\bibinfo{pages}{457--479}.
\bibitem[{Ferdousi et~al.(2017)Ferdousi, Safdari and Omidi}]{Ferdousi_2017}
\bibinfo{author}{Ferdousi\xfnm[ R.]}, \bibinfo{author}{Safdari\xfnm[ R.]},
  \bibinfo{author}{Omidi\xfnm[ Y.]}.
\newblock \bibinfo{title}{Computational prediction of drug-drug interactions
  based on drugs functional similarities}.
\newblock \bibinfo{journal}{Journal of Biomedical Informatics}
  \bibinfo{year}{2017};\bibinfo{volume}{70}:\bibinfo{pages}{54--64}.
\newblock \URLprefix \url{https://doi.org/10.1016%2Fj.jbi.2017.04.021}.
  \DOIprefix\doi{10.1016/j.jbi.2017.04.021}.
\bibitem[{Friedman et~al.(2013)Friedman, Rindflesch and Corn}]{Friedman_2013}
\bibinfo{author}{Friedman\xfnm[ C.]}, \bibinfo{author}{Rindflesch\xfnm[ T.C.]},
  \bibinfo{author}{Corn\xfnm[ M.]}.
\newblock \bibinfo{title}{Natural language processing: State of the art and
  prospects for significant progress, a workshop sponsored by the national
  library of medicine}.
\newblock \bibinfo{journal}{Journal of Biomedical Informatics}
  \bibinfo{year}{2013};\bibinfo{volume}{46}(\bibinfo{number}{5}):\bibinfo{pages}{765--773}.
\newblock \URLprefix \url{https://doi.org/10.1016%2Fj.jbi.2013.06.004}.
  \DOIprefix\doi{10.1016/j.jbi.2013.06.004}.
\bibitem[{Gaizauskas et~al.(2004)Gaizauskas, Hepple and
  Greenwood}]{gaizauskas2004information}
\bibinfo{author}{Gaizauskas\xfnm[ R.]}, \bibinfo{author}{Hepple\xfnm[ M.]},
  \bibinfo{author}{Greenwood\xfnm[ M.]}.
\newblock \bibinfo{title}{Information retrieval for question answering a sigir
  2004 workshop}.
\newblock In: \bibinfo{booktitle}{ACM SIGIR Forum}.
  \bibinfo{organization}{ACM}; volume~\bibinfo{volume}{38};
  \bibinfo{year}{2004}. p. \bibinfo{pages}{41--44}.
\bibitem[{Gobeill et~al.(2009)Gobeill, Patsche, Theodoro, Veuthey, Lovis and
  Ruch}]{gobeill2009question}
\bibinfo{author}{Gobeill\xfnm[ J.]}, \bibinfo{author}{Patsche\xfnm[ E.]},
  \bibinfo{author}{Theodoro\xfnm[ D.]}, \bibinfo{author}{Veuthey\xfnm[ A.L.]},
  \bibinfo{author}{Lovis\xfnm[ C.]}, \bibinfo{author}{Ruch\xfnm[ P.]}.
\newblock \bibinfo{title}{Question answering for biology and medicine}.
\newblock In: \bibinfo{booktitle}{9th International Conference on Information
  Technology and Applications in Biomedicine}. \bibinfo{year}{2009}. p.
  \bibinfo{pages}{1--5}.
\bibitem[{Goetz(2000)}]{goetz2000lucene}
\bibinfo{author}{Goetz\xfnm[ B.]}.
\newblock \bibinfo{title}{The lucene search engine: Powerful, flexible, and
  free}.
\newblock \bibinfo{journal}{Available at http://luceneapacheorg/core/}
  \bibinfo{year}{2000};.
\bibitem[{Green et~al.(1961)Green, Wolf, Chomsky and Laughery}]{Green_1961}
\bibinfo{author}{Green\xfnm[ B.F.]}, \bibinfo{author}{Wolf\xfnm[ A.K.]},
  \bibinfo{author}{Chomsky\xfnm[ C.]}, \bibinfo{author}{Laughery\xfnm[ K.]}.
\newblock \bibinfo{title}{{BASEBALL}: An automatic question-answerer}.
\newblock In: \bibinfo{booktitle}{Papers presented at the May 9-11, 1961,
  western joint {IRE}-{AIEE}-{ACM} computer conference, {IRE}-{AIEE}-{ACM} 61
  (Western)}. \bibinfo{publisher}{Association for Computing Machinery ({ACM})};
  \bibinfo{year}{1961}. p. \bibinfo{pages}{219--224}.
\newblock \URLprefix \url{http://dx.doi.org/10.1145/1460690.1460714}.
  \DOIprefix\doi{10.1145/1460690.1460714}.
\bibitem[{Greengrass(2000)}]{greengrass2000information}
\bibinfo{author}{Greengrass\xfnm[ E.]}.
\newblock \bibinfo{title}{Information retrieval: A survey}.
\newblock \bibinfo{publisher}{University of Maryland, Baltimore County},
  \bibinfo{year}{2000}.
\bibitem[{Gridach(2017)}]{Gridach_2017}
\bibinfo{author}{Gridach\xfnm[ M.]}.
\newblock \bibinfo{title}{Character-level neural network for biomedical named
  entity recognition}.
\newblock \bibinfo{journal}{Journal of Biomedical Informatics}
  \bibinfo{year}{2017};\bibinfo{volume}{70}:\bibinfo{pages}{85--91}.
\newblock \URLprefix \url{https://doi.org/10.1016%2Fj.jbi.2017.05.002}.
  \DOIprefix\doi{10.1016/j.jbi.2017.05.002}.
\bibitem[{Hao et~al.(2017)Hao, Xie, Wu, Weng and Qu}]{Hao_2017}
\bibinfo{author}{Hao\xfnm[ T.]}, \bibinfo{author}{Xie\xfnm[ W.]},
  \bibinfo{author}{Wu\xfnm[ Q.]}, \bibinfo{author}{Weng\xfnm[ H.]},
  \bibinfo{author}{Qu\xfnm[ Y.]}.
\newblock \bibinfo{title}{Leveraging question target word features through
  semantic relation expansion for answer type classification}.
\newblock \bibinfo{journal}{Knowledge-Based Systems}
  \bibinfo{year}{2017};\bibinfo{volume}{133}:\bibinfo{pages}{43--52}.
\newblock \URLprefix \url{https://doi.org/10.1016%2Fj.knosys.2017.06.030}.
  \DOIprefix\doi{10.1016/j.knosys.2017.06.030}.
\bibitem[{Hersh et~al.(2002)Hersh, Crabtree, Hickam, Sacherek, Friedman,
  Tidmarsh, Mosbaek and Kraemer}]{Hersh283}
\bibinfo{author}{Hersh\xfnm[ W.R.]}, \bibinfo{author}{Crabtree\xfnm[ M.K.]},
  \bibinfo{author}{Hickam\xfnm[ D.H.]}, \bibinfo{author}{Sacherek\xfnm[ L.]},
  \bibinfo{author}{Friedman\xfnm[ C.P.]}, \bibinfo{author}{Tidmarsh\xfnm[ P.]},
  \bibinfo{author}{Mosbaek\xfnm[ C.]}, \bibinfo{author}{Kraemer\xfnm[ D.]}.
\newblock \bibinfo{title}{Factors associated with success in searching medline
  and applying evidence to answer clinical questions}.
\newblock \bibinfo{journal}{Journal of the American Medical Informatics
  Association}
  \bibinfo{year}{2002};\bibinfo{volume}{9}(\bibinfo{number}{3}):\bibinfo{pages}{283--293}.
\newblock \URLprefix \url{http://jamia.oxfordjournals.org/content/9/3/283}.
  \DOIprefix\doi{10.1197/jamia.M0996}.
\bibitem[{Hirschman and Gaizauskas(2001)}]{HIRSCHMAN_2001}
\bibinfo{author}{Hirschman\xfnm[ L.]}, \bibinfo{author}{Gaizauskas\xfnm[ R.]}.
\newblock \bibinfo{title}{Natural language question answering: the view from
  here}.
\newblock \bibinfo{journal}{Natural Language Engineering}
  \bibinfo{year}{2001};\bibinfo{volume}{7}(\bibinfo{number}{04}).
\newblock \URLprefix \url{https://doi.org/10.1017%2Fs1351324901002807}.
  \DOIprefix\doi{10.1017/s1351324901002807}.
\bibitem[{Hirschman et~al.(2005)Hirschman, Yeh, Blaschke and
  Valencia}]{Hirschman_2005}
\bibinfo{author}{Hirschman\xfnm[ L.]}, \bibinfo{author}{Yeh\xfnm[ A.]},
  \bibinfo{author}{Blaschke\xfnm[ C.]}, \bibinfo{author}{Valencia\xfnm[ A.]}.
\newblock \bibinfo{title}{Overview of {BioCreAtIvE}: critical assessment of
  information extraction for biology}.
\newblock \bibinfo{journal}{{BMC} Bioinformatics}
  \bibinfo{year}{2005};\bibinfo{volume}{6}(\bibinfo{number}{Suppl
  1}):\bibinfo{pages}{S1}.
\newblock \URLprefix \url{https://doi.org/10.1186%2F1471-2105-6-s1-s1}.
  \DOIprefix\doi{10.1186/1471-2105-6-s1-s1}.
\bibitem[{Hristovski et~al.(2015)Hristovski, Dinevski, Kastrin and
  Rindflesch}]{hristovski2015biomedical}
\bibinfo{author}{Hristovski\xfnm[ D.]}, \bibinfo{author}{Dinevski\xfnm[ D.]},
  \bibinfo{author}{Kastrin\xfnm[ A.]}, \bibinfo{author}{Rindflesch\xfnm[
  T.C.]}.
\newblock \bibinfo{title}{Biomedical question answering using semantic
  relations}.
\newblock \bibinfo{journal}{BMC bioinformatics}
  \bibinfo{year}{2015};\bibinfo{volume}{16}(\bibinfo{number}{1}):\bibinfo{pages}{6}.
\newblock \DOIprefix\doi{10.1186/s12859-014-0365-3}.
\bibitem[{Hu and Liu(2004)}]{Hu_2004}
\bibinfo{author}{Hu\xfnm[ M.]}, \bibinfo{author}{Liu\xfnm[ B.]}.
\newblock \bibinfo{title}{Mining and summarizing customer reviews}.
\newblock In: \bibinfo{booktitle}{Proceedings of the 2004 {ACM} {SIGKDD}
  international conference on Knowledge discovery and data mining - {KDD} 04}.
  \bibinfo{publisher}{{ACM} Press}; \bibinfo{year}{2004}. \URLprefix
  \url{https://doi.org/10.1145%2F1014052.1014073}.
  \DOIprefix\doi{10.1145/1014052.1014073}.
\bibitem[{Huang et~al.(2006)Huang, Lin and
  Demner-Fushman}]{huang2006evaluation}
\bibinfo{author}{Huang\xfnm[ X.]}, \bibinfo{author}{Lin\xfnm[ J.]},
  \bibinfo{author}{Demner-Fushman\xfnm[ D.]}.
\newblock \bibinfo{title}{Evaluation of pico as a knowledge representation for
  clinical questions}.
\newblock In: \bibinfo{booktitle}{AMIA annual symposium proceedings}.
  \bibinfo{organization}{American Medical Informatics Association}; volume
  \bibinfo{volume}{2006}; \bibinfo{year}{2006}. p. \bibinfo{pages}{359}.
\bibitem[{Jackson and Moulinier(2007)}]{Jackson_2007}
\bibinfo{author}{Jackson\xfnm[ P.]}, \bibinfo{author}{Moulinier\xfnm[ I.]}.
\newblock \bibinfo{title}{Natural Language Processing for Online Applications}.
\newblock \bibinfo{publisher}{John Benjamins Publishing Company},
  \bibinfo{year}{2007}.
\newblock \URLprefix \url{https://doi.org/10.1075%2Fnlp.5}.
  \DOIprefix\doi{10.1075/nlp.5}.
\bibitem[{Jacquemart and Zweigenbaum(2003)}]{jacquemart2003towards}
\bibinfo{author}{Jacquemart\xfnm[ P.]}, \bibinfo{author}{Zweigenbaum\xfnm[
  P.]}.
\newblock \bibinfo{title}{Towards a medical question-answering system: a
  feasibility study}.
\newblock \bibinfo{journal}{Studies in health technology and informatics}
  \bibinfo{year}{2003};\bibinfo{volume}{95}:\bibinfo{pages}{463--468}.
\newblock \DOIprefix\doi{https://doi.org/10.3233/978-1-60750-939-4-463}.
\bibitem[{Jang et~al.(2016)Jang, Lee, Lee, Kim and Lee}]{Jang_2016}
\bibinfo{author}{Jang\xfnm[ D.]}, \bibinfo{author}{Lee\xfnm[ S.]},
  \bibinfo{author}{Lee\xfnm[ J.]}, \bibinfo{author}{Kim\xfnm[ K.]},
  \bibinfo{author}{Lee\xfnm[ D.]}.
\newblock \bibinfo{title}{Inferring new drug indications using the
  complementarity between clinical disease signatures and drug effects}.
\newblock \bibinfo{journal}{Journal of Biomedical Informatics}
  \bibinfo{year}{2016};\bibinfo{volume}{59}:\bibinfo{pages}{248--257}.
\newblock \URLprefix \url{https://doi.org/10.1016%2Fj.jbi.2015.12.003}.
  \DOIprefix\doi{10.1016/j.jbi.2015.12.003}.
\bibitem[{Ji et~al.(2017)Ji, Ritter and Yen}]{Ji_2017}
\bibinfo{author}{Ji\xfnm[ X.]}, \bibinfo{author}{Ritter\xfnm[ A.]},
  \bibinfo{author}{Yen\xfnm[ P.Y.]}.
\newblock \bibinfo{title}{Using ontology-based semantic similarity to
  facilitate the article screening process for systematic reviews}.
\newblock \bibinfo{journal}{Journal of Biomedical Informatics}
  \bibinfo{year}{2017};\bibinfo{volume}{69}:\bibinfo{pages}{33--42}.
\newblock \URLprefix \url{https://doi.org/10.1016%2Fj.jbi.2017.03.007}.
  \DOIprefix\doi{10.1016/j.jbi.2017.03.007}.
\bibitem[{Jiang(2013)}]{Jiang2013}
\bibinfo{author}{Jiang\xfnm[ R.]}.
\newblock \bibinfo{title}{Gene-Gene Interaction}; \bibinfo{address}{New York,
  NY}: \bibinfo{publisher}{Springer New York}.
\newblock p. \bibinfo{pages}{841--842}.
\newblock \URLprefix \url{https://doi.org/10.1007/978-1-4419-1005-9_690}.
  \DOIprefix\doi{10.1007/978-1-4419-1005-9_690}.
\bibitem[{Jivani et~al.(2011)}]{jivani2011comparative}
\bibinfo{author}{Jivani\xfnm[ A.G.]}, et~al.
\newblock \bibinfo{title}{A comparative study of stemming algorithms}.
\newblock \bibinfo{journal}{Int J Comp Tech Appl}
  \bibinfo{year}{2011};\bibinfo{volume}{2}(\bibinfo{number}{6}):\bibinfo{pages}{1930--1938}.
\bibitem[{Kaisser(2008)}]{Kaisser_2008}
\bibinfo{author}{Kaisser\xfnm[ M.]}.
\newblock \bibinfo{title}{The {QuALiM} question answering demo}.
\newblock In: \bibinfo{booktitle}{Proceedings of the 46th Annual Meeting of the
  Association for Computational Linguistics on Human Language Technologies Demo
  Session - {HLT} 08}. \bibinfo{publisher}{Association for Computational
  Linguistics ({ACL})}; \bibinfo{year}{2008}. \URLprefix
  \url{http://dx.doi.org/10.3115/1564144.1564153}.
  \DOIprefix\doi{10.3115/1564144.1564153}.
\bibitem[{Kando(2002)}]{kando2002overview}
\bibinfo{author}{Kando\xfnm[ N.]}.
\newblock \bibinfo{title}{Overview of the third ntcir workshop.}
\newblock In: \bibinfo{booktitle}{NTCIR}. \bibinfo{year}{2002}. .
\bibitem[{Katz et~al.(2002)Katz, Felshin, Yuret, Ibrahim, Lin, Marton,
  McFarland and Temelkuran}]{Katz_2002}
\bibinfo{author}{Katz\xfnm[ B.]}, \bibinfo{author}{Felshin\xfnm[ S.]},
  \bibinfo{author}{Yuret\xfnm[ D.]}, \bibinfo{author}{Ibrahim\xfnm[ A.]},
  \bibinfo{author}{Lin\xfnm[ J.]}, \bibinfo{author}{Marton\xfnm[ G.]},
  \bibinfo{author}{McFarland\xfnm[ A.J.]}, \bibinfo{author}{Temelkuran\xfnm[
  B.]}.
\newblock \bibinfo{title}{Omnibase: Uniform access to heterogeneous data for
  question answering}.
\newblock In: \bibinfo{booktitle}{Natural Language Processing and Information
  Systems}. \bibinfo{publisher}{Springer Nature}; \bibinfo{year}{2002}. p.
  \bibinfo{pages}{230--234}.
\newblock \URLprefix \url{http://dx.doi.org/10.1007/3-540-36271-1_23}.
  \DOIprefix\doi{10.1007/3-540-36271-1_23}.
\bibitem[{Khoury(2011{\natexlab{a}})}]{Khoury_2011}
\bibinfo{author}{Khoury\xfnm[ R.]}.
\newblock \bibinfo{title}{Question type classification using a part-of-speech
  hierarchy}.
\newblock In: \bibinfo{booktitle}{Autonomous and Intelligent Systems}.
  \bibinfo{publisher}{Springer Berlin Heidelberg};
  \bibinfo{year}{2011}{\natexlab{a}}. p. \bibinfo{pages}{212--221}.
\newblock \URLprefix \url{https://doi.org/10.1007%2F978-3-642-21538-4_21}.
  \DOIprefix\doi{10.1007/978-3-642-21538-4_21}.
\bibitem[{Khoury(2011{\natexlab{b}})}]{khoury2011question}
\bibinfo{author}{Khoury\xfnm[ R.]}.
\newblock \bibinfo{title}{Question type classification using a part-of-speech
  hierarchy}.
\newblock In: \bibinfo{booktitle}{Autonomous and Intelligent Systems}.
  \bibinfo{year}{2011}{\natexlab{b}}. p. \bibinfo{pages}{212--221}.
\newblock \DOIprefix\doi{10.1007/978-3-642-21538-4_21}.
\bibitem[{Kim et~al.(2012)Kim, Nguyen, Wang, Tsujii, Takagi and
  Yonezawa}]{Kim_2012}
\bibinfo{author}{Kim\xfnm[ J.D.]}, \bibinfo{author}{Nguyen\xfnm[ N.]},
  \bibinfo{author}{Wang\xfnm[ Y.]}, \bibinfo{author}{Tsujii\xfnm[ J.]},
  \bibinfo{author}{Takagi\xfnm[ T.]}, \bibinfo{author}{Yonezawa\xfnm[ A.]}.
\newblock \bibinfo{title}{The genia event and protein coreference tasks of the
  {BioNLP} shared task 2011}.
\newblock \bibinfo{journal}{{BMC} Bioinformatics}
  \bibinfo{year}{2012};\bibinfo{volume}{13}(\bibinfo{number}{Suppl
  11}):\bibinfo{pages}{S1}.
\newblock \URLprefix \url{https://doi.org/10.1186%2F1471-2105-13-s11-s1}.
  \DOIprefix\doi{10.1186/1471-2105-13-s11-s1}.
\bibitem[{Kim et~al.(2003)Kim, Ohta, Tateisi and Tsujii}]{Kim_2003}
\bibinfo{author}{Kim\xfnm[ J.D.]}, \bibinfo{author}{Ohta\xfnm[ T.]},
  \bibinfo{author}{Tateisi\xfnm[ Y.]}, \bibinfo{author}{Tsujii\xfnm[ J.]}.
\newblock \bibinfo{title}{{GENIA} corpus--a semantically annotated corpus for
  bio-textmining}.
\newblock \bibinfo{journal}{Bioinformatics}
  \bibinfo{year}{2003};\bibinfo{volume}{19}(\bibinfo{number}{Suppl
  1}):\bibinfo{pages}{i180--i182}.
\newblock \URLprefix \url{https://doi.org/10.1093%2Fbioinformatics%2Fbtg1023}.
  \DOIprefix\doi{10.1093/bioinformatics/btg1023}.
\bibitem[{Kim et~al.(2011)Kim, Pyysalo, Ohta, Bossy, Nguyen and
  Tsujii}]{kim2011overview}
\bibinfo{author}{Kim\xfnm[ J.D.]}, \bibinfo{author}{Pyysalo\xfnm[ S.]},
  \bibinfo{author}{Ohta\xfnm[ T.]}, \bibinfo{author}{Bossy\xfnm[ R.]},
  \bibinfo{author}{Nguyen\xfnm[ N.]}, \bibinfo{author}{Tsujii\xfnm[ J.]}.
\newblock \bibinfo{title}{Overview of bionlp shared task 2011}.
\newblock In: \bibinfo{booktitle}{Proceedings of the BioNLP Shared Task 2011
  Workshop}. \bibinfo{organization}{Association for Computational Linguistics};
  \bibinfo{year}{2011}. p. \bibinfo{pages}{1--6}.
\bibitem[{Kobayashi and Shyu(2006)}]{kobayashi2006representing}
\bibinfo{author}{Kobayashi\xfnm[ T.]}, \bibinfo{author}{Shyu\xfnm[ C.R.]}.
\newblock \bibinfo{title}{Representing clinical questions by semantic type for
  better classification}.
\newblock In: \bibinfo{booktitle}{AMIA Annual Symposium Proceedings}.
  \bibinfo{organization}{American Medical Informatics Association}; volume
  \bibinfo{volume}{2006}; \bibinfo{year}{2006}. p. \bibinfo{pages}{987}.
\bibitem[{Kopanitsa(2017)}]{Kopanitsa_2017}
\bibinfo{author}{Kopanitsa\xfnm[ G.]}.
\newblock \bibinfo{title}{Integration of hospital information and clinical
  decision support systems to enable the reuse of electronic health record
  data}.
\newblock \bibinfo{journal}{Methods of Information in Medicine}
  \bibinfo{year}{2017};\bibinfo{volume}{56}(\bibinfo{number}{3}):\bibinfo{pages}{238--247}.
\newblock \URLprefix \url{https://doi.org/10.3414%2Fme16-01-0057}.
  \DOIprefix\doi{10.3414/me16-01-0057}.
\bibitem[{Krallinger et~al.(2008)Krallinger, Leitner, Rodriguez-Penagos and
  Valencia}]{Krallinger_2008}
\bibinfo{author}{Krallinger\xfnm[ M.]}, \bibinfo{author}{Leitner\xfnm[ F.]},
  \bibinfo{author}{Rodriguez-Penagos\xfnm[ C.]},
  \bibinfo{author}{Valencia\xfnm[ A.]}.
\newblock \bibinfo{title}{Overview of the protein-protein interaction
  annotation extraction task of {BioCreative} {II}}.
\newblock \bibinfo{journal}{Genome Biology}
  \bibinfo{year}{2008};\bibinfo{volume}{9}(\bibinfo{number}{Suppl
  2}):\bibinfo{pages}{S4--S4}.
\newblock \URLprefix \url{https://doi.org/10.1186%2Fgb-2008-9-s2-s4}.
  \DOIprefix\doi{10.1186/gb-2008-9-s2-s4}.
\bibitem[{Krallinger et~al.(2011)Krallinger, Vazquez, Leitner, Salgado,
  Chatr-aryamontri, Winter, Perfetto, Briganti, Licata, Iannuccelli,
  Castagnoli, Cesareni, Tyers, Schneider, Rinaldi, Leaman, Gonzalez, Matos,
  Kim, Wilbur, Rocha, Shatkay, Tendulkar, Agarwal, Liu, Wang, Rak, Noto, Elkan,
  Lu, Dogan, Fontaine, Andrade-Navarro and Valencia}]{Krallinger_2011}
\bibinfo{author}{Krallinger\xfnm[ M.]}, \bibinfo{author}{Vazquez\xfnm[ M.]},
  \bibinfo{author}{Leitner\xfnm[ F.]}, \bibinfo{author}{Salgado\xfnm[ D.]},
  \bibinfo{author}{Chatr-aryamontri\xfnm[ A.]}, \bibinfo{author}{Winter\xfnm[
  A.]}, \bibinfo{author}{Perfetto\xfnm[ L.]}, \bibinfo{author}{Briganti\xfnm[
  L.]}, \bibinfo{author}{Licata\xfnm[ L.]}, \bibinfo{author}{Iannuccelli\xfnm[
  M.]}, \bibinfo{author}{Castagnoli\xfnm[ L.]}, \bibinfo{author}{Cesareni\xfnm[
  G.]}, \bibinfo{author}{Tyers\xfnm[ M.]}, \bibinfo{author}{Schneider\xfnm[
  G.]}, \bibinfo{author}{Rinaldi\xfnm[ F.]}, \bibinfo{author}{Leaman\xfnm[
  R.]}, \bibinfo{author}{Gonzalez\xfnm[ G.]}, \bibinfo{author}{Matos\xfnm[
  S.]}, \bibinfo{author}{Kim\xfnm[ S.]}, \bibinfo{author}{Wilbur\xfnm[ W.]},
  \bibinfo{author}{Rocha\xfnm[ L.]}, \bibinfo{author}{Shatkay\xfnm[ H.]},
  \bibinfo{author}{Tendulkar\xfnm[ A.V.]}, \bibinfo{author}{Agarwal\xfnm[ S.]},
  \bibinfo{author}{Liu\xfnm[ F.]}, \bibinfo{author}{Wang\xfnm[ X.]},
  \bibinfo{author}{Rak\xfnm[ R.]}, \bibinfo{author}{Noto\xfnm[ K.]},
  \bibinfo{author}{Elkan\xfnm[ C.]}, \bibinfo{author}{Lu\xfnm[ Z.]},
  \bibinfo{author}{Dogan\xfnm[ R.]}, \bibinfo{author}{Fontaine\xfnm[ J.F.]},
  \bibinfo{author}{Andrade-Navarro\xfnm[ M.A.]},
  \bibinfo{author}{Valencia\xfnm[ A.]}.
\newblock \bibinfo{title}{The protein-protein interaction tasks of
  {BioCreative} {III}: classification/ranking of articles and linking
  bio-ontology concepts to full text}.
\newblock \bibinfo{journal}{{BMC} Bioinformatics}
  \bibinfo{year}{2011};\bibinfo{volume}{12}(\bibinfo{number}{Suppl
  8}):\bibinfo{pages}{S3}.
\newblock \URLprefix \url{https://doi.org/10.1186%2F1471-2105-12-s8-s3}.
  \DOIprefix\doi{10.1186/1471-2105-12-s8-s3}.
\bibitem[{Kraus et~al.(2017)Kraus, Niedermeier, Jankrift, Tietbohl, Stachewicz,
  Folkerts, Uflacker and Neves}]{Kraus_2017}
\bibinfo{author}{Kraus\xfnm[ M.]}, \bibinfo{author}{Niedermeier\xfnm[ J.]},
  \bibinfo{author}{Jankrift\xfnm[ M.]}, \bibinfo{author}{Tietbohl\xfnm[ S.]},
  \bibinfo{author}{Stachewicz\xfnm[ T.]}, \bibinfo{author}{Folkerts\xfnm[ H.]},
  \bibinfo{author}{Uflacker\xfnm[ M.]}, \bibinfo{author}{Neves\xfnm[ M.]}.
\newblock \bibinfo{title}{Olelo: a web application for intuitive exploration of
  biomedical literature}.
\newblock \bibinfo{journal}{Nucleic Acids Research}
  \bibinfo{year}{2017};\bibinfo{volume}{45}(\bibinfo{number}{W1}):\bibinfo{pages}{W478--W483}.
\newblock \URLprefix \url{https://doi.org/10.1093%2Fnar%2Fgkx363}.
  \DOIprefix\doi{10.1093/nar/gkx363}.
\bibitem[{Kreimeyer et~al.(2017)Kreimeyer, Foster, Pandey, Arya, Halford,
  Jones, Forshee, Walderhaug and Botsis}]{Kreimeyer_2017}
\bibinfo{author}{Kreimeyer\xfnm[ K.]}, \bibinfo{author}{Foster\xfnm[ M.]},
  \bibinfo{author}{Pandey\xfnm[ A.]}, \bibinfo{author}{Arya\xfnm[ N.]},
  \bibinfo{author}{Halford\xfnm[ G.]}, \bibinfo{author}{Jones\xfnm[ S.F.]},
  \bibinfo{author}{Forshee\xfnm[ R.]}, \bibinfo{author}{Walderhaug\xfnm[ M.]},
  \bibinfo{author}{Botsis\xfnm[ T.]}.
\newblock \bibinfo{title}{Natural language processing systems for capturing and
  standardizing unstructured clinical information: A systematic review}.
\newblock \bibinfo{journal}{Journal of Biomedical Informatics}
  \bibinfo{year}{2017};\bibinfo{volume}{73}:\bibinfo{pages}{14--29}.
\newblock \URLprefix \url{https://doi.org/10.1016%2Fj.jbi.2017.07.012}.
  \DOIprefix\doi{10.1016/j.jbi.2017.07.012}.
\bibitem[{Krithara et~al.(2016)Krithara, Nentidis, Paliouras and
  Kakadiaris}]{krithara2016results}
\bibinfo{author}{Krithara\xfnm[ A.]}, \bibinfo{author}{Nentidis\xfnm[ A.]},
  \bibinfo{author}{Paliouras\xfnm[ G.]}, \bibinfo{author}{Kakadiaris\xfnm[
  I.]}.
\newblock \bibinfo{title}{Results of the 4th edition of {BioASQ} challenge}.
\newblock In: \bibinfo{booktitle}{Proceedings of the Fourth {BioASQ} workshop
  at the Conference of the Association for Computational Linguistics}.
  \bibinfo{year}{2016}. p. \bibinfo{pages}{1--7}.
\bibitem[{Kropf et~al.(2017)Kropf, Kr\"ucken, Mueller and Denecke}]{Kropf_2017}
\bibinfo{author}{Kropf\xfnm[ S.]}, \bibinfo{author}{Kr\"ucken\xfnm[ P.]},
  \bibinfo{author}{Mueller\xfnm[ W.]}, \bibinfo{author}{Denecke\xfnm[ K.]}.
\newblock \bibinfo{title}{Structuring legacy pathology reports by {openEHR}
  archetypes to enable semantic querying}.
\newblock \bibinfo{journal}{Methods of Information in Medicine}
  \bibinfo{year}{2017};\bibinfo{volume}{56}(\bibinfo{number}{2}):\bibinfo{pages}{230--237}.
\newblock \URLprefix \url{https://doi.org/10.3414%2Fme16-01-0073}.
  \DOIprefix\doi{10.3414/me16-01-0073}.
\bibitem[{Krovetz(1993)}]{Krovetz_1993}
\bibinfo{author}{Krovetz\xfnm[ R.]}.
\newblock \bibinfo{title}{Viewing morphology as an inference process}.
\newblock In: \bibinfo{booktitle}{Proceedings of the 16th annual international
  {ACM} {SIGIR} conference on Research and development in information retrieval
  - {SIGIR} 93}. \bibinfo{publisher}{{ACM} Press}; \bibinfo{year}{1993}.
  \URLprefix \url{https://doi.org/10.1145%2F160688.160718}.
  \DOIprefix\doi{10.1145/160688.160718}.
\bibitem[{LeCun et~al.(1989)LeCun, Boser, Denker, Henderson, Howard, Hubbard
  and Jackel}]{lecun1989backpropagation}
\bibinfo{author}{LeCun\xfnm[ Y.]}, \bibinfo{author}{Boser\xfnm[ B.]},
  \bibinfo{author}{Denker\xfnm[ J.S.]}, \bibinfo{author}{Henderson\xfnm[ D.]},
  \bibinfo{author}{Howard\xfnm[ R.E.]}, \bibinfo{author}{Hubbard\xfnm[ W.]},
  \bibinfo{author}{Jackel\xfnm[ L.D.]}.
\newblock \bibinfo{title}{Backpropagation applied to handwritten zip code
  recognition}.
\newblock \bibinfo{journal}{Neural computation}
  \bibinfo{year}{1989};\bibinfo{volume}{1}(\bibinfo{number}{4}):\bibinfo{pages}{541--551}.
\bibitem[{Lee et~al.(2006)Lee, Cimino, Zhu, Sable, Shanker, Ely and
  Yu}]{lee2006beyond}
\bibinfo{author}{Lee\xfnm[ M.]}, \bibinfo{author}{Cimino\xfnm[ J.J.]},
  \bibinfo{author}{Zhu\xfnm[ H.R.]}, \bibinfo{author}{Sable\xfnm[ C.]},
  \bibinfo{author}{Shanker\xfnm[ V.]}, \bibinfo{author}{Ely\xfnm[ J.W.]},
  \bibinfo{author}{Yu\xfnm[ H.]}.
\newblock \bibinfo{title}{Beyond information retrieval-medical question
  answering.}
\newblock In: \bibinfo{booktitle}{AMIA Annual Symposium Proceedings}. volume
  \bibinfo{volume}{2006}; \bibinfo{year}{2006}. p. \bibinfo{pages}{469--473}.
\bibitem[{Levow(2013)}]{levow2013uwcl}
\bibinfo{author}{Levow\xfnm[ G.A.]}.
\newblock \bibinfo{title}{{UWCL} at mediaeval 2013: Similar segments in social
  speech task.}
\newblock In: \bibinfo{booktitle}{MediaEval}. \bibinfo{year}{2013}. .
\bibitem[{Li et~al.(2008)Li, Zhang, Yuan and Zhu}]{li2008classifying}
\bibinfo{author}{Li\xfnm[ F.]}, \bibinfo{author}{Zhang\xfnm[ X.]},
  \bibinfo{author}{Yuan\xfnm[ J.]}, \bibinfo{author}{Zhu\xfnm[ X.]}.
\newblock \bibinfo{title}{Classifying what-type questions by head noun
  tagging}.
\newblock In: \bibinfo{booktitle}{Proceedings of the 22nd International
  Conference on Computational Linguistics-Volume 1}.
  \bibinfo{organization}{Association for Computational Linguistics};
  \bibinfo{year}{2008}. p. \bibinfo{pages}{481--488}.
\bibitem[{Li and Roth(2002)}]{li2002learning}
\bibinfo{author}{Li\xfnm[ X.]}, \bibinfo{author}{Roth\xfnm[ D.]}.
\newblock \bibinfo{title}{Learning question classifiers}.
\newblock In: \bibinfo{booktitle}{Proceedings of the 19th international
  conference on Computational linguistics-Volume 1}. \bibinfo{year}{2002}. p.
  \bibinfo{pages}{1--7}.
\newblock \DOIprefix\doi{10.3115/1072228.1072378}.
\bibitem[{Li and Roth(2005)}]{li2006learning}
\bibinfo{author}{Li\xfnm[ X.]}, \bibinfo{author}{Roth\xfnm[ D.]}.
\newblock \bibinfo{title}{Learning question classifiers: the role of semantic
  information}.
\newblock \bibinfo{journal}{Natural Language Engineering}
  \bibinfo{year}{2005};\bibinfo{volume}{12}(\bibinfo{number}{03}):\bibinfo{pages}{229--249}.
\newblock \DOIprefix\doi{10.1017/s1351324905003955}.
\bibitem[{Li et~al.(2015)Li, Yin, Zhang, Liu, Zhang and Hao}]{ligeneric}
\bibinfo{author}{Li\xfnm[ Y.]}, \bibinfo{author}{Yin\xfnm[ X.C.]},
  \bibinfo{author}{Zhang\xfnm[ B.W.]}, \bibinfo{author}{Liu\xfnm[ T.T.]},
  \bibinfo{author}{Zhang\xfnm[ Z.J.]}, \bibinfo{author}{Hao\xfnm[ H.W.]}.
\newblock \bibinfo{title}{A generic framework for biomedical snippet
  retrieval}.
\newblock In: \bibinfo{booktitle}{Third International Conference on Artificial
  Intelligence, Modelling and Simulation}. \bibinfo{year}{2015}. .
\bibitem[{Lin and Wilbur(2007)}]{lin2007syntactic}
\bibinfo{author}{Lin\xfnm[ J.]}, \bibinfo{author}{Wilbur\xfnm[ W.J.]}.
\newblock \bibinfo{title}{Syntactic sentence compression in the biomedical
  domain: facilitating access to related articles}.
\newblock \bibinfo{journal}{Information Retrieval}
  \bibinfo{year}{2007};\bibinfo{volume}{10}(\bibinfo{number}{4-5}):\bibinfo{pages}{393--414}.
\newblock \URLprefix \url{https://doi.org/10.1007%2Fs10791-007-9029-5}.
  \DOIprefix\doi{10.1007/s10791-007-9029-5}.
\bibitem[{Lindberg et~al.(1993)Lindberg, Humphreys, McCray
  et~al.}]{lindberg1993unified}
\bibinfo{author}{Lindberg\xfnm[ D.A.]}, \bibinfo{author}{Humphreys\xfnm[
  B.L.]}, \bibinfo{author}{McCray\xfnm[ A.T.]}, et~al.
\newblock \bibinfo{title}{The unified medical language system}.
\newblock \bibinfo{journal}{Methods of Information in Medicine}
  \bibinfo{year}{1993};\bibinfo{volume}{32}(\bibinfo{number}{4}):\bibinfo{pages}{281--291}.
\bibitem[{Lingeman and Dietz(2014)}]{lingeman2014umass}
\bibinfo{author}{Lingeman\xfnm[ J.]}, \bibinfo{author}{Dietz\xfnm[ L.]}.
\newblock \bibinfo{title}{{UMASS} at {BioASQ} 2014: Figure-inspired text
  retrieval.}
\newblock In: \bibinfo{booktitle}{CLEF (Working Notes)}. \bibinfo{year}{2014}.
  p. \bibinfo{pages}{1296--1310}.
\bibitem[{Liu et~al.(2011)Liu, Antieau and Yu}]{Liu_2011}
\bibinfo{author}{Liu\xfnm[ F.]}, \bibinfo{author}{Antieau\xfnm[ L.D.]},
  \bibinfo{author}{Yu\xfnm[ H.]}.
\newblock \bibinfo{title}{Toward automated consumer question answering:
  Automatically separating consumer questions from professional questions in
  the healthcare domain}.
\newblock \bibinfo{journal}{Journal of Biomedical Informatics}
  \bibinfo{year}{2011};\bibinfo{volume}{44}(\bibinfo{number}{6}):\bibinfo{pages}{1032--1038}.
\newblock \URLprefix \url{https://doi.org/10.1016%2Fj.jbi.2011.08.008}.
  \DOIprefix\doi{10.1016/j.jbi.2011.08.008}.
\bibitem[{Loni(2011)}]{loni2011survey}
\bibinfo{author}{Loni\xfnm[ B.]}.
\newblock \bibinfo{title}{A survey of state-of-the-art methods on question
  classification}.
\newblock \bibinfo{journal}{Delft University of Technology, Delft}
  \bibinfo{year}{2011};:\bibinfo{pages}{40}.
\bibitem[{Lopez et~al.(2011)Lopez, Uren, Sabou and Motta}]{lopez2011question}
\bibinfo{author}{Lopez\xfnm[ V.]}, \bibinfo{author}{Uren\xfnm[ V.]},
  \bibinfo{author}{Sabou\xfnm[ M.]}, \bibinfo{author}{Motta\xfnm[ E.]}.
\newblock \bibinfo{title}{Is question answering fit for the semantic web?: a
  survey}.
\newblock \bibinfo{journal}{Semantic Web}
  \bibinfo{year}{2011};\bibinfo{volume}{2}(\bibinfo{number}{2}):\bibinfo{pages}{125--155}.
\bibitem[{Magnini et~al.(2005)Magnini, Vallin, Ayache, Erbach, Pe{\~{n}}as,
  de~Rijke, Rocha, Simov and Sutcliffe}]{Magnini_2005}
\bibinfo{author}{Magnini\xfnm[ B.]}, \bibinfo{author}{Vallin\xfnm[ A.]},
  \bibinfo{author}{Ayache\xfnm[ C.]}, \bibinfo{author}{Erbach\xfnm[ G.]},
  \bibinfo{author}{Pe{\~{n}}as\xfnm[ A.]}, \bibinfo{author}{de~Rijke\xfnm[
  M.]}, \bibinfo{author}{Rocha\xfnm[ P.]}, \bibinfo{author}{Simov\xfnm[ K.]},
  \bibinfo{author}{Sutcliffe\xfnm[ R.]}.
\newblock \bibinfo{title}{Overview of the {CLEF} 2004 multilingual question
  answering track}.
\newblock In: \bibinfo{booktitle}{Multilingual Information Access for Text,
  Speech and Images}. \bibinfo{publisher}{Springer Berlin Heidelberg};
  \bibinfo{year}{2005}. p. \bibinfo{pages}{371--391}.
\newblock \URLprefix \url{https://doi.org/10.1007%2F11519645_38}.
  \DOIprefix\doi{10.1007/11519645_38}.
\bibitem[{Manning et~al.(2014)Manning, Surdeanu, Bauer, Finkel, Bethard and
  McClosky}]{Manning_2014}
\bibinfo{author}{Manning\xfnm[ C.]}, \bibinfo{author}{Surdeanu\xfnm[ M.]},
  \bibinfo{author}{Bauer\xfnm[ J.]}, \bibinfo{author}{Finkel\xfnm[ J.]},
  \bibinfo{author}{Bethard\xfnm[ S.]}, \bibinfo{author}{McClosky\xfnm[ D.]}.
\newblock \bibinfo{title}{The stanford {CoreNLP} natural language processing
  toolkit}.
\newblock In: \bibinfo{booktitle}{Proceedings of 52nd Annual Meeting of the
  Association for Computational Linguistics: System Demonstrations}.
  \bibinfo{year}{2014}. \DOIprefix\doi{10.3115/v1/p14-5010}.
\bibitem[{Mao et~al.(2014)Mao, Wei and Lu}]{mao2014ncbi}
\bibinfo{author}{Mao\xfnm[ Y.]}, \bibinfo{author}{Wei\xfnm[ C.H.]},
  \bibinfo{author}{Lu\xfnm[ Z.]}.
\newblock \bibinfo{title}{{NCBI} at the 2014 {BioASQ} challenge task:
  Large-scale biomedical semantic indexing and question answering}
  \bibinfo{year}{2014};:\bibinfo{pages}{1319--1327}.
\bibitem[{Marchand et~al.(2013)Marchand, Ginsca, Besan{\c{c}}on and
  Mesnard}]{marchand2013lvic}
\bibinfo{author}{Marchand\xfnm[ M.]}, \bibinfo{author}{Ginsca\xfnm[ A.]},
  \bibinfo{author}{Besan{\c{c}}on\xfnm[ R.]}, \bibinfo{author}{Mesnard\xfnm[
  O.]}.
\newblock \bibinfo{title}{[lvic-limsi]: Using syntactic features and
  multi-polarity words for sentiment analysis in twitter}.
\newblock In: \bibinfo{booktitle}{Proceedings of the Seventh International
  Workshop on Semantic Evaluation (SemEval 2013)}. volume~\bibinfo{volume}{2};
  \bibinfo{year}{2013}. p. \bibinfo{pages}{418--424}.
\bibitem[{McInnes et~al.(2009)McInnes, Pedersen and Pakhomov}]{mcinnes2009umls}
\bibinfo{author}{McInnes\xfnm[ B.T.]}, \bibinfo{author}{Pedersen\xfnm[ T.]},
  \bibinfo{author}{Pakhomov\xfnm[ S.V.]}.
\newblock \bibinfo{title}{{UMLS}-interface and {UMLS}-similarity: open source
  software for measuring paths and semantic similarity}.
\newblock In: \bibinfo{booktitle}{AMIA Annual Symposium Proceedings}. volume
  \bibinfo{volume}{2009}; \bibinfo{year}{2009}. p. \bibinfo{pages}{431}.
\bibitem[{McRoy et~al.(2015)McRoy, Jones and Kurmally}]{McRoy_2016}
\bibinfo{author}{McRoy\xfnm[ S.]}, \bibinfo{author}{Jones\xfnm[ S.]},
  \bibinfo{author}{Kurmally\xfnm[ A.]}.
\newblock \bibinfo{title}{Toward automated classification of consumers'
  cancer-related questions with a new taxonomy of expected answer types}.
\newblock \bibinfo{journal}{Health Informatics Journal}
  \bibinfo{year}{2015};\bibinfo{volume}{22}(\bibinfo{number}{3}):\bibinfo{pages}{523--535}.
\newblock \URLprefix \url{https://doi.org/10.1177%2F1460458215571643}.
  \DOIprefix\doi{10.1177/1460458215571643}.
\bibitem[{Metzler and Croft(2005)}]{metzler2005analysis}
\bibinfo{author}{Metzler\xfnm[ D.]}, \bibinfo{author}{Croft\xfnm[ W.B.]}.
\newblock \bibinfo{title}{Analysis of statistical question classification for
  fact-based questions}.
\newblock \bibinfo{journal}{Information Retrieval}
  \bibinfo{year}{2005};\bibinfo{volume}{8}(\bibinfo{number}{3}):\bibinfo{pages}{481--504}.
\newblock \DOIprefix\doi{10.1007/s10791-005-6995-3}.
\bibitem[{Mikheev(1996)}]{Mikheev_1996}
\bibinfo{author}{Mikheev\xfnm[ A.]}.
\newblock \bibinfo{title}{Learning part-of-speech guessing rules from lexicon}.
\newblock In: \bibinfo{booktitle}{Proceedings of the 16th conference on
  Computational linguistics -}. \bibinfo{publisher}{Association for
  Computational Linguistics}; \bibinfo{year}{1996}. \URLprefix
  \url{https://doi.org/10.3115%2F993268.993302}.
  \DOIprefix\doi{10.3115/993268.993302}.
\bibitem[{Miller(1995)}]{Miller_1995}
\bibinfo{author}{Miller\xfnm[ G.A.]}.
\newblock \bibinfo{title}{{WordNet}: a lexical database for english}.
\newblock \bibinfo{journal}{Communications of the {ACM}}
  \bibinfo{year}{1995};\bibinfo{volume}{38}(\bibinfo{number}{11}):\bibinfo{pages}{39--41}.
\newblock \URLprefix \url{https://doi.org/10.1145%2F219717.219748}.
  \DOIprefix\doi{10.1145/219717.219748}.
\bibitem[{Mishra and Jain(2016)}]{mishra2016survey}
\bibinfo{author}{Mishra\xfnm[ A.]}, \bibinfo{author}{Jain\xfnm[ S.K.]}.
\newblock \bibinfo{title}{A survey on question answering systems with
  classification}.
\newblock \bibinfo{journal}{Journal of King Saud University-Computer and
  Information Sciences}
  \bibinfo{year}{2016};\bibinfo{volume}{28}(\bibinfo{number}{3}):\bibinfo{pages}{345--361}.
\newblock \DOIprefix\doi{https://doi.org/10.1109/ictrc.2015.7156484}.
\bibitem[{Mitra and Craswell(2017)}]{mitra2017neural}
\bibinfo{author}{Mitra\xfnm[ B.]}, \bibinfo{author}{Craswell\xfnm[ N.]}.
\newblock \bibinfo{title}{Neural models for information retrieval}.
\newblock \bibinfo{journal}{CoRR abs/170501509}
  \bibinfo{year}{2017};\DOIprefix\doi{https://arxiv.org/abs/1705.01509}.
\bibitem[{Mitra and Chaudhuri(2000)}]{mitra2000information}
\bibinfo{author}{Mitra\xfnm[ M.]}, \bibinfo{author}{Chaudhuri\xfnm[ B.]}.
\newblock \bibinfo{title}{Information retrieval from documents: A survey}.
\newblock \bibinfo{journal}{Information retrieval}
  \bibinfo{year}{2000};\bibinfo{volume}{2}(\bibinfo{number}{2-3}):\bibinfo{pages}{141--163}.
\bibitem[{Mohan et~al.(2017)Mohan, Fiorini, Kim and Lu}]{mohan2017deep}
\bibinfo{author}{Mohan\xfnm[ S.]}, \bibinfo{author}{Fiorini\xfnm[ N.]},
  \bibinfo{author}{Kim\xfnm[ S.]}, \bibinfo{author}{Lu\xfnm[ Z.]}.
\newblock \bibinfo{title}{Deep learning for biomedical information retrieval:
  Learning textual relevance from click logs}.
\newblock \bibinfo{journal}{BioNLP 2017}
  \bibinfo{year}{2017};:\bibinfo{pages}{222--231}.
\bibitem[{Moldovan et~al.(2003{\natexlab{a}})Moldovan, Pa{\c{s}}ca, Harabagiu
  and Surdeanu}]{moldovan2003performance}
\bibinfo{author}{Moldovan\xfnm[ D.]}, \bibinfo{author}{Pa{\c{s}}ca\xfnm[ M.]},
  \bibinfo{author}{Harabagiu\xfnm[ S.]}, \bibinfo{author}{Surdeanu\xfnm[ M.]}.
\newblock \bibinfo{title}{Performance issues and error analysis in an
  open-domain question answering system}.
\newblock \bibinfo{journal}{ACM Transactions on Information Systems (TOIS)}
  \bibinfo{year}{2003}{\natexlab{a}};\bibinfo{volume}{21}(\bibinfo{number}{2}):\bibinfo{pages}{133--154}.
\newblock \DOIprefix\doi{10.1145/763693.763694}.
\bibitem[{Moldovan et~al.(2003{\natexlab{b}})Moldovan, Pa{\c{s}}ca, Harabagiu
  and Surdeanu}]{Moldovan_2003}
\bibinfo{author}{Moldovan\xfnm[ D.]}, \bibinfo{author}{Pa{\c{s}}ca\xfnm[ M.]},
  \bibinfo{author}{Harabagiu\xfnm[ S.]}, \bibinfo{author}{Surdeanu\xfnm[ M.]}.
\newblock \bibinfo{title}{Performance issues and error analysis in an
  open-domain question answering system}.
\newblock \bibinfo{journal}{{ACM} Transactions on Information Systems}
  \bibinfo{year}{2003}{\natexlab{b}};\bibinfo{volume}{21}(\bibinfo{number}{2}):\bibinfo{pages}{133--154}.
\newblock \URLprefix \url{https://doi.org/10.1145%2F763693.763694}.
  \DOIprefix\doi{10.1145/763693.763694}.
\bibitem[{Moll{\'a} and Vicedo(2007)}]{molla2007question}
\bibinfo{author}{Moll{\'a}\xfnm[ D.]}, \bibinfo{author}{Vicedo\xfnm[ J.L.]}.
\newblock \bibinfo{title}{Question answering in restricted domains: An
  overview}.
\newblock \bibinfo{journal}{Computational Linguistics}
  \bibinfo{year}{2007};\bibinfo{volume}{33}(\bibinfo{number}{1}):\bibinfo{pages}{41--61}.
\bibitem[{Monz(2003)}]{Monz_2003}
\bibinfo{author}{Monz\xfnm[ C.]}.
\newblock \bibinfo{title}{Document retrieval in the context of question
  answering}.
\newblock In: \bibinfo{booktitle}{Lecture Notes in Computer Science}.
  \bibinfo{publisher}{Springer Berlin Heidelberg}; \bibinfo{year}{2003}. p.
  \bibinfo{pages}{571--579}.
\newblock \URLprefix \url{https://doi.org/10.1007%2F3-540-36618-0_44}.
  \DOIprefix\doi{10.1007/3-540-36618-0_44}.
\bibitem[{Morante et~al.(2012)Morante, Krallinger, Valencia and
  Daelemans}]{morante2012machine}
\bibinfo{author}{Morante\xfnm[ R.]}, \bibinfo{author}{Krallinger\xfnm[ M.]},
  \bibinfo{author}{Valencia\xfnm[ A.]}, \bibinfo{author}{Daelemans\xfnm[ W.]}.
\newblock \bibinfo{title}{Machine reading of biomedical texts about
  alzheimer’s disease 1}.
\newblock \bibinfo{journal}{CLEF} \bibinfo{year}{2012};.
\bibitem[{Morante et~al.(2013)Morante, Krallinger, Valencia and
  Daelemans}]{morante2013machine}
\bibinfo{author}{Morante\xfnm[ R.]}, \bibinfo{author}{Krallinger\xfnm[ M.]},
  \bibinfo{author}{Valencia\xfnm[ A.]}, \bibinfo{author}{Daelemans\xfnm[ W.]}.
\newblock \bibinfo{title}{Machine reading of biomedical texts about
  alzheimer’s disease 1}.
\newblock \bibinfo{journal}{CLEF} \bibinfo{year}{2013};.
\bibitem[{Morid et~al.(2016)Morid, Fiszman, Raja, Jonnalagadda and
  Fiol}]{Morid_2016}
\bibinfo{author}{Morid\xfnm[ M.A.]}, \bibinfo{author}{Fiszman\xfnm[ M.]},
  \bibinfo{author}{Raja\xfnm[ K.]}, \bibinfo{author}{Jonnalagadda\xfnm[ S.R.]},
  \bibinfo{author}{Fiol\xfnm[ G.D.]}.
\newblock \bibinfo{title}{Classification of clinically useful sentences in
  clinical evidence resources}.
\newblock \bibinfo{journal}{Journal of Biomedical Informatics}
  \bibinfo{year}{2016};\bibinfo{volume}{60}:\bibinfo{pages}{14--22}.
\newblock \URLprefix \url{https://doi.org/10.1016%2Fj.jbi.2016.01.003}.
  \DOIprefix\doi{10.1016/j.jbi.2016.01.003}.
\bibitem[{van Mulligen et~al.(2012)van Mulligen, Fourrier-Reglat, Gurwitz,
  Molokhia, Nieto, Trifiro, Kors and Furlong}]{van_Mulligen_2012}
\bibinfo{author}{van Mulligen\xfnm[ E.M.]},
  \bibinfo{author}{Fourrier-Reglat\xfnm[ A.]}, \bibinfo{author}{Gurwitz\xfnm[
  D.]}, \bibinfo{author}{Molokhia\xfnm[ M.]}, \bibinfo{author}{Nieto\xfnm[
  A.]}, \bibinfo{author}{Trifiro\xfnm[ G.]}, \bibinfo{author}{Kors\xfnm[
  J.A.]}, \bibinfo{author}{Furlong\xfnm[ L.I.]}.
\newblock \bibinfo{title}{The {EU}-{ADR} corpus: Annotated drugs, diseases,
  targets, and their relationships}.
\newblock \bibinfo{journal}{Journal of Biomedical Informatics}
  \bibinfo{year}{2012};\bibinfo{volume}{45}(\bibinfo{number}{5}):\bibinfo{pages}{879--884}.
\newblock \URLprefix \url{https://doi.org/10.1016%2Fj.jbi.2012.04.004}.
  \DOIprefix\doi{10.1016/j.jbi.2012.04.004}.
\bibitem[{Nastase et~al.(2007)Nastase, Shirabad and
  Caropreso}]{nastase2007using}
\bibinfo{author}{Nastase\xfnm[ V.]}, \bibinfo{author}{Shirabad\xfnm[ J.S.]},
  \bibinfo{author}{Caropreso\xfnm[ M.F.]}.
\newblock \bibinfo{title}{Using dependency relations for text classification}.
\newblock \bibinfo{journal}{University of Ottawa SITE Technical Report
  TR-2007-12} \bibinfo{year}{2007};\bibinfo{volume}{13}.
\bibitem[{N{\.e}dellec et~al.(2016)N{\.e}dellec, Bossy and
  Kim}]{nedellec2016proceedings}
\bibinfo{author}{N{\.e}dellec\xfnm[ C.]}, \bibinfo{author}{Bossy\xfnm[ R.]},
  \bibinfo{author}{Kim\xfnm[ J.D.]}.
\newblock \bibinfo{title}{Proceedings of the 4th {BioNLP} shared task
  workshop}.
\newblock In: \bibinfo{booktitle}{Proceedings of the 4th BioNLP Shared Task
  Workshop}. \bibinfo{year}{2016}. \URLprefix
  \url{https://doi.org/10.18653%2Fv1%2Fw16-30}.
  \DOIprefix\doi{10.18653/v1/w16-30}.
\bibitem[{N{\'e}dellec et~al.(2013)N{\'e}dellec, Bossy, Kim, Kim, Ohta, Pyysalo
  and Zweigenbaum}]{nedellec2013overview}
\bibinfo{author}{N{\'e}dellec\xfnm[ C.]}, \bibinfo{author}{Bossy\xfnm[ R.]},
  \bibinfo{author}{Kim\xfnm[ J.D.]}, \bibinfo{author}{Kim\xfnm[ J.J.]},
  \bibinfo{author}{Ohta\xfnm[ T.]}, \bibinfo{author}{Pyysalo\xfnm[ S.]},
  \bibinfo{author}{Zweigenbaum\xfnm[ P.]}.
\newblock \bibinfo{title}{Overview of bionlp shared task 2013}.
\newblock In: \bibinfo{booktitle}{Proceedings of the BioNLP Shared Task 2013
  Workshop}. \bibinfo{organization}{Association for Computational Linguistics
  Sofia, Bulgaria}; \bibinfo{year}{2013}. p. \bibinfo{pages}{1--7}.
\bibitem[{Nentidis et~al.(2017)Nentidis, Bougiatiotis, Krithara, Paliouras and
  Kakadiaris}]{Nentidis_2017}
\bibinfo{author}{Nentidis\xfnm[ A.]}, \bibinfo{author}{Bougiatiotis\xfnm[ K.]},
  \bibinfo{author}{Krithara\xfnm[ A.]}, \bibinfo{author}{Paliouras\xfnm[ G.]},
  \bibinfo{author}{Kakadiaris\xfnm[ I.]}.
\newblock \bibinfo{title}{Results of the fifth edition of the {BioASQ}
  challenge}.
\newblock In: \bibinfo{booktitle}{{BioNLP} 2017}.
  \bibinfo{publisher}{Association for Computational Linguistics};
  \bibinfo{year}{2017}. \URLprefix
  \url{https://doi.org/10.18653%2Fv1%2Fw17-2306}.
  \DOIprefix\doi{10.18653/v1/w17-2306}.
\bibitem[{Neves(2014)}]{neves2014hpi}
\bibinfo{author}{Neves\xfnm[ M.]}.
\newblock \bibinfo{title}{{HPI} in-memory-based database system in {Task} 2b of
  {BioASQ}}.
\newblock \bibinfo{journal}{Proceedings of Question Answering Lab at {CLEF}}
  \bibinfo{year}{2014};.
\bibitem[{Neves(2015)}]{neves2015hpi}
\bibinfo{author}{Neves\xfnm[ M.]}.
\newblock \bibinfo{title}{{HPI} question answering system in the {BioASQ} 2015
  challenge}.
\newblock In: \bibinfo{booktitle}{Working Notes for the Conference and Labs of
  the Evaluation Forum ({CLEF})}. \bibinfo{year}{2015}. .
\bibitem[{Neves and Leser(2015)}]{neves2015question}
\bibinfo{author}{Neves\xfnm[ M.]}, \bibinfo{author}{Leser\xfnm[ U.]}.
\newblock \bibinfo{title}{Question answering for biology}.
\newblock \bibinfo{journal}{Methods}
  \bibinfo{year}{2015};\bibinfo{volume}{74}:\bibinfo{pages}{36--46}.
\newblock \DOIprefix\doi{10.1016/j.ymeth.2014.10.023}.
\bibitem[{Niu and Hirst(2004)}]{niu2004analysis}
\bibinfo{author}{Niu\xfnm[ Y.]}, \bibinfo{author}{Hirst\xfnm[ G.]}.
\newblock \bibinfo{title}{Analysis of semantic classes in medical text for
  question answering}.
\newblock In: \bibinfo{booktitle}{Proceedings of the ACL 2004 Workshop on
  Question Answering in Restricted Domains}. \bibinfo{organization}{Association
  for Computational Linguistics}; \bibinfo{year}{2004}. p.
  \bibinfo{pages}{54--61}.
\bibitem[{Niu et~al.(2003)Niu, Hirst, McArthur and
  Rodriguez-Gianolli}]{niu2003answering}
\bibinfo{author}{Niu\xfnm[ Y.]}, \bibinfo{author}{Hirst\xfnm[ G.]},
  \bibinfo{author}{McArthur\xfnm[ G.]},
  \bibinfo{author}{Rodriguez-Gianolli\xfnm[ P.]}.
\newblock \bibinfo{title}{Answering clinical questions with role
  identification}.
\newblock In: \bibinfo{booktitle}{Proceedings of the ACL 2003 workshop on
  Natural language processing in biomedicine-Volume 13}.
  \bibinfo{organization}{Association for Computational Linguistics};
  \bibinfo{year}{2003}. p. \bibinfo{pages}{73--80}.
\bibitem[{Niu et~al.(2006)Niu, Zhu and Hirst}]{niu2006using}
\bibinfo{author}{Niu\xfnm[ Y.]}, \bibinfo{author}{Zhu\xfnm[ X.]},
  \bibinfo{author}{Hirst\xfnm[ G.]}.
\newblock \bibinfo{title}{Using outcome polarity in sentence extraction for
  medical question-answering}.
\newblock In: \bibinfo{booktitle}{AMIA Annual Symposium Proceedings}.
  \bibinfo{organization}{American Medical Informatics Association}; volume
  \bibinfo{volume}{2006}; \bibinfo{year}{2006}. p. \bibinfo{pages}{599}.
\bibitem[{Niu et~al.(2005)Niu, Zhu, Li and Hirst}]{niu2005analysis}
\bibinfo{author}{Niu\xfnm[ Y.]}, \bibinfo{author}{Zhu\xfnm[ X.]},
  \bibinfo{author}{Li\xfnm[ J.]}, \bibinfo{author}{Hirst\xfnm[ G.]}.
\newblock \bibinfo{title}{Analysis of polarity information in medical text}.
\newblock In: \bibinfo{booktitle}{AMIA annual symposium proceedings}.
  \bibinfo{organization}{American Medical Informatics Association}; volume
  \bibinfo{volume}{2005}; \bibinfo{year}{2005}. p. \bibinfo{pages}{570}.
\bibitem[{Othman and Faiz(2016)}]{Othman_2016}
\bibinfo{author}{Othman\xfnm[ N.]}, \bibinfo{author}{Faiz\xfnm[ R.]}.
\newblock \bibinfo{title}{A multi-lingual approach to improve passage retrieval
  for automatic question answering}.
\newblock In: \bibinfo{booktitle}{Natural Language Processing and Information
  Systems}. \bibinfo{year}{2016}. p. \bibinfo{pages}{127--139}.
\newblock \DOIprefix\doi{10.1007/978-3-319-41754-7_11}.
\bibitem[{Otterbacher et~al.(2009)Otterbacher, Erkan and
  Radev}]{Otterbacher_2009}
\bibinfo{author}{Otterbacher\xfnm[ J.]}, \bibinfo{author}{Erkan\xfnm[ G.]},
  \bibinfo{author}{Radev\xfnm[ D.R.]}.
\newblock \bibinfo{title}{Biased {LexRank}: Passage retrieval using random
  walks with question-based priors}.
\newblock \bibinfo{journal}{Information Processing and Management}
  \bibinfo{year}{2009};\bibinfo{volume}{45}(\bibinfo{number}{1}):\bibinfo{pages}{42--54}.
\newblock \DOIprefix\doi{10.1016/j.ipm.2008.06.004}.
\bibitem[{Patrick and Li(2012)}]{Patrick_2012}
\bibinfo{author}{Patrick\xfnm[ J.]}, \bibinfo{author}{Li\xfnm[ M.]}.
\newblock \bibinfo{title}{An ontology for clinical questions about the contents
  of patient notes}.
\newblock \bibinfo{journal}{Journal of Biomedical Informatics}
  \bibinfo{year}{2012};\bibinfo{volume}{45}(\bibinfo{number}{2}):\bibinfo{pages}{292--306}.
\newblock \URLprefix \url{http://dx.doi.org/10.1016/j.jbi.2011.11.008}.
  \DOIprefix\doi{10.1016/j.jbi.2011.11.008}.
\bibitem[{Peng et~al.(2015)Peng, You, Xie, Zhang and Zhu}]{peng2015fudan}
\bibinfo{author}{Peng\xfnm[ S.]}, \bibinfo{author}{You\xfnm[ R.]},
  \bibinfo{author}{Xie\xfnm[ Z.]}, \bibinfo{author}{Zhang\xfnm[ Y.]},
  \bibinfo{author}{Zhu\xfnm[ S.]}.
\newblock \bibinfo{title}{The {Fudan} participation in the 2015 {BioASQ}
  challenge: Large-scale biomedical semantic indexing and question answering}.
\newblock In: \bibinfo{booktitle}{Working Notes for the Conference and Labs of
  the Evaluation Forum (CLEF), Toulouse, France}. \bibinfo{year}{2015}. .
\bibitem[{Porter(1980)}]{Porter_1980}
\bibinfo{author}{Porter\xfnm[ M.]}.
\newblock \bibinfo{title}{An algorithm for suffix stripping}.
\newblock \bibinfo{journal}{Program: electronic library and information
  systems}
  \bibinfo{year}{1980};\bibinfo{volume}{14}(\bibinfo{number}{3}):\bibinfo{pages}{130--137}.
\newblock \URLprefix \url{http://dx.doi.org/10.1108/eb046814}.
  \DOIprefix\doi{10.1108/eb046814}.
\bibitem[{Richardson et~al.(1995)Richardson, Wilson, Nishikawa and
  Hayward}]{richardson1995well}
\bibinfo{author}{Richardson\xfnm[ W.S.]}, \bibinfo{author}{Wilson\xfnm[ M.C.]},
  \bibinfo{author}{Nishikawa\xfnm[ J.]}, \bibinfo{author}{Hayward\xfnm[ R.S.]}.
\newblock \bibinfo{title}{The well-built clinical question: a key to
  evidence-based decisions}.
\newblock \bibinfo{journal}{ACP journal club}
  \bibinfo{year}{1995};\bibinfo{volume}{123}(\bibinfo{number}{3}):\bibinfo{pages}{A12--A12}.
\bibitem[{Rindflesch and Fiszman(2003)}]{Rindflesch_2003}
\bibinfo{author}{Rindflesch\xfnm[ T.C.]}, \bibinfo{author}{Fiszman\xfnm[ M.]}.
\newblock \bibinfo{title}{The interaction of domain knowledge and linguistic
  structure in natural language processing: interpreting hypernymic
  propositions in biomedical text}.
\newblock \bibinfo{journal}{Journal of Biomedical Informatics}
  \bibinfo{year}{2003};\bibinfo{volume}{36}(\bibinfo{number}{6}):\bibinfo{pages}{462--477}.
\newblock \URLprefix \url{https://doi.org/10.1016%2Fj.jbi.2003.11.003}.
  \DOIprefix\doi{10.1016/j.jbi.2003.11.003}.
\bibitem[{Roberts(2002)}]{roberts2002information}
\bibinfo{author}{Roberts\xfnm[ I.]}.
\newblock \bibinfo{title}{Information Retrieval for question answering}.
\newblock Ph.D. thesis; MSc Dissertation, Department of Computer Science, The
  University of Sheffield, UK. Available, Februray 2003, from http://www. dcs.
  shef. ac. uk/teaching/eproj/msc2002/abs/m1ir. htm; \bibinfo{year}{2002}.
\bibitem[{Roberts and Demner-Fushman(2016)}]{Roberts_2016}
\bibinfo{author}{Roberts\xfnm[ K.]}, \bibinfo{author}{Demner-Fushman\xfnm[
  D.]}.
\newblock \bibinfo{title}{Interactive use of online health resources: a
  comparison of consumer and professional questions}.
\newblock \bibinfo{journal}{Journal of the American Medical Informatics
  Association}
  \bibinfo{year}{2016};\bibinfo{volume}{23}(\bibinfo{number}{4}):\bibinfo{pages}{802--811}.
\newblock \URLprefix \url{https://doi.org/10.1093%2Fjamia%2Focw024}.
  \DOIprefix\doi{10.1093/jamia/ocw024}.
\bibitem[{Roberts et~al.(2014)Roberts, Kilicoglu, Fiszman and
  Demner-Fushman}]{roberts2014automatically}
\bibinfo{author}{Roberts\xfnm[ K.]}, \bibinfo{author}{Kilicoglu\xfnm[ H.]},
  \bibinfo{author}{Fiszman\xfnm[ M.]}, \bibinfo{author}{Demner-Fushman\xfnm[
  D.]}.
\newblock \bibinfo{title}{Automatically classifying question types for consumer
  health questions}.
\newblock In: \bibinfo{booktitle}{AMIA Annual Symposium Proceedings}.
  \bibinfo{organization}{American Medical Informatics Association}; volume
  \bibinfo{volume}{2014}; \bibinfo{year}{2014}. p. \bibinfo{pages}{1018}.
\bibitem[{Roberts et~al.(2016)Roberts, Rodriguez, Shooshan and
  Demner-Fushman}]{roberts2016resource}
\bibinfo{author}{Roberts\xfnm[ K.]}, \bibinfo{author}{Rodriguez\xfnm[ L.]},
  \bibinfo{author}{Shooshan\xfnm[ S.E.]}, \bibinfo{author}{Demner-Fushman\xfnm[
  D.]}.
\newblock \bibinfo{title}{Resource classification for medical questions}.
\newblock In: \bibinfo{booktitle}{AMIA Annual Symposium Proceedings}.
  \bibinfo{organization}{American Medical Informatics Association}; volume
  \bibinfo{volume}{2016}; \bibinfo{year}{2016}. p. \bibinfo{pages}{1040}.
\bibitem[{Robertson et~al.(1996)Robertson, Walker, Jones, Hancock-Beaulieu and
  Gatford}]{Robertson96okapiat}
\bibinfo{author}{Robertson\xfnm[ S.]}, \bibinfo{author}{Walker\xfnm[ S.]},
  \bibinfo{author}{Jones\xfnm[ S.]}, \bibinfo{author}{Hancock-Beaulieu\xfnm[
  M.]}, \bibinfo{author}{Gatford\xfnm[ M.]}.
\newblock \bibinfo{title}{Okapi at {TREC-3}}.
\newblock \bibinfo{year}{1996}. p. \bibinfo{pages}{109--126}.
\bibitem[{Robertson et~al.(2004)Robertson, Zaragoza and
  Taylor}]{Robertson_2004}
\bibinfo{author}{Robertson\xfnm[ S.]}, \bibinfo{author}{Zaragoza\xfnm[ H.]},
  \bibinfo{author}{Taylor\xfnm[ M.]}.
\newblock \bibinfo{title}{Simple {BM}25 extension to multiple weighted fields}.
\newblock In: \bibinfo{booktitle}{Proceedings of the Thirteenth {ACM}
  conference on Information and knowledge management - {CIKM} 04}.
  \bibinfo{publisher}{{ACM} Press}; \bibinfo{year}{2004}. \URLprefix
  \url{https://doi.org/10.1145%2F1031171.1031181}.
  \DOIprefix\doi{10.1145/1031171.1031181}.
\bibitem[{Ryu et~al.(2014)Ryu, Jang and Kim}]{Ryu_2014}
\bibinfo{author}{Ryu\xfnm[ P.M.]}, \bibinfo{author}{Jang\xfnm[ M.G.]},
  \bibinfo{author}{Kim\xfnm[ H.K.]}.
\newblock \bibinfo{title}{Open domain question answering using wikipedia-based
  knowledge model}.
\newblock \bibinfo{journal}{Information Processing {\&} Management}
  \bibinfo{year}{2014};\bibinfo{volume}{50}(\bibinfo{number}{5}):\bibinfo{pages}{683--692}.
\newblock \URLprefix \url{https://doi.org/10.1016%2Fj.ipm.2014.04.007}.
  \DOIprefix\doi{10.1016/j.ipm.2014.04.007}.
\bibitem[{Salton(1968)}]{salton1968automatic}
\bibinfo{author}{Salton\xfnm[ G.]}.
\newblock \bibinfo{title}{Automatic information organization and retrieval}.
\newblock \bibinfo{publisher}{McGraw-Hill}, \bibinfo{year}{1968}.
\bibitem[{Salton and McGill(1986)}]{salton1986introduction}
\bibinfo{author}{Salton\xfnm[ G.]}, \bibinfo{author}{McGill\xfnm[ M.J.]}.
\newblock \bibinfo{title}{Introduction to modern information retrieval}.
\newblock \bibinfo{publisher}{McGraw-Hill, Inc.}, \bibinfo{year}{1986}.
\bibitem[{Saneifar et~al.(2014)Saneifar, Bonniol, Poncelet and
  Roche}]{Saneifar_2014}
\bibinfo{author}{Saneifar\xfnm[ H.]}, \bibinfo{author}{Bonniol\xfnm[ S.]},
  \bibinfo{author}{Poncelet\xfnm[ P.]}, \bibinfo{author}{Roche\xfnm[ M.]}.
\newblock \bibinfo{title}{Enhancing passage retrieval in log files by query
  expansion based on explicit and pseudo relevance feedback}.
\newblock \bibinfo{journal}{Computers in Industry}
  \bibinfo{year}{2014};\bibinfo{volume}{65}(\bibinfo{number}{6}):\bibinfo{pages}{937--951}.
\newblock \URLprefix \url{https://doi.org/10.1016%2Fj.compind.2014.02.010}.
  \DOIprefix\doi{10.1016/j.compind.2014.02.010}.
\bibitem[{Sarker et~al.(2016)Sarker, Moll{\'{a}} and Paris}]{Sarker_2016}
\bibinfo{author}{Sarker\xfnm[ A.]}, \bibinfo{author}{Moll{\'{a}}\xfnm[ D.]},
  \bibinfo{author}{Paris\xfnm[ C.]}.
\newblock \bibinfo{title}{Query-oriented evidence extraction to support
  evidence-based medicine practice}.
\newblock \bibinfo{journal}{Journal of Biomedical Informatics}
  \bibinfo{year}{2016};\bibinfo{volume}{59}:\bibinfo{pages}{169--184}.
\newblock \URLprefix \url{https://doi.org/10.1016%2Fj.jbi.2015.11.010}.
  \DOIprefix\doi{10.1016/j.jbi.2015.11.010}.
\bibitem[{Sarrouti and Alaoui(2016)}]{Sarrouti_2016}
\bibinfo{author}{Sarrouti\xfnm[ M.]}, \bibinfo{author}{Alaoui\xfnm[ S.O.E.]}.
\newblock \bibinfo{title}{A generic document retrieval framework based on
  {UMLS} similarity for biomedical question answering system}.
\newblock In: \bibinfo{booktitle}{Intelligent Decision Technologies 2016}.
  \bibinfo{year}{2016}. p. \bibinfo{pages}{207--216}.
\newblock \DOIprefix\doi{10.1007/978-3-319-39627-9_18}.
\bibitem[{Sarrouti and Alaoui(2017{\natexlab{a}})}]{Sarrouti_bioasq_2017}
\bibinfo{author}{Sarrouti\xfnm[ M.]}, \bibinfo{author}{Alaoui\xfnm[ S.O.E.]}.
\newblock \bibinfo{title}{A biomedical question answering system in {BioASQ}
  2017}.
\newblock \bibinfo{journal}{Biomedical Natural Language Processing (BioNLP)
  Workshop at Association for Computational Linguistics (ACL'17)}
  \bibinfo{year}{2017}{\natexlab{a}};\URLprefix
  \url{https://aclweb.org/aclwiki/index.php?title=BioNLP_Workshop}.
  \DOIprefix\doi{10.18653/v1/w17-2337}.
\bibitem[{Sarrouti and Alaoui(2017{\natexlab{b}})}]{Sarrouti_MIM_2017}
\bibinfo{author}{Sarrouti\xfnm[ M.]}, \bibinfo{author}{Alaoui\xfnm[ S.O.E.]}.
\newblock \bibinfo{title}{A machine learning-based method for question type
  classification in biomedical question answering}.
\newblock \bibinfo{journal}{Methods of Information in Medicine}
  \bibinfo{year}{2017}{\natexlab{b}};\bibinfo{volume}{56}(\bibinfo{number}{3}).
\newblock \URLprefix \url{https://doi.org/10.3414%2Fme16-01-0116}.
  \DOIprefix\doi{10.3414/me16-01-0116}.
\bibitem[{Sarrouti and Alaoui(2017{\natexlab{c}})}]{Sarrouti_2017}
\bibinfo{author}{Sarrouti\xfnm[ M.]}, \bibinfo{author}{Alaoui\xfnm[ S.O.E.]}.
\newblock \bibinfo{title}{A passage retrieval method based on probabilistic
  information retrieval and {UMLS} concepts in biomedical question answering}.
\newblock \bibinfo{journal}{Journal of Biomedical Informatics}
  \bibinfo{year}{2017}{\natexlab{c}};\bibinfo{volume}{68}:\bibinfo{pages}{96--103}.
\newblock \URLprefix \url{https://doi.org/10.1016%2Fj.jbi.2017.03.001}.
  \DOIprefix\doi{10.1016/j.jbi.2017.03.001}.
\bibitem[{Sarrouti and Alaoui(2017{\natexlab{d}})}]{Sarrouti_yes_2017}
\bibinfo{author}{Sarrouti\xfnm[ M.]}, \bibinfo{author}{Alaoui\xfnm[ S.O.E.]}.
\newblock \bibinfo{title}{A yes/no answer generator based on sentiment-word
  scores in biomedical question answering}.
\newblock \bibinfo{journal}{International Journal of Healthcare Information
  Systems and Informatics}
  \bibinfo{year}{2017}{\natexlab{d}};\bibinfo{volume}{12}(\bibinfo{number}{3}):\bibinfo{pages}{62--74}.
\newblock \URLprefix \url{https://doi.org/10.4018%2Fijhisi.2017070104}.
  \DOIprefix\doi{10.4018/ijhisi.2017070104}.
\bibitem[{Sarrouti and Lachkar(2017)}]{Sarrouti_IBRA_2017}
\bibinfo{author}{Sarrouti\xfnm[ M.]}, \bibinfo{author}{Lachkar\xfnm[ A.]}.
\newblock \bibinfo{title}{A new and efficient method based on syntactic
  dependency relations features for ad hoc clinical question classification}.
\newblock \bibinfo{journal}{International Journal of Bioinformatics Research
  and Applications}
  \bibinfo{year}{2017};\bibinfo{volume}{13}(\bibinfo{number}{2}):\bibinfo{pages}{161--177}.
\newblock \DOIprefix\doi{10.1504/ijbra.2017.10003490}.
\bibitem[{Sarrouti et~al.(2015)Sarrouti, Lachkar and Ouatik}]{kdir15}
\bibinfo{author}{Sarrouti\xfnm[ M.]}, \bibinfo{author}{Lachkar\xfnm[ A.]},
  \bibinfo{author}{Ouatik\xfnm[ S.E.A.]}.
\newblock \bibinfo{title}{Biomedical question types classification using
  syntactic and rule based approach}.
\newblock In: \bibinfo{booktitle}{Proceedings of the 7th International Joint
  Conference on Knowledge Discovery, Knowledge Engineering and Knowledge
  Management}. \bibinfo{year}{2015}. p. \bibinfo{pages}{265--272}.
\newblock \DOIprefix\doi{10.5220/0005598002650272}.
\bibitem[{Schulze et~al.(2016)Schulze, Schuler, Draeger, Dummer, Ernst,
  Flemming, Perscheid and Neves}]{schulze2016hpi}
\bibinfo{author}{Schulze\xfnm[ F.]}, \bibinfo{author}{Schuler\xfnm[ R.]},
  \bibinfo{author}{Draeger\xfnm[ T.]}, \bibinfo{author}{Dummer\xfnm[ D.]},
  \bibinfo{author}{Ernst\xfnm[ A.]}, \bibinfo{author}{Flemming\xfnm[ P.]},
  \bibinfo{author}{Perscheid\xfnm[ C.]}, \bibinfo{author}{Neves\xfnm[ M.]}.
\newblock \bibinfo{title}{{HPI} question answering system in {BioASQ} 2016}.
\newblock In: \bibinfo{booktitle}{Proceedings of the Fourth BioASQ workshop at
  the Conference of the Association for Computational Linguistics}.
  \bibinfo{year}{2016}. p. \bibinfo{pages}{38--44}.
\bibitem[{Seol et~al.(2004)Seol, Kaufman, Mendon{\c{c}}a, Cimino and
  Johnson}]{seol2004scenario}
\bibinfo{author}{Seol\xfnm[ Y.H.]}, \bibinfo{author}{Kaufman\xfnm[ D.R.]},
  \bibinfo{author}{Mendon{\c{c}}a\xfnm[ E.A.]}, \bibinfo{author}{Cimino\xfnm[
  J.J.]}, \bibinfo{author}{Johnson\xfnm[ S.B.]}.
\newblock \bibinfo{title}{Scenario-based assessment of physicians' information
  needs.}
\newblock In: \bibinfo{booktitle}{Medinfo}. volume~\bibinfo{volume}{11};
  \bibinfo{year}{2004}. p. \bibinfo{pages}{306--310}.
\bibitem[{Simpson and Demner-Fushman(2012)}]{Simpson_2012}
\bibinfo{author}{Simpson\xfnm[ M.S.]}, \bibinfo{author}{Demner-Fushman\xfnm[
  D.]}.
\newblock \bibinfo{title}{Biomedical text mining: A survey of recent progress}.
\newblock In: \bibinfo{booktitle}{Mining Text Data}.
  \bibinfo{publisher}{Springer {US}}; \bibinfo{year}{2012}. p.
  \bibinfo{pages}{465--517}.
\newblock \URLprefix \url{https://doi.org/10.1007%2F978-1-4614-3223-4_14}.
  \DOIprefix\doi{10.1007/978-1-4614-3223-4_14}.
\bibitem[{Stearns et~al.(2001)Stearns, Price, Spackman and
  Wang}]{stearns2001snomed}
\bibinfo{author}{Stearns\xfnm[ M.Q.]}, \bibinfo{author}{Price\xfnm[ C.]},
  \bibinfo{author}{Spackman\xfnm[ K.A.]}, \bibinfo{author}{Wang\xfnm[ A.Y.]}.
\newblock \bibinfo{title}{{SNOMED} clinical terms: overview of the development
  process and project status.}
\newblock In: \bibinfo{booktitle}{Proceedings of the AMIA Symposium}.
  \bibinfo{organization}{American Medical Informatics Association};
  \bibinfo{year}{2001}. p. \bibinfo{pages}{662--666}.
\bibitem[{Strohman et~al.(2005)Strohman, Metzler, Turtle and
  Croft}]{strohman2005indri}
\bibinfo{author}{Strohman\xfnm[ T.]}, \bibinfo{author}{Metzler\xfnm[ D.]},
  \bibinfo{author}{Turtle\xfnm[ H.]}, \bibinfo{author}{Croft\xfnm[ W.B.]}.
\newblock \bibinfo{title}{Indri: A language model-based search engine for
  complex queries}.
\newblock In: \bibinfo{booktitle}{Proceedings of the International Conference
  on Intelligent Analysis}. \bibinfo{organization}{Citeseer};
  volume~\bibinfo{volume}{2}; \bibinfo{year}{2005}. p. \bibinfo{pages}{2--6}.
\bibitem[{Sun et~al.(2017)Sun, Luo and Chen}]{Sun_2017}
\bibinfo{author}{Sun\xfnm[ S.]}, \bibinfo{author}{Luo\xfnm[ C.]},
  \bibinfo{author}{Chen\xfnm[ J.]}.
\newblock \bibinfo{title}{A review of natural language processing techniques
  for opinion mining systems}.
\newblock \bibinfo{journal}{Information Fusion}
  \bibinfo{year}{2017};\bibinfo{volume}{36}:\bibinfo{pages}{10--25}.
\newblock \URLprefix \url{https://doi.org/10.1016%2Fj.inffus.2016.10.004}.
  \DOIprefix\doi{10.1016/j.inffus.2016.10.004}.
\bibitem[{Tellex et~al.(2003)Tellex, Katz, Lin, Fernandes and
  Marton}]{Tellex_2003}
\bibinfo{author}{Tellex\xfnm[ S.]}, \bibinfo{author}{Katz\xfnm[ B.]},
  \bibinfo{author}{Lin\xfnm[ J.]}, \bibinfo{author}{Fernandes\xfnm[ A.]},
  \bibinfo{author}{Marton\xfnm[ G.]}.
\newblock \bibinfo{title}{Quantitative evaluation of passage retrieval
  algorithms for question answering}.
\newblock In: \bibinfo{booktitle}{Proceedings of the 26th annual international
  {ACM} {SIGIR} conference on Research and development in informaion retrieval
  - {SIGIR} 03}. \bibinfo{publisher}{{ACM} Press}; \bibinfo{year}{2003}.
  \URLprefix \url{https://doi.org/10.1145%2F860435.860445}.
  \DOIprefix\doi{10.1145/860435.860445}.
\bibitem[{Teufel(2007)}]{Teufel}
\bibinfo{author}{Teufel\xfnm[ S.]}.
\newblock \bibinfo{title}{An overview of evaluation methods in {{TREC}} ad hoc
  information retrieval and {{TREC}} question answering}.
\newblock In: \bibinfo{booktitle}{Text, Speech and Language Technology}.
  \bibinfo{year}{2007}. p. \bibinfo{pages}{163--186}.
\newblock \DOIprefix\doi{10.1007/978-1-4020-5817-2_6}.
\bibitem[{Tsatsaronis et~al.(2015)Tsatsaronis, Balikas, Malakasiotis, Partalas,
  Zschunke, Alvers, Weissenborn, Krithara, Petridis, Polychronopoulos,
  Almirantis, Pavlopoulos, Baskiotis, Gallinari, Arti{\'e}res, Ngomo, Heino,
  Gaussier, Barrio-Alvers, Schroeder, Androutsopoulos and
  Paliouras}]{tsatsaronis2012bioasq}
\bibinfo{author}{Tsatsaronis\xfnm[ G.]}, \bibinfo{author}{Balikas\xfnm[ G.]},
  \bibinfo{author}{Malakasiotis\xfnm[ P.]}, \bibinfo{author}{Partalas\xfnm[
  I.]}, \bibinfo{author}{Zschunke\xfnm[ M.]}, \bibinfo{author}{Alvers\xfnm[
  M.R.]}, \bibinfo{author}{Weissenborn\xfnm[ D.]},
  \bibinfo{author}{Krithara\xfnm[ A.]}, \bibinfo{author}{Petridis\xfnm[ S.]},
  \bibinfo{author}{Polychronopoulos\xfnm[ D.]},
  \bibinfo{author}{Almirantis\xfnm[ Y.]}, \bibinfo{author}{Pavlopoulos\xfnm[
  J.]}, \bibinfo{author}{Baskiotis\xfnm[ N.]}, \bibinfo{author}{Gallinari\xfnm[
  P.]}, \bibinfo{author}{Arti{\'e}res\xfnm[ T.]}, \bibinfo{author}{Ngomo\xfnm[
  A.C.N.]}, \bibinfo{author}{Heino\xfnm[ N.]}, \bibinfo{author}{Gaussier\xfnm[
  E.]}, \bibinfo{author}{Barrio-Alvers\xfnm[ L.]},
  \bibinfo{author}{Schroeder\xfnm[ M.]}, \bibinfo{author}{Androutsopoulos\xfnm[
  I.]}, \bibinfo{author}{Paliouras\xfnm[ G.]}.
\newblock \bibinfo{title}{An overview of the {BIOASQ} large-scale biomedical
  semantic indexing and question answering competition}.
\newblock \bibinfo{journal}{BMC Bioinformatics}
  \bibinfo{year}{2015};\bibinfo{volume}{16}(\bibinfo{number}{1}):\bibinfo{pages}{1--28}.
\newblock \DOIprefix\doi{10.1186/s12859-015-0564-6}.
\bibitem[{Özlem Uzuner et~al.(2011)Özlem Uzuner, South, Shen and
  DuVall}]{Uzuner_2011}
\bibinfo{author}{Özlem Uzuner\xfnm[]}, \bibinfo{author}{South\xfnm[ B.R.]},
  \bibinfo{author}{Shen\xfnm[ S.]}, \bibinfo{author}{DuVall\xfnm[ S.L.]}.
\newblock \bibinfo{title}{2010 i2b2/{VA} challenge on concepts, assertions, and
  relations in clinical text}.
\newblock \bibinfo{journal}{Journal of the American Medical Informatics
  Association}
  \bibinfo{year}{2011};\bibinfo{volume}{18}(\bibinfo{number}{5}):\bibinfo{pages}{552--556}.
\newblock \URLprefix \url{https://doi.org/10.1136%2Famiajnl-2011-000203}.
  \DOIprefix\doi{10.1136/amiajnl-2011-000203}.
\bibitem[{Voorhees and Harman(2005)}]{voorhees2005trec}
\bibinfo{author}{Voorhees\xfnm[ E.M.]}, \bibinfo{author}{Harman\xfnm[ D.K.]}.
\newblock \bibinfo{title}{{TREC}: Experiment and Evaluation in Information
  Retrieval (Digital Libraries and Electronic Publishing)}.
\newblock \bibinfo{publisher}{The MIT Press}, \bibinfo{year}{2005}.
\bibitem[{Voorhees and Tice(1999)}]{voorhees1999trec}
\bibinfo{author}{Voorhees\xfnm[ E.M.]}, \bibinfo{author}{Tice\xfnm[ D.M.]}.
\newblock \bibinfo{title}{The trec-8 question answering track evaluation.}
\newblock In: \bibinfo{booktitle}{Proceedings of the Eigth Text REtrieval
  Conference (TREC-8)}. \bibinfo{address}{Gaithersburg, Maryland}:
  \bibinfo{publisher}{NIST}; volume \bibinfo{volume}{1999};
  \bibinfo{year}{1999}. p.~\bibinfo{pages}{82}.
\bibitem[{Wei et~al.(2013)Wei, Kao and Lu}]{Wei_2013}
\bibinfo{author}{Wei\xfnm[ C.H.]}, \bibinfo{author}{Kao\xfnm[ H.Y.]},
  \bibinfo{author}{Lu\xfnm[ Z.]}.
\newblock \bibinfo{title}{{PubTator}: a web-based text mining tool for
  assisting biocuration}.
\newblock \bibinfo{journal}{Nucleic Acids Research}
  \bibinfo{year}{2013};\bibinfo{volume}{41}(\bibinfo{number}{W1}):\bibinfo{pages}{W518--W522}.
\newblock \URLprefix \url{https://doi.org/10.1093%2Fnar%2Fgkt441}.
  \DOIprefix\doi{10.1093/nar/gkt441}.
\bibitem[{Weissenborn et~al.(2013)Weissenborn, Tsatsaronis and
  Schroeder}]{weissenborn2013answering}
\bibinfo{author}{Weissenborn\xfnm[ D.]}, \bibinfo{author}{Tsatsaronis\xfnm[
  G.]}, \bibinfo{author}{Schroeder\xfnm[ M.]}.
\newblock \bibinfo{title}{Answering factoid questions in the biomedical
  domain}.
\newblock \bibinfo{journal}{BioASQ@CLEF}
  \bibinfo{year}{2013};\bibinfo{volume}{1094}.
\bibitem[{Wiese et~al.(2017)Wiese, Weissenborn and Neves}]{wiese2017neural}
\bibinfo{author}{Wiese\xfnm[ G.]}, \bibinfo{author}{Weissenborn\xfnm[ D.]},
  \bibinfo{author}{Neves\xfnm[ M.]}.
\newblock \bibinfo{title}{Neural question answering at bioasq 5b}.
\newblock \bibinfo{journal}{arXiv preprint arXiv:170608568}
  \bibinfo{year}{2017};.
\bibitem[{William et~al.(2006)William, Aaron, Lynn and Phoebe}]{voorheestrec}
\bibinfo{author}{William\xfnm[ H.]}, \bibinfo{author}{Aaron\xfnm[ C.]},
  \bibinfo{author}{Lynn\xfnm[ R.]}, \bibinfo{author}{Phoebe\xfnm[ R.]}.
\newblock \bibinfo{title}{Trec 2006 genomics track overview}
  \bibinfo{year}{2006};.
\bibitem[{William et~al.(2007)William, Aaron, Lynn and
  Phoebe}]{voorheestrec2007}
\bibinfo{author}{William\xfnm[ H.]}, \bibinfo{author}{Aaron\xfnm[ C.]},
  \bibinfo{author}{Lynn\xfnm[ R.]}, \bibinfo{author}{Phoebe\xfnm[ R.]}.
\newblock \bibinfo{title}{Trec 2007 genomics track overview}
  \bibinfo{year}{2007};.
\bibitem[{Wilson et~al.(2005)Wilson, Wiebe and Hoffmann}]{Wilson_2005}
\bibinfo{author}{Wilson\xfnm[ T.]}, \bibinfo{author}{Wiebe\xfnm[ J.]},
  \bibinfo{author}{Hoffmann\xfnm[ P.]}.
\newblock \bibinfo{title}{Recognizing contextual polarity in phrase-level
  sentiment analysis}.
\newblock In: \bibinfo{booktitle}{Proceedings of the conference on Human
  Language Technology and Empirical Methods in Natural Language Processing -
  {HLT} 05}. \bibinfo{publisher}{Association for Computational Linguistics};
  \bibinfo{year}{2005}. \URLprefix
  \url{https://doi.org/10.3115%2F1220575.1220619}.
  \DOIprefix\doi{10.3115/1220575.1220619}.
\bibitem[{Woods(1973)}]{Woods_1973}
\bibinfo{author}{Woods\xfnm[ W.A.]}.
\newblock \bibinfo{title}{Progress in natural language understanding: An
  application to lunar geology}.
\newblock In: \bibinfo{booktitle}{Proceedings of the June 4-8, 1973, National
  Computer Conference and Exposition}. \bibinfo{address}{New York, NY, USA}:
  \bibinfo{publisher}{ACM}; AFIPS '73; \bibinfo{year}{1973}. p.
  \bibinfo{pages}{441--450}.
\newblock \URLprefix \url{http://doi.acm.org/10.1145/1499586.1499695}.
  \DOIprefix\doi{10.1145/1499586.1499695}.
\bibitem[{Wren(2011)}]{Wren_2011}
\bibinfo{author}{Wren\xfnm[ J.D.]}.
\newblock \bibinfo{title}{Question answering systems in biology and
  medicine--the time is now}.
\newblock \bibinfo{journal}{Bioinformatics}
  \bibinfo{year}{2011};\bibinfo{volume}{27}(\bibinfo{number}{14}):\bibinfo{pages}{2025--2026}.
\newblock \URLprefix \url{https://doi.org/10.1093%2Fbioinformatics%2Fbtr327}.
  \DOIprefix\doi{10.1093/bioinformatics/btr327}.
\bibitem[{Xu et~al.(2012)Xu, Zhou and Wang}]{Xu_2012}
\bibinfo{author}{Xu\xfnm[ J.]}, \bibinfo{author}{Zhou\xfnm[ Y.]},
  \bibinfo{author}{Wang\xfnm[ Y.]}.
\newblock \bibinfo{title}{A classification of questions using {SVM} and
  semantic similarity analysis}.
\newblock In: \bibinfo{booktitle}{2012 Sixth International Conference on
  Internet Computing for Science and Engineering}. \bibinfo{publisher}{{IEEE}};
  \bibinfo{year}{2012}. \URLprefix
  \url{https://doi.org/10.1109%2Ficicse.2012.49}.
  \DOIprefix\doi{10.1109/icicse.2012.49}.
\bibitem[{Xu et~al.(2006)Xu, Zhang, Ting and Jin-Shan}]{xu2006syntactic}
\bibinfo{author}{Xu\xfnm[ W.]}, \bibinfo{author}{Zhang\xfnm[ Y.]},
  \bibinfo{author}{Ting\xfnm[ L.]}, \bibinfo{author}{Jin-Shan\xfnm[ M.]}.
\newblock \bibinfo{title}{Syntactic structure parsing based {Chinese} question
  classification}.
\newblock \bibinfo{journal}{Journal of Chinese information processing}
  \bibinfo{year}{2006};\bibinfo{volume}{20}:\bibinfo{pages}{006}.
\bibitem[{Yang et~al.(2015)Yang, Gupta, Sun, Xu, Zhang and
  Nyberg}]{yang2015learning}
\bibinfo{author}{Yang\xfnm[ Z.]}, \bibinfo{author}{Gupta\xfnm[ N.]},
  \bibinfo{author}{Sun\xfnm[ X.]}, \bibinfo{author}{Xu\xfnm[ D.]},
  \bibinfo{author}{Zhang\xfnm[ C.]}, \bibinfo{author}{Nyberg\xfnm[ E.]}.
\newblock \bibinfo{title}{Learning to answer biomedical factoid and list
  questions oaqa at bioasq 3b}.
\newblock In: \bibinfo{booktitle}{Working Notes for the Conference and Labs of
  the Evaluation Forum (CLEF)}. \bibinfo{year}{2015}. .
\bibitem[{Yeh et~al.(2005)Yeh, Morgan, Colosimo and Hirschman}]{Yeh_2005}
\bibinfo{author}{Yeh\xfnm[ A.]}, \bibinfo{author}{Morgan\xfnm[ A.]},
  \bibinfo{author}{Colosimo\xfnm[ M.]}, \bibinfo{author}{Hirschman\xfnm[ L.]}.
\newblock \bibinfo{title}{{BioCreAtIvE} task 1a: gene mention finding
  evaluation}.
\newblock \bibinfo{journal}{{BMC} Bioinformatics}
  \bibinfo{year}{2005};\bibinfo{volume}{6}(\bibinfo{number}{Suppl
  1}):\bibinfo{pages}{S2}.
\newblock \URLprefix \url{https://doi.org/10.1186%2F1471-2105-6-s1-s2}.
  \DOIprefix\doi{10.1186/1471-2105-6-s1-s2}.
\bibitem[{Yenala et~al.(2015)Yenala, Kamineni, Shrivastava and
  Chinnakotla}]{yenalaiiith}
\bibinfo{author}{Yenala\xfnm[ H.]}, \bibinfo{author}{Kamineni\xfnm[ A.]},
  \bibinfo{author}{Shrivastava\xfnm[ M.]}, \bibinfo{author}{Chinnakotla\xfnm[
  M.]}.
\newblock \bibinfo{title}{{IIITH} at {BioASQ} challange 2015 {Task} 3b
  bio-medical question answering system}.
\newblock In: \bibinfo{booktitle}{Working Notes for the Conference and Labs of
  the Evaluation Forum (CLEF), Toulouse, France}. \bibinfo{year}{2015}. .
\bibitem[{Yepes(2017)}]{Jimeno_Yepes_2017}
\bibinfo{author}{Yepes\xfnm[ A.J.]}.
\newblock \bibinfo{title}{Word embeddings and recurrent neural networks based
  on long-short term memory nodes in supervised biomedical word sense
  disambiguation}.
\newblock \bibinfo{journal}{Journal of Biomedical Informatics}
  \bibinfo{year}{2017};\bibinfo{volume}{73}:\bibinfo{pages}{137--147}.
\newblock \URLprefix \url{https://doi.org/10.1016%2Fj.jbi.2017.08.001}.
  \DOIprefix\doi{10.1016/j.jbi.2017.08.001}.
\bibitem[{Yu and Cao(2008)}]{yu2008automatically}
\bibinfo{author}{Yu\xfnm[ H.]}, \bibinfo{author}{Cao\xfnm[ Y.g.]}.
\newblock \bibinfo{title}{Automatically extracting information needs from ad
  hoc clinical questions}.
\newblock In: \bibinfo{booktitle}{AMIA annual symposium proceedings}.
  \bibinfo{organization}{American Medical Informatics Association}; volume
  \bibinfo{volume}{2008}; \bibinfo{year}{2008}. p.~\bibinfo{pages}{96}.
\bibitem[{Yu et~al.(2007)Yu, Lee, Kaufman, Ely, Osheroff, Hripcsak and
  Cimino}]{Yu_2007}
\bibinfo{author}{Yu\xfnm[ H.]}, \bibinfo{author}{Lee\xfnm[ M.]},
  \bibinfo{author}{Kaufman\xfnm[ D.]}, \bibinfo{author}{Ely\xfnm[ J.]},
  \bibinfo{author}{Osheroff\xfnm[ J.A.]}, \bibinfo{author}{Hripcsak\xfnm[ G.]},
  \bibinfo{author}{Cimino\xfnm[ J.]}.
\newblock \bibinfo{title}{Development, implementation, and a cognitive
  evaluation of a definitional question answering system for physicians}.
\newblock \bibinfo{journal}{Journal of Biomedical Informatics}
  \bibinfo{year}{2007};\bibinfo{volume}{40}(\bibinfo{number}{3}):\bibinfo{pages}{236--251}.
\newblock \URLprefix \url{https://doi.org/10.1016%2Fj.jbi.2007.03.002}.
  \DOIprefix\doi{10.1016/j.jbi.2007.03.002}.
\bibitem[{Yu and Sable(2005)}]{yu2005being}
\bibinfo{author}{Yu\xfnm[ H.]}, \bibinfo{author}{Sable\xfnm[ C.]}.
\newblock \bibinfo{title}{Being erlang shen: identifying answerable questions}.
\newblock In: \bibinfo{booktitle}{In IJCAI Workshop on Knowledge and Reasoning
  for Answering Questions}. \bibinfo{year}{2005}. .
\bibitem[{Yu et~al.(2005{\natexlab{a}})Yu, Sable and Zhu}]{yu2005classifying}
\bibinfo{author}{Yu\xfnm[ H.]}, \bibinfo{author}{Sable\xfnm[ C.]},
  \bibinfo{author}{Zhu\xfnm[ H.R.]}.
\newblock \bibinfo{title}{Classifying medical questions based on an evidence
  taxonomy}.
\newblock In: \bibinfo{booktitle}{Proceedings of the AAAI 2005 workshop on
  question answering in restricted domains}.
  \bibinfo{year}{2005}{\natexlab{a}}. \URLprefix
  \url{http://www.aaai.org/Papers/Workshops/2005/WS-05-10/WS05-10-005.pdf}.
\bibitem[{Yu et~al.(2005{\natexlab{b}})Yu, Ting and Xu}]{yu2005modified}
\bibinfo{author}{Yu\xfnm[ Z.]}, \bibinfo{author}{Ting\xfnm[ L.]},
  \bibinfo{author}{Xu\xfnm[ W.]}.
\newblock \bibinfo{title}{Modified {Bayesian} model based question
  classification}.
\newblock \bibinfo{journal}{Journal of Chinese information processing}
  \bibinfo{year}{2005}{\natexlab{b}};\bibinfo{volume}{19}(\bibinfo{number}{2}):\bibinfo{pages}{100--105}.
\bibitem[{Zhang et~al.(2015)Zhang, Peng, You, Xie, Wang and
  Zhu}]{zhang2015fudan}
\bibinfo{author}{Zhang\xfnm[ Y.]}, \bibinfo{author}{Peng\xfnm[ S.]},
  \bibinfo{author}{You\xfnm[ R.]}, \bibinfo{author}{Xie\xfnm[ Z.]},
  \bibinfo{author}{Wang\xfnm[ B.]}, \bibinfo{author}{Zhu\xfnm[ S.]}.
\newblock \bibinfo{title}{The {Fudan} participation in the 2015 {BioASQ}
  challenge: Large-scale biomedical semantic indexing and question answering}.
\newblock In: \bibinfo{booktitle}{In Working Notes for the Conference and Labs
  of the Evaluation Forum (CLEF)}. \bibinfo{year}{2015}. .
\bibitem[{Zhang et~al.(2016)Zhang, Rahman, Braylan, Dang, Chang, Kim, McNamara,
  Angert, Banner, Khetan et~al.}]{zhang2016neural}
\bibinfo{author}{Zhang\xfnm[ Y.]}, \bibinfo{author}{Rahman\xfnm[ M.M.]},
  \bibinfo{author}{Braylan\xfnm[ A.]}, \bibinfo{author}{Dang\xfnm[ B.]},
  \bibinfo{author}{Chang\xfnm[ H.L.]}, \bibinfo{author}{Kim\xfnm[ H.]},
  \bibinfo{author}{McNamara\xfnm[ Q.]}, \bibinfo{author}{Angert\xfnm[ A.]},
  \bibinfo{author}{Banner\xfnm[ E.]}, \bibinfo{author}{Khetan\xfnm[ V.]},
  et~al.
\newblock \bibinfo{title}{Neural information retrieval: A literature review}.
\newblock \bibinfo{journal}{CoRRabs/161106792}
  \bibinfo{year}{2016};\DOIprefix\doi{http://arxiv.org/abs/1611.06792}.
\bibitem[{Zheng et~al.(2017)Zheng, Lin, Luo, Zhao, Li, Zhang, Yang and
  Wang}]{Zheng_2017}
\bibinfo{author}{Zheng\xfnm[ W.]}, \bibinfo{author}{Lin\xfnm[ H.]},
  \bibinfo{author}{Luo\xfnm[ L.]}, \bibinfo{author}{Zhao\xfnm[ Z.]},
  \bibinfo{author}{Li\xfnm[ Z.]}, \bibinfo{author}{Zhang\xfnm[ Y.]},
  \bibinfo{author}{Yang\xfnm[ Z.]}, \bibinfo{author}{Wang\xfnm[ J.]}.
\newblock \bibinfo{title}{An attention-based effective neural model for
  drug-drug interactions extraction}.
\newblock \bibinfo{journal}{{BMC} Bioinformatics}
  \bibinfo{year}{2017};\bibinfo{volume}{18}(\bibinfo{number}{1}).
\newblock \URLprefix \url{https://doi.org/10.1186%2Fs12859-017-1855-x}.
  \DOIprefix\doi{10.1186/s12859-017-1855-x}.
\bibitem[{Zhou et~al.(2007)Zhou, Yu, Smalheiser, Torvik and Hong}]{Zhou_2007}
\bibinfo{author}{Zhou\xfnm[ W.]}, \bibinfo{author}{Yu\xfnm[ C.]},
  \bibinfo{author}{Smalheiser\xfnm[ N.]}, \bibinfo{author}{Torvik\xfnm[ V.]},
  \bibinfo{author}{Hong\xfnm[ J.]}.
\newblock \bibinfo{title}{Knowledge-intensive conceptual retrieval and passage
  extraction of biomedical literature}.
\newblock In: \bibinfo{booktitle}{Proceedings of the 30th annual international
  {ACM} {SIGIR} conference on Research and development in information retrieval
  - {SIGIR} 07}. \bibinfo{publisher}{{ACM} Press}; \bibinfo{year}{2007}.
  \URLprefix \url{https://doi.org/10.1145%2F1277741.1277853}.
  \DOIprefix\doi{10.1145/1277741.1277853}.
\bibitem[{Zweigenbaum(2003)}]{zweigenbaum2003question}
\bibinfo{author}{Zweigenbaum\xfnm[ P.]}.
\newblock \bibinfo{title}{Question answering in biomedicine}.
\newblock In: \bibinfo{booktitle}{Proceedings Workshop on Natural Language
  Processing for Question Answering, EACL}. volume \bibinfo{volume}{2005};
  \bibinfo{year}{2003}. p. \bibinfo{pages}{1--4}.
\bibitem[{Zweigenbaum et~al.(2007)Zweigenbaum, Demner-Fushman, Yu and
  Cohen}]{Zweigenbaum_2007}
\bibinfo{author}{Zweigenbaum\xfnm[ P.]}, \bibinfo{author}{Demner-Fushman\xfnm[
  D.]}, \bibinfo{author}{Yu\xfnm[ H.]}, \bibinfo{author}{Cohen\xfnm[ K.B.]}.
\newblock \bibinfo{title}{Frontiers of biomedical text mining: current
  progress}.
\newblock \bibinfo{journal}{Briefings in Bioinformatics}
  \bibinfo{year}{2007};\bibinfo{volume}{8}(\bibinfo{number}{5}):\bibinfo{pages}{358--375}.
\newblock \URLprefix \url{https://doi.org/10.1093%2Fbib%2Fbbm045}.
  \DOIprefix\doi{10.1093/bib/bbm045}.

\end{thebibliography}
\includepdf[pages=1,pagecommand={\thispagestyle{empty}}, fitpaper=true,offset=-0.0cm -0.0cm]{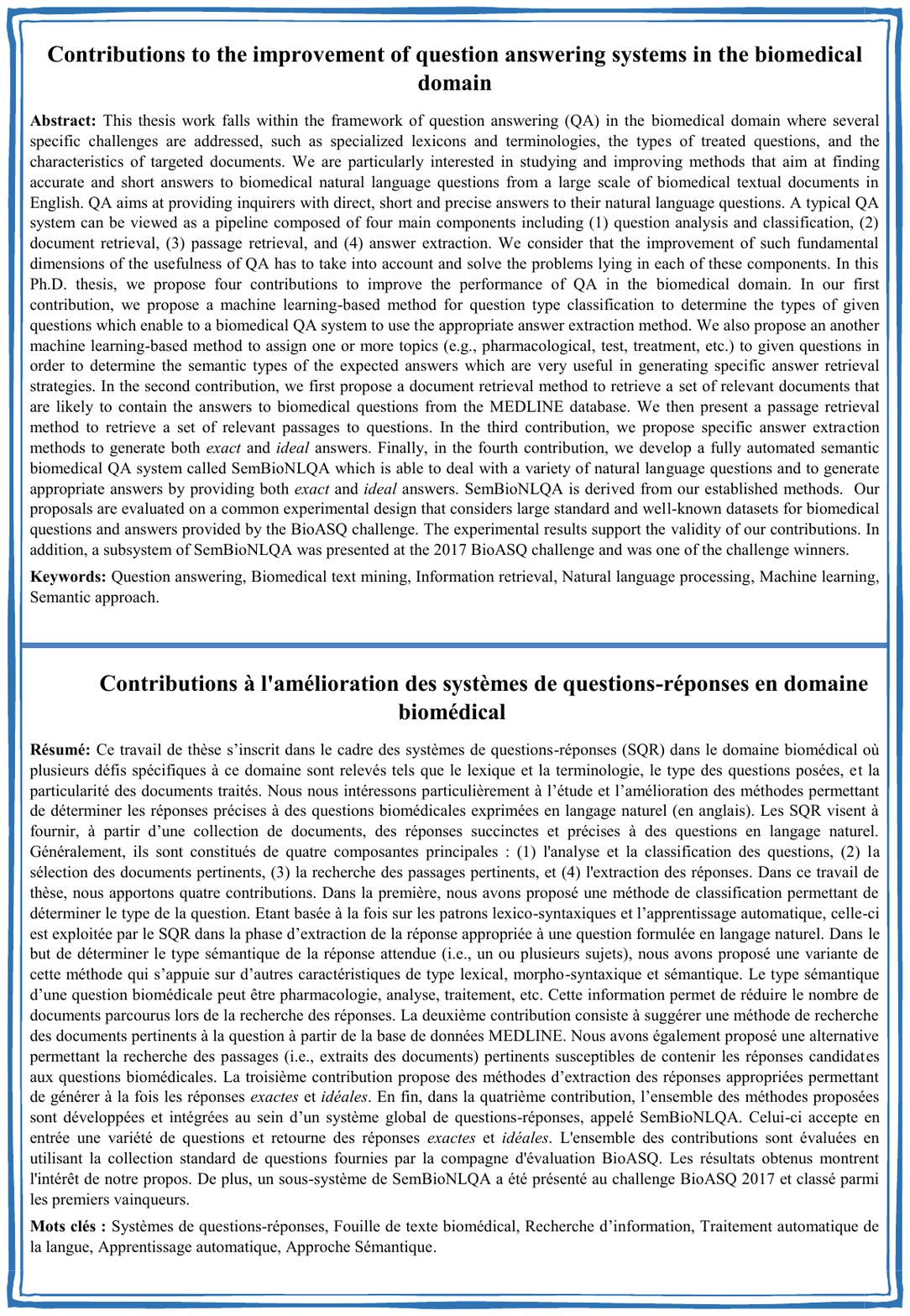}
\thispagestyle{empty}
\cleardoublepage
%

\end{document}